\documentclass[10pt,twocolumn,letterpaper]{article}

\usepackage{style/cvpr2025}              % To produce the CAMERA-READY version

\usepackage[accsupp]{axessibility}  % Improves PDF readability for those with disabilities.

\definecolor{cvprblue}{rgb}{0.21,0.49,0.74}
\usepackage[pagebackref,breaklinks,colorlinks,allcolors=cvprblue]{hyperref}

\usepackage[export]{adjustbox}

\usepackage{subcaption}  % To get subfigures and subcaptions
\usepackage{makecell}   % Used for multi-line table cell
\usepackage{amsmath}
\usepackage{nicefrac}
\usepackage{tikz}
\usetikzlibrary{positioning,shapes}
\usepackage{multirow}
\usepackage{float}

\usetikzlibrary{tikzmark}
\usepackage{enumitem}

\usepackage{mathtools}
\usepackage{caption}
\usepackage{verbatim} % For the comment environment % New (approved) packages
\definecolor{LIGHTPINK}{RGB}{237,157,202}
\definecolor{LIGHTRED}{RGB}{210,121,121}
\definecolor{LIGHTORANGE}{RGB}{230,170,50}
\definecolor{LIGHTGOLD}{RGB}{210,194,121}
\definecolor{LIGHTGREEN}{RGB}{121,210,121}
\definecolor{LIGHTAQUA}{RGB}{121,206,210}
\definecolor{LIGHTBLUE}{RGB}{121,124,210}
\definecolor{LIGHTPURPLE}{RGB}{153,102,255}
\definecolor{RED}{RGB}{178,34,34}
\definecolor{GRAY}{RGB}{166,166,166}
\definecolor{WHITE}{RGB}{255,255,255}

\newcommand{\showrevise}[1]{}  % Hide comments
\def\paperID{*****} % *** Enter the Paper ID here
\def\confName{CVPR}
\def\confYear{2026}

\title{Intrinsic Image Fusion for Multi-View 3D Material Reconstruction}

\author{
Peter Kocsis \quad
Lukas H{\"o}llein \quad
Matthias Nie{\ss}ner \\
Technical University of Munich
}

\newcommand{\myfigure}{
    \centering
    \vspace{-18pt}
    \tt\href{https://peter-kocsis.github.io/IntrinsicImageFusion/}{peter-kocsis.github.io/IntrinsicImageFusion/} \\
    \vspace{9pt}
    \includegraphics[width=\textwidth]{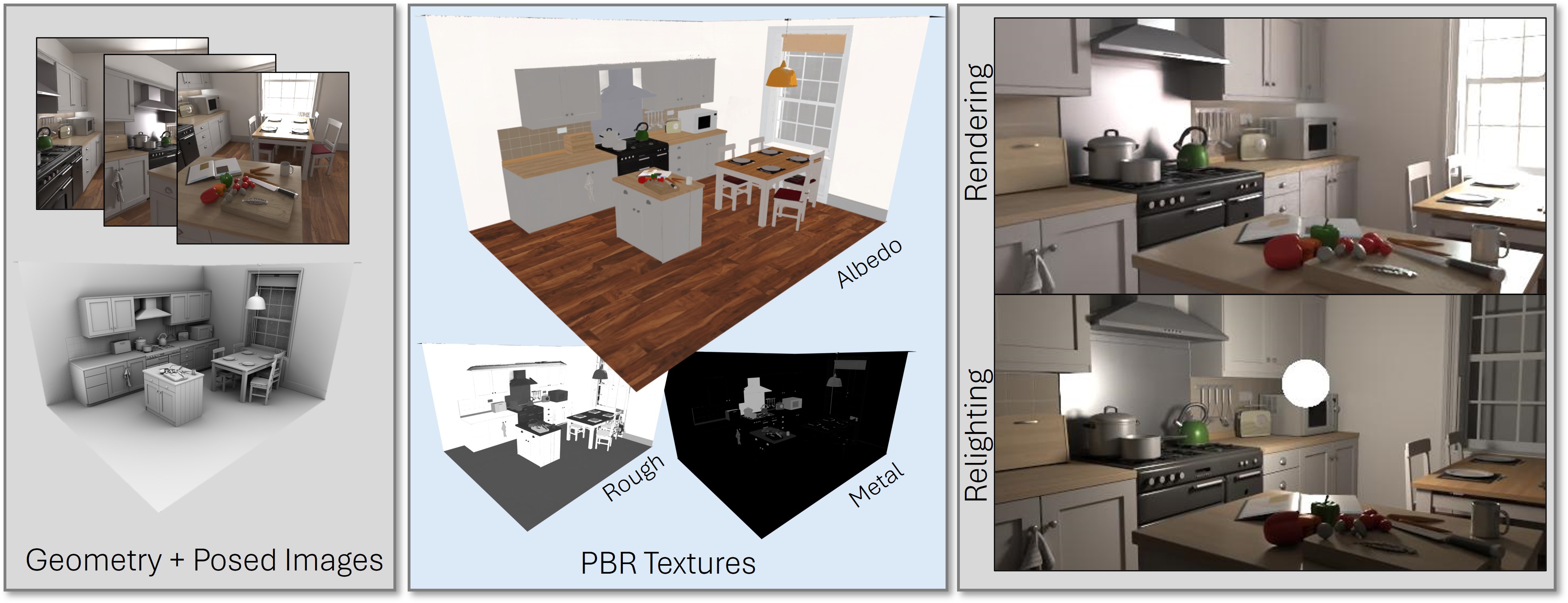}
    \vspace{-15pt}
    \captionsetup{type=figure}\caption{\textbf{Intrinsic Image Fusion (IIF).} 
    Our method reconstructs room-scale 3D physically based rendering (PBR) materials. 
    Given multi-view images with reconstructed geometry, we obtain high-quality PBR textures by distilling 2D decomposition priors into a 3D-consistent space via inverse path-tracing.
    This enables downstream applications for content creation, suitable for rendering and relighting. 
        }
    \label{fig:teaser}
    \vspace{12pt}
}
\makeatletter
\apptocmd\@maketitle{{\myfigure{}\par}}{}{}
\makeatother

\begin{document}
\maketitle

\begin{abstract} \vspace*{-21pt}

We introduce Intrinsic Image Fusion, a method that reconstructs high-quality physically based materials from multi-view images.
Material reconstruction is highly underconstrained and typically relies on analysis-by-synthesis, which requires expensive and noisy path tracing. 
To better constrain the optimization, we incorporate single-view priors into the reconstruction process. 
We leverage a diffusion-based material estimator that produces multiple, but often inconsistent, candidate decompositions per view.
To reduce the inconsistency, we fit an explicit low-dimensional parametric function to the predictions.
We then propose a robust optimization framework using soft per-view prediction selection together with confidence-based soft multi-view inlier set to fuse the most consistent predictions of the most confident views into a consistent parametric material space. 
Finally, we use inverse path tracing to optimize for the low-dimensional parameters. 
Our results outperform state-of-the-art methods in material disentanglement on both synthetic and real scenes, producing sharp and clean reconstructions suitable for high-quality relighting.
\end{abstract}       % Abstract (in whatever position required by the conference template)

%!TEX root = main.tex

%!TEX root = main.tex

%%
%% NOTE: Assign collaboration badges and section labels to all sections and 
%% subsects when created (badges include: \incomplete, \underRevision,
%% \readyForFeedback, \feedbackProvided, \complete, \locked)
%%

\begin{figure*}[t]
    \centering
    \setlength\tabcolsep{1.pt}
    \resizebox{\linewidth}{!}{
    \fboxsep=0pt
    \begin{tabular}{ccc|ccc|c}

        % ----- Row 1: Single-view inconsistency -----
        % \rotatebox[origin=c]{90}{\parbox{0cm}{\centering Single-view\\Inconsistent}} &
        \rotatebox{90}{Single-View mismatch} &
        \fbox{\includegraphics[width=0.20\linewidth]{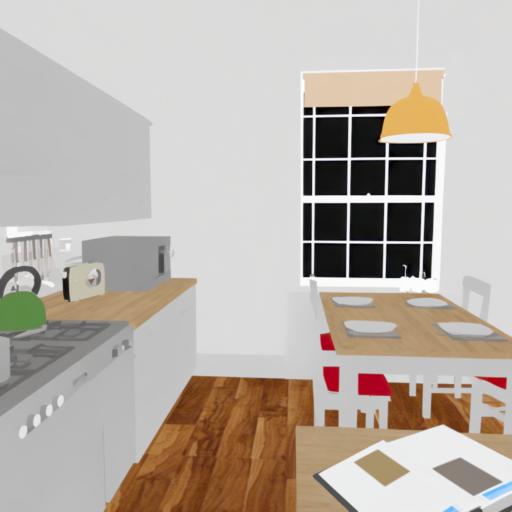}} &
        \fbox{\includegraphics[width=0.20\linewidth]{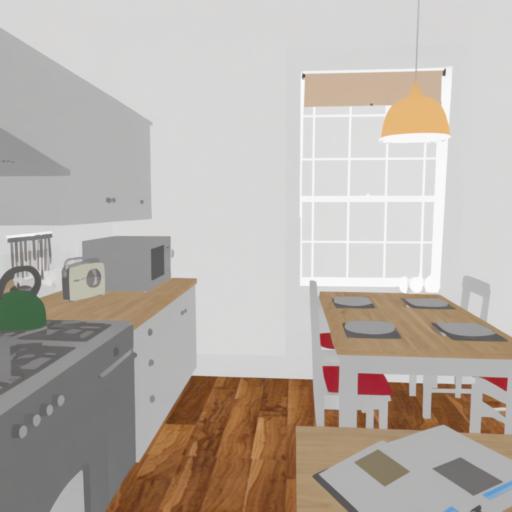}} &
        \fbox{\includegraphics[width=0.20\linewidth]{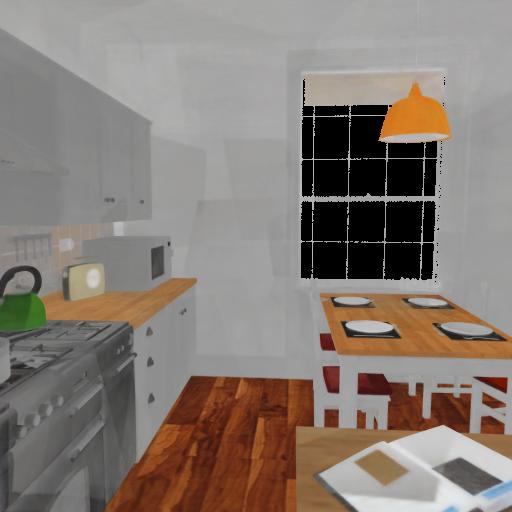}} &
        \fbox{\includegraphics[width=0.20\linewidth]{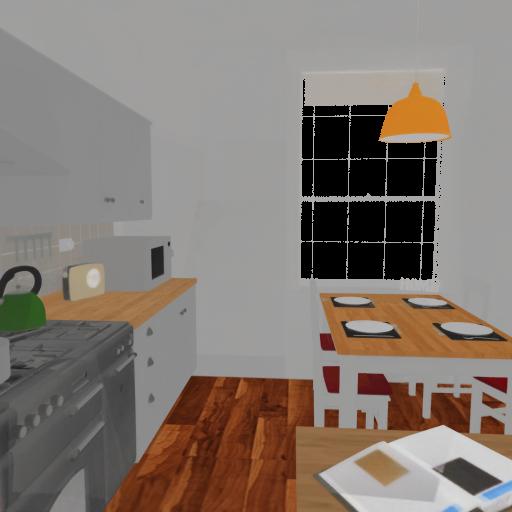}} &
        \fbox{\includegraphics[width=0.20\linewidth]{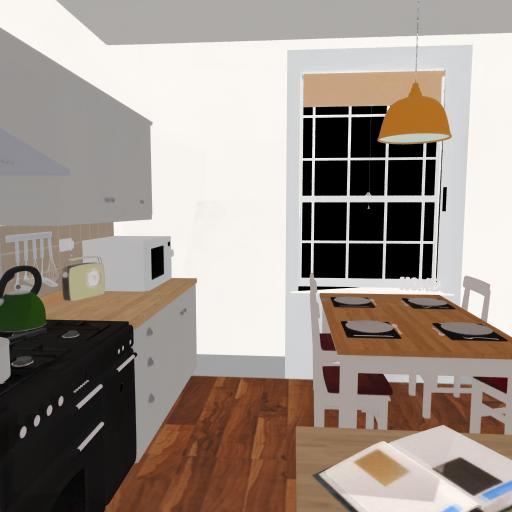}} &
        \fbox{\includegraphics[width=0.20\linewidth]{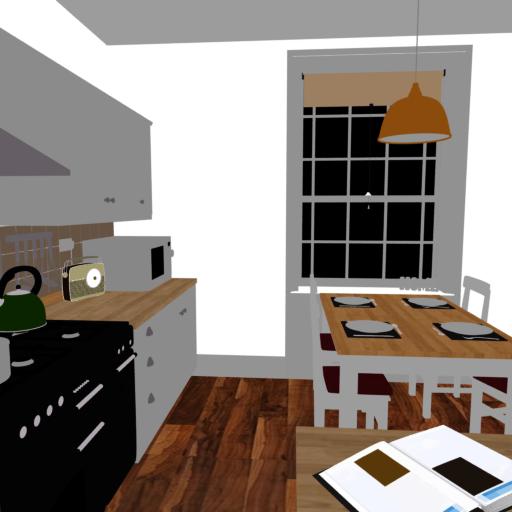}} \\

        % Label row under single-view
        % \multicolumn{2}{c}{Single-view inconsistent} & & & \\

        % ----- Row 2: Multi-view inconsistency -----
        % \rotatebox[origin=c]{90}{\parbox{0cm}{\centering Multi-view\\Inconsistent}} &
        \rotatebox{90}{Cross-View mismatch}&
        \fbox{\includegraphics[width=0.20\linewidth]{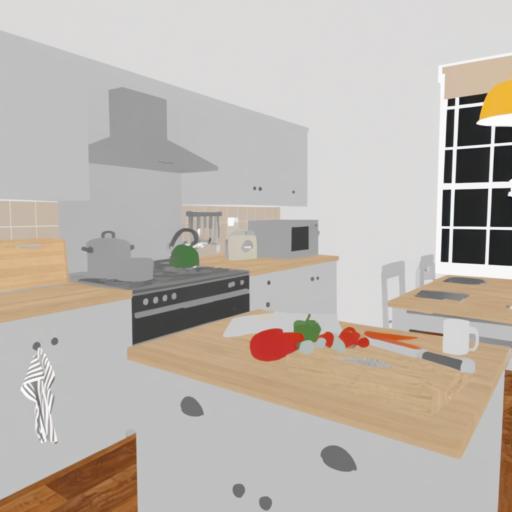}} &
        \fbox{\includegraphics[width=0.20\linewidth]{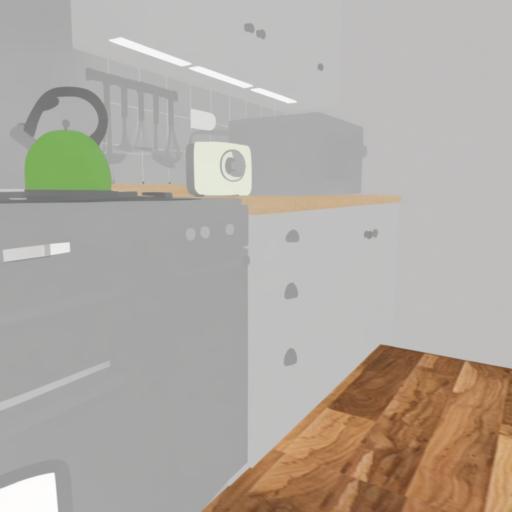}} &
        \fbox{\includegraphics[width=0.20\linewidth]{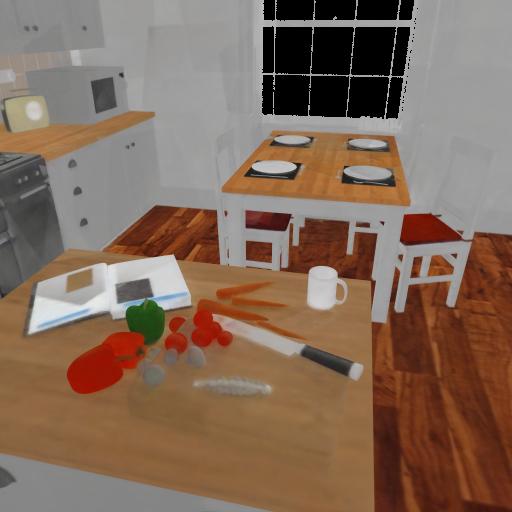}} &
        \fbox{\includegraphics[width=0.20\linewidth]{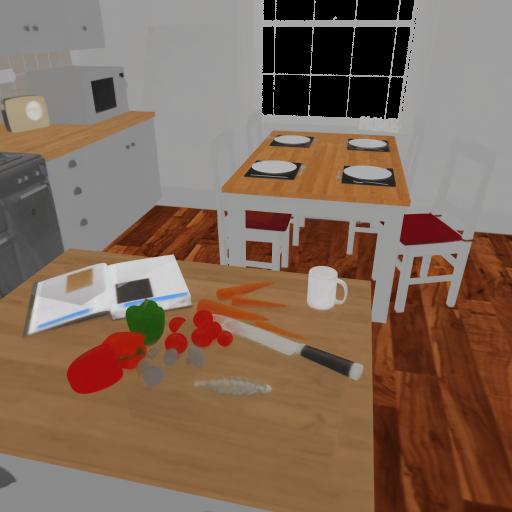}} &
        \fbox{\includegraphics[width=0.20\linewidth]{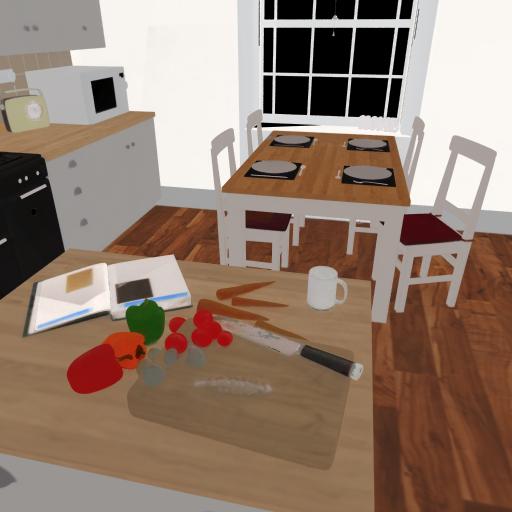}} &
        \fbox{\includegraphics[width=0.20\linewidth]{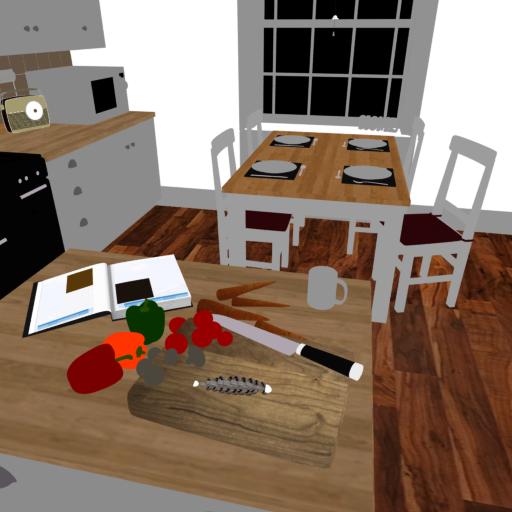}} \\

        % Label row under multi-view
        % \multicolumn{2}{c}{Multi-view inconsistent} & & & \\

        % ----- Bottom row: captions -----
        \multicolumn{3}{c}{(a) RGBX \cite{RGBX} Inconsistent Predictions} &
        (b) Naive Aggregation &
        (c) Our Base Aggregation &
        (d) Ours Fitted &
        (e) Ground Truth \\

    \end{tabular}}
    % \vspace{-9pt}
    \caption{\textbf{Parametric PBR Aggregation.}
    (a) RGBX \cite{RGBX} predicts PBR maps from 2D images, but they are inconsistent within the same view (top) or across views (bottom).
    (b) Naively using them to texture an entire 3D scene results in visible continuity artifacts and blurred-out details.
    (c) We propose a parametric aggregation, that distills the rich prior of 2D models into 3D-consistent and sharp PBR textures.
    (d) Using inverse path tracing we further optimize for the free parameters of our textures to find a physically grounded decomposition. 
    }
    \label{fig:method:motivation}
    % \vspace{-15pt}
\end{figure*}

\section{Introduction}
\label{sec:introduction}
\vspace{6pt}
%\pet{This is the retreat version. Ppl liked it, Angie also said it reads well. Their main recommendation was to add more details so ppl from other fields also understand (we need to highlight why is this a difficult task). The story is still similar but the probablistic inverse path tracing is dropped. So now the story: 1. Inverse Path Tracing is important for ... applications. 2. It's highly underconstrained, we need priors. 3. Prior predictions are ambiguos and inconsistent, we need our robust aggregation. 4. The prior predictions are reasonable, but physically incorrect, to ground them, we use inverse path tracing in a reduced parameter space. }
Accurately decomposing indoor scenes into physically based rendering (PBR) components—such as albedo, roughness, metallic properties, and illumination—is a core task in computer graphics and vision. 
High-quality PBR decomposition enables a wide range of applications, including relighting, material editing, and virtual object insertion. 
Despite its importance, recovering these properties at room scale remains challenging. 

Many inverse rendering approaches rely on path tracing to simulate the light transport in complex 3D scenes \cite{IPT,FIPT,IRIS}, but realistic renderings are computationally expensive and inherently noisy. 
This noise propagates into the optimization process, making it difficult to recover stable and accurate material estimates. 
In addition, appearance decomposition is fundamentally ambiguous: diffuse, specular, and lighting components are tightly coupled, particularly in complex indoor scenes with diverse geometries and light sources.

On the other hand, single image material estimators \cite{IID,RGBX} have seen unprecedented improvements in the recent years. 
Utilizing these 2D image priors of generative models have brought strong generalization capabilities, which enables high-quality material predictions solely from a single image.
These models act probabilistically and are thus able to sample possible solutions of the ambiguous problem of appearance decomposition.
Despite their high-quality results, patterns and relative reflectances are still inconsistently predicted across views, which is inherited from the probabilistic formulation.
In contrast, explicit re-rendering from only a single image is still challenging due to the unobserved regions in complex 3D scenes.

To this end, we introduce Intrinsic Image Fusion, a method that embeds single-view decomposition models into an inverse rendering optimization scheme.
This allows us to distill high-quality PBR textures for entire 3D scenes from strong 2D image priors.
First, we use RGBX \cite{RGBX}, a diffusion-based material estimator, to generate multiple candidate materials for each observed view of the 3D scene. 
From these candidates, we construct an explicit parametric distribution that captures the space of plausible materials.
Specifically, we express the texture of each object as the combination of a linear basis function and an ambiguity-invariant base pattern.
% Specifically, we segment the images with an off-the-shelf model \cite{SAM2} and express each object as the combination of a linear basis function and an ambiguity-invariant base pattern.
Furthermore, we model the variance of complex patterns by fitting a Laplacian distribution to multiple candidate predictions of the same object.
Next, we aggregate these single-view distributions into a 3D-consistent texture through our novel parametric distribution matching.
Finally, we use inverse rendering to optimize only for the remaining per-object parameters. 
%This way, we can balance between relying on the pretrained model and the rerendering; additionally enabling to use better rendering-based estimates on the uncertain regions. 
Our hybrid approach yields sharp and consistent material estimates thank to the pretrained predictions, while still being re-renderable thanks to the inverse path tracing.
In summary, our main contributions are:
\begin{itemize}[leftmargin=*,topsep=1pt, noitemsep]
    \item We model the solution space of plausible materials using an explicit parametric distribution, drastically reducing the number of free parameters and thereby limiting the impact of rendering noise of inverse path tracing.
    \item We aggregate single-view material predictions into a consistent 3D parametric distributional texture using consistent distribution matching, which enables us to utilize the most consistent predictions, instead of averaging them. 
\end{itemize}
%!TEX root = main.tex

%%
%% NOTE: Assign collaboration badges and section labels to all sections and 
%% subsects when created (badges include: \incomplete, \underRevision,
%% \readyForFeedback, \feedbackProvided, \complete, \locked)
%%

\begin{figure*}[t]
    \centering
    \includegraphics[width=\textwidth]{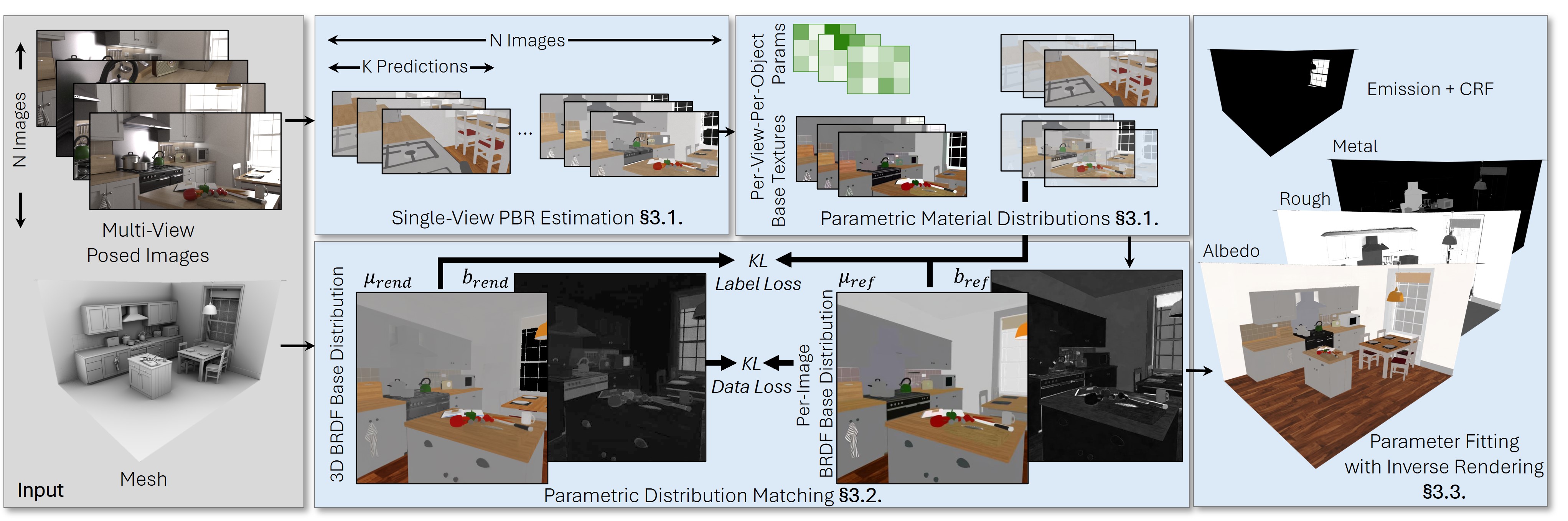}
    \vspace{-12pt}
    \caption{\textbf{Method Overview}. 
    IIF takes multi-view posed input images and their reconstructed geometry as input. 
    First, we estimate the materials for all the views using a monocular probabilistic estimator (\cref{sec:method:single_view}). 
    We define an explicit parametric distribution over the predictions consisting of a base texture together with a set of per-object parameters and aggregate the bases into a consistent 3D texture (\cref{sec:method:cross_view}).
    Then, we optimize for the free per-object parameters using inverse path tracing to get physically grounded albedo, rough and metal maps together with an emission texture and camera response function (CRF) suitable for complete rerendering. 
    }
    \label{fig:method:pipeline}
    \vspace{-12pt}
\end{figure*}

\newcommand{\mypar}[1]{\vspace{1mm}\noindent\textbf{#1.}}

\section{Related Work}
\label{sec:relatedwork}

\mypar{Inverse Rendering}
The goal of inverse rendering is to reconstruct a virtual renderable representation of a 3D scene, usually from multi-view images. 
Traditional methods formulate an optimization problem using analysis-by-synthesis. 
The key challenge is how to efficiently synthesize the image in this process, because light transport estimation is generally very costly. 
Recent methods have shown really impressive results using inverse path tracing and using brilliant engineering solutions to make this expensive optimization tractable by optimizing hierarchically \cite{IPT,MerlinMitsuba,Mitsuba3,MILO,I2SDF,NeILF, NeilfPP}, alternating between light transport estimation and reflectance evaluation \cite{FIPT}, introducing neural components \cite{INR}.
The recent work of IRIS \cite{IRIS} even managed to extend to LDR inputs and has introduced a proxy material using per-object aggregation of a single-view material estimator \cite{IRISFormer}. 
% NeRF-based \cite{mildenhall2020nerf} methods often optimize for cached incident radiance maps \cite{NeILF, NeilfPP}.  
However, rendering noise still poses a challenge for these methods and baked-in lighting, specular effects and geometric shading details are often noticeable in the diffuse materials. 
Our goal is to further reduce this noise by aggregating single-view predictions in a consistent manner and reduce the number of optimizable parameters. 

\mypar{Intrinsic Image Decomposition}
Predicting material and lighting information only from a single image has a longstanding history.
Early approaches focus on separating the reflectance from shading \cite{land1971lightness, horn1974determining, wu2023MAW} using various heuristics, such as sparsity in reflectance properties \cite{Entropy2004, shen2008NonLocal, grosse2009MITIntrinsics, zhang22SelfSimilarity}, or smoothness \cite{bell2014IIW}.
Later, deep-learning methods \cite{ComplexInvIndoor, LiS18a, IRISFormer} train decomposition networks on synthetic datasets, such as \cite{zhu2022learning, LiS18, Openrooms2021}.
Recently, \cite{IID,RGBX} formulate the ambiguous decomposition problem probabilistically using a diffusion model, thereby sampling from the solution space. 
Furthermore, utilizing the strong image prior of pretrained models enables unprecedented generalization to complex scenes and lighting conditions. 
Although these methods are able to produce impressive and sharp materials, their predictions are not necessarily physically correct and often prone to hallucination artifacts.  
Our method combines these model predictions with a rendering-based optimization to ensure physically correct and consistent textures for room-scale scenes. 

%!TEX root = main.tex

%%
%% NOTE: Assign collaboration badges and section labels to all sections and 
%% subsects when created (badges include: \incomplete, \underRevision,
%% \readyForFeedback, \feedbackProvided, \complete, \locked)
%%

\section{Method}
\label{sec:method}
% Given a set of $N$ multi-view images ${I_i}$ of an indoor scene, our goal is to reconstruct a full PBR representation: albedo, roughness, and metallic textures, together with an emission map and $M$ point lights. 
% We first reconstruct the scene geometry using an off-the-shelf method (e.g., 2DGS or BakedSDF) and treat it as fixed throughout the optimization. 
% Using this geometry, we combine single-view material priors with inverse rendering to produce high-quality, disentangled material and lighting estimates.
% Since path tracing usually gives noisy estimates, we aim to reduce the number of free material parameters. 
% We fit a parametric distribution to the single-view predictions (\cref{sec:method:single_view}) and then aggregate them in 3D to get a consistent distribution of possible materials (\cref{sec:method:cross_view}). 
% Finally, we optimize for the free parameters using factorized inverse path tracing (\cite{FIPT}), while accounting for LDR inputs (\cite{IRIS}) (\cref{sec:method:decomposition}). 
% We show our overall pipeline in \cref{fig:method:pipeline}

Our method optimizes a full PBR representation of a 3D scene: albedo, roughness, and metallic textures, together with an emission map and camera response function (CRF).
Given a set of posed image observations and the corresponding 3D scene reconstruction as input, we turn this RGB-space representation into PBR-space.
We combine single-view material priors ~\cite{RGBX} with inverse rendering to produce high-quality, disentangled material and lighting estimates.
Concretely, we first define a parametric distribution of the single-view predictions (\cref{sec:method:single_view}) and then aggregate them into a consistent 3D PBR representation (\cref{sec:method:cross_view}). 
% Concretely, we propose to exploit 2D material priors ~\cite{RGBX} to model a parametric per-view PBR distribution (\cref{sec:method:single_view}).
% Our novel distribution matching optimization then distills these model predictions into a consistent 3D PBR representation (\cref{sec:method:cross_view}).
Finally, we optimize for the free parameters using factorized inverse path tracing (\cite{FIPT}), while accounting for LDR inputs (\cite{IRIS}) (\cref{sec:method:decomposition}). 
We show our overall pipeline in \cref{fig:method:pipeline}

\subsection{Parametric single-view material distributions}
\label{sec:method:single_view}

Estimating materials requires disentangling reflectance from illumination while remaining consistent with physically accurate light transport. However, directly optimizing these quantities is difficult: photo-realistic evaluation of the rendering equation is computationally expensive, typically resulting in Monte-Carlo noise.
This leads to high-variance gradients and instabilities in the recovered materials.

To this end we propose to leverage the powerful priors of 2D decomposition methods like RGBX~\cite{RGBX}.
Since decomposing from RGB-space into PBR-space is an underconstrained problem, such 2D methods model a distribution of possible solutions.
As a result of this ambiguity, these predictions are not necessarily consistent within a view or across views (\Cref{fig:method:motivation}a), which limits their direct applicability to 3D material reconstruction (\Cref{fig:method:motivation}b).
Our core idea is to transform these predictions into a consistent parametric space and model their variance as a distribution.
This allows us to distill the predicted 2D PBR maps into consistent 3D PBR textures (\Cref{fig:method:motivation}c).

% To this end we propose to leverage the powerful priors of 2D decomposition methods like RGBX~\cite{RGBX}.
% Since decomposing from RGB-space into PBR-space is an underconstrained problem, such 2D methods model a distribution of possible solutions.
% This results in multiple valid decompositions (~\Cref{fig:method:motivation}a).
% However, these predictions are  inconsistent within a view or across views.
% This limits the applicability of directly utilizing such 2D methods for 3D decomposition (\Cref{fig:method:motivation}b).
% Our core idea is to transform these predictions into a consistent parametric space and aggregate the predicted distributions there (\Cref{fig:method:motivation}c). 
% % Our core idea is to model this uncertainty as \textit{per-image-per-object} distributions, which can be leveraged to distill consistent 3D decompositions (\Cref{fig:method:motivation}c).

\mypar{Parametric BRDF}
% \paragraph{Reference Transformations}
For each image observation of the scene, we predict $K$ reference decompositions of albedo, roughness, and metallic maps using RGBX \cite{RGBX}.  
% Inspired by the white-balancing literature, we define our parametric model to be linear with an additional bias to further increase the expressivity. 
% This model can already handle color shifts that might happen if the model incorrectly estimates the white-balance, but it's not too expressive to make the pattern over-parametrized. 
Since these raw material predictions are lying in the ambiguity-variant solution space, naive aggregation yields inconsistent textures (\Cref{fig:method:motivation}b). 
The main ambiguity between lighting and reflectance is the the scale invariance: it is hard to tell, whether the observed RGB color is caused by the reflectance or by the lighting. For example, multiple albedo colors are plausible for the kettle in \Cref{fig:method:motivation}a.
Inspired by the color calibration literature, we parametrize the solution space via learnable affine transformations, that capture these ambiguities.
Specifically, for each object in each prediction, we define a parametric model:
%
%
% Instead, we aggregate the ambiguity-invariant based texture distributions $B_{i,j}$ using Bayesian inference. 
% We parametrize this solution space via learnable affine transformations.
% These transformations model simple differences between predictions, such as different green-tones of the kettle in \Cref{fig:method:motivation}a.
% That is, for each object in each image, we can convert the predicted materials into a unified space by applying these transformations:
\vspace{-0.8em}
\begin{equation}
\label{eq:param-brdf}
\begin{gathered}
\bar{\mathbf{a}}_{i,k} = T^a_{i,k} \begin{bmatrix} \mathbf{a}_{i,k}, & 1 \end{bmatrix}, \quad T^a_{i,k} \in \mathbb{R}^{3 \times 4}, \\
\bar{\mathbf{r}}_{i,k} = T^r_{i,k} \begin{bmatrix} \mathbf{r}_{i,k}, & 1 \end{bmatrix}, \quad T^r_{i,k} \in \mathbb{R}^{1 \times 2}, \\
\bar{\mathbf{m}}_{i,k} = T^m_{i,k} \begin{bmatrix} \mathbf{m}_{i,k}, & 1 \end{bmatrix}, \quad T^m_{i,k} \in \mathbb{R}^{1 \times 2}
\end{gathered}
\end{equation}
where $\mathbf{a_{i,k}}$ are the albedo pixels of the $i$-th object in the $k$-th prediction of a given image observation, $T^a_{i,k}$ its corresponding affine transformation, and $\bar{\mathbf{a}}_{i,k}$ we call the albedo base texture. %the albedo in unified space.
We similarly denote the quantities for roughness and metallic as $\mathbf{r}$ and $\mathbf{m}$, respectively.

\mypar{BRDF Distribution}
Our parametric BRDF formulation can account for the per-object global inconsistencies across predictions, but it cannot solve inconsistencies in high-frequency patterns, such as on the countertop, which leads to oversmoothing, when aggregated in 3D (\Cref{fig:method:motivation}b).
Our goal is to find a distribution, whose mean is the most consistent prediction across all views.
To this end, we model the solution space of a single view as a \textit{per-image-per-object} Laplacian distribution.

% Next to simple tone-differences, 2D decomposition models can also produce more complex differences between predictions.
% For example, \textbf{TODO make an example e.g. texure in albedo vs reflectance/light combination reproducing it?}.
% To this end, we further model the solution space of PBR materials as a \textit{per-image-per-object} Laplacian distribution.

First, we introduce learnable assignment logits for each object in each reference image, per material: $z_{i,k}^{a} {\in} \mathbb{R}$ for albedo and $z_{i,k}^{r}, z_{i,k}^{m}$ for roughness and metallic, respectively. 
We then compute the pixel-wise weighted mixture of reference predictions for each material property:
\vspace{-0.8em}
\begin{equation}
\alpha_{i,k}^a = \frac{\exp(z_{i,k}^a / \tau_{\mathrm{logit}})}{\sum_{j=1}^{K} \exp(z_{n,j}^a / \tau_{\mathrm{logit}})}
\end{equation}
where $\tau_{\mathrm{logit}}$ is a fixed temperature. 
We then define the location of the distribution as $\boldsymbol{\mu}^\mathrm{ref}_i {=} [\bar{\mathbf{a}}_{i}, \bar{\mathbf{r}}_{i}, \bar{\mathbf{m}}_{i}]$, where $\bar{\mathbf{a}}_{i} {=} \sum_{k=1}^{K} \alpha_{i,k}^a \cdot \bar{\mathbf{a}}_{i,k}$, and $\bar{\mathbf{r}}_{i}$ and $\bar{\mathbf{m}}_{i}$ are obtained similarly.

To estimate the scale of the mixture, we compute the median deviation of the reference predictions from the mixture:
% \vspace{-0.1em}
\begin{equation}
\mathbf{b}^\mathrm{ref}_i = \mathrm{median}_{k=1}^K \big| \boldsymbol{\mu}^\mathrm{ref}_i - [\bar{\mathbf{a}}_{i,k}, \bar{\mathbf{r}}_{i,k}, \bar{\mathbf{m}}_{i,k}] \big|
\end{equation}

We then define the Laplacian distribution of the reference predictions as $p^\mathrm{ref}_i \sim \mathrm{Laplace}(\boldsymbol{\mu}^\mathrm{ref}_i, \mathbf{b}^\mathrm{ref}_i)$.

\subsection{Distribution Matching Optimization}
\label{sec:method:cross_view}

In order to obtain a 3D-consistent PBR texture of the whole scene, we utilize the \textit{per-image-per-object} PBR distributions in a distribution matching optimization.
In other words, the 3D PBR texture is similarly modeled as a Laplacian distribution and should be identical to the respective 2D distribution when rendered into images.
This allows us to distill the rich 2D prior of models such as RGBX \cite{RGBX} into a consistent 3D PBR texture.

\mypar{PBR Texture Model}
During optimization, we randomly sample $N$ pixels across all input images and retrieve the corresponding 3D coordinates $\mathbf{x}_n \in \mathbb{R}^3$ on the reconstructed mesh.
A BRDF network $f_\theta$ based on InstantNGP~\cite{InstantNGP} predicts material properties and associated uncertainties at these 3D positions:
\vspace{-0.8em}
\begin{equation}
(\mathbf{a}_n, \boldsymbol{\sigma}^a_n), (\mathbf{r}_n, \boldsymbol{\sigma}^r_n), (\mathbf{m}_n, \boldsymbol{\sigma}^m_n) = f_\theta(\mathbf{x}_n)
\end{equation}
We utilize these outputs to define the Laplacian distribution of the PBR texture as:
\vspace{-0.8em}
\begin{equation}
\begin{gathered}
\boldsymbol{\mu}^\mathrm{pred}_n = [\mathbf{a}_n, \mathbf{r}_n, \mathbf{m}_n], \\
\mathbf{b}^\mathrm{pred}_n = [\boldsymbol{\sigma}^a_n, \boldsymbol{\sigma}^r_n, \boldsymbol{\sigma}^m_n], \\
p^\mathrm{pred}_n \sim \mathrm{Laplace}(\boldsymbol{\mu}^\mathrm{pred}_n, \mathbf{b}^\mathrm{pred}_n)
\end{gathered}
\end{equation}

\mypar{Data Loss via Laplacian Distributions}
We encourage the BRDF network predictions to match the reference mixture under the Laplacian distribution assumption, by computing the KL-divergence:
\vspace{-1.2em}
\begin{equation}
\mathcal{L}_{\mathrm{data}} = \frac{1}{N} \sum_{n=1}^{N}  D_{\mathrm{KL}} \big( p^\mathrm{ref}_{i_n} \parallel p^\mathrm{pred}_n \big)
\end{equation}
where $i_n$ specifies which of the \textit{per-image-per-object} 2D distributions to pick for the $n$-th sampled pixel.

\mypar{Label Loss for Assignment Logits}
In practice, we found that the assignment logits need to be regularized to reach stable results.
We choose a simple L2 regularization between the rendered materials and each reference prediction:
\vspace{-0.8em}
\begin{equation}
\begin{gathered}
\mathbf{E}_{n,k} = \mathcal{L}_2(\boldsymbol{\mu}^\mathrm{pred}_n, [\bar{\mathbf{a}}_{i_n,k}, \bar{\mathbf{r}}_{i_n,k}, \bar{\mathbf{m}}_{i_n,k}]) \\
q_{n,k} = \frac{\exp(- \mathbf{E}_{n,k} / \tau_{\mathrm{err}})}{\sum_{j=1}^{K} \exp(- \mathbf{E}_{n,j} / \tau_{\mathrm{err}})}
\end{gathered}
\end{equation}

The label loss is then defined as:
\vspace{-0.9em}
\begin{equation}
\mathcal{L}_{\mathrm{label}} = - \frac{1}{N} \sum_{n=1}^{N} \sum_{k=1}^{K} q_{n,k} \log [\alpha_{i_n,k}^a, \alpha_{i_n,k}^r, \alpha_{i_n,k}^m]
\end{equation}
\vspace{-1.5em}

\mypar{Optimization}
The total loss combines the BRDF data term, the label regularization, and an identity-regularizer for the affine transformations:
\vspace{-0.9em}
\begin{equation}
\mathcal{L}_{\mathrm{total}} = w_{\mathrm{data}} \, \mathcal{L}_{\mathrm{data}} 
+ w_{\mathrm{label}} \, \mathcal{L}_{\mathrm{label}} 
+ w_{\mathrm{reg}} \, \mathcal{L}_{\mathrm{reg}}(T)
\end{equation}

The optimization jointly updates the BRDF network parameters $\theta$, the assignment logits $z$, and the affine transformations $T$ to distill a consistent 3D material texture from uncertain multi-view 2D predictions.

\subsection{Parameter Fitting with Inverse Rendering}
\label{sec:method:decomposition}

After aggregating the parametrized single-view predictions, we obtain a 3D texture of base distributions.  
However, these distributions are ambiguity-invariant per object; that is, they are consistent across views for each object, but they represent independent samples from the solution space.  
To obtain a physically-grounded decomposition, we optimize for per-object PBR parameters $T_{o}^a,T_{o}^r,T_{o}^m$ of every 3D object $o$, which we achieve with analysis-by-synthesis.

To optimize through photo-realistic rendering, we use path tracing to solve the rendering equation \cite{Rendering}:
\vspace{-0.7em}
\begin{equation}
\begin{aligned}
L_o(x, \omega_o) 
&= L_e(x, \omega_o) \\
&\quad + 
\int_{\Omega} 
f_r(x, \omega_i, \omega_o) \,
L_i(x, \omega_i) \,
(\omega_i \cdot n_x) \, d\omega_i
\end{aligned}
\end{equation}
where $n_x$ is the surface normal at point $x$.

We model the surface reflectance using the Cook-Torrance microfacet model \cite{CookTorrance}:
\vspace{-0.5em}
\begin{equation}
\begin{aligned}
f_r(\mathbf{x}, \mathbf{\omega_i}, \mathbf{\omega_o}) &= 
f_\mathrm{diff}(\mathbf{x}) + f_\mathrm{spec}(\mathbf{x}, \mathbf{\omega_i}, \mathbf{\omega_o}) \\
f_\mathrm{diff}(x) &= \frac{\mathbf{k_d}(x)}{\pi} \\
f_\mathrm{spec}(\mathbf{x}, \mathbf{\omega_i}, \mathbf{\omega_o}) &= 
\frac{D(\mathbf{h}) \, F(\mathbf{\omega_i}, \mathbf{h}, \mathbf{k_s}) \, G(\mathbf{\omega_i}, \mathbf{\omega_o}, \mathbf{h})}{4 (\mathbf{n}_x \cdot \mathbf{\omega_i})(\mathbf{n}_x \cdot \mathbf{\omega_o})}
\end{aligned}
\end{equation}
where $\mathbf{k_d}(\mathbf{x}) {=} (1 {-} m(\mathbf{x})) \mathbf{a}(\mathbf{x})$, $\mathbf{k_s}(\mathbf{x}) {=} 0.04\cdot(1 {-} m(\mathbf{x}))+\mathbf{a}(\mathbf{x})m(\mathbf{x})$, $h$ is the half-angle, $D(h)$ is the microfacet normal distribution function, $F(\omega_i, h)$ is the Fresnel term, $G(\omega_i, \omega_o, h)$ is the geometric visibility term, $\mathbf{a}(x)$ is the albedo, $m(x)$ is the metallic at $x$, and $\mathbf{n}_x$ is the surface normal.
The roughness $\mathbf{r}(x)$ and metallic $\mathbf{m}(x)$ textures modulate the specular lobe and Fresnel behavior according to standard PBR conventions~\cite{CookTorrance}.

Path tracing approximates the rendering equation with Monte-Carlo sampling of the rendering integral, usually resulting in noisy estimates. 
This noise is then backpropagated into the parameters, often causing backed-in shading effects. 
Our method regularizes this optimization by drastically reducing the total number of trainable parameters.
Instead of optimizing for the full BRDF texture, we only need to find the per-object transformations. 
To further reduce the variance, we follow \cite{FIPT} and do an alternating optimization. 

\mypar{1. Lighting Optimization} 
The scene illumination is represented as per-triangle uniform emission $\mathbf{E} {=} \{E_t\}_{t \in \mathcal{T}}$, where $\mathcal{T}$ is the set of emissive triangles.  
We pre-filter triangles that are constantly oversaturated and optimize the remaining ones via inverse path tracing while keeping the BRDF $f_\theta$ fixed:
\begin{equation}
\mathcal{L}_{\mathrm{light}} = \sum_{i,p} \big\| \hat{L}_{o}(\mathbf{x},\mathbf{\omega_i;f_\theta},\mathbf{E}) - \mathbf{I}_{i}(p) \big\|_2,
\end{equation}
where $\mathbf{I}_i$ is the radiance of view $i$, pixel $p$ with position $x$.  
A volumetric diffuse radiance cache is used to approximate multi-bounce lighting $L_i$ after the final bounce, and low-intensity triangles are pruned after several iterations to accelerate convergence.

\mypar{2. Light Transport Caching}  
Second, we cache the light transport. 
We render diffuse and specular shading maps, containing per-pixel pre-integrated lighting information \cite{FIPT}. 
Since this step does not involve any optimization, we can allow an estimation with higher sample count. 

\mypar{3. BRDF Parameter Fitting} 
Finally, we optimize for the per-object parameters  $T_{o}^a,T_{o}^r,T_{o}^m$ of our BRDF texture. 
We render the scene using the equation from FIPT. 
To account for LDR inputs, we jointly optimize for CRF parameters, following \cite{IRIS}. 

%!TEX root = main.tex

%%
%% NOTE: Assign collaboration badges and section labels to all sections and 
%% subsects when created (badges include: \incomplete, \underRevision,
%% \readyForFeedback, \feedbackProvided, \complete, \locked)
%%

\section{Experiments}
\label{sec:experiments}
\mypar{Implementation details}
% Our material textures are represented using a multi-resolution hashgrid encoding \cite{InstantNGP}, which allows the texture field to remain smooth in 3D without relying on a specific UV parameterization.
% This choice avoids artifacts from imperfect UV unwrapping and ensures consistent optimization across the scene volume.
% We provide all the hyperparameters and runtime usage in the supplemental. 
We use $K{=}16$ predictions per view and aggregate them using Adam \cite{Adam} optimizer (bs${=} 65536$, lr ${=} 1e{-}2$, decayed by $0.5$ every 2 epochs) with weights $w_{data}{=}w_{label}{=}1$,$w_{reg}{=}1e{+}2$.
We anneal the temperatures $\tau_{err}{=}\tau_{logit}{=}1$ every $100$ iterations by $0.85$. 
We run the distribution matching optimization for $10$ epochs, which takes approximately 5 minutes. 
Our parameter fitting follows the implementation in FIPT \cite{FIPT} using Mitsuba 3\cite{Mitsuba3} and converges in approximately 55 minutes on a single A6000 GPU.

% \textbf{\textcolor{red}{mention: how we obtain the 3d segmentation in practice, how many pixels sampled per iteration (batch-size), loss weights, lr, total GPU memory, total optimization runtime, softmax temperatures}}
% For all of our experiments, we used LDR inputs. 

\mypar{Test Scenes}
\label{sec:exp:dataset}
We provide evaluations on the synthetic scenes from \cite{BitterliScenes}. 
We transform the scenes into Blender \cite{Blender} and unify all the BRDFs to make sure that the renderings follows our BRDF. 
Finally, we render all the modalities together with instance segmentations from multiple viewpoints at $512$x$512$ resolution. 

For our real evaluations, we use ScanNet++ \cite{ScanNet++} (2a1b555966, 651dc6b4f1). 
As geometry, we use the laser scan mesh. 
Although the laser scan provides high-quality surface information, it contains holes and reflective surfaces, such as windows are not reconstructed. 
To account for this, we extend our lighting representation for these scenes with an environment map of resolution $16$x$32$ wrapped around a scene. 
To obtain 3D consistent instance segmentation, we use MaskClustering \cite{MaskClustering} with SAM \cite{SAM2} predictions. 
We provide additional details on the data pre-processing in our supplemental.

\subsection{Comparisons}
\label{sec:experiments:comparisons}
\begin{table}[t]
    \centering\setlength{\tabcolsep}{4pt}
    \resizebox{\columnwidth}{!}{%
    \begin{tabular}{l|ccc|c|c|c}
    \toprule
        & \multicolumn{3}{c}{Albedo} & \multicolumn{1}{c}{Rough} & \multicolumn{1}{c}{Metal} & \multicolumn{1}{c}{Emit}\\
        Method & PSNR $\uparrow$ & SSIM $\uparrow$ & LPIPS $\downarrow$ & L2 $\downarrow$ & L2 $\downarrow$ & L2 $\downarrow$ \\
    \midrule
       NeILF++~\cite{NeilfPP} & 13.18 & 0.733 & 0.375 & 0.103 & 0.047 & N/A \\
       FIPT~\cite{FIPT} & 10.63 & 0.661 & 0.403 & 0.110 & \textbf{0.006} & 2.208 \\
       IRIS~\cite{IRIS} & 15.86 & 0.735 & 0.307 & 0.056 & 0.040 & 2.046 \\
       IIF (Ours) & \textbf{20.72} & \textbf{0.846} & \textbf{0.201} & \textbf{0.028} & 0.007 & \textbf{0.384} \\
    \bottomrule
    \end{tabular}%
    }
    \vspace{-3pt}
    \caption{\textbf{Baseline comparisons.}
        We show quantitative results averaged over all the views of our four synthetic scenes (\Cref{sec:exp:dataset}).
        Our parametric formulation gives more constraints to inverse path tracing, yielding consistent and sharp predictions, outperforming the baselines by a high margin. 
    }
    \label{tab:exp:comparisons}
\end{table}

\begin{comment}
Albedo: 
| Method | l2 | lpips | psnr | ssim |
|---|---|---|---|---|
| v0_render | **0.011** | **0.201** | **20.718** | **0.846** | 
| neilfpp | 0.057 | 0.375 | 13.184 | 0.733 | 
| fipt_hq | 0.093 | 0.403 | 10.628 | 0.661 | 
| iris_hq | 0.042 | 0.307 | 15.858 | 0.735 | 

Rough:
| Method | l2 | lpips | psnr | ssim |
|---|---|---|---|---|
| v0_render | **0.028** | 0.188 | 25.204 | 0.840 | 
| neilfpp | 0.103 | 0.391 | 10.095 | 0.817 | 
| fipt_hq | 0.110 | 0.456 | 10.095 | 0.673 | 
| iris_hq | 0.056 | **0.181** | **26.328** | 0.843 | 

Metal:
| Method | l2 | lpips | psnr | ssim |
|---|---|---|---|---|
| v0_render | 0.007 | 0.162 | 29.507 | 0.821 | 
| neilfpp | 0.047 | **0.046** | **41.740** | 0.942 | 
| fipt_hq | **0.006** | 0.046 | 41.476 | **0.979** | 
| iris_hq | 0.040 | 0.170 | 26.634 | 0.827 | 

Emit:
| Method | iou | l2 | log_l2 | ssim |
|---|---|---|---|---|
| v0_render | **0.773** | **0.384** | **0.018** | **0.989** | 
| fipt_hq | 0.664 | 2.208 | 0.042 | 0.971 | 
| iris_hq | 0.558 | 2.046 | 0.091 | 0.963 | 
\end{comment}

% Huge table, left: conditioning (text + component, in and out of domain), right: all the components
\begin{figure*}[t]
    \centering
    \setlength\tabcolsep{1.25pt}
    \resizebox{\textwidth}{!}{
    \fboxsep=0pt
    
    \begin{tabular}{c|ccc|c|c}    
        \begin{tabular}{c}
            \fbox{\includegraphics[width=0.19\textwidth]{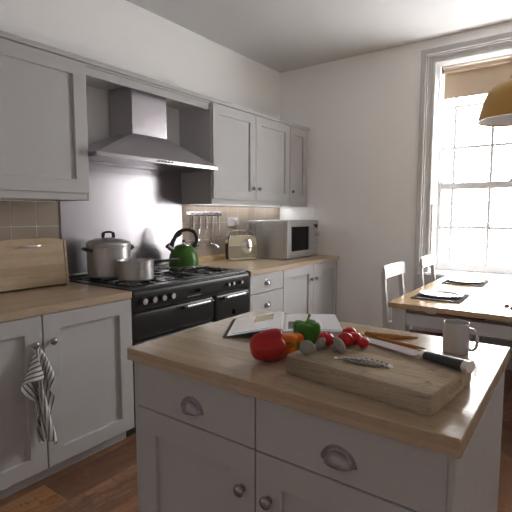}}
        \end{tabular} 
        &
        
        \begin{tabular}{c}
            \fbox{\includegraphics[width=0.13\textwidth]{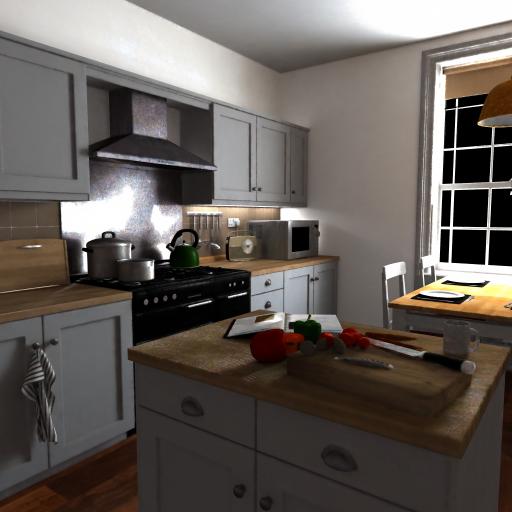}} \\
            \fbox{\includegraphics[width=0.06\textwidth]{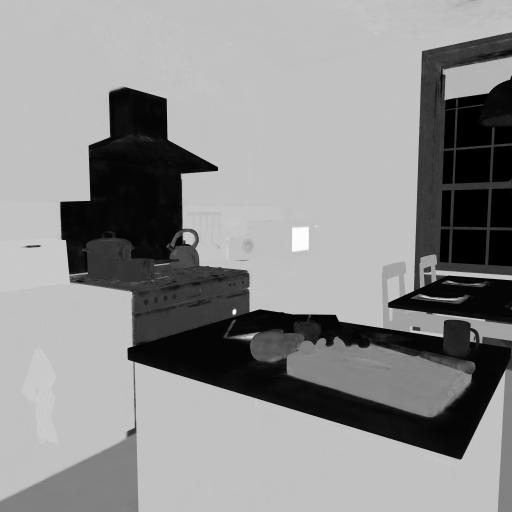}} 
            \fbox{\includegraphics[width=0.06\textwidth]{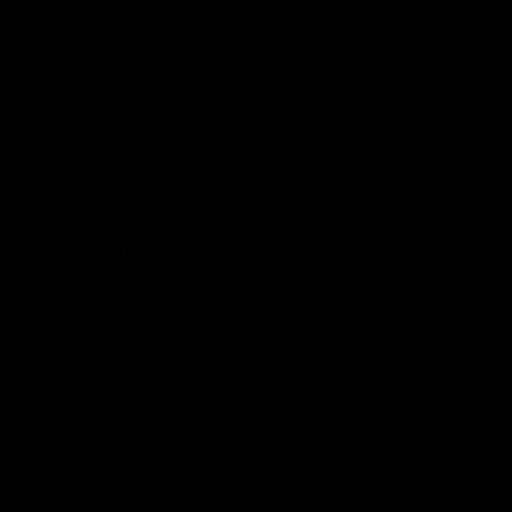}}
        \end{tabular} 
        &

        \begin{tabular}{c}
            \fbox{\includegraphics[width=0.13\textwidth]{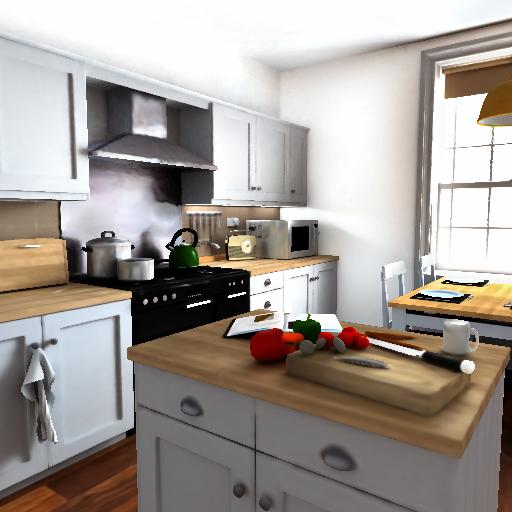}} \\
            \fbox{\includegraphics[width=0.06\textwidth]{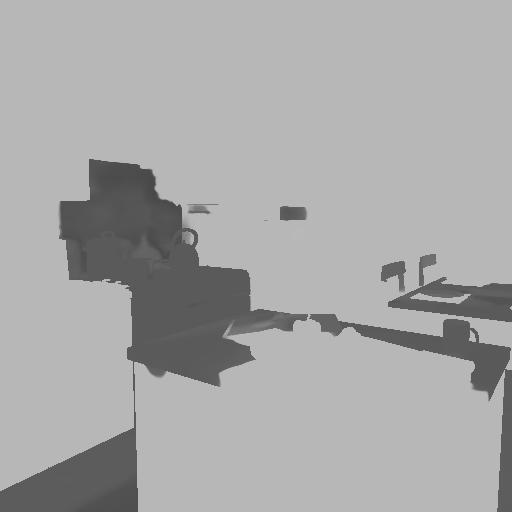}} 
            \fbox{\includegraphics[width=0.06\textwidth]{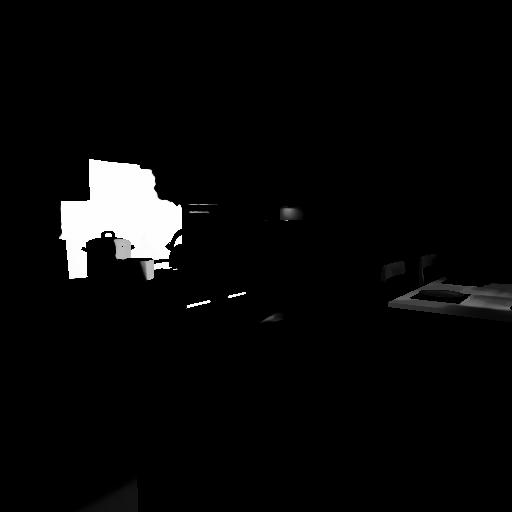}}
        \end{tabular} 
        &

        \begin{tabular}{c}
            \fbox{\includegraphics[width=0.13\textwidth]{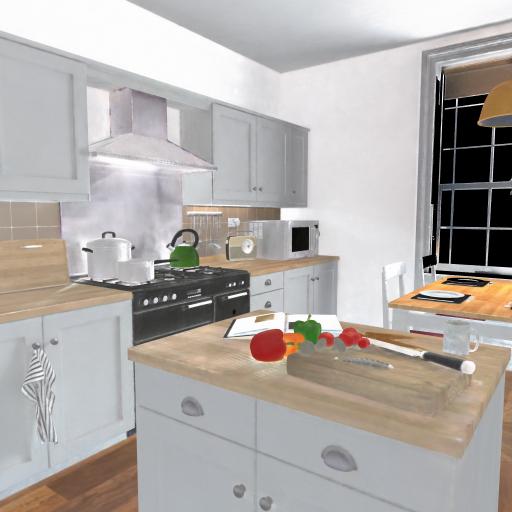}} \\
            \fbox{\includegraphics[width=0.06\textwidth]{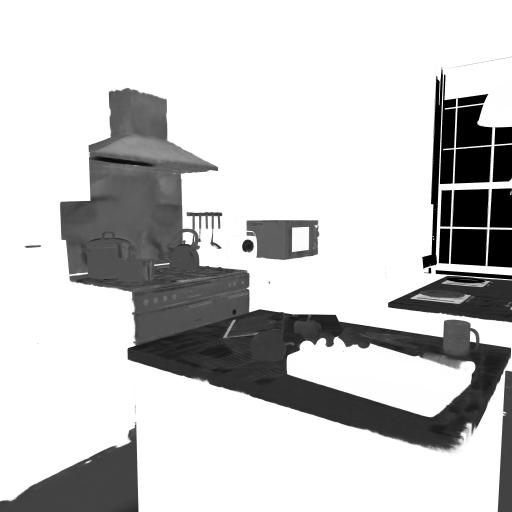}} 
            \fbox{\includegraphics[width=0.06\textwidth]{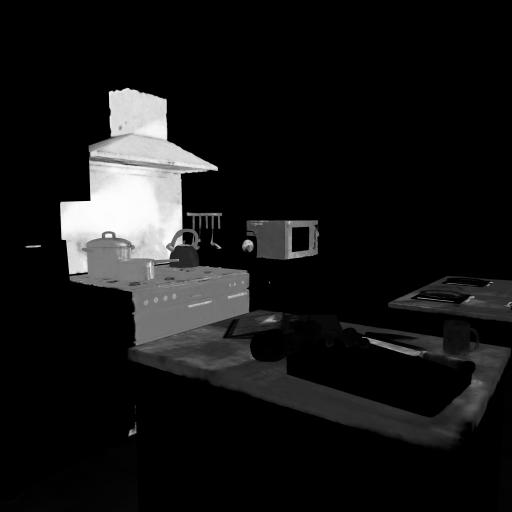}}
        \end{tabular} 
        &

        \begin{tabular}{c}
            \fbox{\includegraphics[width=0.13\textwidth]{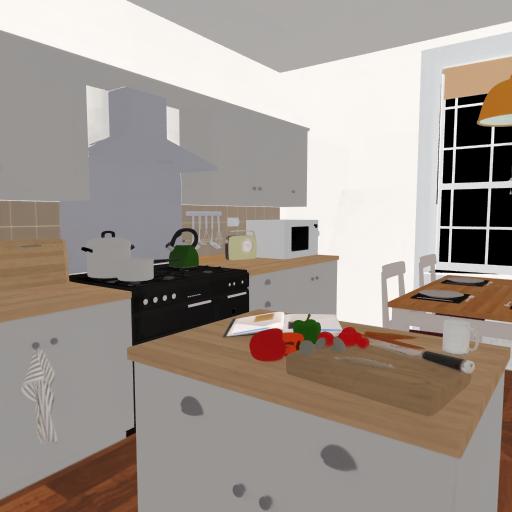}} \\
            \fbox{\includegraphics[width=0.06\textwidth]{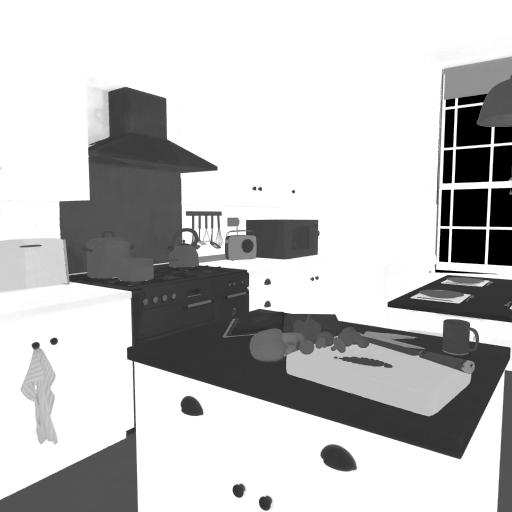}} 
            \fbox{\includegraphics[width=0.06\textwidth]{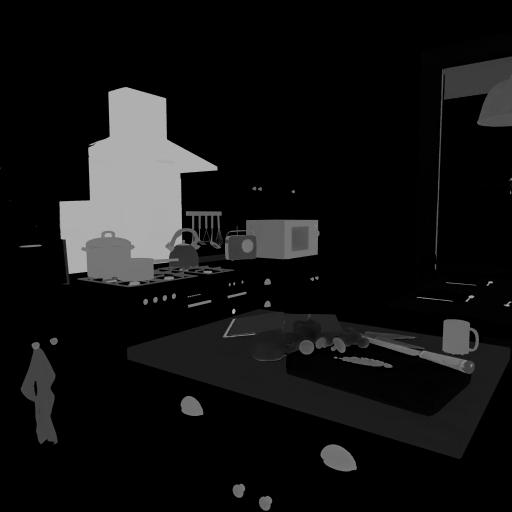}}
        \end{tabular} 
        &

        \begin{tabular}{c}
            \fbox{\includegraphics[width=0.13\textwidth]{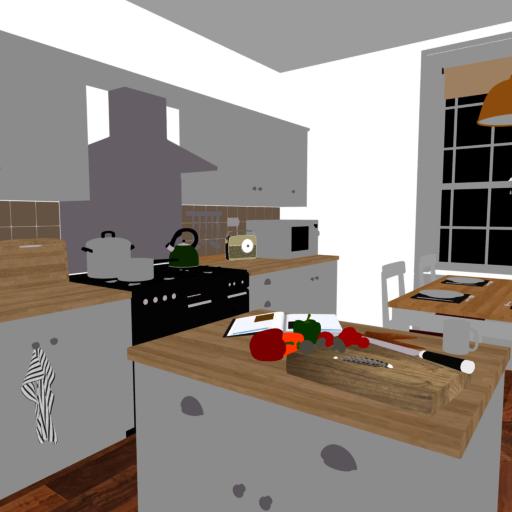}} \\
            \fbox{\includegraphics[width=0.06\textwidth]{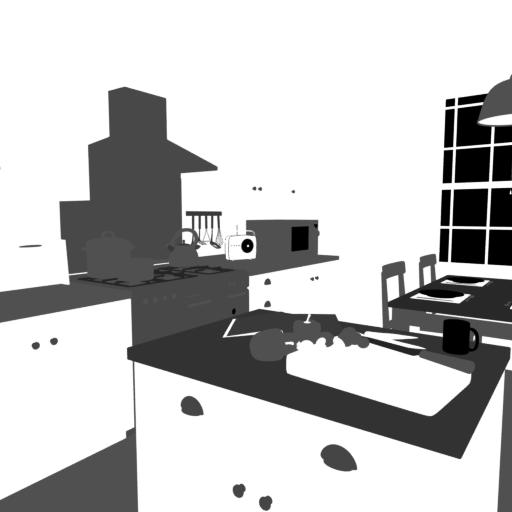}} 
            \fbox{\includegraphics[width=0.06\textwidth]{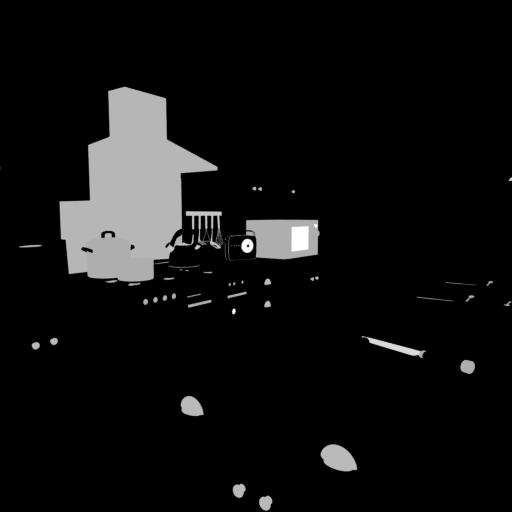}}
        \end{tabular} 
         \\

        \begin{tabular}{c}
            \fbox{\includegraphics[width=0.19\textwidth]{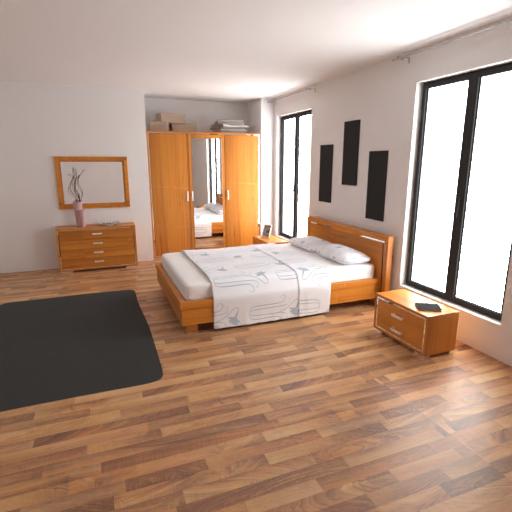}}
        \end{tabular} 
        &
        
        \begin{tabular}{c}
            \fbox{\includegraphics[width=0.13\textwidth]{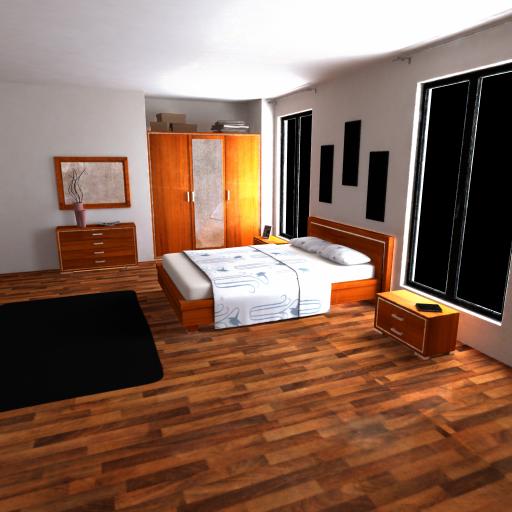}} \\
            \fbox{\includegraphics[width=0.06\textwidth]{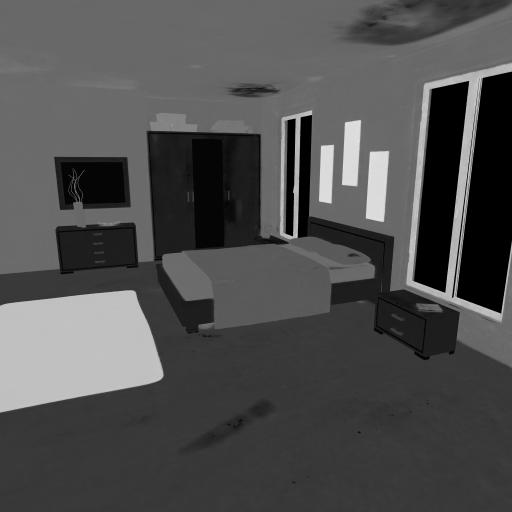}} 
            \fbox{\includegraphics[width=0.06\textwidth]{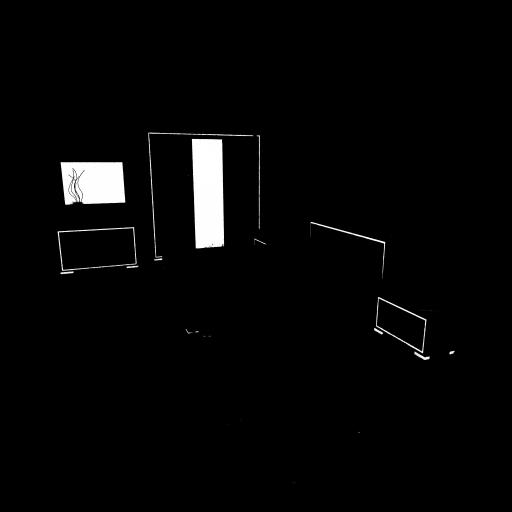}}
        \end{tabular} 
        &

        \begin{tabular}{c}
            \fbox{\includegraphics[width=0.13\textwidth]{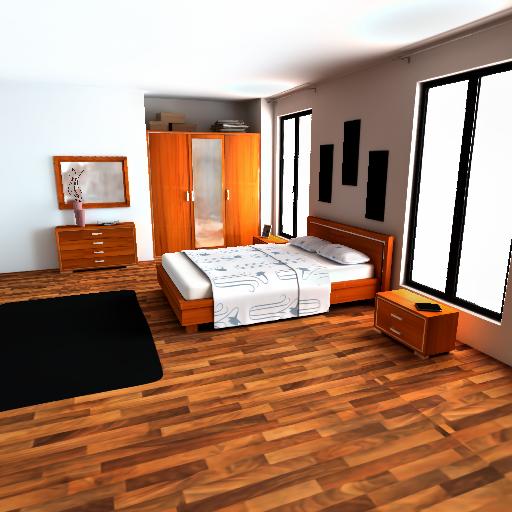}} \\
            \fbox{\includegraphics[width=0.06\textwidth]{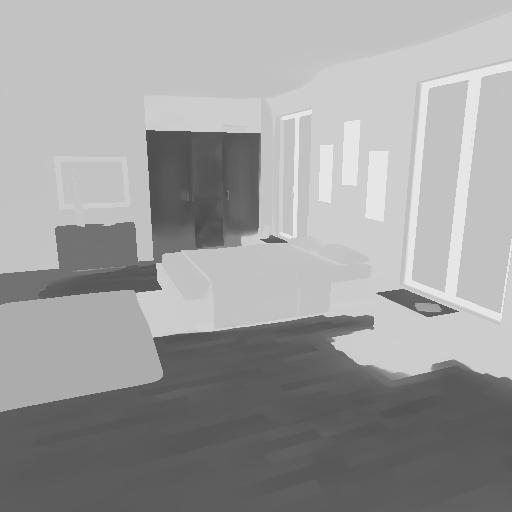}} 
            \fbox{\includegraphics[width=0.06\textwidth]{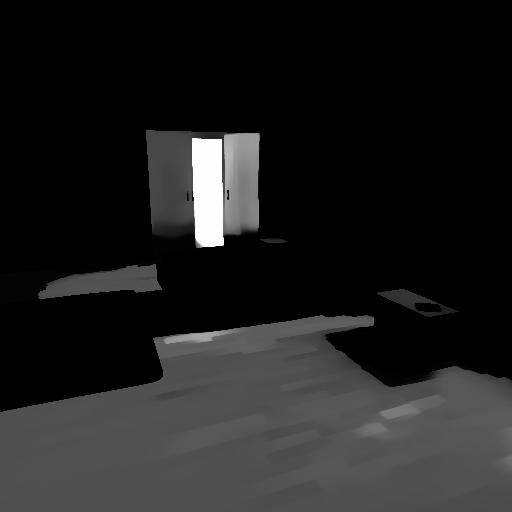}}
        \end{tabular} 
        &

        \begin{tabular}{c}
            \fbox{\includegraphics[width=0.13\textwidth]{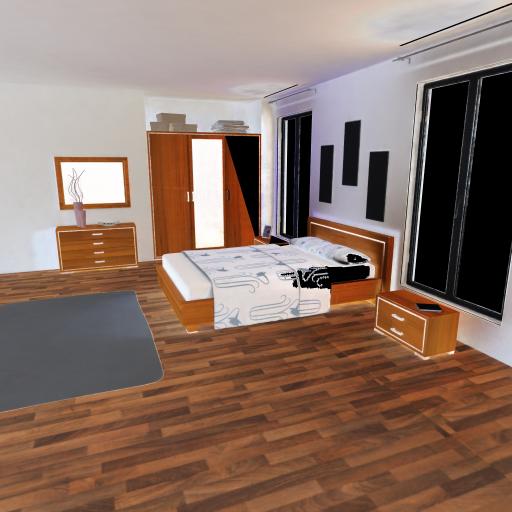}} \\
            \fbox{\includegraphics[width=0.06\textwidth]{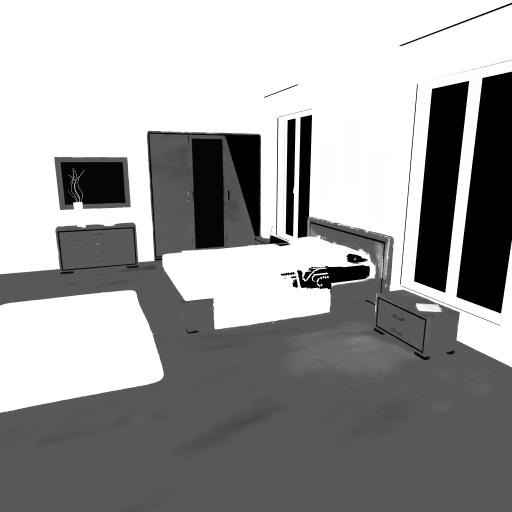}} 
            \fbox{\includegraphics[width=0.06\textwidth]{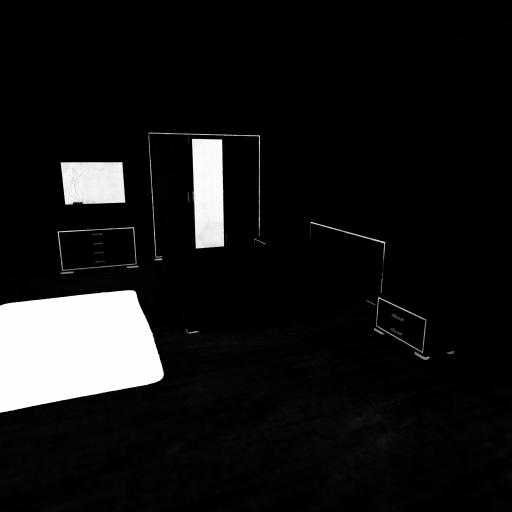}}
        \end{tabular} 
        &

        \begin{tabular}{c}
            \fbox{\includegraphics[width=0.13\textwidth]{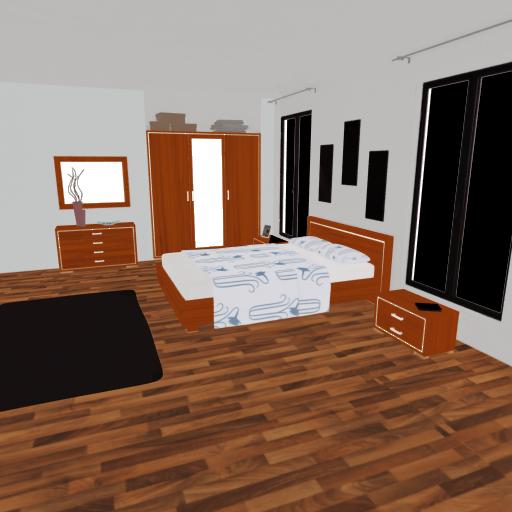}} \\
            \fbox{\includegraphics[width=0.06\textwidth]{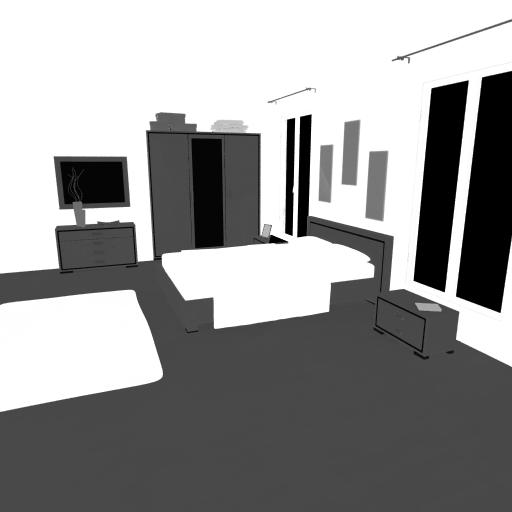}} 
            \fbox{\includegraphics[width=0.06\textwidth]{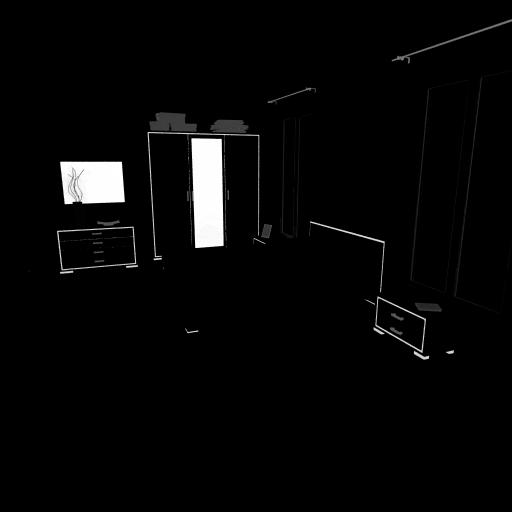}}
        \end{tabular} 
        &

        \begin{tabular}{c}
            \fbox{\includegraphics[width=0.13\textwidth]{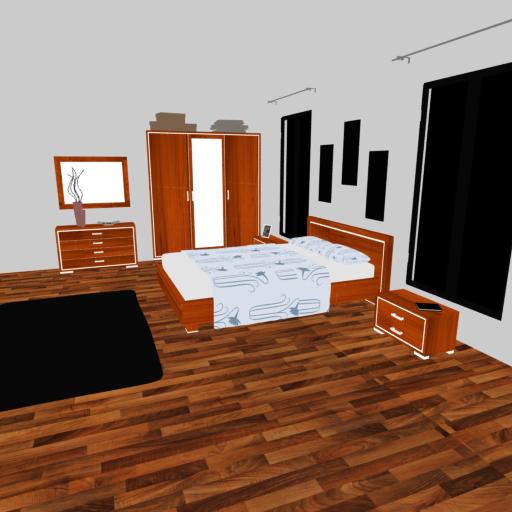}} \\
            \fbox{\includegraphics[width=0.06\textwidth]{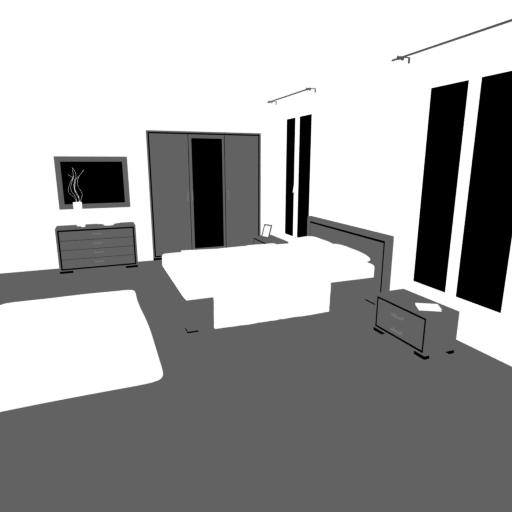}}
            \fbox{\includegraphics[width=0.06\textwidth]{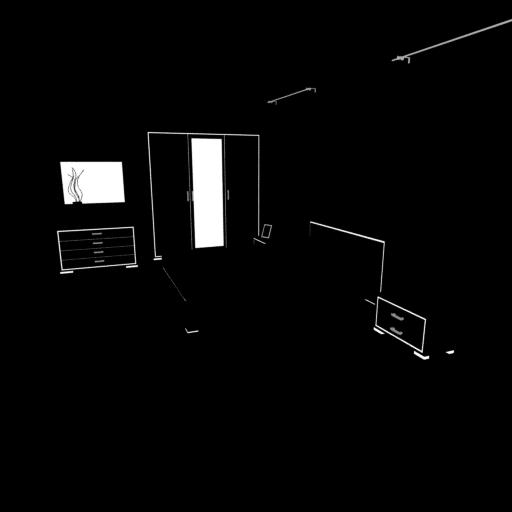}}
        \end{tabular} 
         \\[-4pt]

        \smash{\scriptsize RGB} &
        \smash{\scriptsize FIPT \cite{FIPT}} &
        \smash{\scriptsize NeILF++ \cite{NeilfPP}} &
        \smash{\scriptsize IRIS \cite{IRIS}} &
        \smash{\scriptsize IIF (Ours)} &
        \smash{\scriptsize GT}
    \end{tabular}}
    % \vspace{-10pt}
    \caption{\textbf{Synthetic comparisons.}
    Previous inverse-rendering approaches often struggle to separate shading from reflectance due to the noise inherent in light-transport estimation, resulting in baked-in illumination and biased specular parameters~\cite{FIPT, NeilfPP}.
    In contrast, our method restricts path-traced optimization to a low-dimensional set of per-object transformations, yielding a more constrained and stable objective. 
    This leads to clean, sharp, and physically consistent material predictions.
    }
    \label{fig:exp:synthetic_comparisons}
\end{figure*}

% Huge table, left: conditioning (text + component, in and out of domain), right: all the components
\begin{figure*}[t]
    \centering
    \setlength\tabcolsep{1.25pt}
    \resizebox{\textwidth}{!}{
    \fboxsep=0pt
    \begin{tabular}{c|ccc|c}    
        \begin{tabular}{c}
            \fbox{\includegraphics[width=0.19\textwidth]{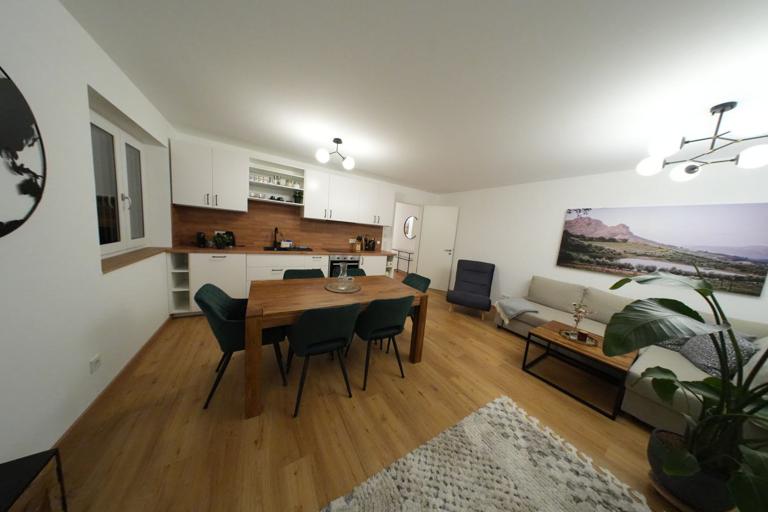}}
        \end{tabular} 
        &
        
        \begin{tabular}{c}
            \fbox{\includegraphics[width=0.13\textwidth]{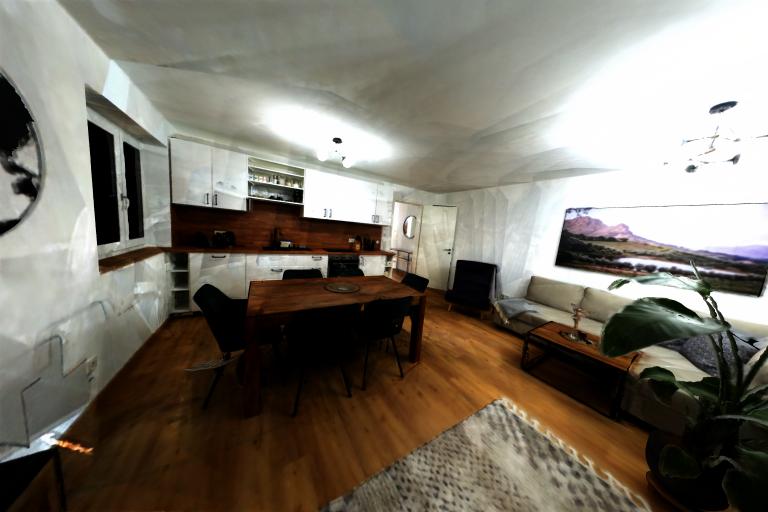}} \\
            \fbox{\includegraphics[width=0.06\textwidth]{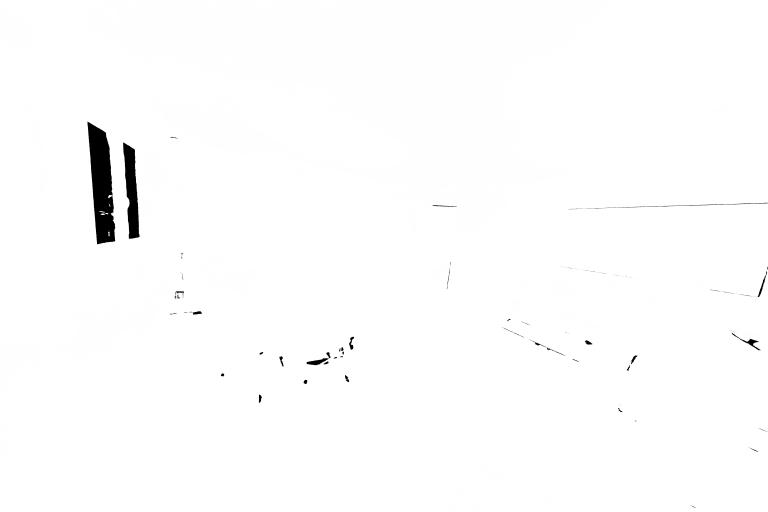}} 
            \fbox{\includegraphics[width=0.06\textwidth]{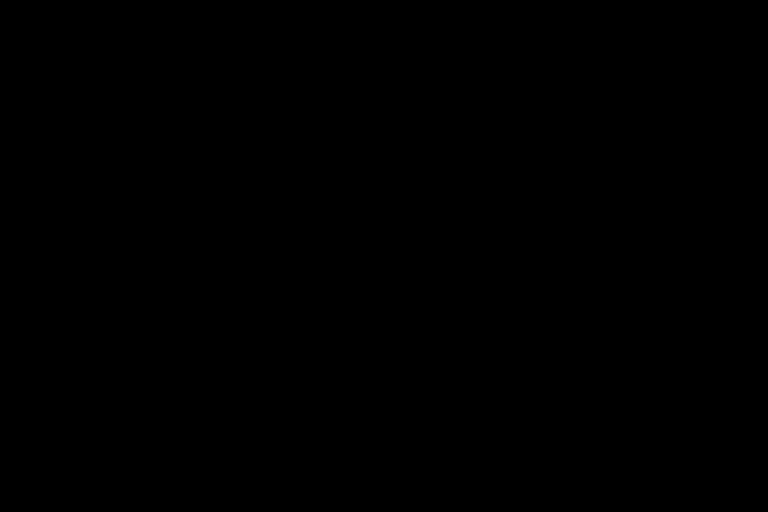}}
        \end{tabular} 
        &

        \begin{tabular}{c}
            \fbox{\includegraphics[width=0.13\textwidth]{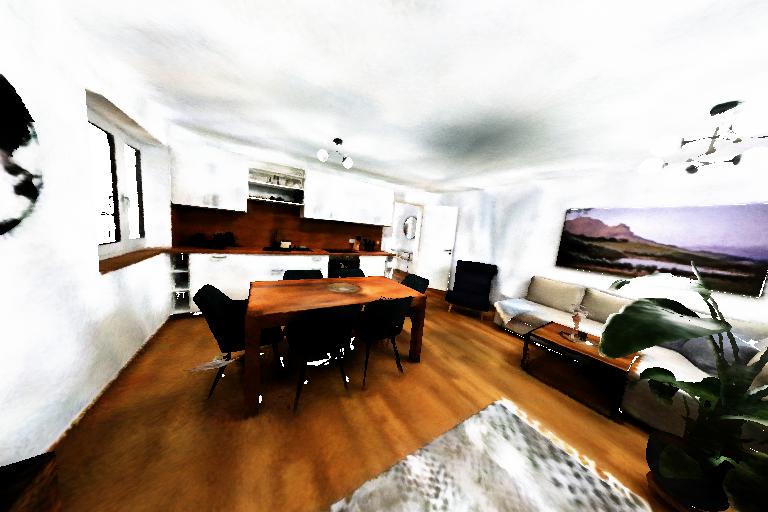}} \\
            \fbox{\includegraphics[width=0.06\textwidth]{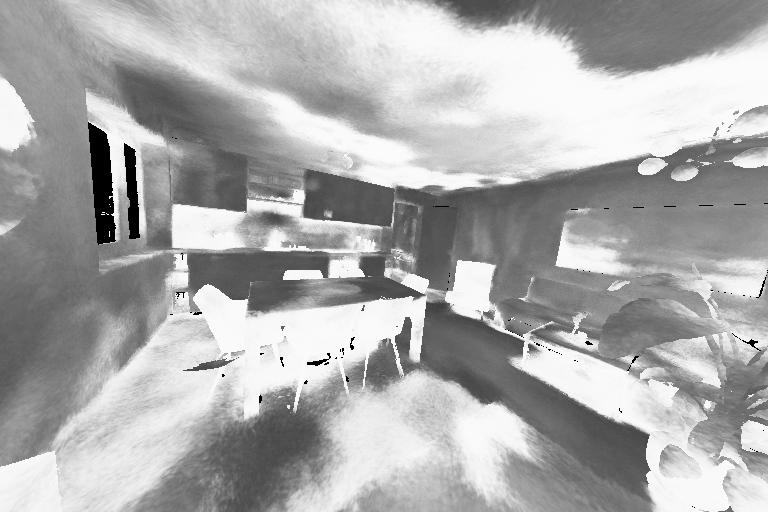}} 
            \fbox{\includegraphics[width=0.06\textwidth]{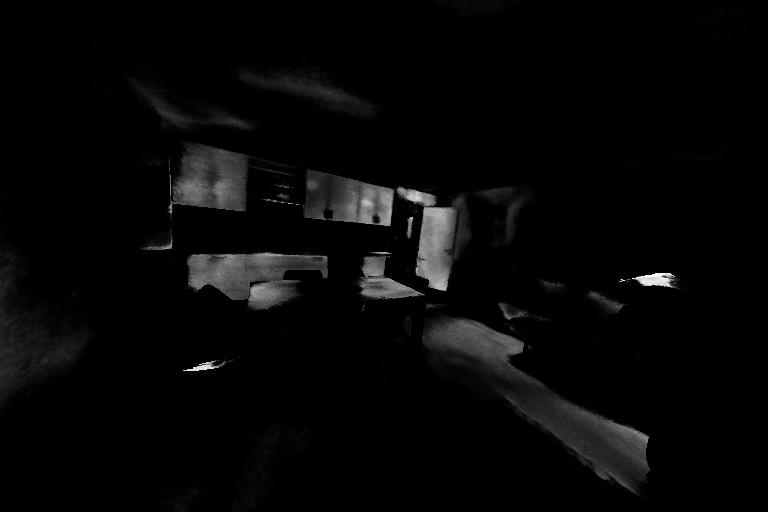}}
        \end{tabular} 
        &

        \begin{tabular}{c}
            \fbox{\includegraphics[width=0.13\textwidth]{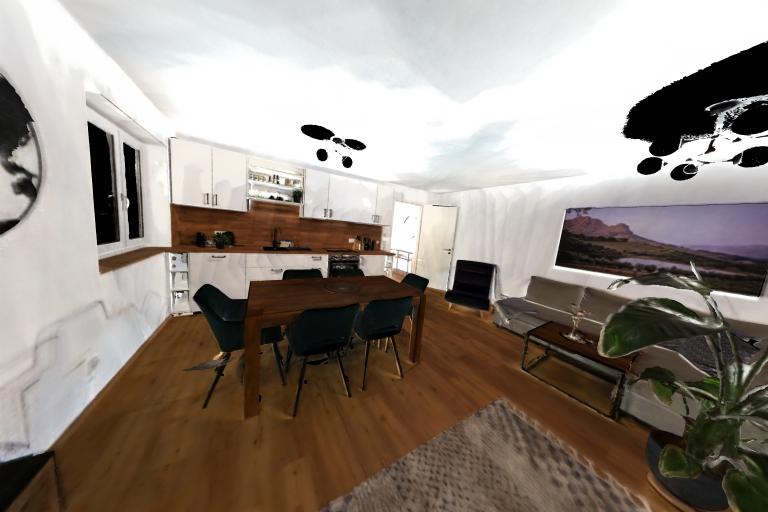}} \\
            \fbox{\includegraphics[width=0.06\textwidth]{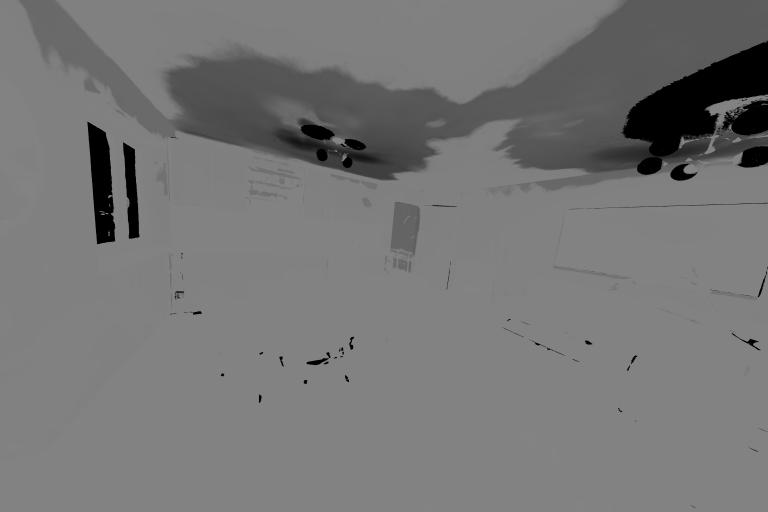}} 
            \fbox{\includegraphics[width=0.06\textwidth]{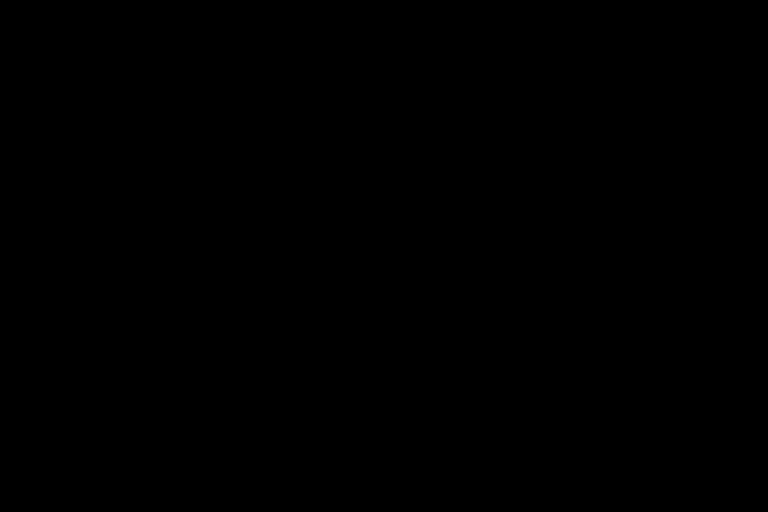}}
        \end{tabular} 
        &
        
        \begin{tabular}{c}
            \fbox{\includegraphics[width=0.13\textwidth]{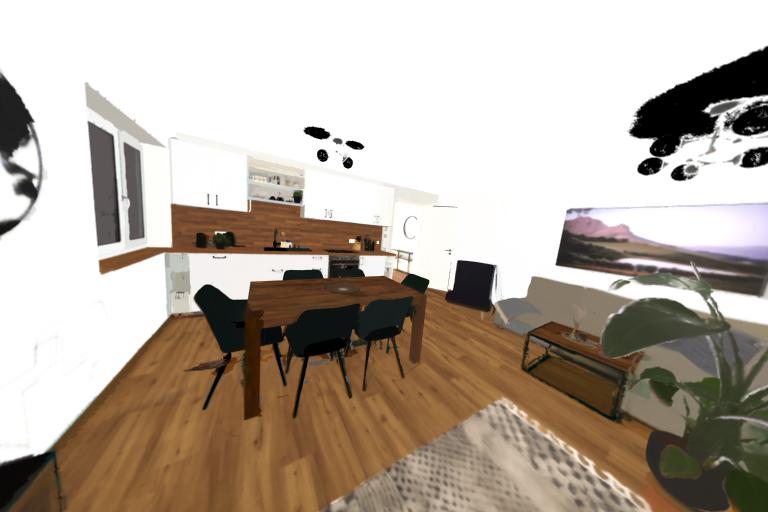}} \\
            \fbox{\includegraphics[width=0.06\textwidth]{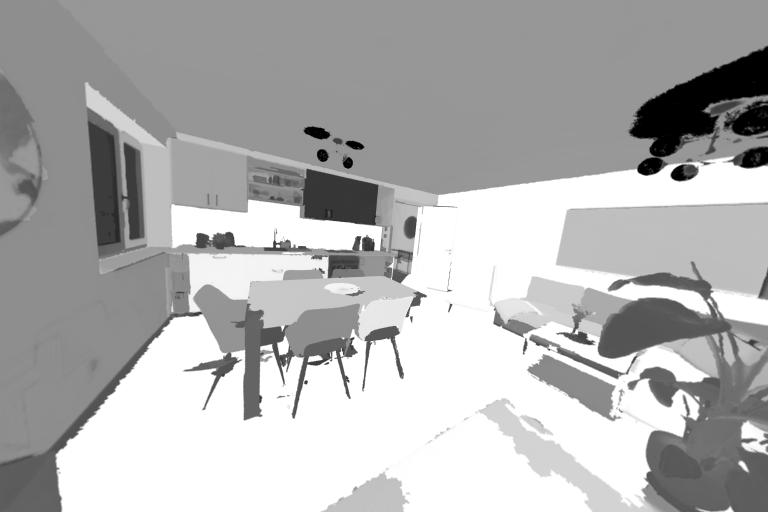}}
            \fbox{\includegraphics[width=0.06\textwidth]{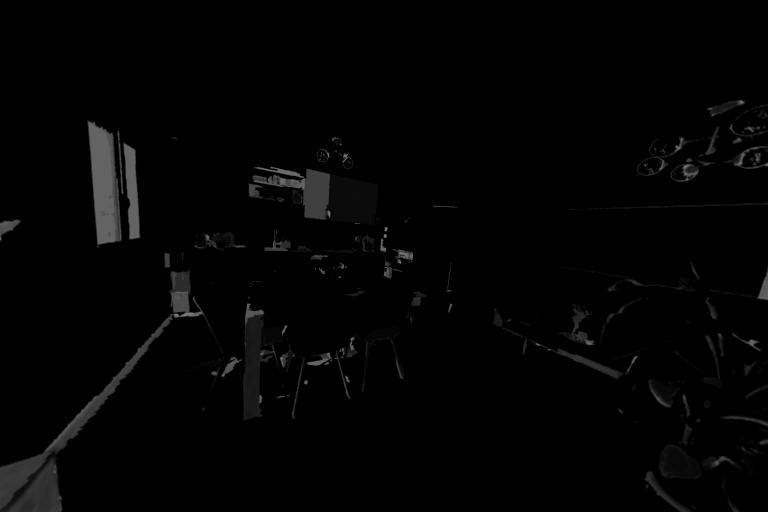}}
        \end{tabular} 
         \\

        \begin{tabular}{c}
            \fbox{\includegraphics[width=0.19\textwidth]{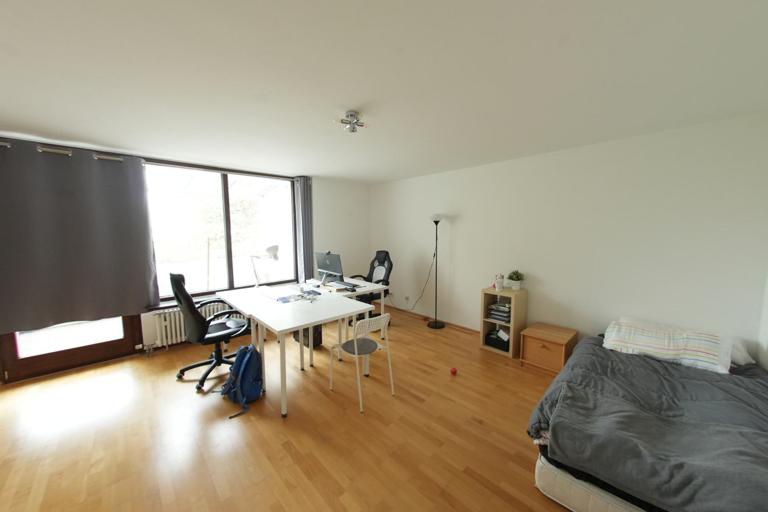}}
        \end{tabular} 
        &
        
        \begin{tabular}{c}
            \fbox{\includegraphics[width=0.13\textwidth]{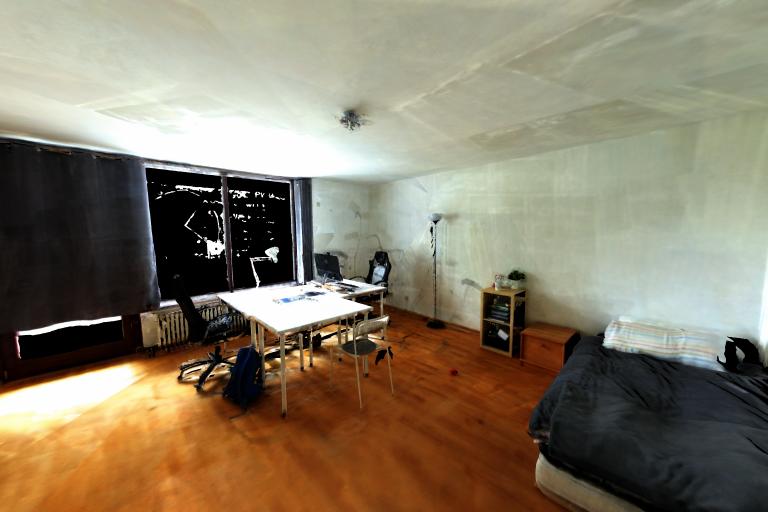}} \\
            \fbox{\includegraphics[width=0.06\textwidth]{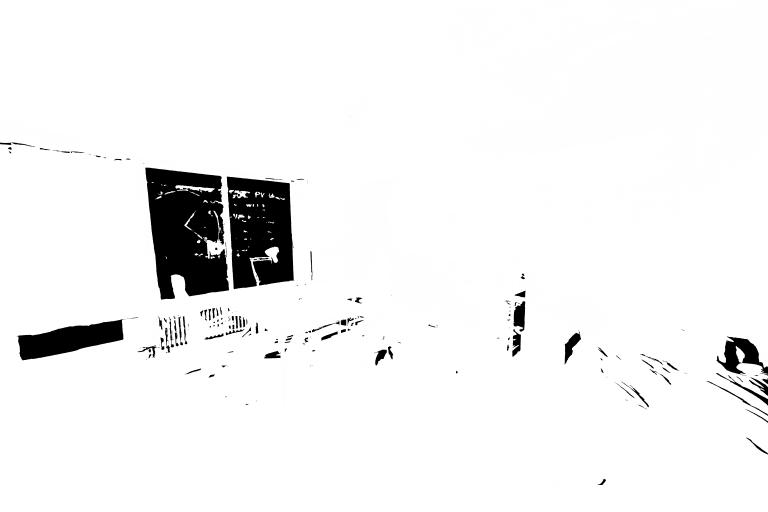}} 
            \fbox{\includegraphics[width=0.06\textwidth]{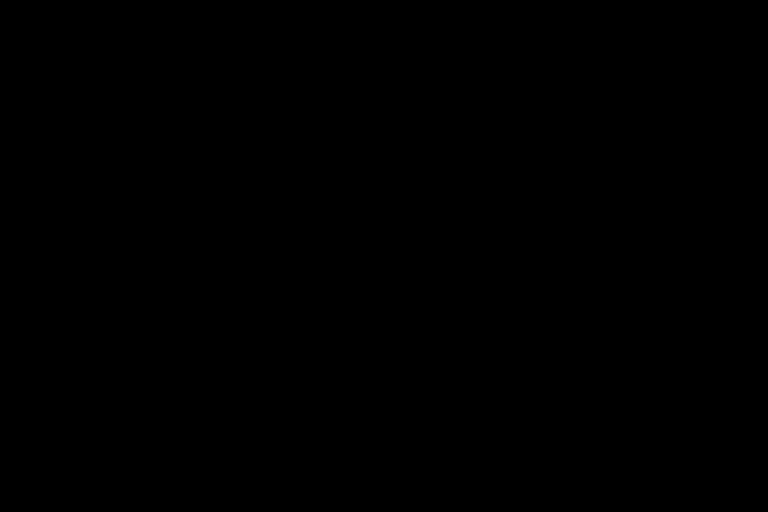}}
        \end{tabular} 
        &
        
        \begin{tabular}{c}
            \fbox{\includegraphics[width=0.13\textwidth]{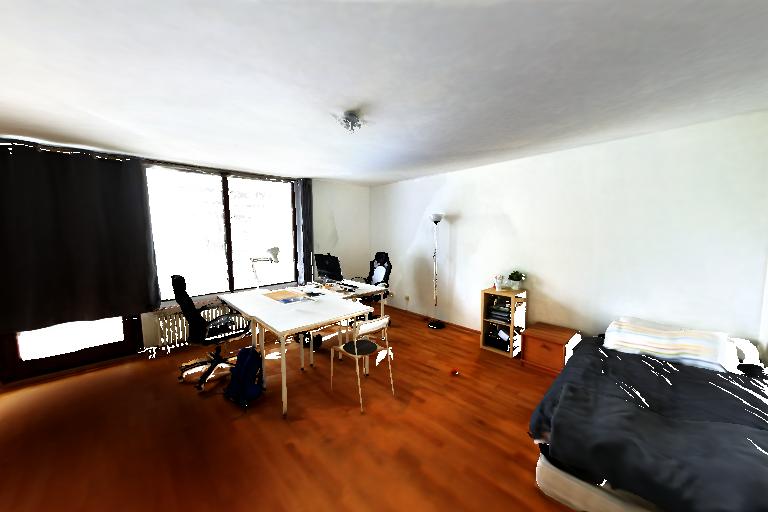}} \\
            \fbox{\includegraphics[width=0.06\textwidth]{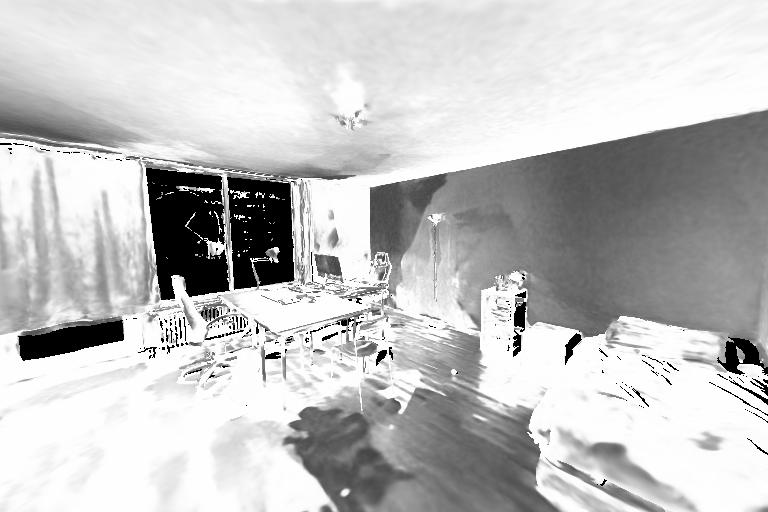}} 
            \fbox{\includegraphics[width=0.06\textwidth]{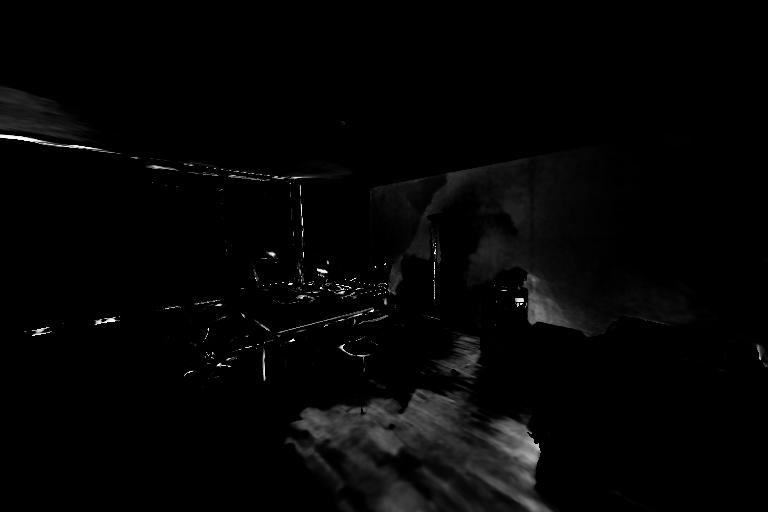}}
        \end{tabular} 
        &
        
        \begin{tabular}{c}
            \fbox{\includegraphics[width=0.13\textwidth]{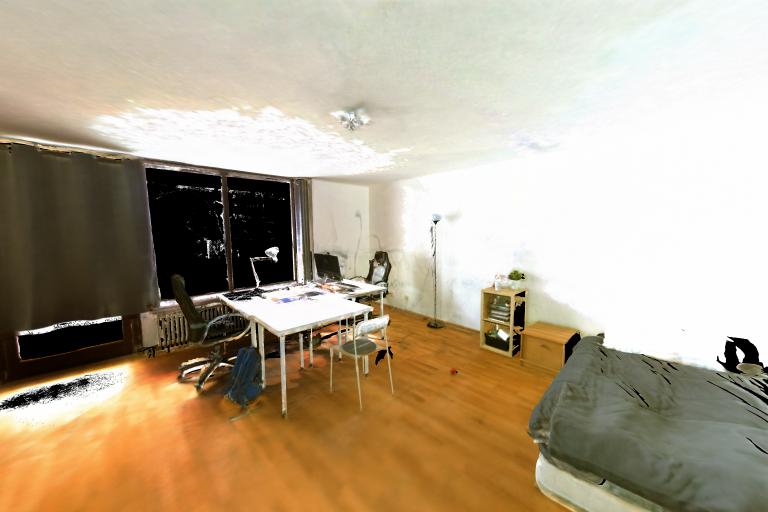}} \\
            \fbox{\includegraphics[width=0.06\textwidth]{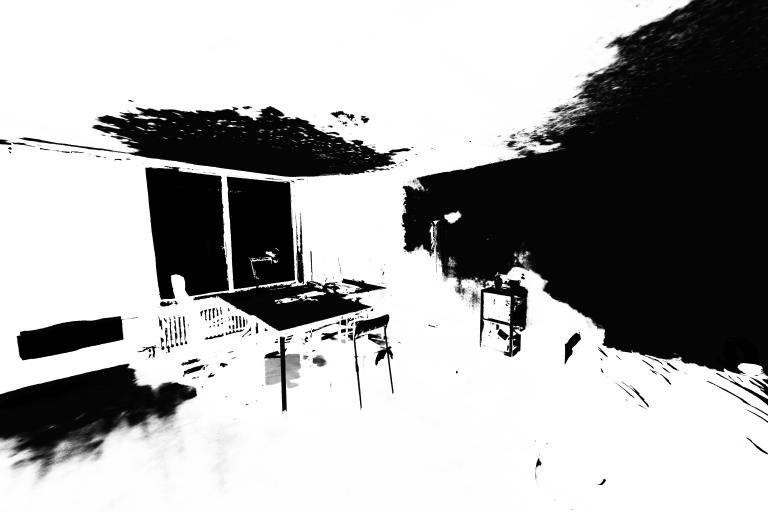}} 
            \fbox{\includegraphics[width=0.06\textwidth]{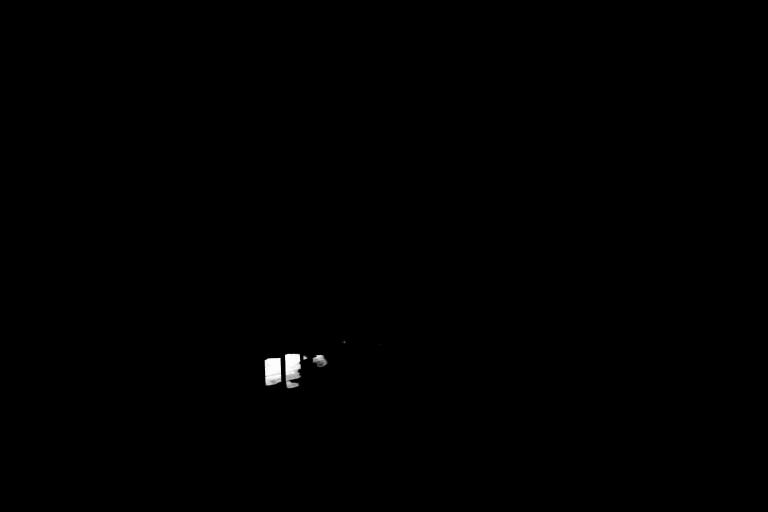}}
        \end{tabular} 
        &
        
        \begin{tabular}{c}
            \fbox{\includegraphics[width=0.13\textwidth]{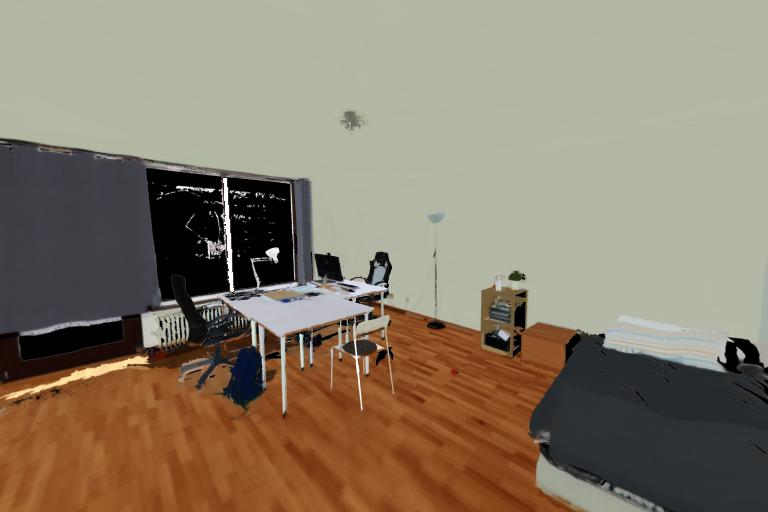}} \\
            \fbox{\includegraphics[width=0.06\textwidth]{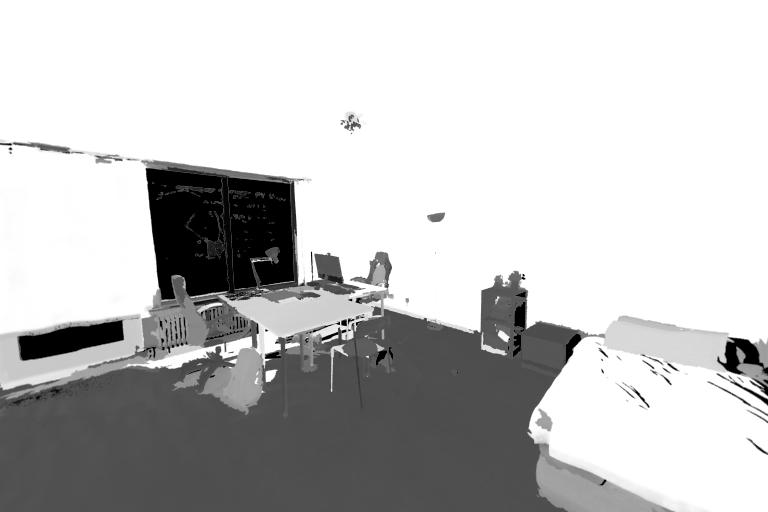}}
            \fbox{\includegraphics[width=0.06\textwidth]{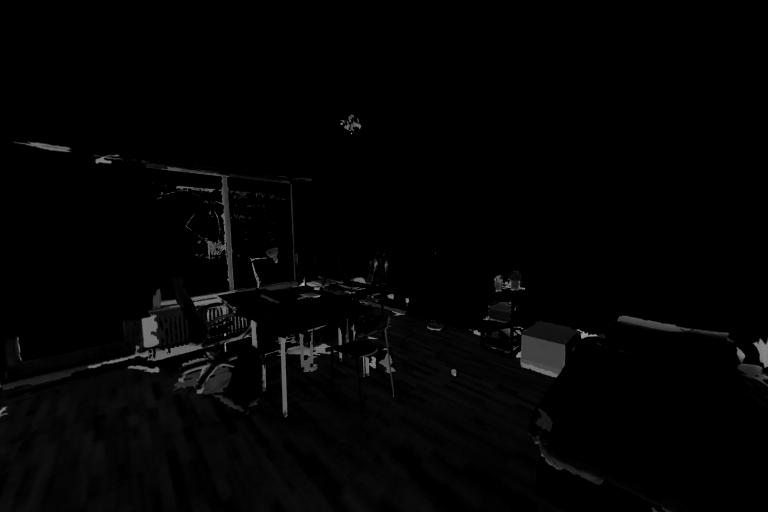}}
        \end{tabular} 
         \\[-4pt]

        \smash{\scriptsize Input} &
        \smash{\scriptsize FIPT \cite{FIPT}} &
        \smash{\scriptsize NeILF++ \cite{NeilfPP}} &
        \smash{\scriptsize IRIS \cite{IRIS}} &
        \smash{\scriptsize IIF (Ours)}
    \end{tabular}}
    % \vspace{-10pt}
    \caption{\textbf{Real-world comparisons.} 
    Incomplete and nosiy geometry poses a challenge for previous methods, yielding projected contour artifacts in the final textures (e.g., chair borders visible on the walls on the top).
    Our method uses a much more constrained objective able to maintain clean and consistent predictions. 
    Please see the supplemental for animated flythroughs.
    }
    \label{fig:exp:real_comparisons}
\end{figure*}

We compare our method against the recent inverse rendering methods using inter-reflection constraints (Neilf++ \cite{NeilfPP}), inverse path tracing (FIPT \cite{FIPT}) and using pre-trained priors (IRIS \cite{IRIS}). 
For all baselines, we use their released implementations and retrain/optimize on our datasets for fair comparisons. 
For IRIS \cite{IRIS}, we used our RGBX \cite{RGBX} predictions for their regularization to ensure consistency with our method. 
For the qualitative and quantitative evaluations, we ignore the BRDF parameters for the emitters, and the albedo for perfectly specular surfaces. 

We provide qualitative comparisons on both synthetic (\Cref{fig:exp:synthetic_comparisons}) and real scenes (\Cref{fig:exp:real_comparisons}).
Due to the noisy estimation of light transport, earlier methods struggle with decoupling shading from reflectance, leading to strong baked-in lighting and shifted specular parameters \cite{FIPT, NeilfPP}. 
IRIS \cite{IRIS} constrains the optimization by introducing pre-trained prior predictions as regularizer. 
However, they are still optimizing the whole texture via the noisy rendering loss, causing baked-in shading effects. 
In contrast, our method optimizes only for the low-dimensional per-object transformations with path-tracing, while fixing the BRDF network $f_\theta$.
This leads to clean/sharp materials, even on challenging real-world scenes (additional results in the supplemental). 

We provide quantitative comparisons on the synthetic scenes.
First, we render all PBR modalities for all views of the four synthetic datasets. 
Then, we report averaged comparisons against the respective ground-truth in \Cref{tab:exp:comparisons}.
Our method outperforms previous approaches by a high margin and yields competitive results on the sparse metallic maps.

\subsection{Ablations}
\label{sec:experiments:ablations}
% Huge table, left: conditioning (text + component, in and out of domain), right: all the components
\begin{figure}
    \centering
    \setlength\tabcolsep{1.25pt}
    \resizebox{\columnwidth}{!}{
    \fboxsep=0pt
    \begin{tabular}{ccc}
        \rotatebox{90}{RGBX \cite{RGBX}} &
        \fbox{\includegraphics[width=0.25\textwidth]{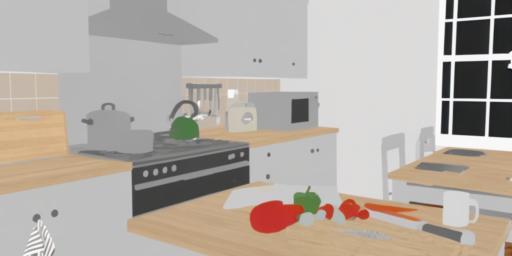}} &
        \fbox{\includegraphics[width=0.25\textwidth]{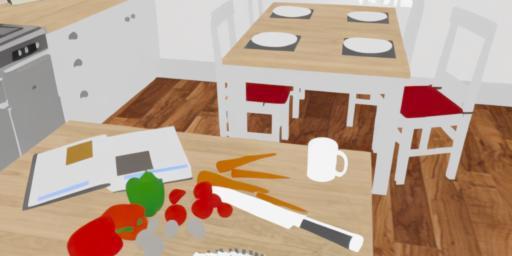}} \\
        
        \rotatebox{90}{Per-Object \cite{IRIS}} &
        \fbox{\includegraphics[width=0.25\textwidth]{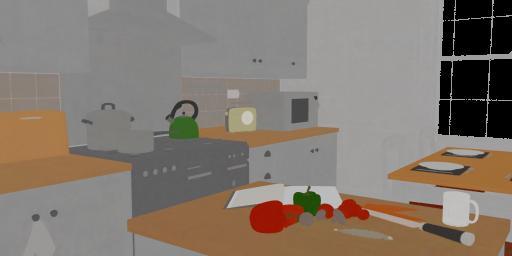}} &
        \fbox{\includegraphics[width=0.25\textwidth]{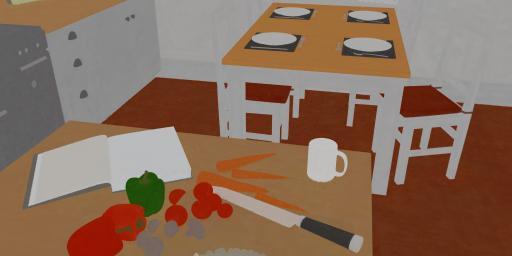}} \\

        \rotatebox{90}{Per-Texel} &
        \fbox{\includegraphics[width=0.25\textwidth]{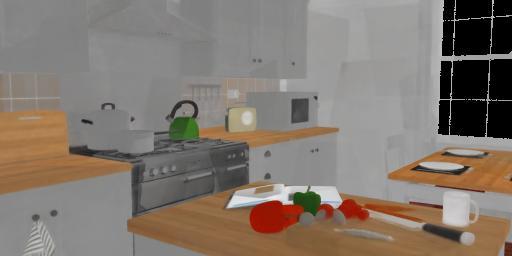}} &
        \fbox{\includegraphics[width=0.25\textwidth]{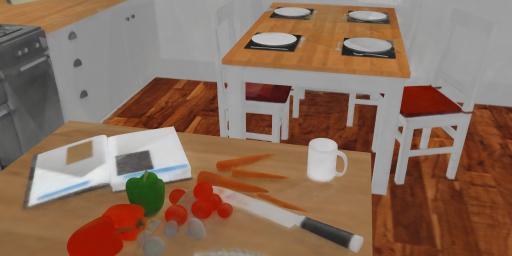}} \\

        \hline

        \rotatebox{90}{w/ Parametric} &
        \begin{tikzpicture}
          \node[inner sep=0pt] (a)
            {\fbox{\includegraphics[width=0.25\textwidth]{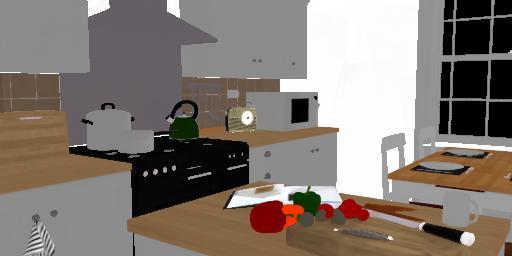}}};
          \begin{scope}
            \clip (a.south west) rectangle (a.north east);
            \node[draw=red, line width=1pt, circle,
                  minimum width=30pt, minimum height=30pt,
                  xshift=-45pt, yshift=-10pt] at (a.center) {};
          \end{scope}
        \end{tikzpicture}
        &
        \begin{tikzpicture}
          \node[inner sep=0pt] (a)
            {\fbox{\includegraphics[width=0.25\textwidth]{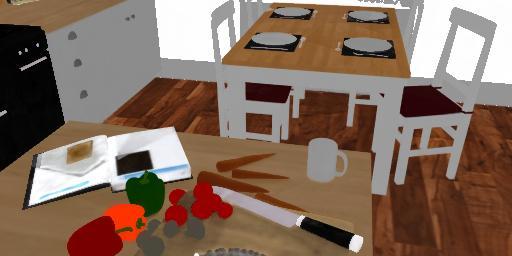}}};
          \begin{scope}
            \clip (a.south west) rectangle (a.north east);
            \node[draw=red, line width=1pt, circle,
                  minimum width=40pt, minimum height=40pt,
                  xshift=10pt, yshift=-10pt] at (a.center) {};
          \end{scope}
        \end{tikzpicture}
        \\

        \rotatebox{90}{Ours} &
        \begin{tikzpicture}
          \node[inner sep=0pt] (a)
            {\fbox{\includegraphics[width=0.25\textwidth]{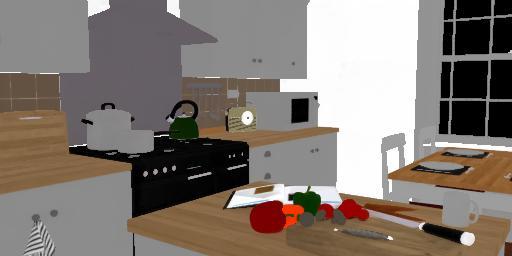}}};
          \begin{scope}
            \clip (a.south west) rectangle (a.north east);
            \node[draw=green, line width=1pt, circle,
                  minimum width=30pt, minimum height=30pt,
                  xshift=-45pt, yshift=-10pt] at (a.center) {};
          \end{scope}
        \end{tikzpicture}
        &
        \begin{tikzpicture}
          \node[inner sep=0pt] (a)
            {\fbox{\includegraphics[width=0.25\textwidth]{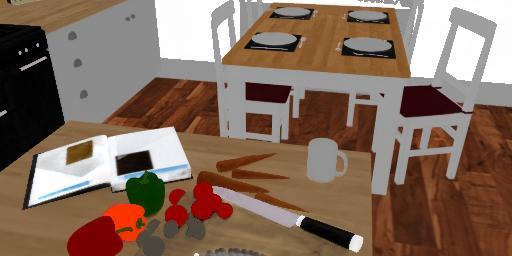}}};
          \begin{scope}
            \clip (a.south west) rectangle (a.north east);
            \node[draw=green, line width=1pt, circle,
                  minimum width=40pt, minimum height=40pt,
                  xshift=10pt, yshift=-10pt] at (a.center) {};
          \end{scope}
        \end{tikzpicture}
        \\

        \hline

        \rotatebox{90}{GT} &
        \fbox{\includegraphics[width=0.25\textwidth]{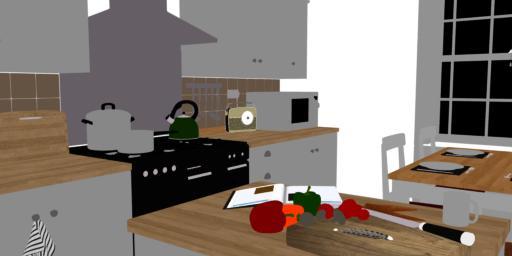}} &
        \fbox{\includegraphics[width=0.25\textwidth]{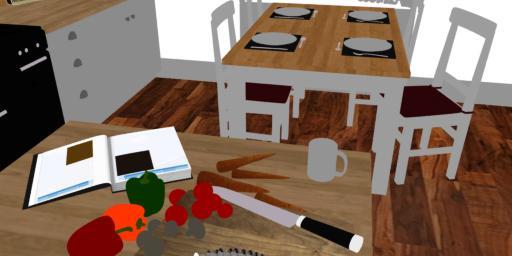}} 
    \end{tabular}
    }
    % \vspace{-10pt}
    
    \caption{\textbf{Cross-view aggregation.} 
    Single-view material estimation can yield detailed, but inconsistent predictions (\cref{fig:method:motivation}), which are physically not well-grounded.
    IRIS \cite{IRIS} uses a per-object aggregation as a material proxy, causing loss of the patterns. 
    Per-texel aggregation could maintain the patterns, but yields texture seams due to the inconsistencies. 
    Our parametric modeling (\cref{sec:method:single_view}) introduces an expressive, but low-dimensional space of possible consistent 3D aggregations. 
    To avoid oversmoothing of local patterns, our distribution matching (\cref{sec:method:cross_view}) aims to use a single best prediction per view, giving more fine-grained details.
    }
    \label{fig:exp:cross_view}

    \vspace{-6pt}
\end{figure}

\mypar{Parametric model complexity}
\begin{figure}
    \centering
    
    \setlength\tabcolsep{0.25pt}
    \resizebox{\columnwidth}{!}{
    \fboxsep=0pt
        \begin{tabular}{cccc}

        \fbox{\includegraphics[width=0.3\columnwidth]{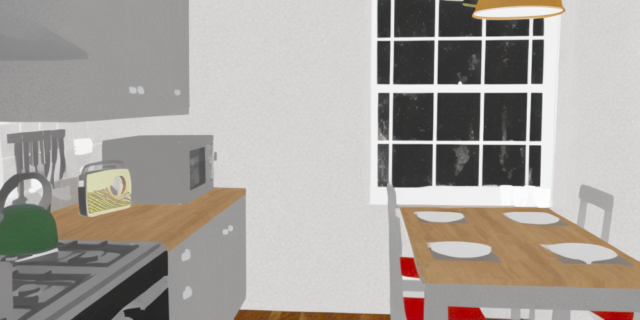}} 
        &
        \fbox{\includegraphics[width=0.3\columnwidth]{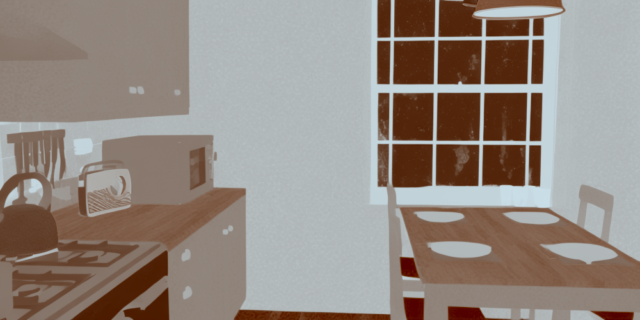}} 
        &
        \fbox{\includegraphics[width=0.3\columnwidth]{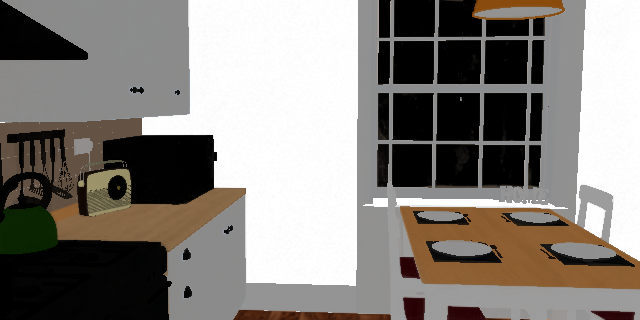}} 
        &
        \fbox{\includegraphics[width=0.3\columnwidth]{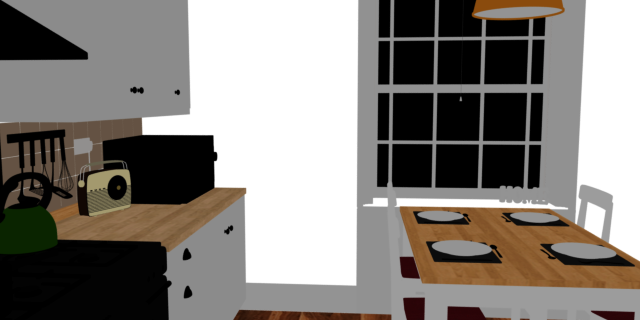}}
        \\

        \fbox{\includegraphics[width=0.3\columnwidth]{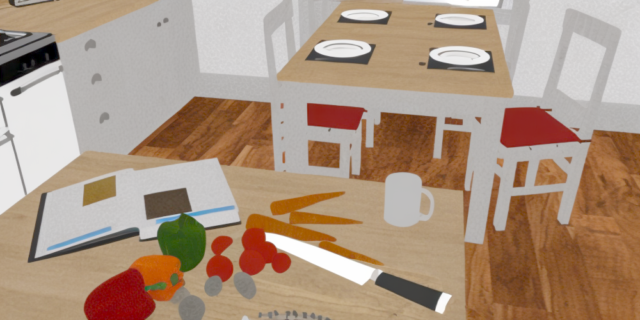}} 
        &
        \fbox{\includegraphics[width=0.3\columnwidth]{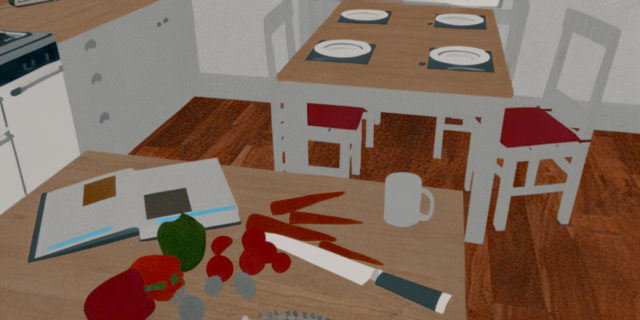}} 
        &
        \fbox{\includegraphics[width=0.3\columnwidth]{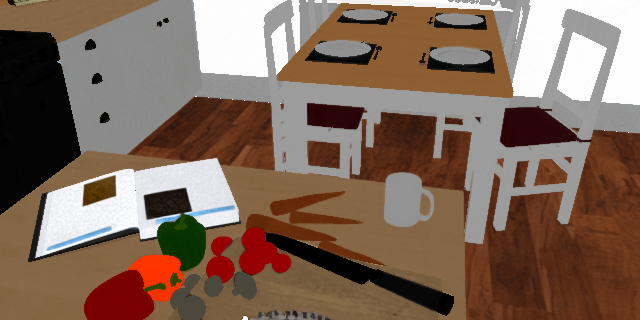}}
        &
        \fbox{\includegraphics[width=0.3\columnwidth]{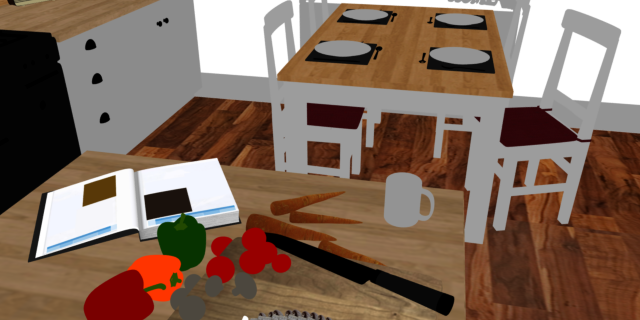}}
        \\

        RGBX \cite{RGBX} &
        Image-level &
        Object-level (Ours) &
        GT 
    \end{tabular}}
    \vspace{-5pt}
    \caption{\textbf{Parametric Model}. 
    We visualize the expressivity of different parametric model choices.
    Image-level fitting cannot correct erroneous relative reflectances between objects, while optimizing object segments independently provides significantly higher expressivity.
    }
    \label{fig:method:object_fitting}
\end{figure}
We study the importance of the parametric model in \cref{fig:method:object_fitting} by comparing our \textit{per-image-per-object} model against a simpler \textit{per-image} model that does not require object segmentations.
While this simpler model also yields a stable optimization, it is not expressive enough and underfits (e.g., \cref{fig:method:object_fitting} shows an average color tone across objects).
This hinders finding consistent 3D PBR textures in the subsequent optimization, leading to oversmoothed results.
% More expressive models are prone to overfitting to noisy gradients.
% We additionally show the need of a parametric model for the aggregation in \cref{sec:method:cross_view}, which is essential to achieve consistent aggregation. 
In contrast, our choice of \textit{per-image-per-object} linear function captures detailed textures.

% \mypar{Parametric model vs Direct texture optimization}
% Direct optimization exhibits severe instability due to noisy path-tracing gradients, often converging to noisy solutions with baked-in lighting. 
% In contrast, our parametric model provides a compact and well-conditioned optimization space that leads to superior reconstruction quality.

\mypar{Cross-view aggregation}
We ablate our design choices on the aggregation of the single-view predictions (\Cref{fig:exp:cross_view}, \Cref{tab:exp:ablations}).
The single-view predictions of RGBX \cite{RGBX} provide a good prior over the patterns, but cannot be directly used for 3D texturing (\Cref{fig:method:motivation}b).
Per-object single color aggregation as in \cite{IRIS, IPT} can make the predictions multi-view consistent, but removes a lot of details. 
Per-texel aggregation enables 3D consistent details, but suffers from texture seams caused by inconsistent predictions. 
We show that introducing our parametric texture model (\Cref{eq:param-brdf}) drastically improves the expressivity of the prior texture. 
% Since this model introduces free parameters, which are optimized during the second stage inverse path tracing, we use visualizations using the optimal parameters as an upper bound. 
Additionally, using our Laplace distribution matching optimization (\cref{sec:method:cross_view}) can further improve the quality by avoiding oversmoothing of local patterns.

\begin{table}

    \centering\setlength{\tabcolsep}{4pt}
    \resizebox{1.0\linewidth}{!}{%
    \begin{tabular}{l|ccc|c|c}
    \toprule
        & \multicolumn{3}{c}{Albedo} & \multicolumn{1}{c}{Rough} & \multicolumn{1}{c}{Metal}\\
        \#Preds & PSNR $\uparrow$ & SSIM $\uparrow$ & LPIPS $\downarrow$ & L2 $\downarrow$ & L2 $\downarrow$ \\
    \midrule
     RGBX~\cite{RGBX} & 13.11 & 0.787 & 0.228 & 0.187 & 0.306 \\
     \hline
     
     Per-Object Mean~\cite{IRIS} & 13.21 & 0.641 & 0.563 & 0.169 & 0.307 \\
     Per-Texel Mean & 13.43 & 0.753 & 0.42 & 0.170 & 0.308 \\

     \hline
     
     w/ Parametric (\cref{sec:method:single_view}) & 29.53 & 0.909 & 0.176 & 8.16e-4 & \textbf{1.36e-4} \\
     Ours full (\cref{sec:method:cross_view}) & \textbf{30.79} & \textbf{0.931} & \textbf{0.160} & \textbf{7.86e-4} & \textbf{1.34e-4} \\
       
    \bottomrule
    \end{tabular}%
    }
    \vspace{-3pt}
    \caption{\textbf{Effect of our aggregation.} 
    We evaluate the expressivity of different aggregations of single-view RGBX \cite{RGBX} predictions on the synthetic scenes. 
    Per-object and per-texel aggregation improve consistency and PSNR but fail to preserve fine patterns (SSIM, LPIPS).
    Our parametric formulation can drastically improve the quality, while maintaining low parameter count. 
    Our full method using distribution matching (\cref{sec:method:cross_view}) aims to use a single prediction per image to avoid oversmoothing, yielding improved quality. 
    }
    \label{tab:exp:ablations}
\end{table}

\begin{figure*}
    \centering
    \setlength\tabcolsep{1.25pt}
    \resizebox{\textwidth}{!}{
    \fboxsep=0pt
    
    \begin{tabular}{ccc|ccc}    
        \begin{tikzpicture}[every node/.style={anchor=north west,inner sep=0pt},x=1pt, y=-1pt,]  
             \node (fig1) at (0,0)
               {\fbox{\includegraphics[width=0.25\textwidth]{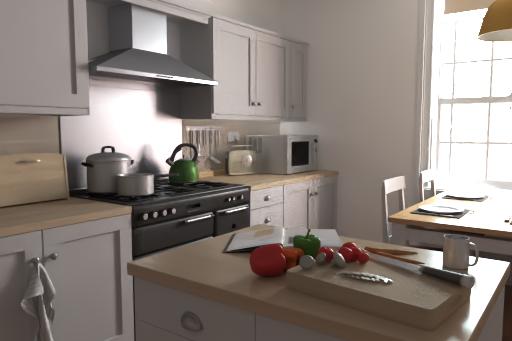}}};   
             \node (fig2) at (-6,-6)
               {\fbox{\includegraphics[width=0.08\textwidth]{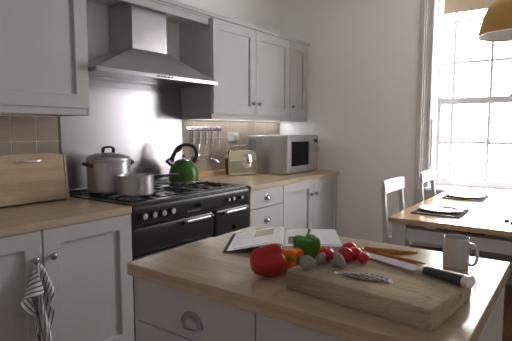}}};
        \end{tikzpicture}
        &
        \includegraphics[width=0.25\textwidth]{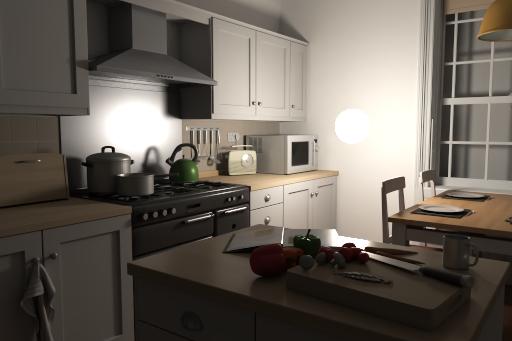} &
        \includegraphics[width=0.25\textwidth]{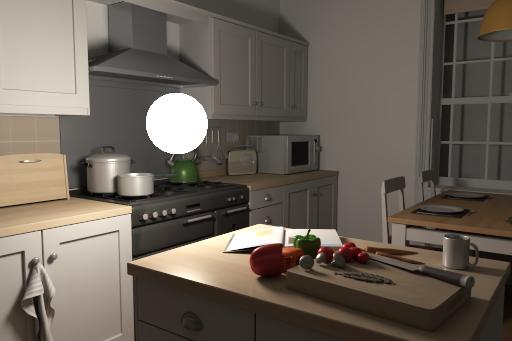} &

        \begin{tikzpicture}[every node/.style={anchor=north west,inner sep=0pt},x=1pt, y=-1pt,]  
             \node (fig1) at (0,0)
               {\fbox{\includegraphics[width=0.25\textwidth]{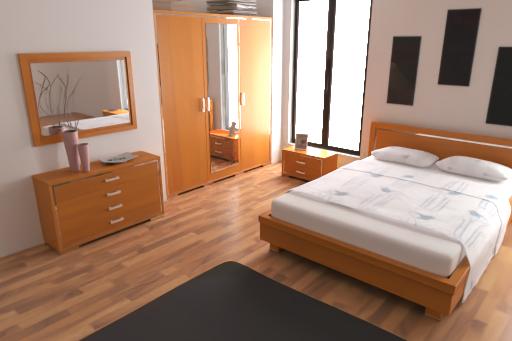}}};   
             \node (fig2) at (-6,-6)
               {\fbox{\includegraphics[width=0.08\textwidth]{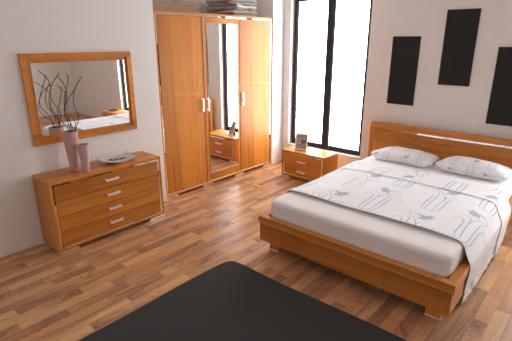}}};
        \end{tikzpicture}
        &
        \includegraphics[width=0.25\textwidth]{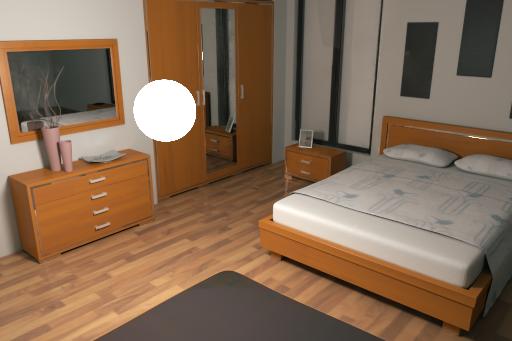} &
        \includegraphics[width=0.25\textwidth]{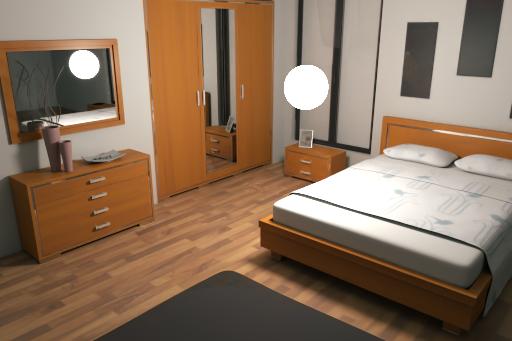} \\

        \begin{tikzpicture}[every node/.style={anchor=north west,inner sep=0pt},x=1pt, y=-1pt,]  
             \node (fig1) at (0,0)
               {\fbox{\includegraphics[width=0.25\textwidth]{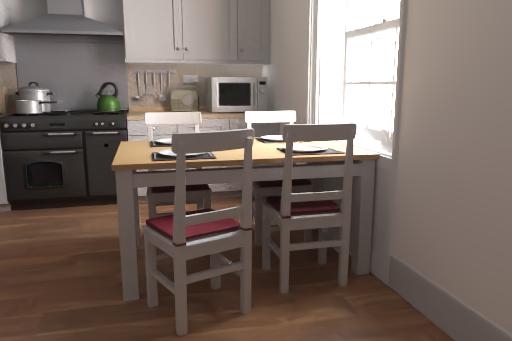}}};   
             \node (fig2) at (-6,-6)
               {\fbox{\includegraphics[width=0.08\textwidth]{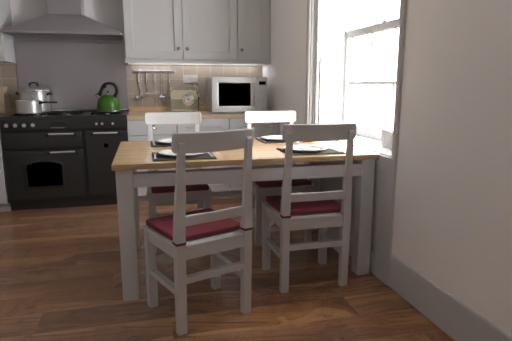}}};
        \end{tikzpicture}
        &
        \includegraphics[width=0.25\textwidth]{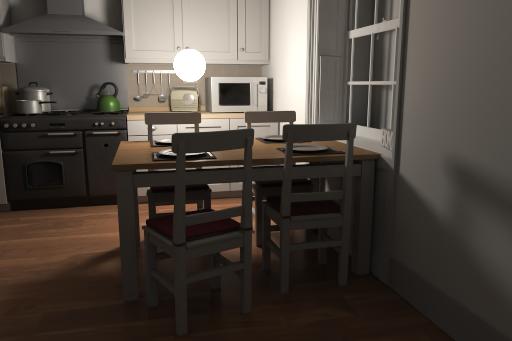} &
        \includegraphics[width=0.25\textwidth]{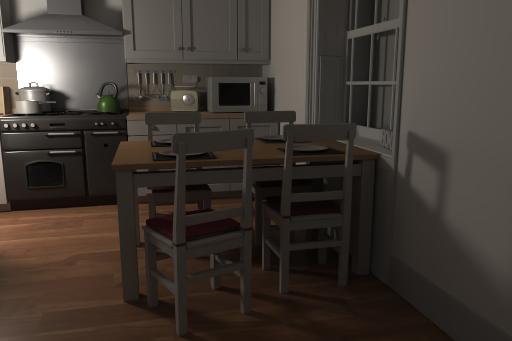} &

        \begin{tikzpicture}[every node/.style={anchor=north west,inner sep=0pt},x=1pt, y=-1pt,]  
             \node (fig1) at (0,0)
               {\fbox{\includegraphics[width=0.25\textwidth]{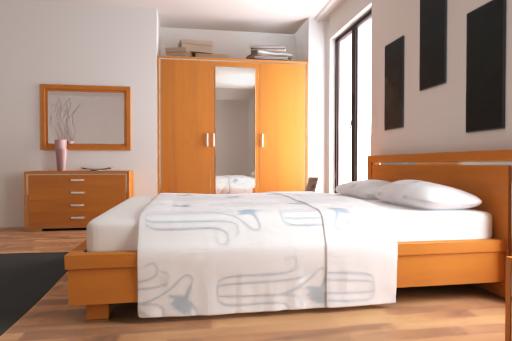}}};   
             \node (fig2) at (-6,-6)
               {\fbox{\includegraphics[width=0.08\textwidth]{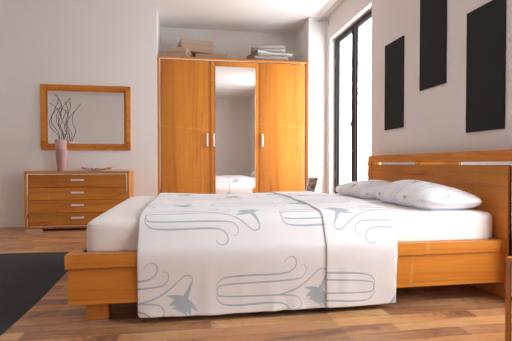}}};
        \end{tikzpicture}
        &
        \includegraphics[width=0.25\textwidth]{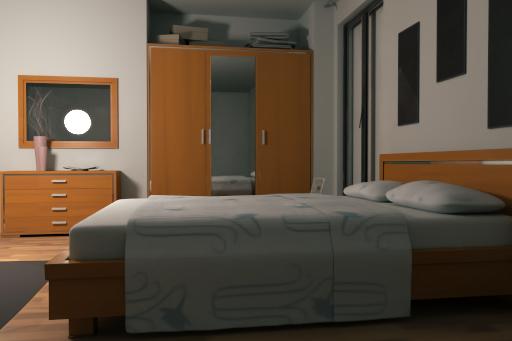} &
        \includegraphics[width=0.25\textwidth]{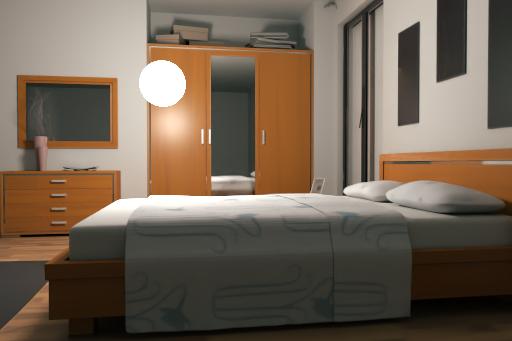} 
        \\

        \hline

        \begin{tikzpicture}[every node/.style={anchor=north west,inner sep=0pt},x=1pt, y=-1pt,]  
             \node (fig1) at (0,0)
               {\fbox{\includegraphics[width=0.25\textwidth]{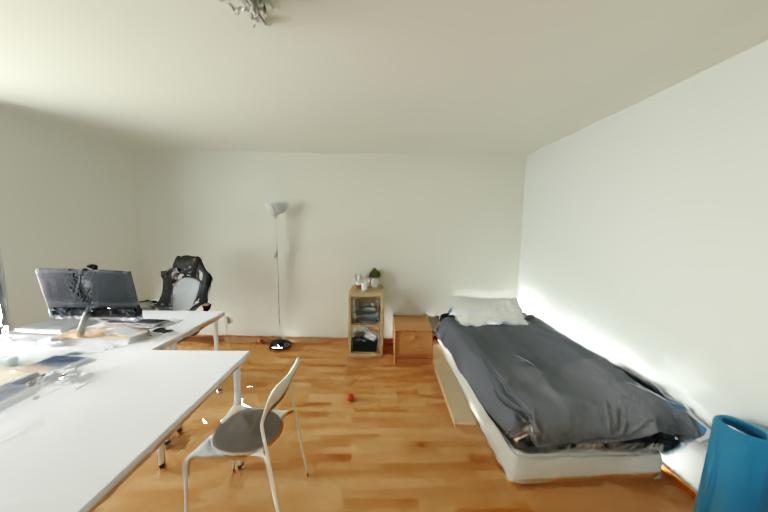}}};   
             \node (fig2) at (-6,-6)
               {\fbox{\includegraphics[width=0.08\textwidth]{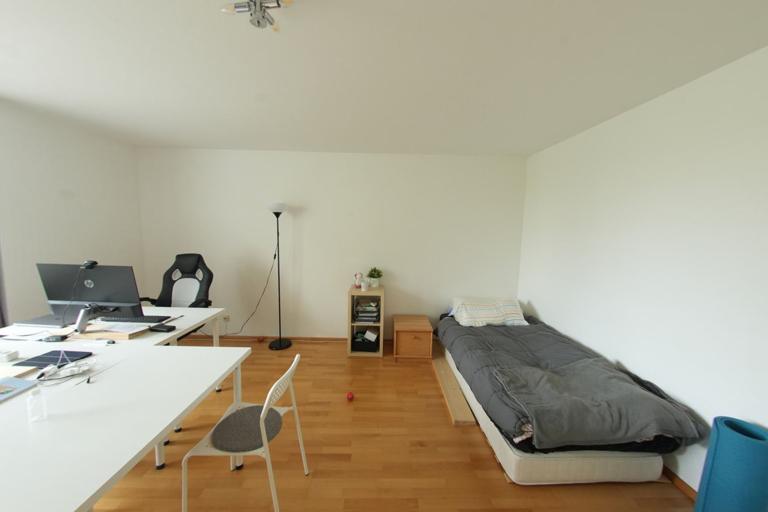}}};
        \end{tikzpicture}
        &
        \includegraphics[width=0.25\textwidth]{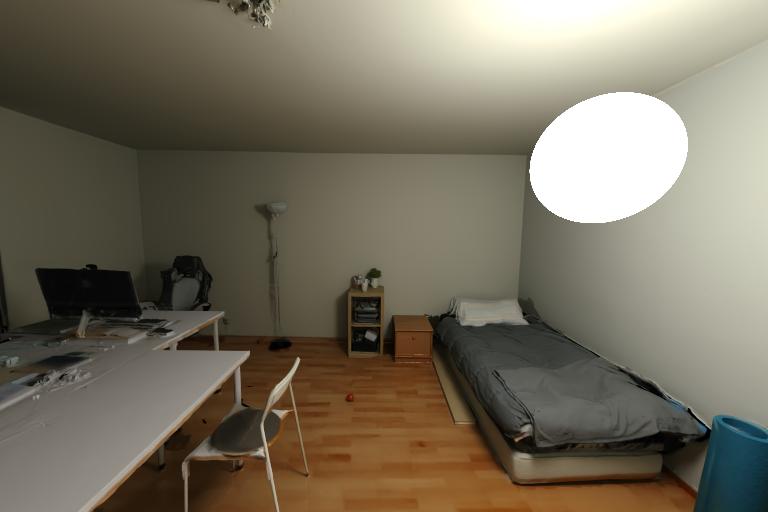} &
        \includegraphics[width=0.25\textwidth]{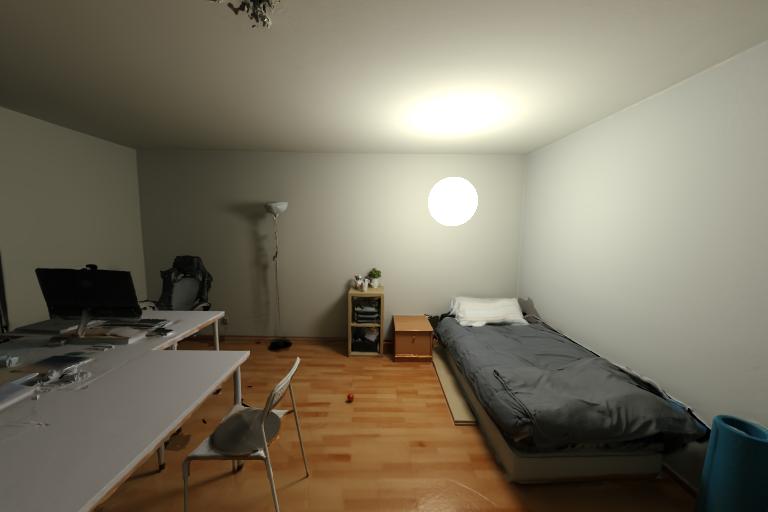}  &

        \begin{tikzpicture}[every node/.style={anchor=north west,inner sep=0pt},x=1pt, y=-1pt,]  
             \node (fig1) at (0,0)
               {\fbox{\includegraphics[width=0.25\textwidth]{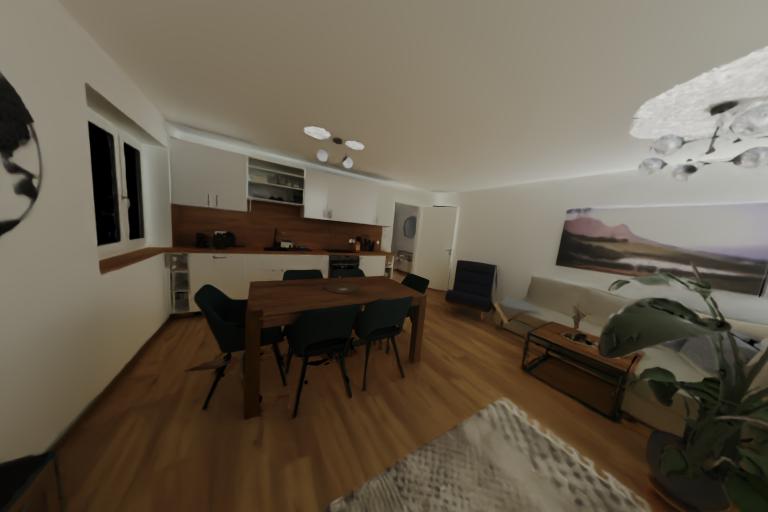}}};   
             \node (fig2) at (-6,-6)
               {\fbox{\includegraphics[width=0.08\textwidth]{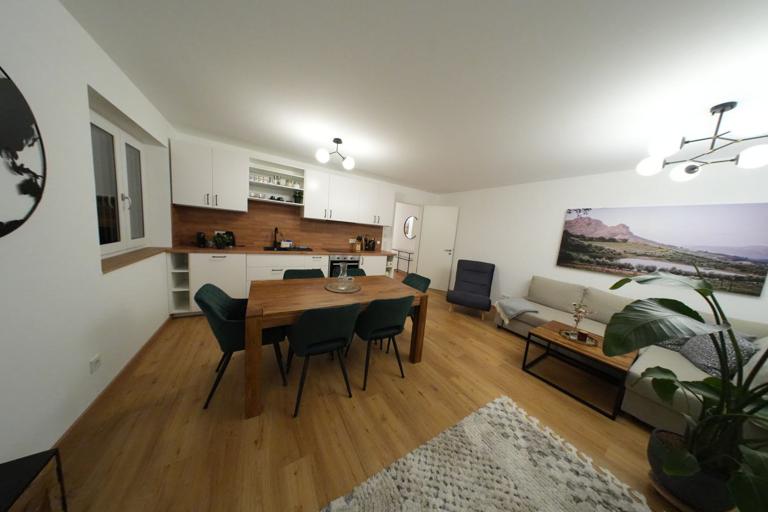}}};
        \end{tikzpicture}
        &
        \includegraphics[width=0.25\textwidth]{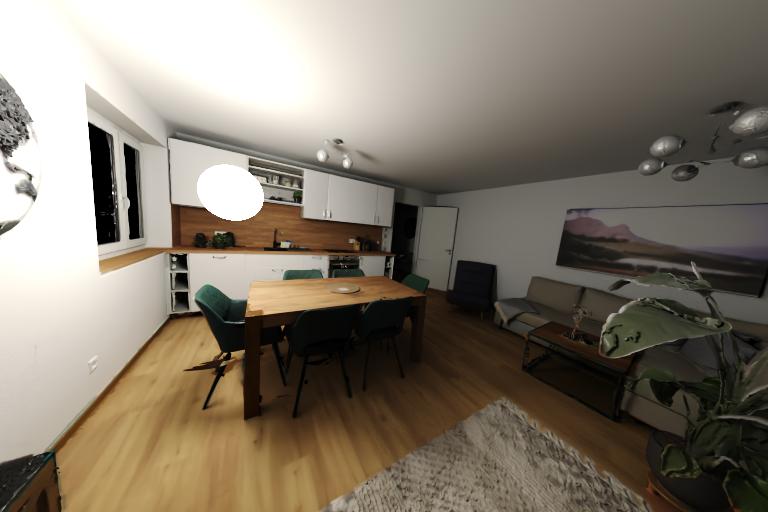} &
        \includegraphics[width=0.25\textwidth]{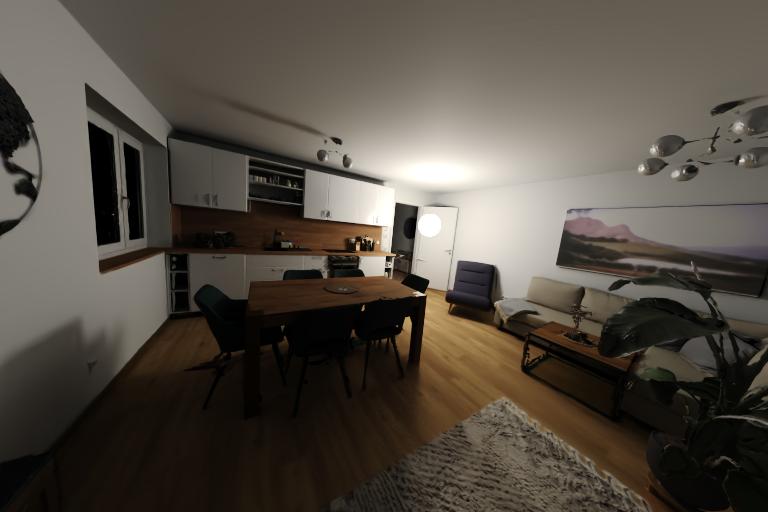} 
        \\
        
        \begin{tikzpicture}[every node/.style={anchor=north west,inner sep=0pt},x=1pt, y=-1pt,]  
             \node (fig1) at (0,0)
               {\fbox{\includegraphics[width=0.25\textwidth]{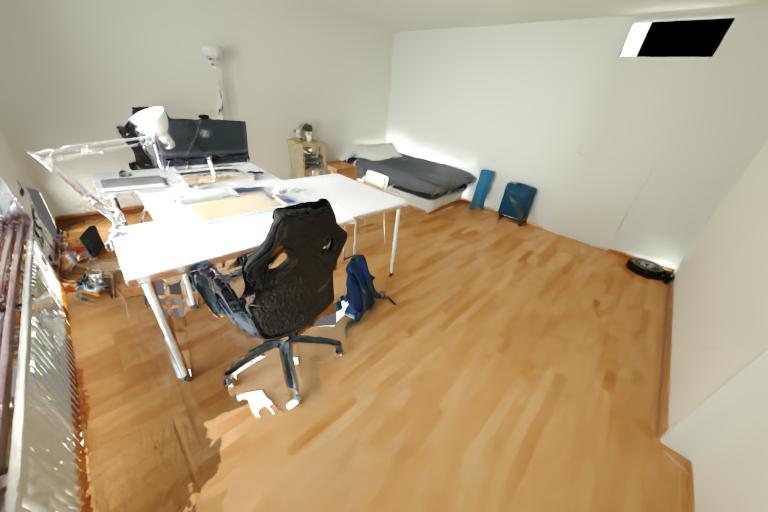}}};   
             \node (fig2) at (-6,-6)
               {\fbox{\includegraphics[width=0.08\textwidth]{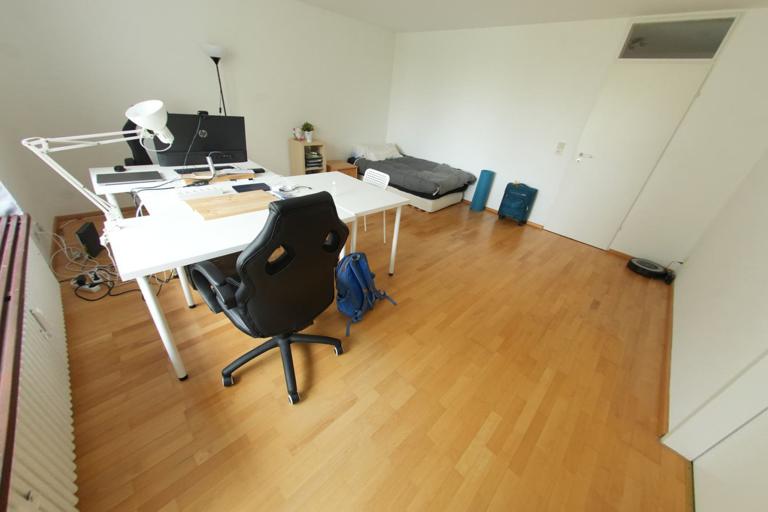}}};
        \end{tikzpicture}
        &
        \includegraphics[width=0.25\textwidth]{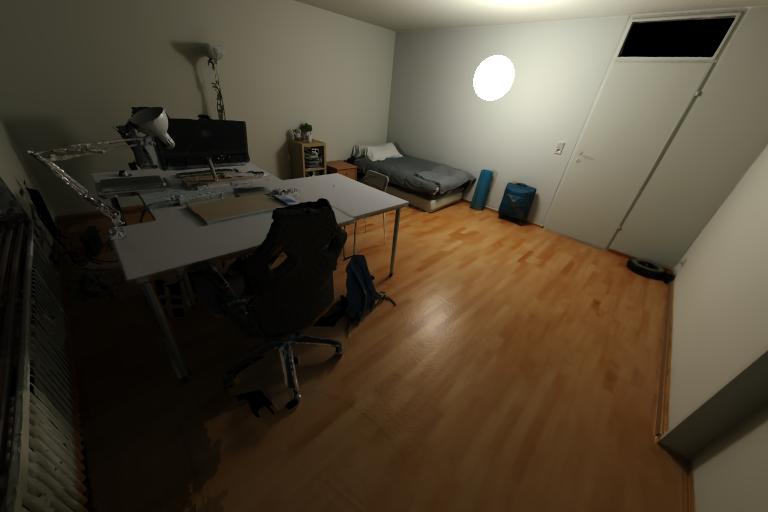} &
        \includegraphics[width=0.25\textwidth]{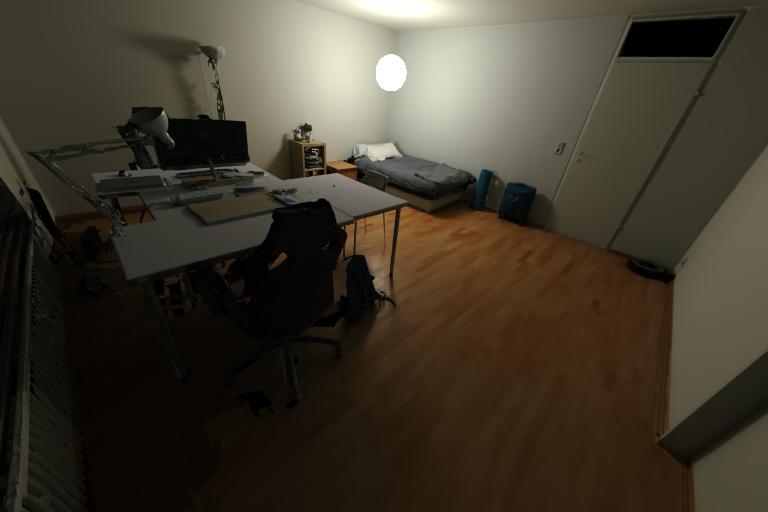} &

        \begin{tikzpicture}[every node/.style={anchor=north west,inner sep=0pt},x=1pt, y=-1pt,]  
             \node (fig1) at (0,0)
               {\fbox{\includegraphics[width=0.25\textwidth]{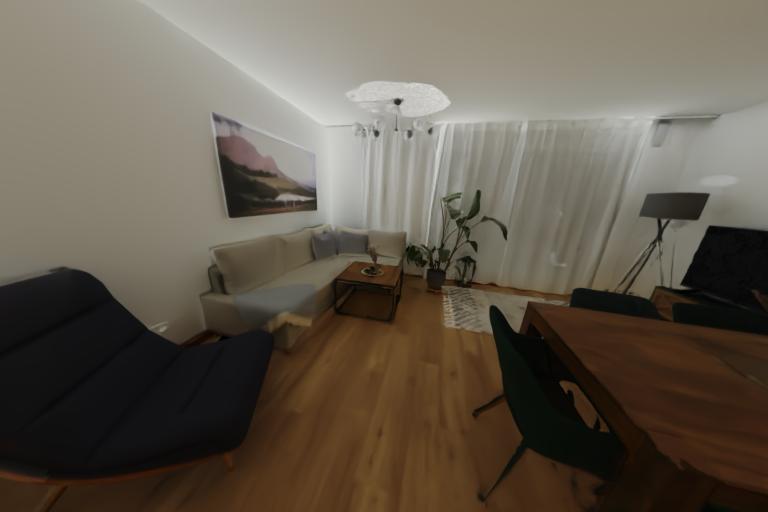}}};   
             \node (fig2) at (-6,-6)
               {\fbox{\includegraphics[width=0.08\textwidth]{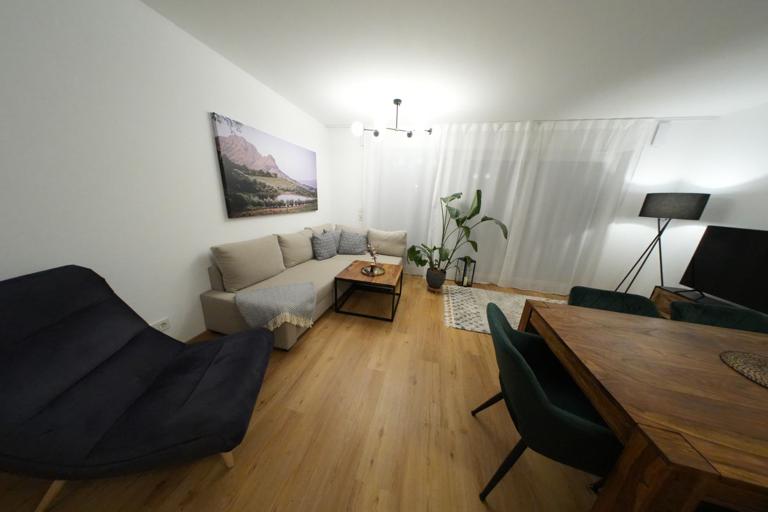}}};
        \end{tikzpicture}
        &
        \includegraphics[width=0.25\textwidth]{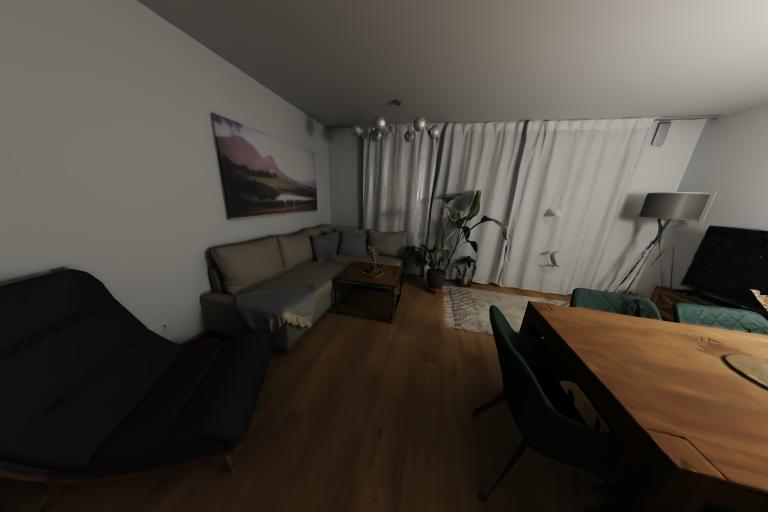} &
        \includegraphics[width=0.25\textwidth]{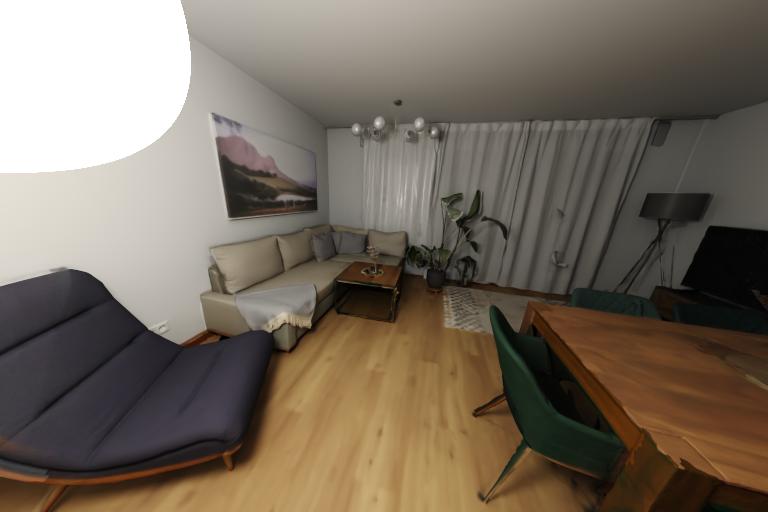} 
        \\

        Rendering (GT top) & 
        Relighting 1 &
        Relighting 2 &

        Rendering (GT top) & 
        Relighting 1 &
        Relighting 2 
    \end{tabular}}
    \vspace{-9pt}
    \caption{\textbf{Applications.} 
    We show rerenderings of the scene and relighting by inserting emissive spheres into the scene.
    Our clean material decompositions enable hiqh-quality renderings without baked-in lighting effects and correct specular highlights.
    }
    \label{fig:exp:applications}
    \vspace{-6pt}
\end{figure*}

\mypar{Number of predictions}
We quantify the effect of our Laplace distribution matching optimization (\cref{sec:method:cross_view}) by evaluating the aggregation quality using different number of predictions (\Cref{tab:exp:num_preds}).
Using more predictions improves the quality, showing that our aggregation does not oversmooth if having more predictions, but aims to find the most consistent one out of them.
In other words, modelling the PBR distribution via many separate predictions is beneficial for finding a high-quality and 3D-consistent PBR texture.

\begin{table}
    
    \centering\setlength{\tabcolsep}{4pt}
    \resizebox{1.0\linewidth}{!}{%
    \begin{tabular}{l|ccc|c|c}
    \toprule
        & \multicolumn{3}{c}{Albedo} & \multicolumn{1}{c}{Rough} & \multicolumn{1}{c}{Metal}\\
        $\#$Preds & PSNR $\uparrow$ & SSIM $\uparrow$ & LPIPS $\downarrow$ & L2 ($\times$ 1000) $\downarrow$ & L2 ($\times$ 1000) $\downarrow$ \\
    \midrule
       1 & 29.62 & 0.908 & 0.177 & 0.817 & 0.134 \\
       2 & 29.77 &  0.910 &  0.168 & 0.799 & \textbf{0.132} \\
       4 & 30.37 & 0.926 & \textbf{0.160} & 0.795 & 0.134 \\
       8 & 30.72 & 0.930 & \textbf{0.160} & 0.788 & 0.135 \\
       (Ours) 16 & \textbf{30.79} & \textbf{0.931} & \textbf{0.160} & \textbf{0.786} & 0.134 \\
    \bottomrule
    \end{tabular}%
    }
    \vspace{-3pt}
    \caption{\textbf{Effect of number of predictions.} 
    We ablate the effect of multiple predictions on the synthetic scenes. The distilled 3D PBR textures improve with more 2D predictions without oversmoothing. This shows the advantage of our distribution matching (\cref{sec:method:cross_view}).
    }
    \label{tab:exp:num_preds}
\end{table}

\subsection{Applications}
\label{sec:experiments:applications}
% \mypar{Relighting}
By explicitly disentangling the PBR materials and lighting, our method enables high-quality relighting of room-scale 3D scenes (\Cref{fig:exp:applications}).
Our clean and sharp material textures do not contain any diffuse baked-in lighting patterns (e.g., smooth walls) and produce realistic specular reflections (e.g., bedroom mirror, around the kitchen oven).
Please see the supplemental for animated relighting effects.

\subsection{Limitations}
\label{sec:experiments:limitations}
While our method significantly improves material reconstruction, we rely on fixed geometry and inherit artifacts from the underlying mesh. 
Jointly optimizing for the geometry is possible and an interesting direction. 
Furthermore, sampling multiple possible material predictions is computationally costly. 
Incorporating the pre-trained prior directly into the optimization process could lead to a more compact approach. 
Finally, our method relies on the quality of pre-trained material estimators. 
Incorporating prediction uncertainties could be used to ignore incorrect predictions.

%!TEX root = main.tex

%%
%% NOTE: Assign collaboration badges and section labels to all sections and 
%% subsects when created (badges include: \incomplete, \underRevision,
%% \readyForFeedback, \feedbackProvided, \complete, \locked)
%%
\section{Conclusion}
\label{sec:conclusion}

We have presented Intrinsic Image Fusion, a probabilistic inverse rendering framework for reconstructing high-quality PBR materials in room-scale indoor scenes. 
Our approach combines single-view generative priors with multi-view optimization, introducing a parametric material representation that reduces the impact of rendering noise while preserving fine details. 
By formulating the problem probabilistically, we explicitly account for uncertainty in both the priors and the path-traced gradients, enabling robust and consistent reconstructions. 
Our experiments demonstrate that Intrinsic Image Fusion outperforms state-of-the-art inverse rendering and intrinsic decomposition methods, delivering sharp, disentangled materials suitable for relighting, editing, and virtual object insertion. 
We believe Intrinsic Image Fusion provides a step toward practical, physically faithful scene decomposition. 
% Text content to add within the document (individual paper sections)

\vspace{3pt}
\noindent\textbf{Acknowledgements.}
This work was supported by Huawei as well as the ERC Starting Grant Scan2CAD (804724). We thank Angela Dai for the video voice-over and David Rozenberszki for the valuable discussions about the real-world instance segmentations.  % Acknoweldgements (positioned according to conference guidelines)

{
    \small
    \bibliographystyle{style/ieeenat_fullname}
    \bibliography{sections/bibliography}
}

\maketitlesupplementary
\appendix
\setcounter{page}{1}

\section{Additional Implementational Details}
\paragraph{Dataset Details. }
We specify the scale of the test set in the inset table.
We calculate PBR decomposition metrics against the ground-truth on all input views and report averaged results for the 4 synthetic scenes.

\paragraph{Number of parameters.}
Our parametric texture representations uses a base texture and a set of per-object affine transformations. 
The base texture uses an neural hashgrid \cite{InstantNGP} with $32$ levels, $2$ features per level and $2^{19}$ hashmap size to store albedo, rough, metallic means and scales, which amounts to $28$~M parameters.
These base texture parameters are optimized only during the aggregation phase.
During the inverse path tracing, we only need to optimize the per-object affine transformations, which amounts to $O$x$3$x$4$, where $O$ is the number of 3D objects, giving a total of $1092$ parameters for the kitchen scene. 
Optimizing on such a low-dimensional manifold makes the inverse path tracing more constrained, making it more robust against the rendering noise.  

\paragraph{Real data pre-processing. }
Our method relies on instance segmentation, which we obtain in a pre-porcessing step for the ScanNet++ \cite{ScanNet++} scenes. 
We use SAM2 \cite{SAM2} to estimate a per-image segmentation.
To aggregate the segmentations, we use MaskClustering \cite{MaskClustering}. 
This way, we get a per-face instance id. 
Then, we rasterize the mesh into all the views and render the instance ids to get per-pixel instances. 
These images are only used for evaluating the baseline IRIS \cite{IRIS}. 

\paragraph{Real-world lighting representation. }
To account for missing geometry and emission coming from outside of the scene, we additionally define an environment map of resolution $16$x$32$.
Similarly to the mesh emission, we filter the potentially emissive environment map pixels by thresholding their aggregated observed radiance values ($t=0.85$). 

\paragraph{Lighting optimization. }
Our lighting optimization follows FIPT \cite{FIPT} and has four steps. 
First, we initialize the emissive triangles by filtering the aggregated observed radiance with a threshold of $t=0.99$. 
Second, we optimize for the emission values using frozen base material textures with inverse path tracing using $128$ samples per pixel with a single bounce.
We use SGD optimizer for $1$ epoch and batch size $8192$ rays with initial learning rate of $lr=1e+2$.
After $1000$ iterations, we prune all the emitters, which has an intensity lower then $5\%$ of the overall maximum intensity. This stage takes $35$ minutes using $20$~GB of VRAM on the kitchen scene. 
Then, we cache the light transport into diffuse ($spp=256$) and specular shading maps ($spp=128$). This stage takes $5$ minutes on the kitchen scene. 
Finally, we optimize for the BRDF and CRF parameters by rerendering with the cached shading maps. This stage takes $20$ minutes using $11$~GB of VRAM on the kitchen scene. 
Here, we use SGD optimizer for $3$ epoch and batch size $32768$ rays with initial learning rate of $lr=1e+1$ and decay after every epoch by a factor of $0.2$.
Following IRIS \cite{IRIS}, we regularize the roughness and metallic channels to stay close to diffuse materials ($w_{rough}=1e-3, w_{metal}=5e-3$).

\begin{table}[t]
    \centering
    \setlength{\tabcolsep}{1pt}
    \small
    \begin{tabular}{ccc}
        \toprule
        Dataset & Scene & Views \\
        \midrule
        FIPT \cite{FIPT} & Kitchen & 202 \\
        FIPT \cite{FIPT} & Bedroom & 208 \\
        FIPT \cite{FIPT} & Livingroom & 213 \\
        FIPT \cite{FIPT} & Bathroom & 109 \\
        \midrule
        FIPT \cite{FIPT} & Conferenceroom & 191 \\
        FIPT \cite{FIPT} & Classroom & 278 \\
        ScanNet++ \cite{ScanNet++} & 2a1b555966 & 349 \\
        ScanNet++ \cite{ScanNet++} & 651dc6b4f1 & 64 \\
        ScanNet++ \cite{ScanNet++} & a003a6585e & 106 \\
        \bottomrule
    \end{tabular}
    \label{tab:supp:test_scale}
    \caption{\textbf{Dataset Details.} We summarize the used scenes.
    }
\end{table}

\begin{figure}[t]
    \centering
    \includegraphics[height=2.5cm]{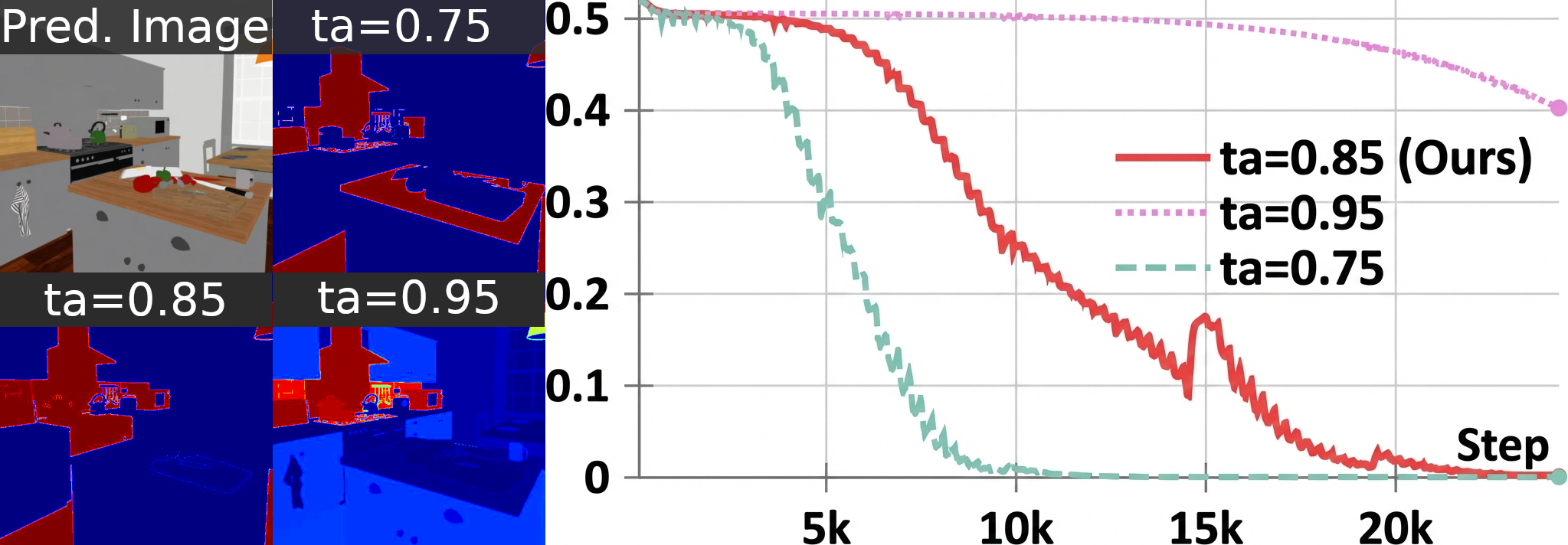}
    % \vspace{-2mm}
    \caption{\textbf{Temperature Ablation.} Lower temperature values motivate quick convergence to hard assignment, but can get stuck in local optima, while too high value keeps soft assignment leading to oversmoothing. We use a value in-between to balance. }
    \label{fig:supp:tau}
\end{figure}

\begin{figure}[t]
    \centering
    \setlength\tabcolsep{1pt}
    \renewcommand{\arraystretch}{0.4}
    \resizebox{\linewidth}{!}{
    \fboxsep=0pt
    \begin{tabular}{cc}

% ---------- TOP LEFT ----------
\begin{tikzpicture}
  \node[inner sep=0pt] (a) {
    \adjustbox{trim=0 5 0 10,clip}{
      \includegraphics[width=0.25\linewidth]{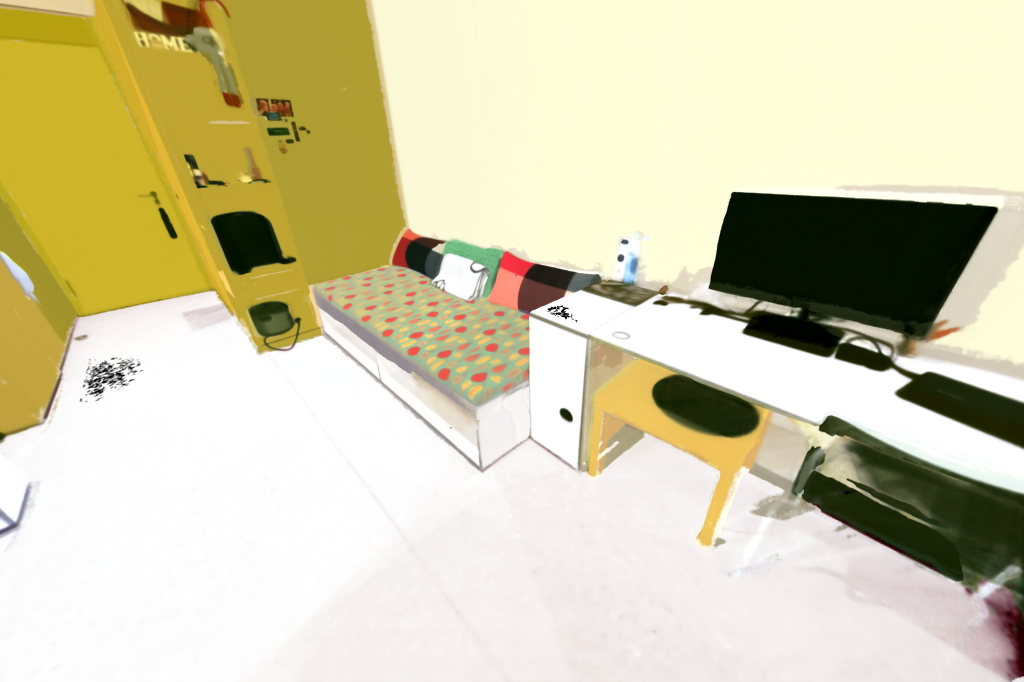}
    }
  };
  \begin{scope}
    \clip (a.south west) rectangle (a.north east);
    \node[draw=red, line width=0.3pt, circle,
          minimum size=13pt,
          xshift=18pt, yshift=-5pt] at (a.center) {};
  \end{scope}
\end{tikzpicture}
&
% ---------- TOP RIGHT ----------
\begin{tikzpicture}
  \node[inner sep=0pt] (a) {
    \adjustbox{trim=0 5 0 10,clip}{
      \includegraphics[width=0.25\linewidth]{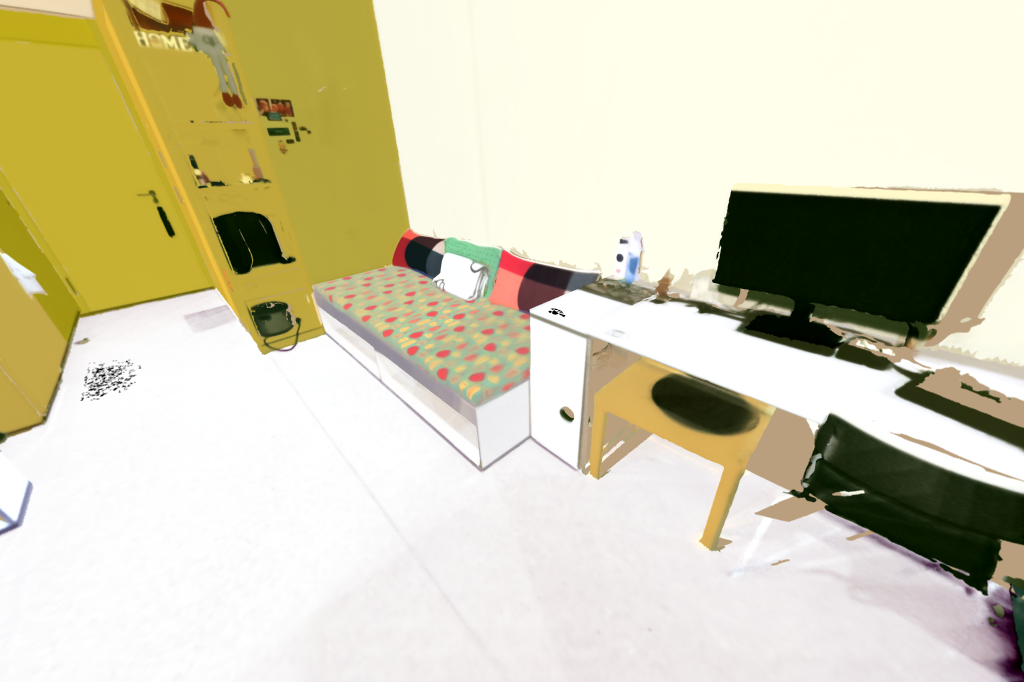}
    }
  };
  \begin{scope}
    \clip (a.south west) rectangle (a.north east);
    \node[draw=orange, line width=0.3pt, circle,
          minimum size=13pt,
          xshift=18pt, yshift=-5pt] at (a.center) {};
  \end{scope}
\end{tikzpicture}
\\

% ---------- BOTTOM LEFT ----------
\begin{tikzpicture}
  \node[inner sep=0pt] (a) {
    \adjustbox{trim=0 5 0 10,clip}{
      \includegraphics[width=0.25\linewidth]{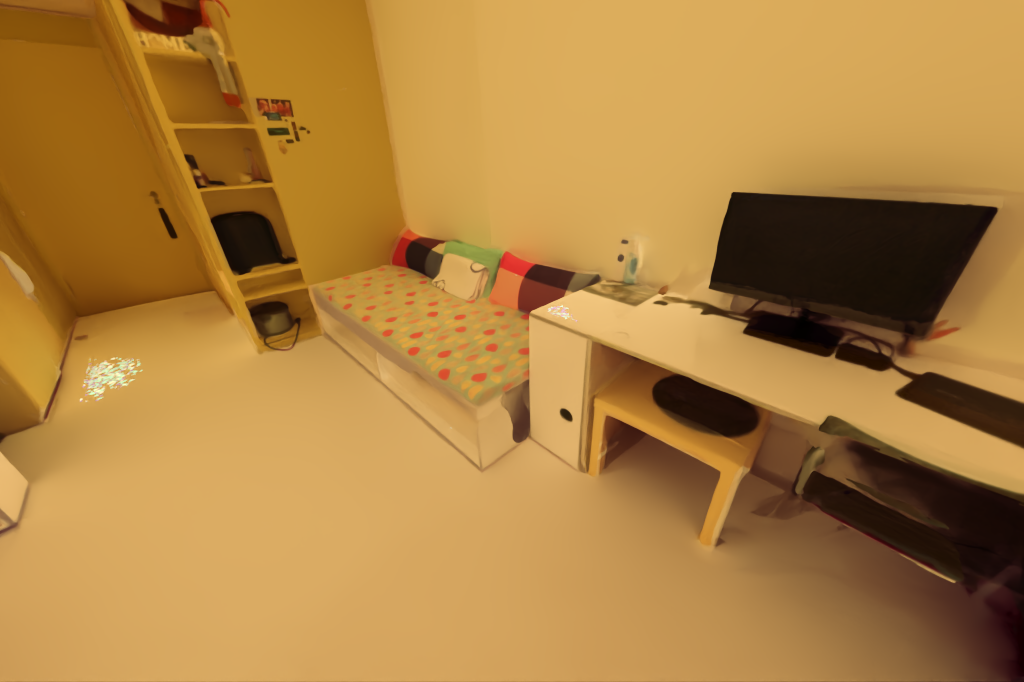}
    }
  };
  \begin{scope}
    \clip (a.south west) rectangle (a.north east);
    \node[draw=green, line width=0.3pt, circle,
          minimum size=10pt,
          xshift=24pt, yshift=5pt] at (a.center) {};
  \end{scope}
\end{tikzpicture}
&
% ---------- BOTTOM RIGHT ----------
\begin{tikzpicture}
  \node[inner sep=0pt] (a) {
    \adjustbox{trim=0 5 0 10,clip}{
      \includegraphics[width=0.25\linewidth]{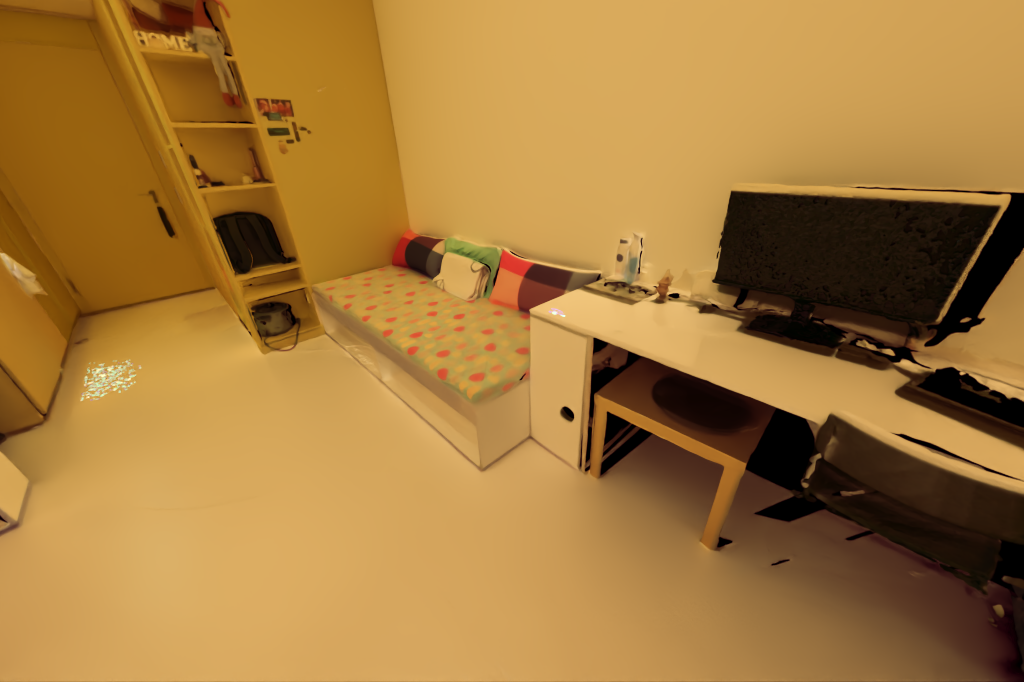}
    }
  };
  \begin{scope}
    \clip (a.south west) rectangle (a.north east);
    \node[draw=red, line width=0.3pt, circle,
          minimum size=10pt,
          xshift=24pt, yshift=5pt] at (a.center) {};
  \end{scope}
\end{tikzpicture}

    \end{tabular}}
    \vspace{-4mm}
    \caption{\textbf{Geometry Ablation.} The estimated albedo is similar with laser scan (left) and photometric reconstruction (right \cite{IRIS}) reconstruction.
    }
    \label{fig:supp:geom}
\end{figure}

\begin{figure}[t]
    \centering
    \setlength\tabcolsep{1pt}
    \renewcommand{\arraystretch}{0.5}
    \resizebox{\linewidth}{!}{
    \fboxsep=0pt
    \begin{tabular}{cc}

% ---------- TOP LEFT ----------
\begin{tikzpicture}
  \node[inner sep=0pt] (a) {
    \adjustbox{trim=0 5 0 10,clip}{
      \includegraphics[width=0.5\linewidth]{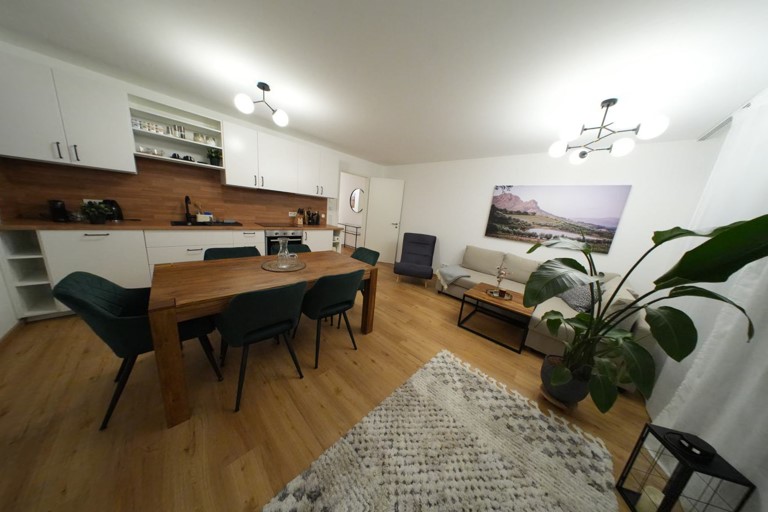}
    }
  };
  \begin{scope}
    \clip (a.south west) rectangle (a.north east);
    \node[draw=red, line width=0.3pt, circle,
          minimum size=13pt,
          xshift=-15pt, yshift=3pt] at (a.center) {};
    \node[draw=red, line width=0.3pt, circle,
          minimum size=13pt,
          xshift=-42pt, yshift=-15pt] at (a.center) {};
  \end{scope}
\end{tikzpicture}
&
% ---------- TOP RIGHT ----------
\begin{tikzpicture}
  \node[inner sep=0pt] (a) {
    \adjustbox{trim=0 5 0 10,clip}{
      \includegraphics[width=0.5\linewidth]{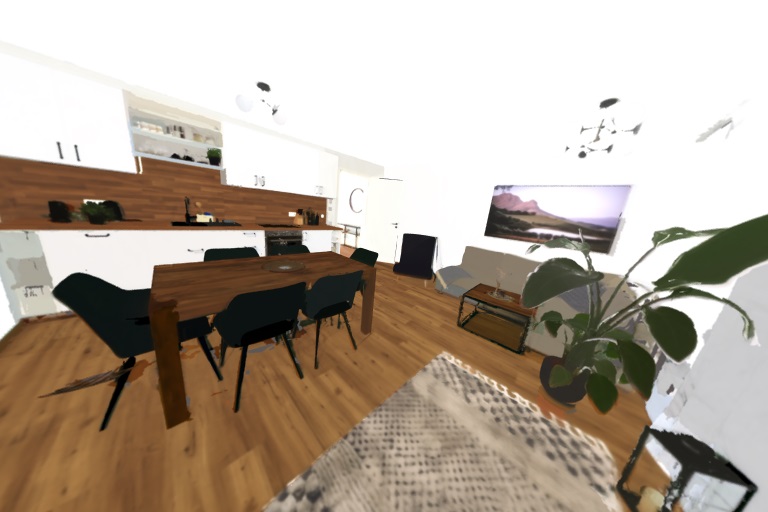}
    }
  };
  \begin{scope}
    \clip (a.south west) rectangle (a.north east);
    \node[draw=red, line width=0.3pt, circle,
          minimum size=13pt,
          xshift=-15pt, yshift=3pt] at (a.center) {};
    \node[draw=red, line width=0.3pt, circle,
          minimum size=13pt,
          xshift=-42pt, yshift=-15pt] at (a.center) {};
  \end{scope}
\end{tikzpicture}
\\
Input & Ours (albedo)

    \end{tabular}}
    \caption{\textbf{Failure Cases.} Even though our parametric formulation makes our method robust, geometric reconstruction errors propagate into our final reconstruction, such as missing thin structures, transparent objects or floating artifacts. }
    \label{fig:supp:fail}
\end{figure}

\section{Additional Results}
\label{sec:supp:experiments}
\begin{figure*}[t]
    \centering
    \setlength\tabcolsep{1.25pt}
    \resizebox{\textwidth}{!}{
    \fboxsep=0pt
    \begin{tabular}{c|cccc|c}
\begin{tabular}[b]{c}
\fbox{\includegraphics[width=0.13\textwidth]{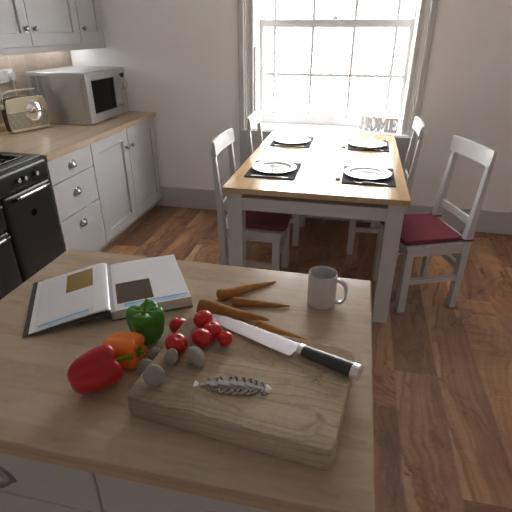}}
\end{tabular} &
\begin{tabular}[b]{cc}
    \fbox{\includegraphics[width=0.13\textwidth]{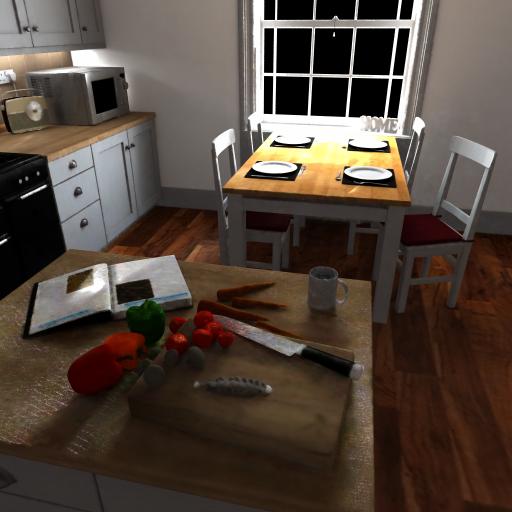}} &
    \begin{tabular}[b]{c}
    \fbox{\includegraphics[width=0.06\textwidth]{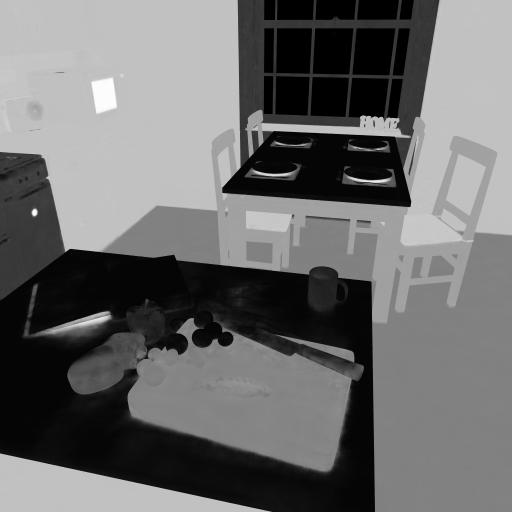}} \\
    \fbox{\includegraphics[width=0.06\textwidth]{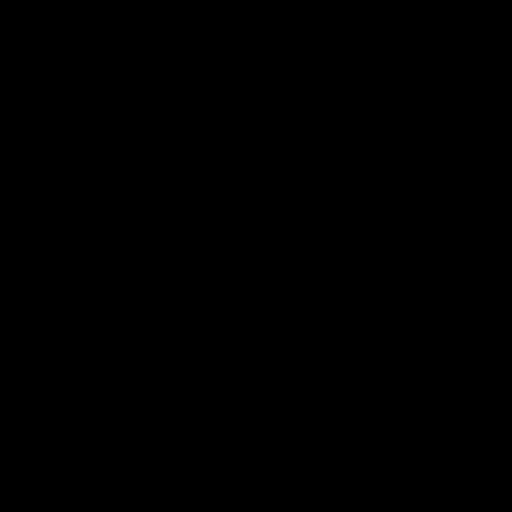}}
    \end{tabular}
    \end{tabular} &
\begin{tabular}[b]{cc}
    \fbox{\includegraphics[width=0.13\textwidth]{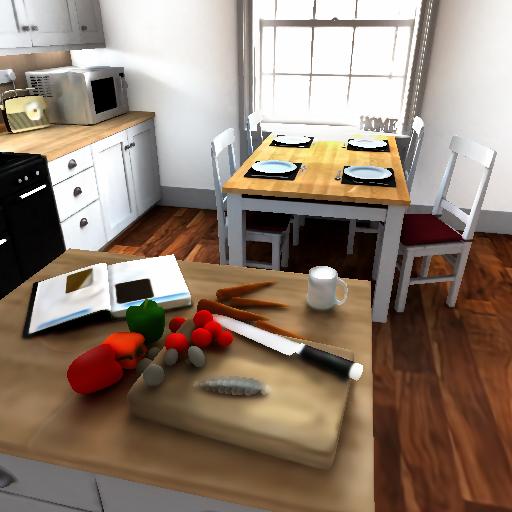}} &
    \begin{tabular}[b]{c}
    \fbox{\includegraphics[width=0.06\textwidth]{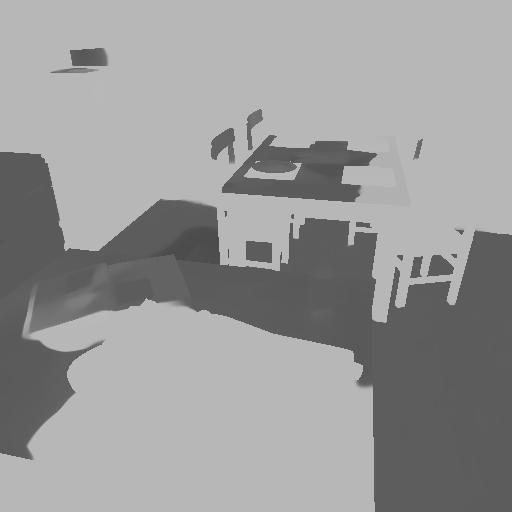}} \\
    \fbox{\includegraphics[width=0.06\textwidth]{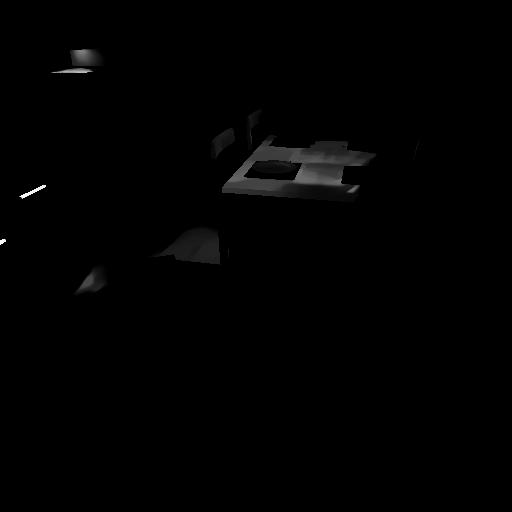}}
    \end{tabular}
    \end{tabular} &
\begin{tabular}[b]{cc}
    \fbox{\includegraphics[width=0.13\textwidth]{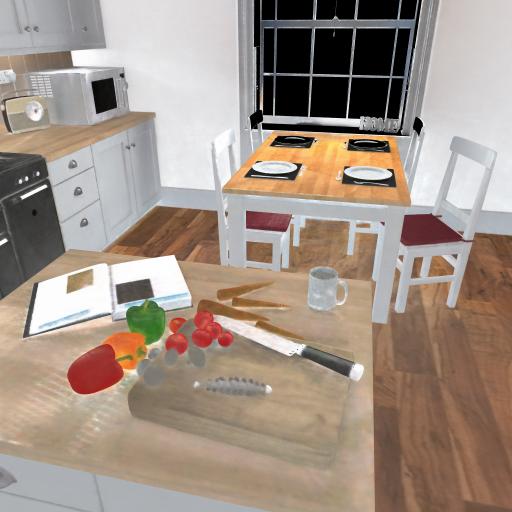}} &
    \begin{tabular}[b]{c}
    \fbox{\includegraphics[width=0.06\textwidth]{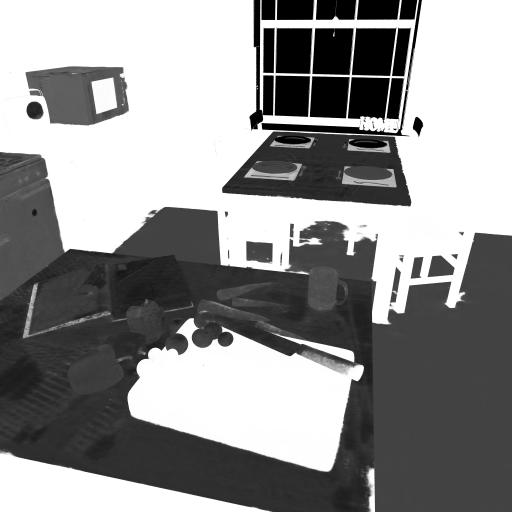}} \\
    \fbox{\includegraphics[width=0.06\textwidth]{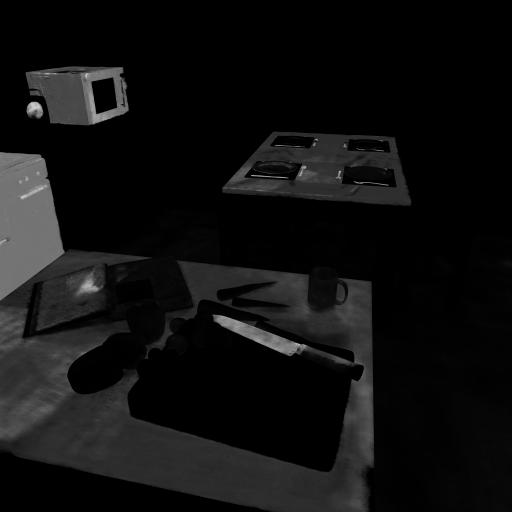}}
    \end{tabular}
    \end{tabular} &
\begin{tabular}[b]{cc}
    \fbox{\includegraphics[width=0.13\textwidth]{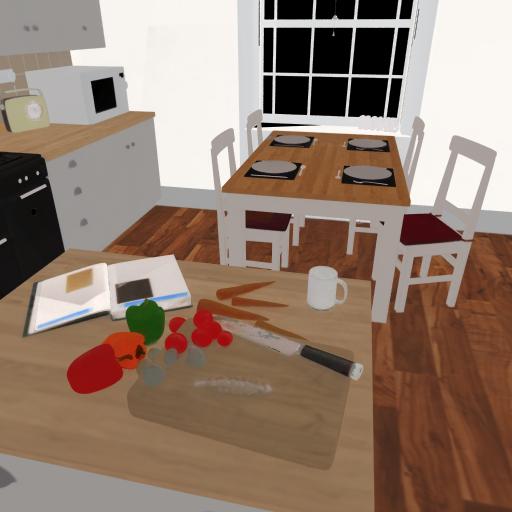}} &
    \begin{tabular}[b]{c}
    \fbox{\includegraphics[width=0.06\textwidth]{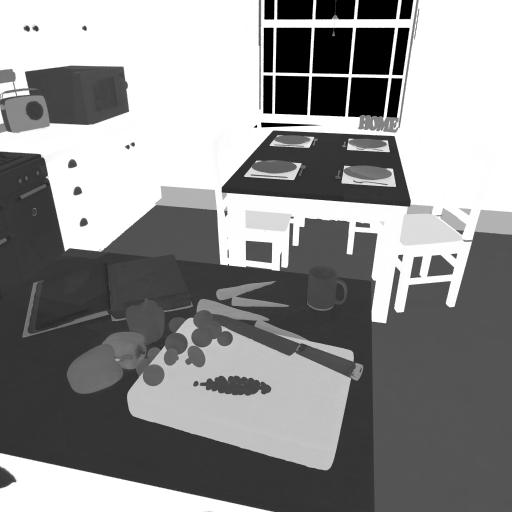}} \\
    \fbox{\includegraphics[width=0.06\textwidth]{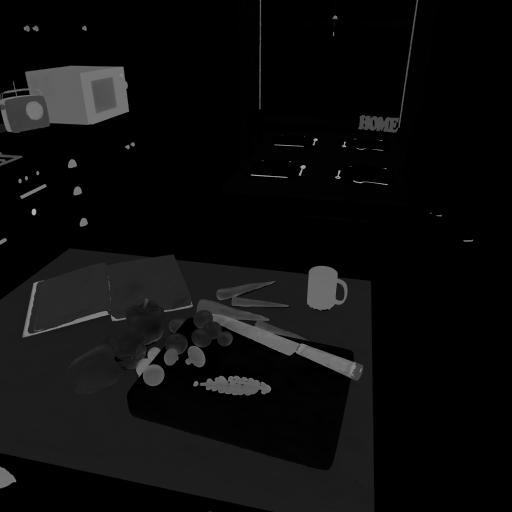}}
    \end{tabular}
    \end{tabular} &
\begin{tabular}[b]{c}
\fbox{\includegraphics[width=0.13\textwidth]{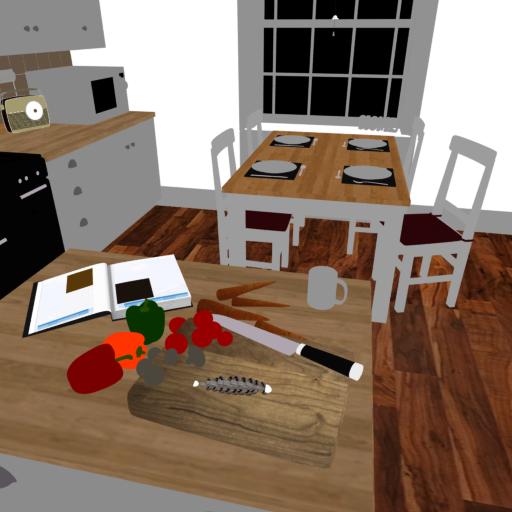}}
\end{tabular} \\
\begin{tabular}[b]{c}
\fbox{\includegraphics[width=0.13\textwidth]{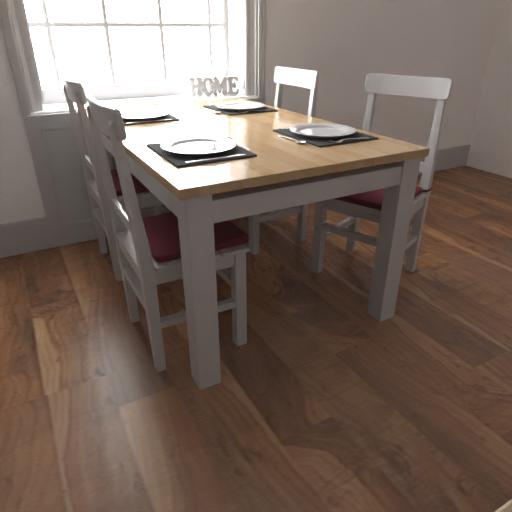}}
\end{tabular} &
\begin{tabular}[b]{cc}
    \fbox{\includegraphics[width=0.13\textwidth]{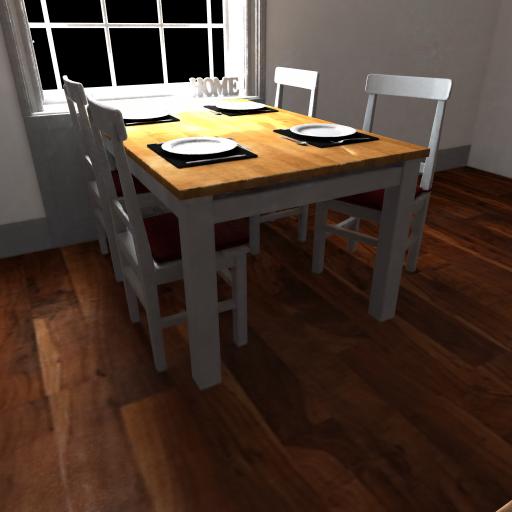}} &
    \begin{tabular}[b]{c}
    \fbox{\includegraphics[width=0.06\textwidth]{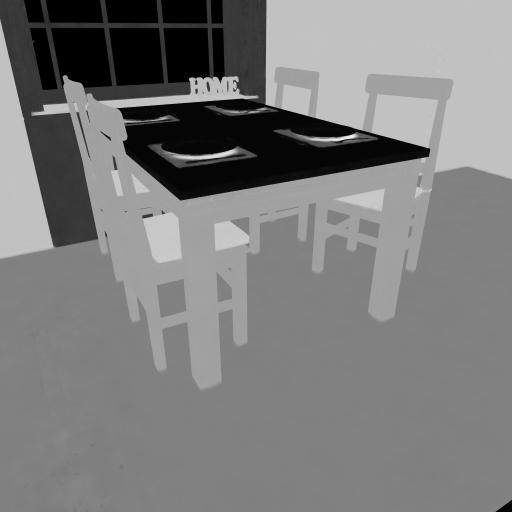}} \\
    \fbox{\includegraphics[width=0.06\textwidth]{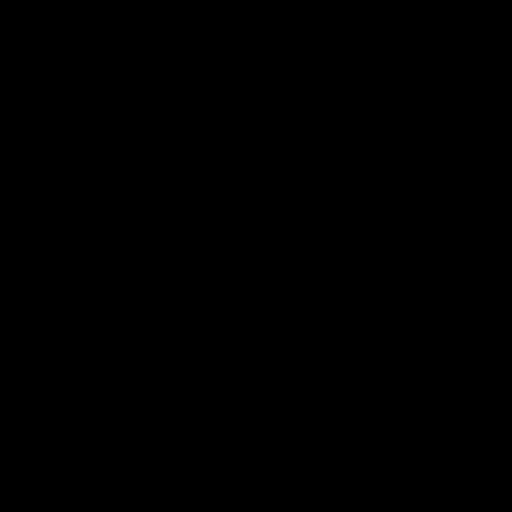}}
    \end{tabular}
    \end{tabular} &
\begin{tabular}[b]{cc}
    \fbox{\includegraphics[width=0.13\textwidth]{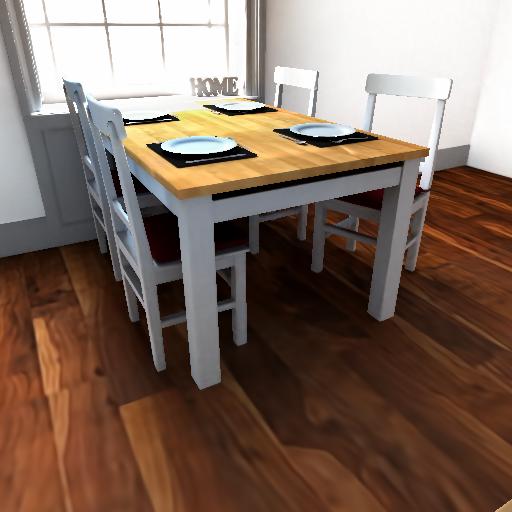}} &
    \begin{tabular}[b]{c}
    \fbox{\includegraphics[width=0.06\textwidth]{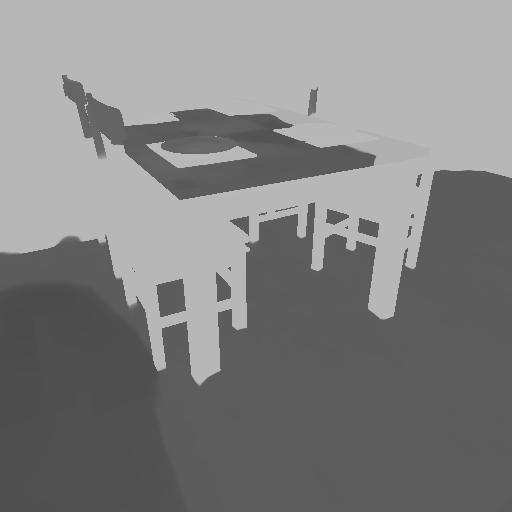}} \\
    \fbox{\includegraphics[width=0.06\textwidth]{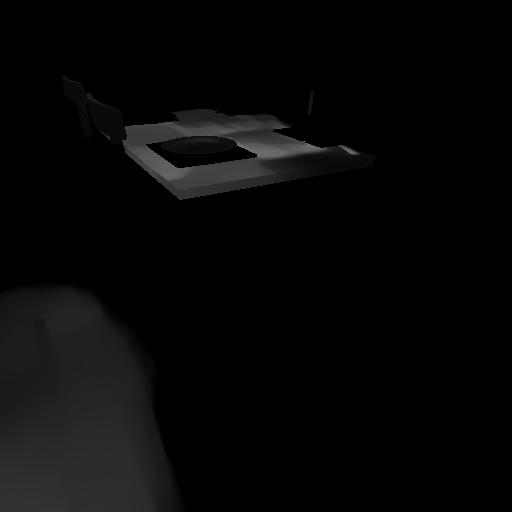}}
    \end{tabular}
    \end{tabular} &
\begin{tabular}[b]{cc}
    \fbox{\includegraphics[width=0.13\textwidth]{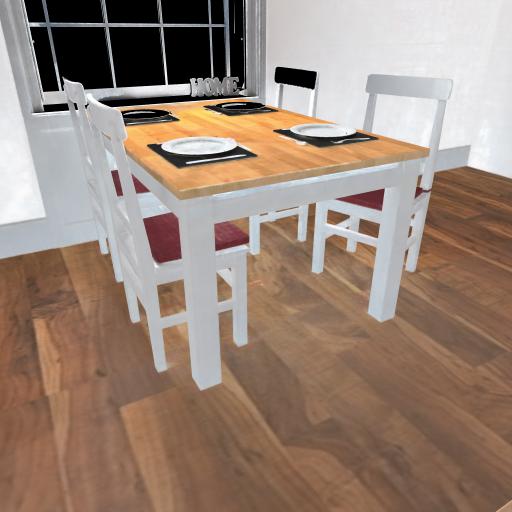}} &
    \begin{tabular}[b]{c}
    \fbox{\includegraphics[width=0.06\textwidth]{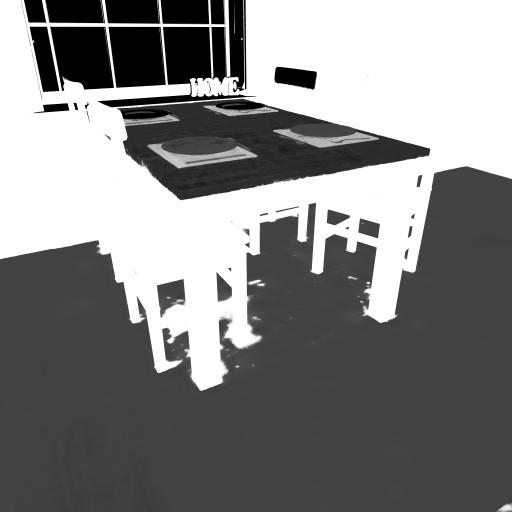}} \\
    \fbox{\includegraphics[width=0.06\textwidth]{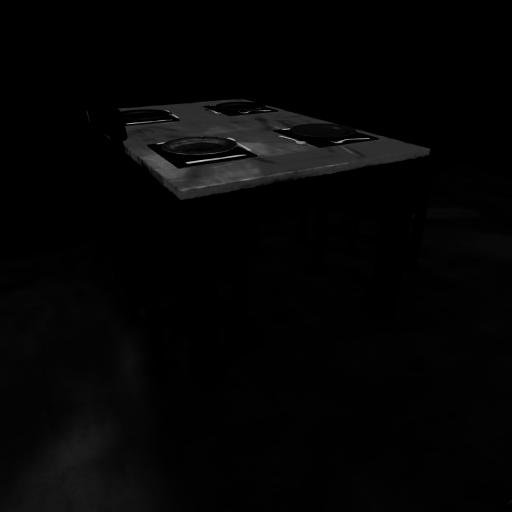}}
    \end{tabular}
    \end{tabular} &
\begin{tabular}[b]{cc}
    \fbox{\includegraphics[width=0.13\textwidth]{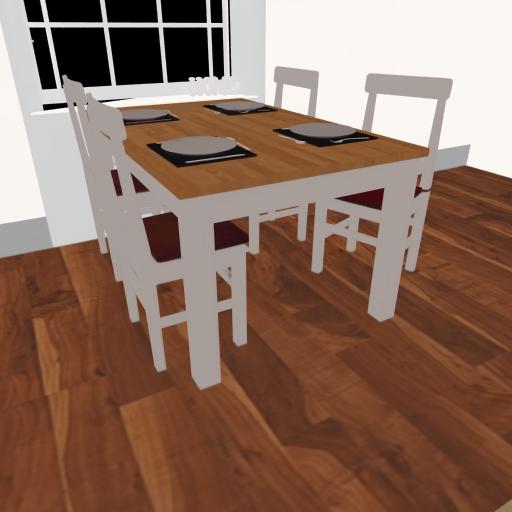}} &
    \begin{tabular}[b]{c}
    \fbox{\includegraphics[width=0.06\textwidth]{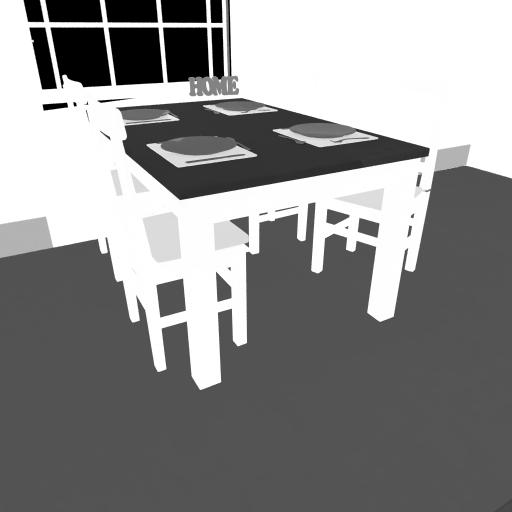}} \\
    \fbox{\includegraphics[width=0.06\textwidth]{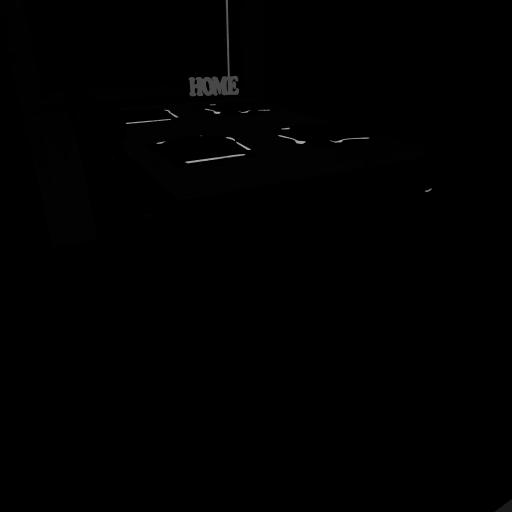}}
    \end{tabular}
    \end{tabular} &
\begin{tabular}[b]{c}
\fbox{\includegraphics[width=0.13\textwidth]{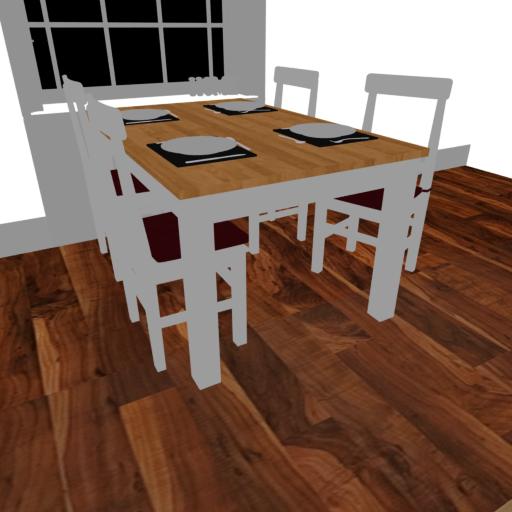}}
\end{tabular} \\
\hline
\begin{tabular}[b]{c}
\fbox{\includegraphics[width=0.13\textwidth]{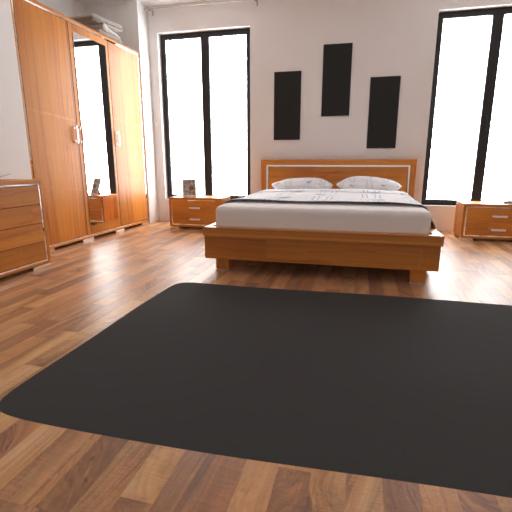}}
\end{tabular} &
\begin{tabular}[b]{cc}
    \fbox{\includegraphics[width=0.13\textwidth]{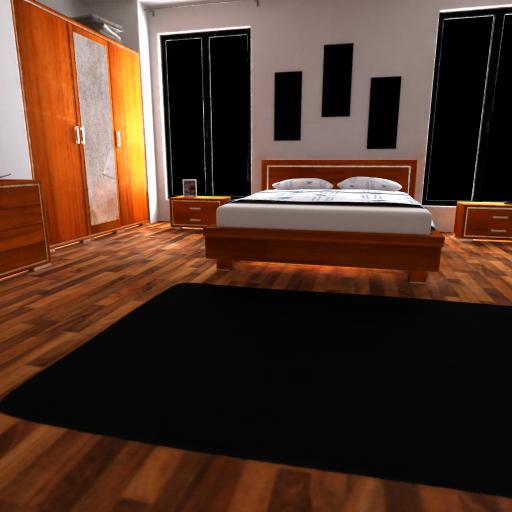}} &
    \begin{tabular}[b]{c}
    \fbox{\includegraphics[width=0.06\textwidth]{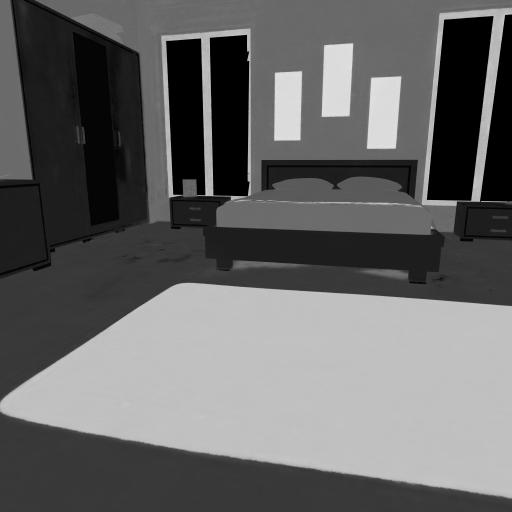}} \\
    \fbox{\includegraphics[width=0.06\textwidth]{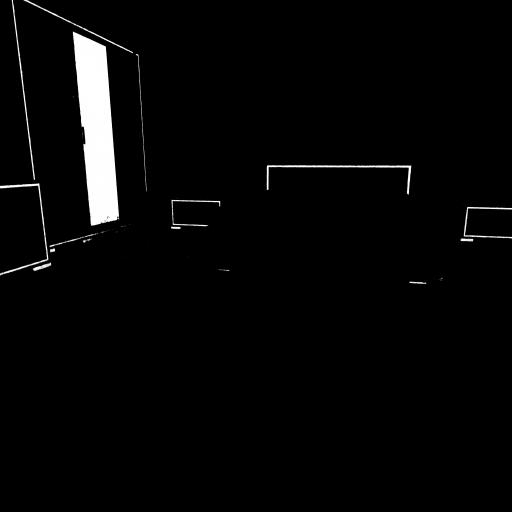}}
    \end{tabular}
    \end{tabular} &
\begin{tabular}[b]{cc}
    \fbox{\includegraphics[width=0.13\textwidth]{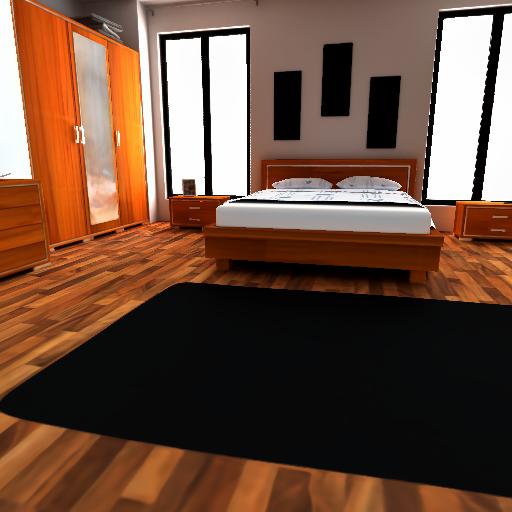}} &
    \begin{tabular}[b]{c}
    \fbox{\includegraphics[width=0.06\textwidth]{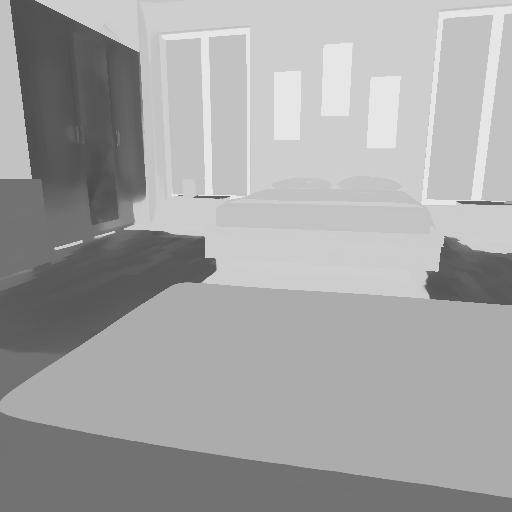}} \\
    \fbox{\includegraphics[width=0.06\textwidth]{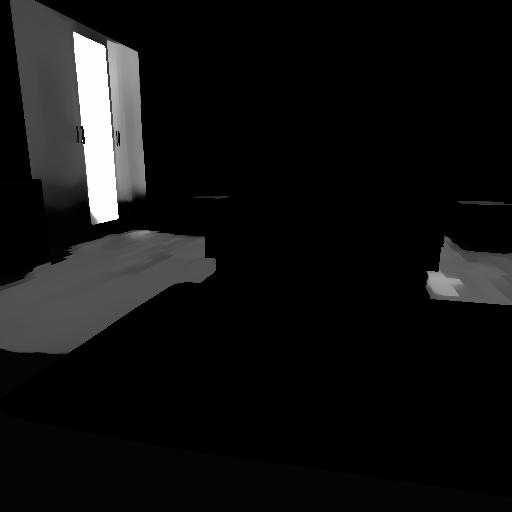}}
    \end{tabular}
    \end{tabular} &
\begin{tabular}[b]{cc}
    \fbox{\includegraphics[width=0.13\textwidth]{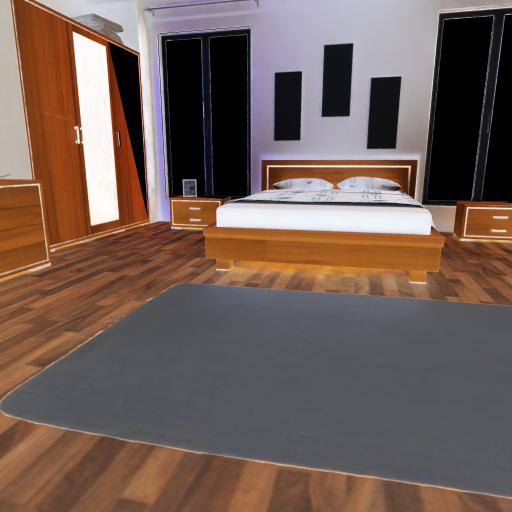}} &
    \begin{tabular}[b]{c}
    \fbox{\includegraphics[width=0.06\textwidth]{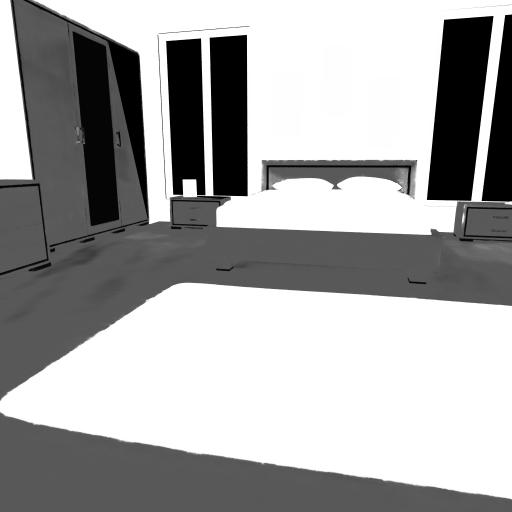}} \\
    \fbox{\includegraphics[width=0.06\textwidth]{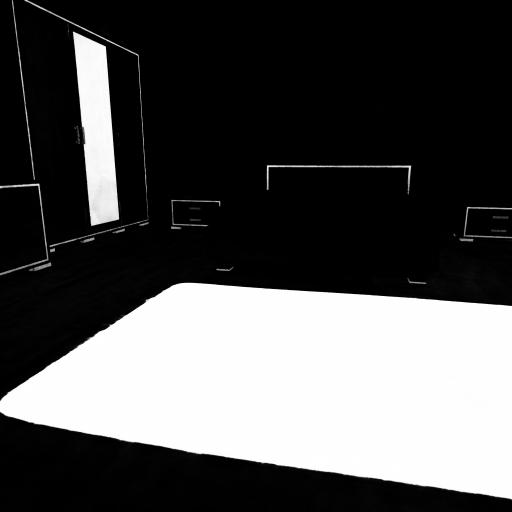}}
    \end{tabular}
    \end{tabular} &
\begin{tabular}[b]{cc}
    \fbox{\includegraphics[width=0.13\textwidth]{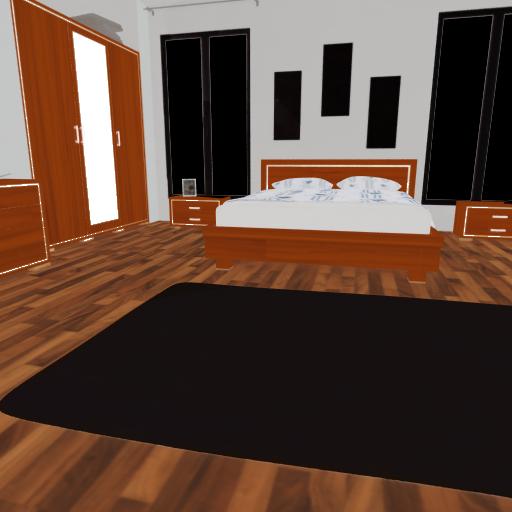}} &
    \begin{tabular}[b]{c}
    \fbox{\includegraphics[width=0.06\textwidth]{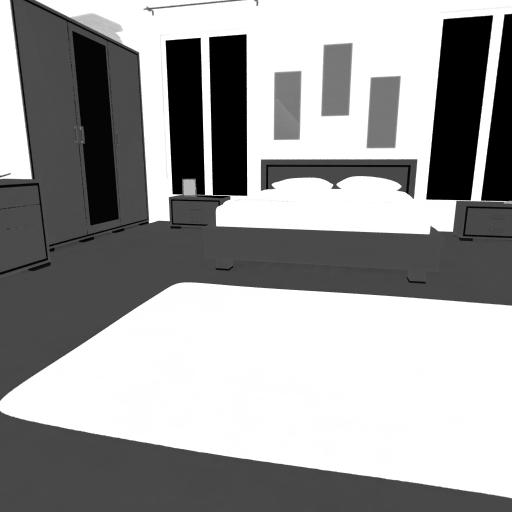}} \\
    \fbox{\includegraphics[width=0.06\textwidth]{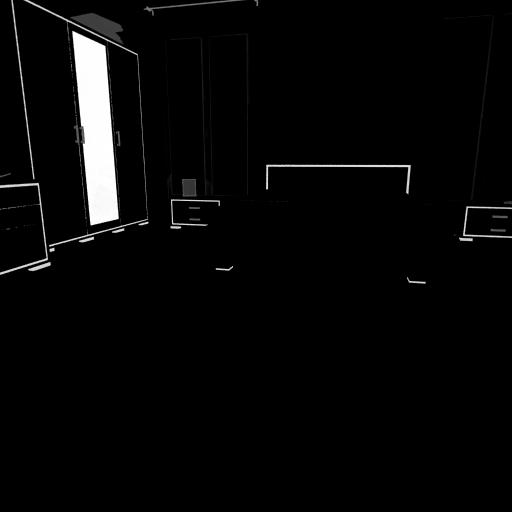}}
    \end{tabular}
    \end{tabular} &
\begin{tabular}[b]{c}
\fbox{\includegraphics[width=0.13\textwidth]{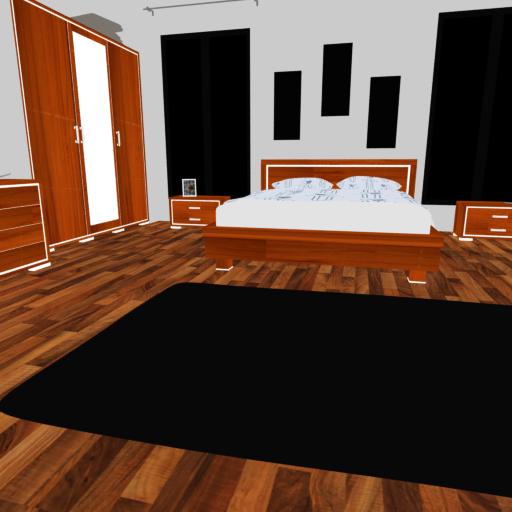}}
\end{tabular} \\
\begin{tabular}[b]{c}
\fbox{\includegraphics[width=0.13\textwidth]{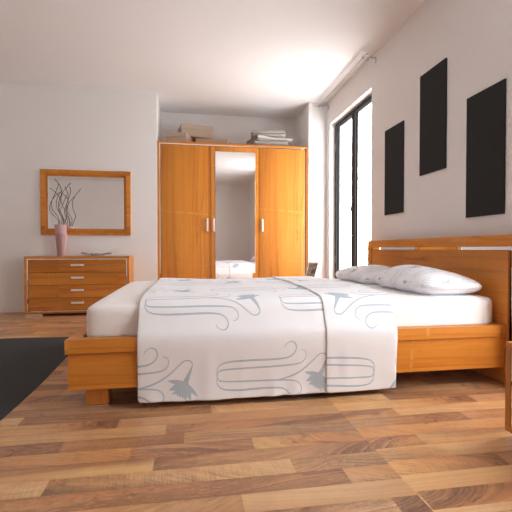}}
\end{tabular} &
\begin{tabular}[b]{cc}
    \fbox{\includegraphics[width=0.13\textwidth]{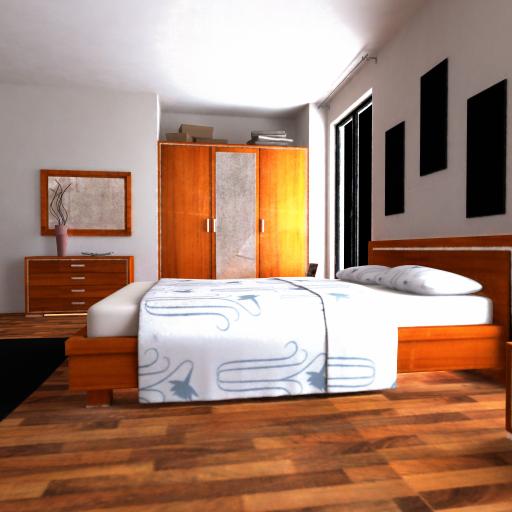}} &
    \begin{tabular}[b]{c}
    \fbox{\includegraphics[width=0.06\textwidth]{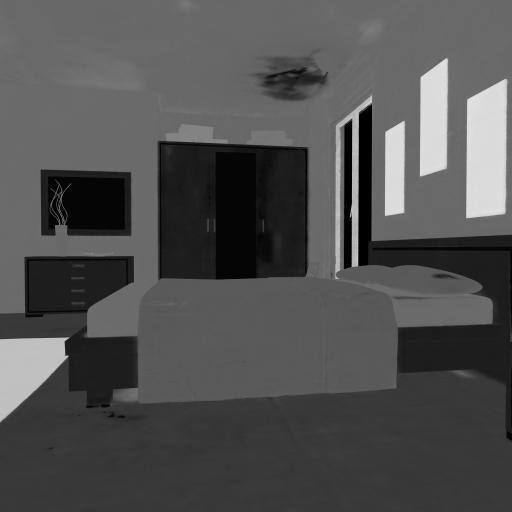}} \\
    \fbox{\includegraphics[width=0.06\textwidth]{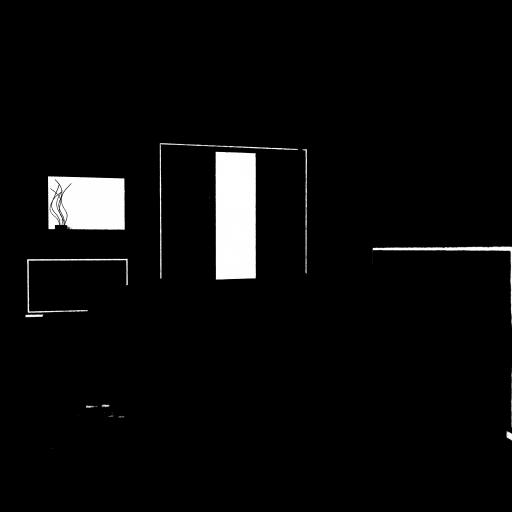}}
    \end{tabular}
    \end{tabular} &
\begin{tabular}[b]{cc}
    \fbox{\includegraphics[width=0.13\textwidth]{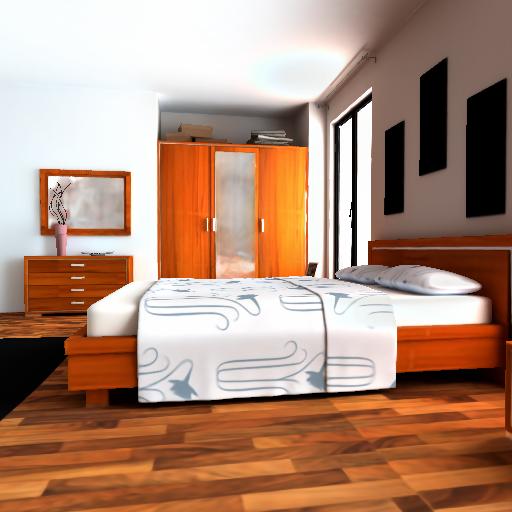}} &
    \begin{tabular}[b]{c}
    \fbox{\includegraphics[width=0.06\textwidth]{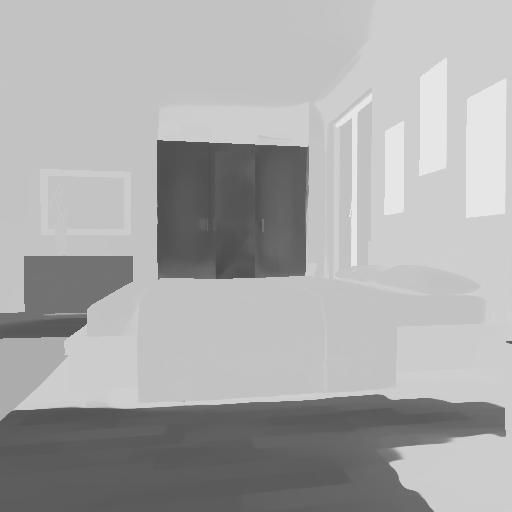}} \\
    \fbox{\includegraphics[width=0.06\textwidth]{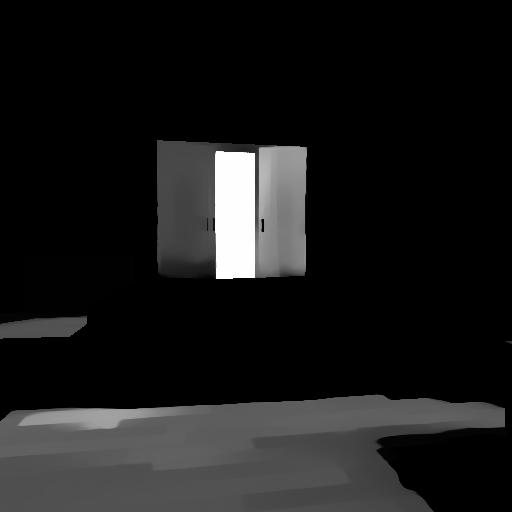}}
    \end{tabular}
    \end{tabular} &
\begin{tabular}[b]{cc}
    \fbox{\includegraphics[width=0.13\textwidth]{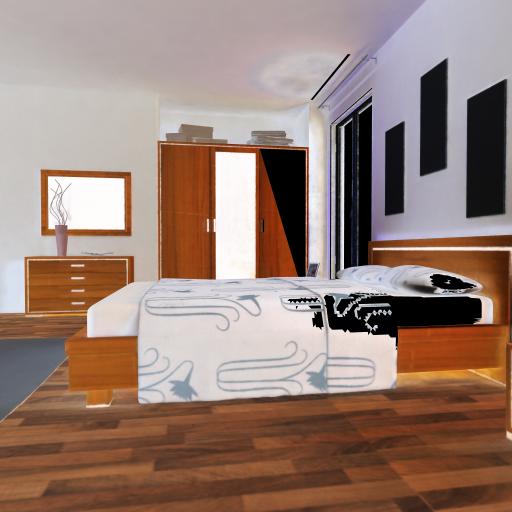}} &
    \begin{tabular}[b]{c}
    \fbox{\includegraphics[width=0.06\textwidth]{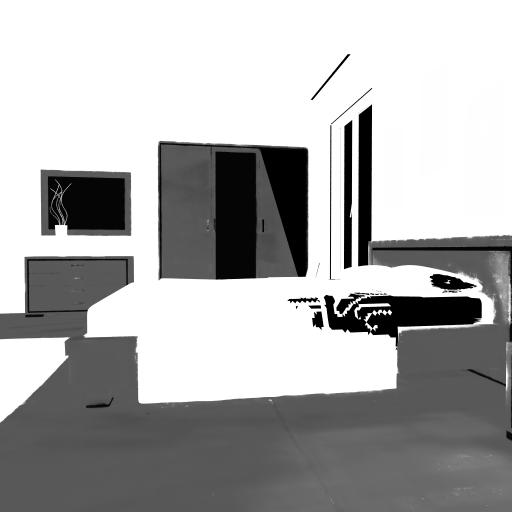}} \\
    \fbox{\includegraphics[width=0.06\textwidth]{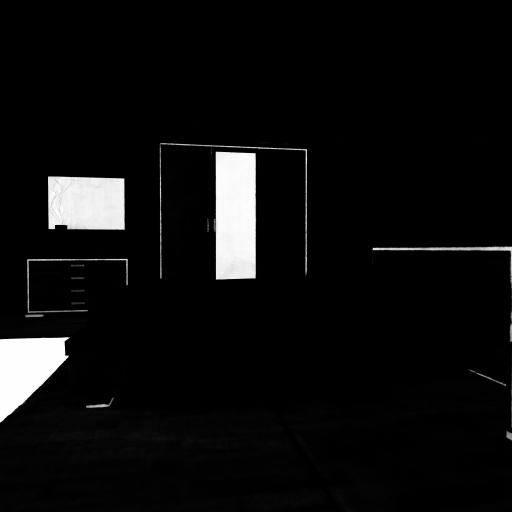}}
    \end{tabular}
    \end{tabular} &
\begin{tabular}[b]{cc}
    \fbox{\includegraphics[width=0.13\textwidth]{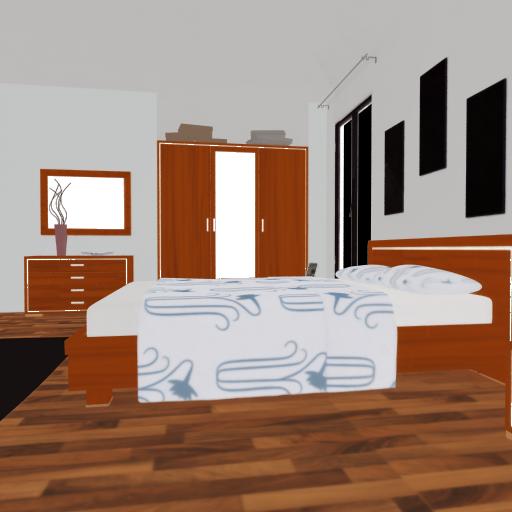}} &
    \begin{tabular}[b]{c}
    \fbox{\includegraphics[width=0.06\textwidth]{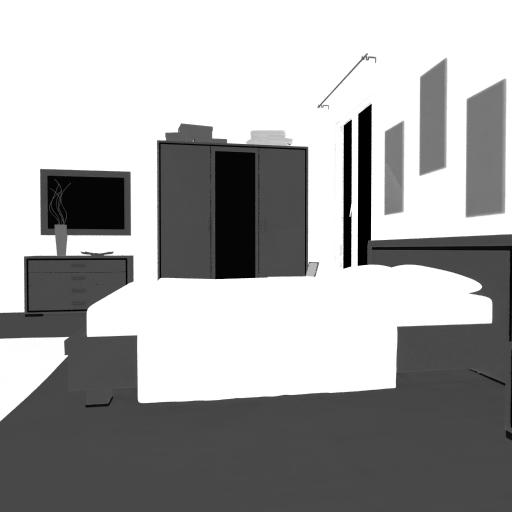}} \\
    \fbox{\includegraphics[width=0.06\textwidth]{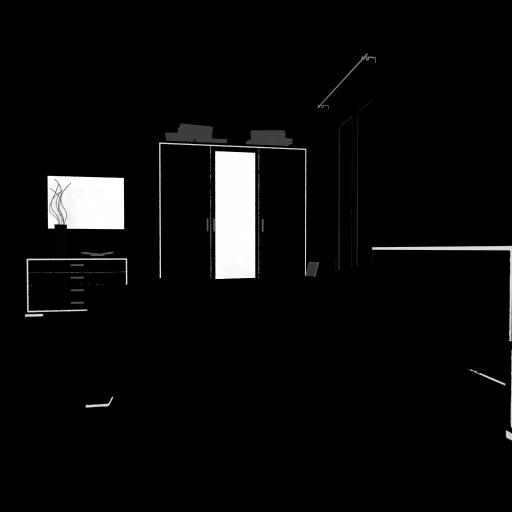}}
    \end{tabular}
    \end{tabular} &
\begin{tabular}[b]{c}
\fbox{\includegraphics[width=0.13\textwidth]{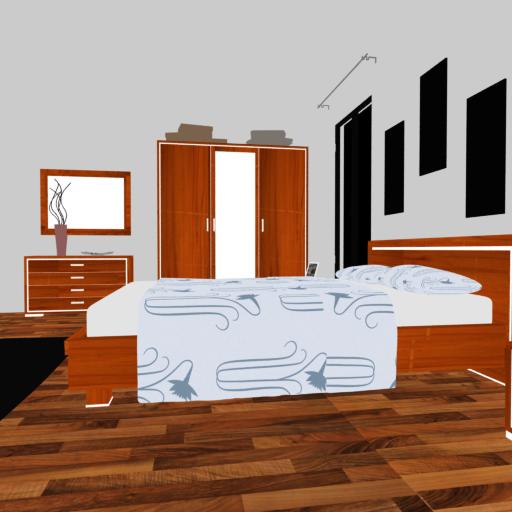}}
\end{tabular} \\
\hline
\begin{tabular}[b]{c}
\fbox{\includegraphics[width=0.13\textwidth]{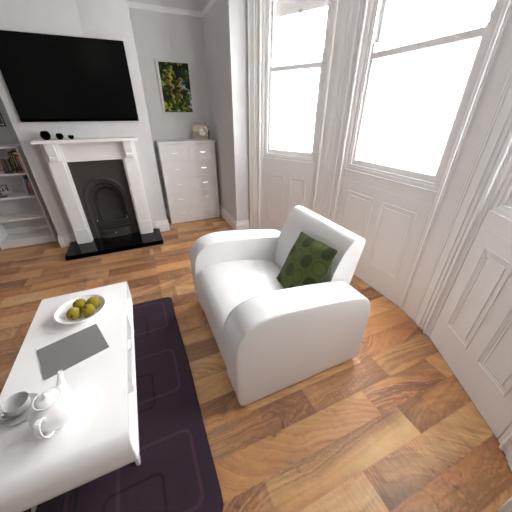}}
\end{tabular} &
\begin{tabular}[b]{cc}
    \fbox{\includegraphics[width=0.13\textwidth]{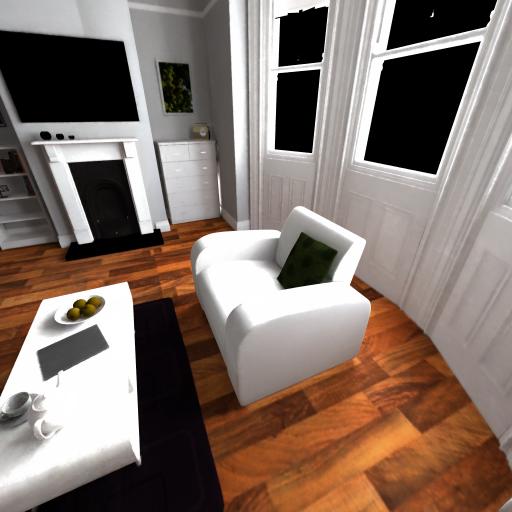}} &
    \begin{tabular}[b]{c}
    \fbox{\includegraphics[width=0.06\textwidth]{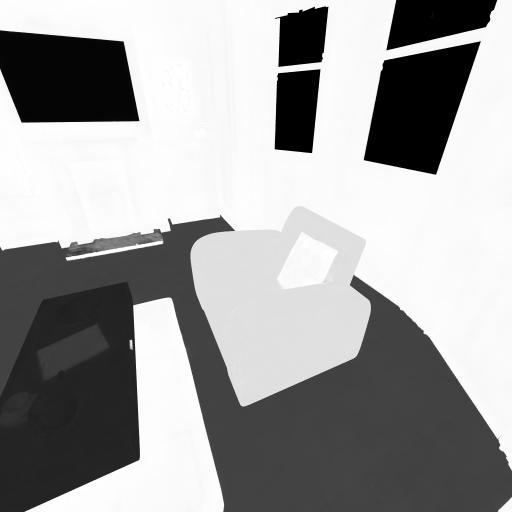}} \\
    \fbox{\includegraphics[width=0.06\textwidth]{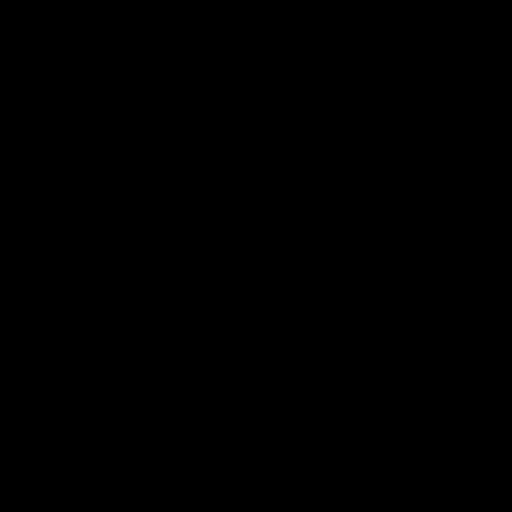}}
    \end{tabular}
    \end{tabular} &
\begin{tabular}[b]{cc}
    \fbox{\includegraphics[width=0.13\textwidth]{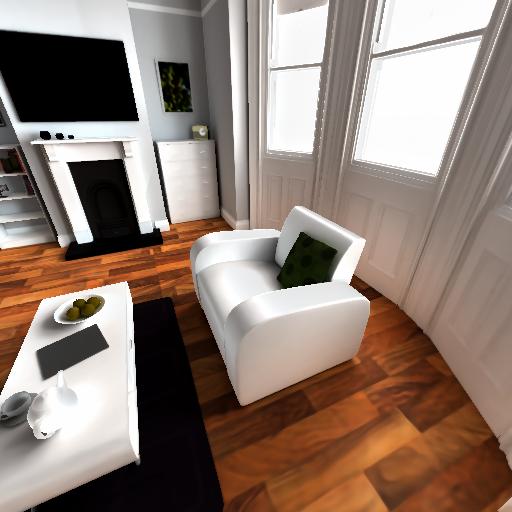}} &
    \begin{tabular}[b]{c}
    \fbox{\includegraphics[width=0.06\textwidth]{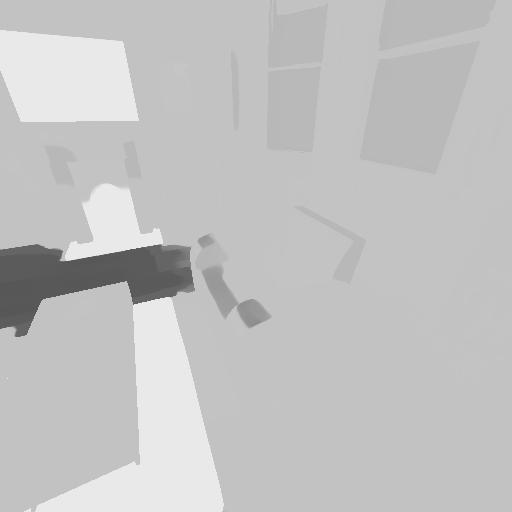}} \\
    \fbox{\includegraphics[width=0.06\textwidth]{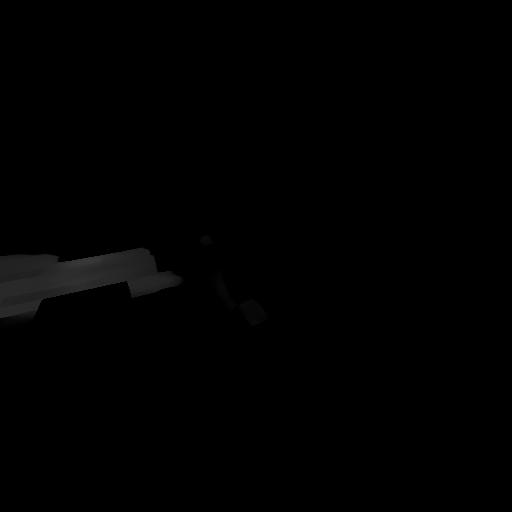}}
    \end{tabular}
    \end{tabular} &
\begin{tabular}[b]{cc}
    \fbox{\includegraphics[width=0.13\textwidth]{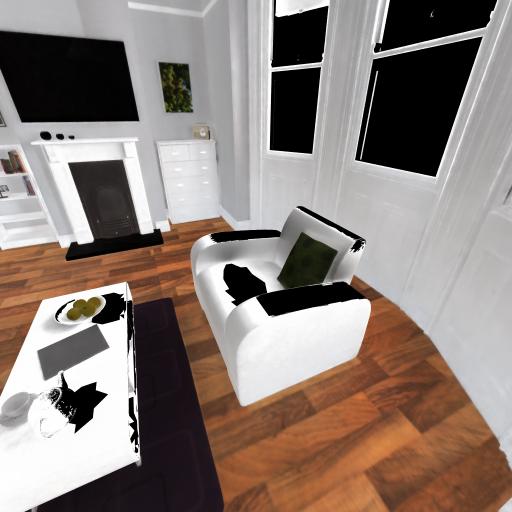}} &
    \begin{tabular}[b]{c}
    \fbox{\includegraphics[width=0.06\textwidth]{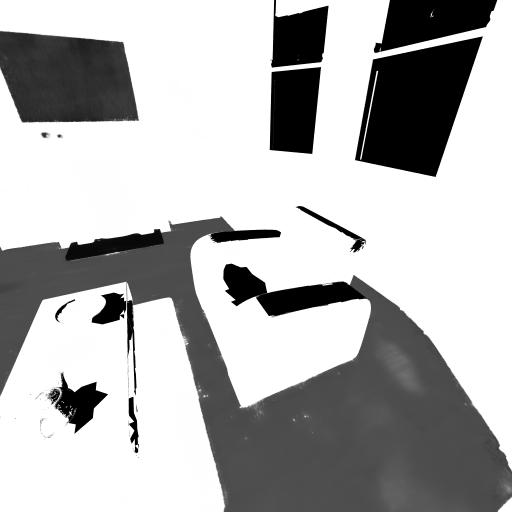}} \\
    \fbox{\includegraphics[width=0.06\textwidth]{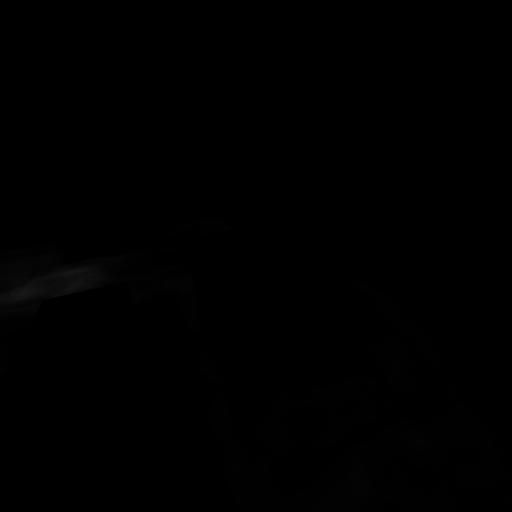}}
    \end{tabular}
    \end{tabular} &
\begin{tabular}[b]{cc}
    \fbox{\includegraphics[width=0.13\textwidth]{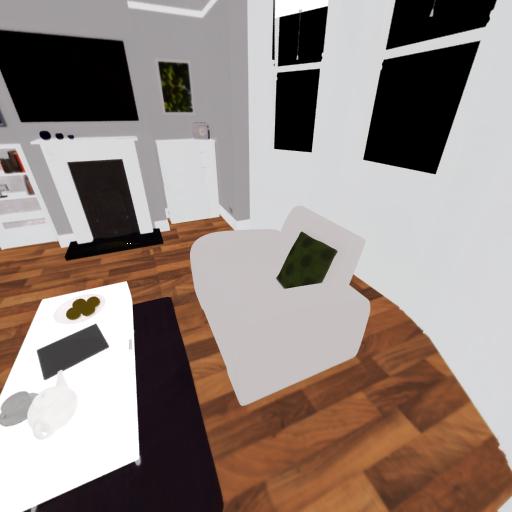}} &
    \begin{tabular}[b]{c}
    \fbox{\includegraphics[width=0.06\textwidth]{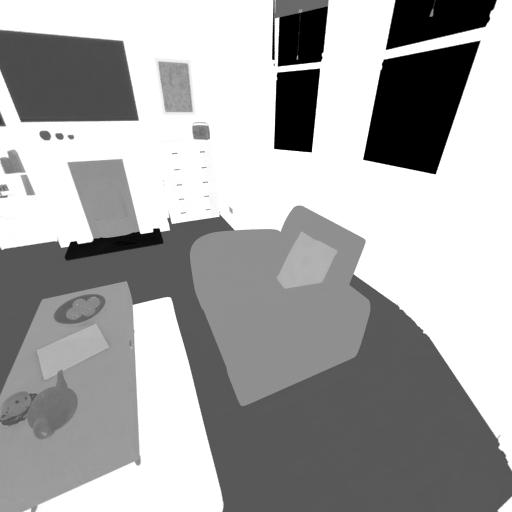}} \\
    \fbox{\includegraphics[width=0.06\textwidth]{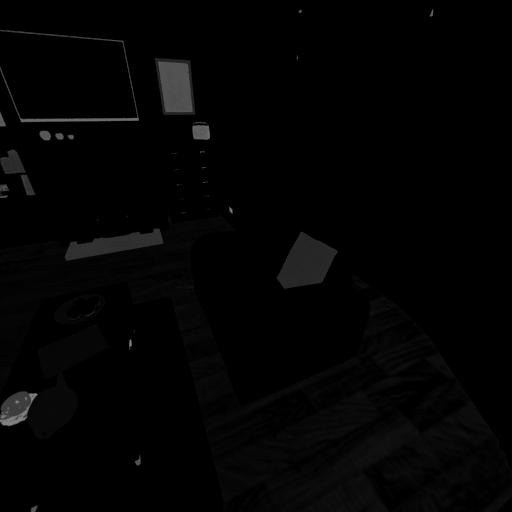}}
    \end{tabular}
    \end{tabular} &
\begin{tabular}[b]{c}
\fbox{\includegraphics[width=0.13\textwidth]{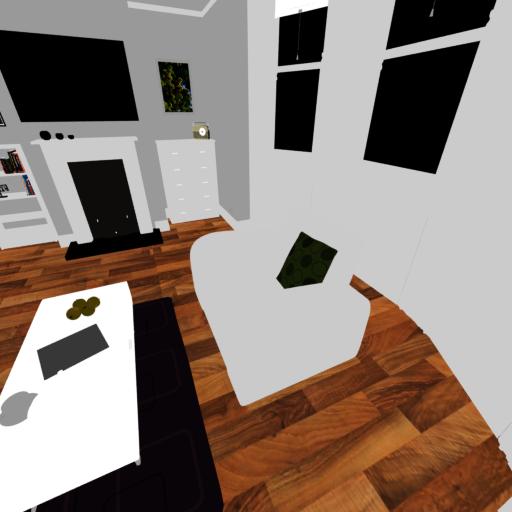}}
\end{tabular} \\
\begin{tabular}[b]{c}
\fbox{\includegraphics[width=0.13\textwidth]{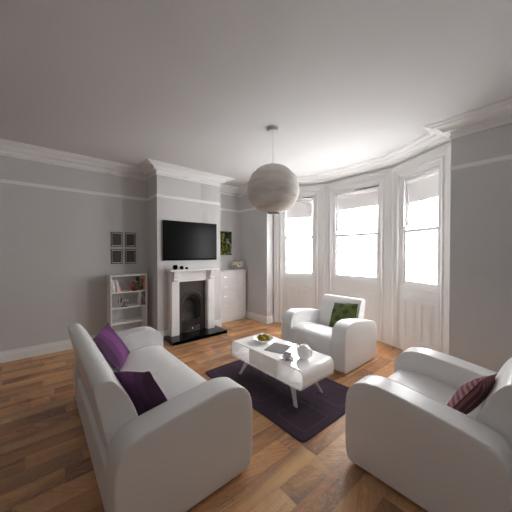}}
\end{tabular} &
\begin{tabular}[b]{cc}
    \fbox{\includegraphics[width=0.13\textwidth]{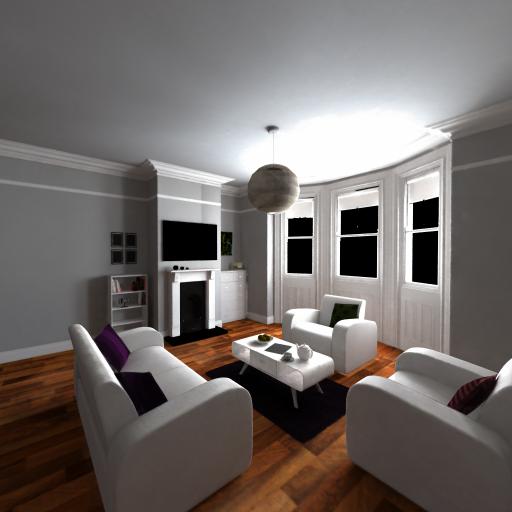}} &
    \begin{tabular}[b]{c}
    \fbox{\includegraphics[width=0.06\textwidth]{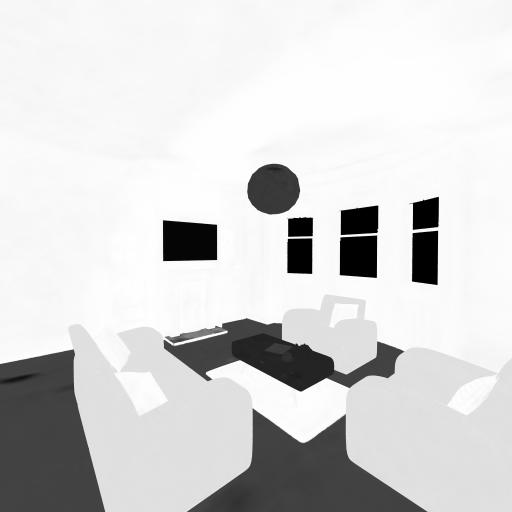}} \\
    \fbox{\includegraphics[width=0.06\textwidth]{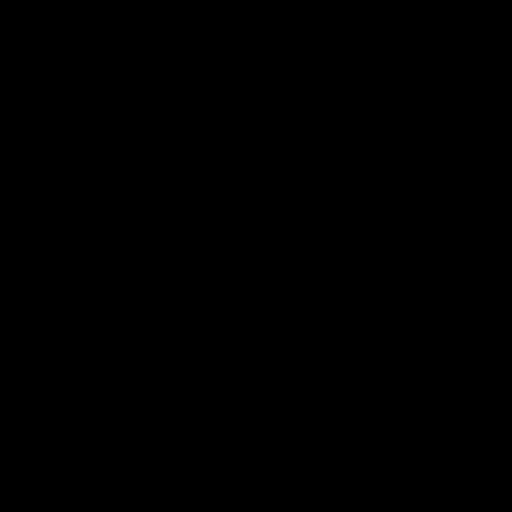}}
    \end{tabular}
    \end{tabular} &
\begin{tabular}[b]{cc}
    \fbox{\includegraphics[width=0.13\textwidth]{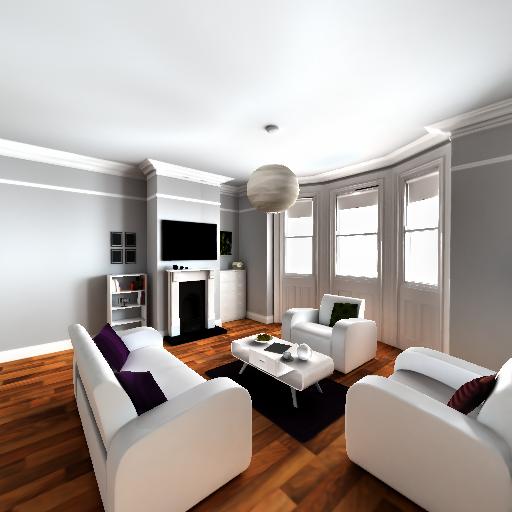}} &
    \begin{tabular}[b]{c}
    \fbox{\includegraphics[width=0.06\textwidth]{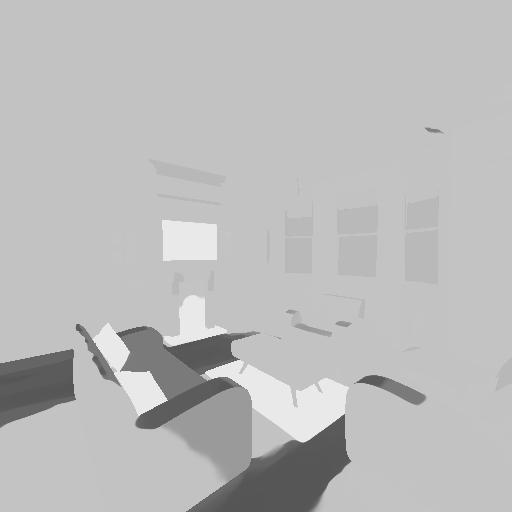}} \\
    \fbox{\includegraphics[width=0.06\textwidth]{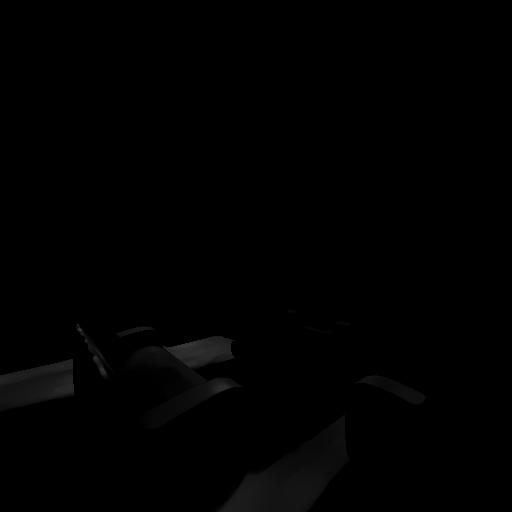}}
    \end{tabular}
    \end{tabular} &
\begin{tabular}[b]{cc}
    \fbox{\includegraphics[width=0.13\textwidth]{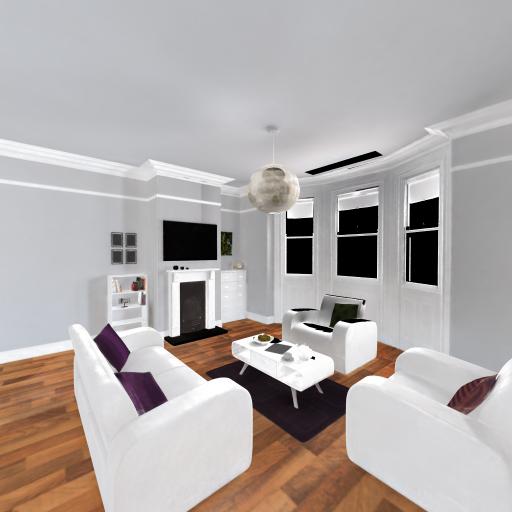}} &
    \begin{tabular}[b]{c}
    \fbox{\includegraphics[width=0.06\textwidth]{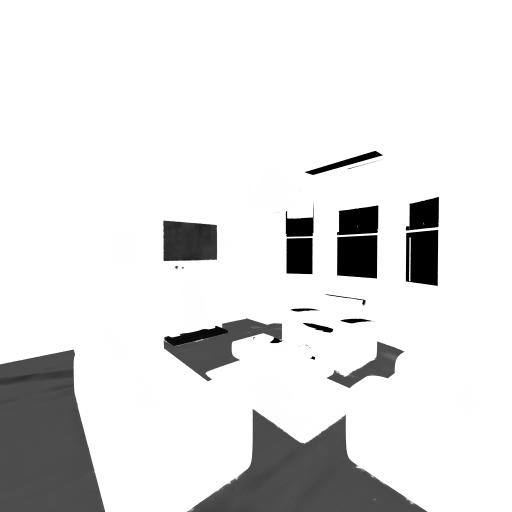}} \\
    \fbox{\includegraphics[width=0.06\textwidth]{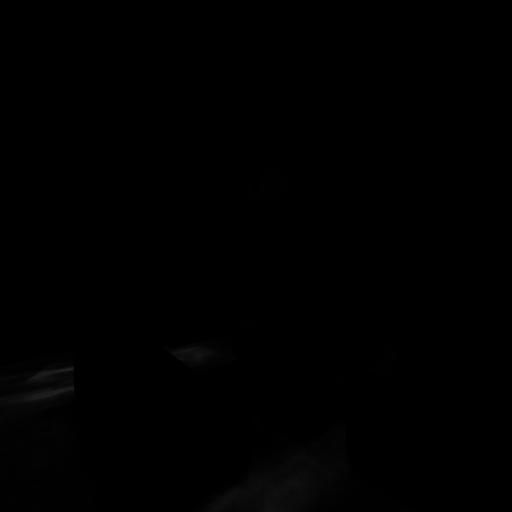}}
    \end{tabular}
    \end{tabular} &
\begin{tabular}[b]{cc}
    \fbox{\includegraphics[width=0.13\textwidth]{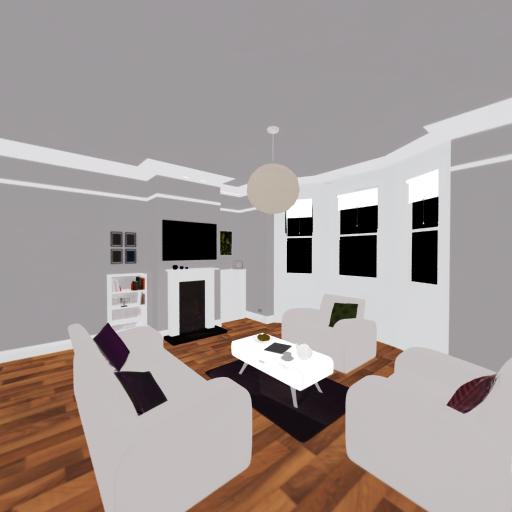}} &
    \begin{tabular}[b]{c}
    \fbox{\includegraphics[width=0.06\textwidth]{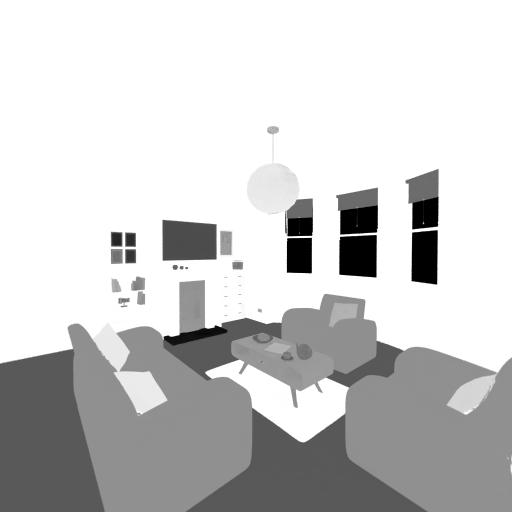}} \\
    \fbox{\includegraphics[width=0.06\textwidth]{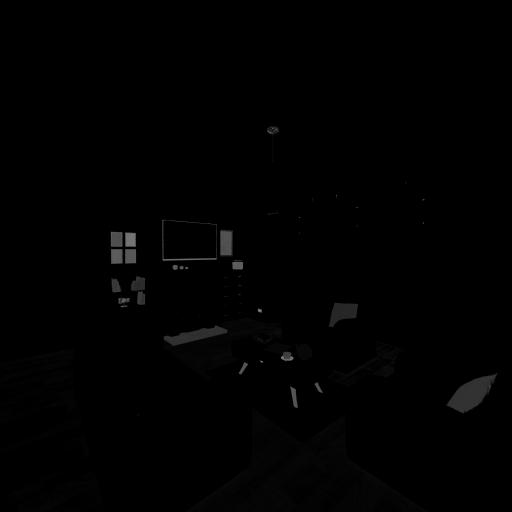}}
    \end{tabular}
    \end{tabular} &
\begin{tabular}[b]{c}
\fbox{\includegraphics[width=0.13\textwidth]{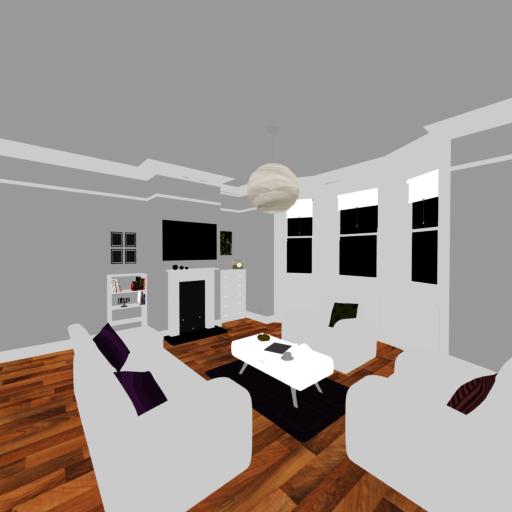}}
\end{tabular} \\
\hline
\begin{tabular}[b]{c}
\fbox{\includegraphics[width=0.13\textwidth]{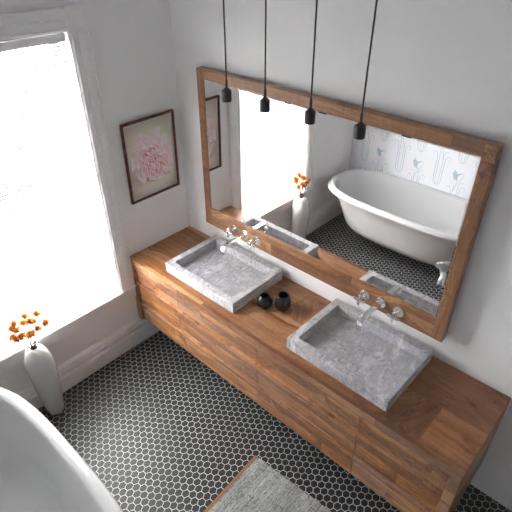}}
\end{tabular} &
\begin{tabular}[b]{cc}
    \fbox{\includegraphics[width=0.13\textwidth]{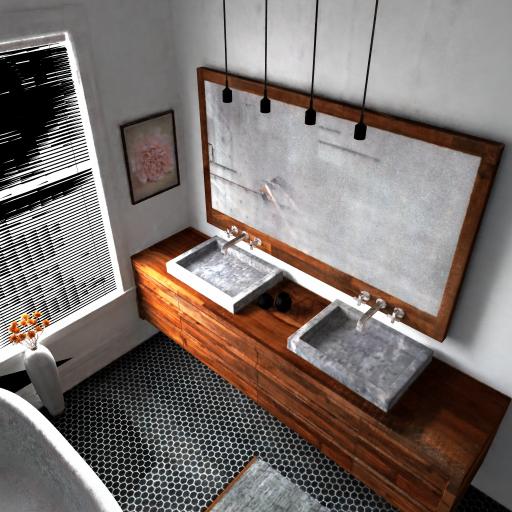}} &
    \begin{tabular}[b]{c}
    \fbox{\includegraphics[width=0.06\textwidth]{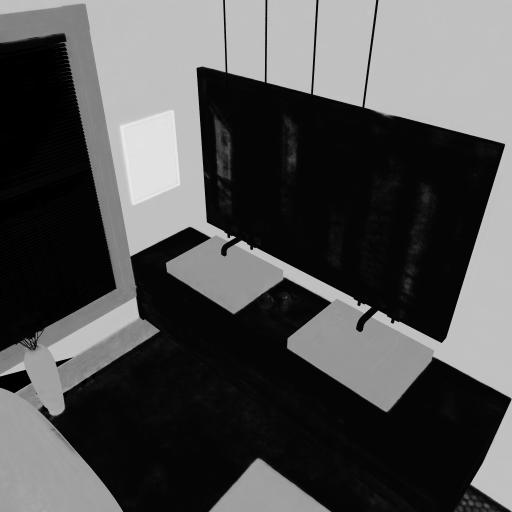}} \\
    \fbox{\includegraphics[width=0.06\textwidth]{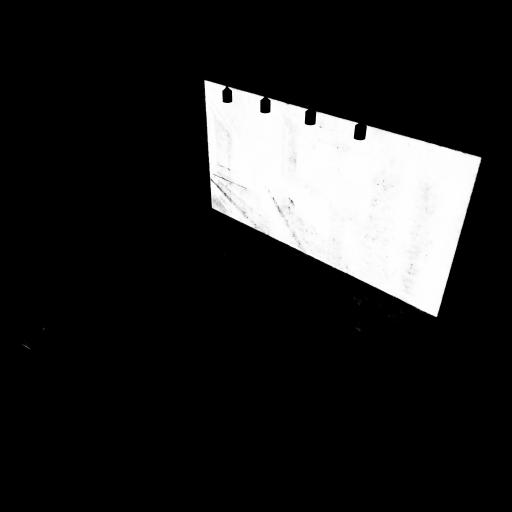}}
    \end{tabular}
    \end{tabular} &
\begin{tabular}[b]{cc}
    \fbox{\includegraphics[width=0.13\textwidth]{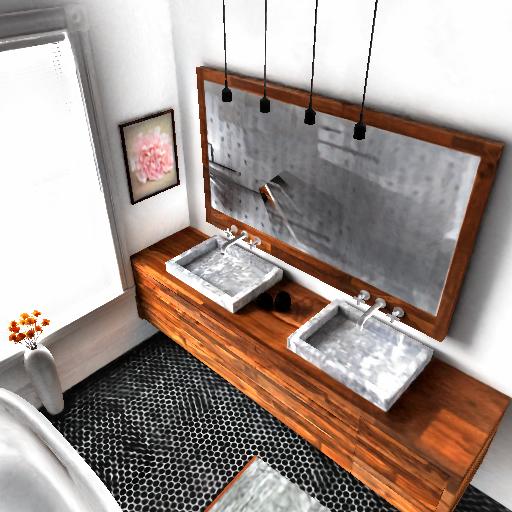}} &
    \begin{tabular}[b]{c}
    \fbox{\includegraphics[width=0.06\textwidth]{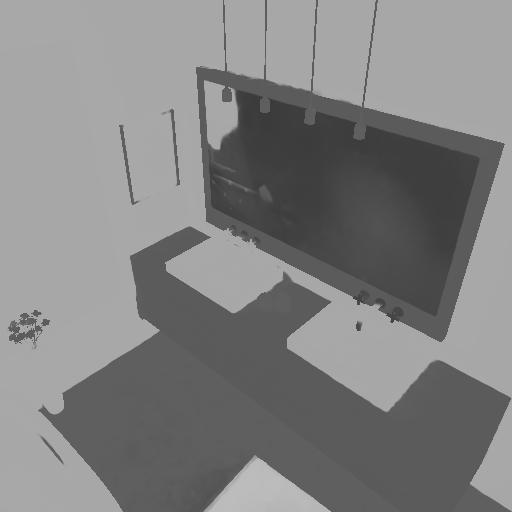}} \\
    \fbox{\includegraphics[width=0.06\textwidth]{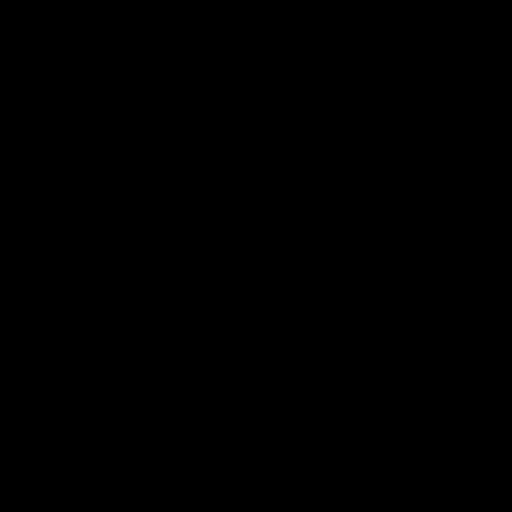}}
    \end{tabular}
    \end{tabular} &
\begin{tabular}[b]{cc}
    \fbox{\includegraphics[width=0.13\textwidth]{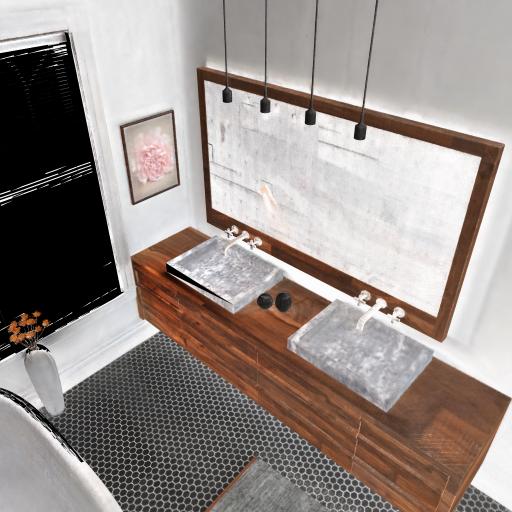}} &
    \begin{tabular}[b]{c}
    \fbox{\includegraphics[width=0.06\textwidth]{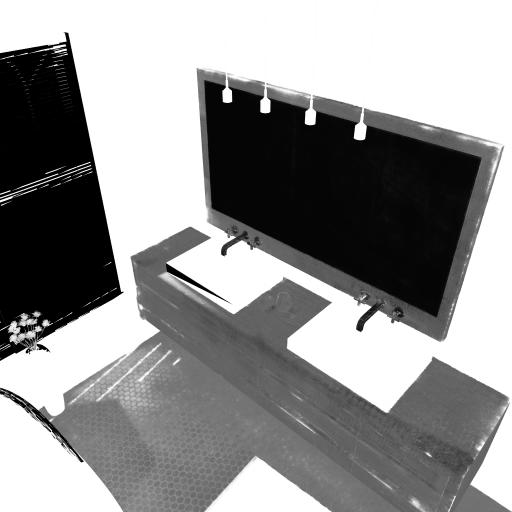}} \\
    \fbox{\includegraphics[width=0.06\textwidth]{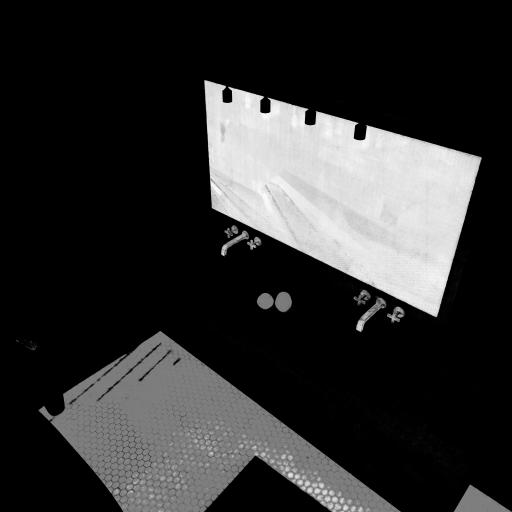}}
    \end{tabular}
    \end{tabular} &
\begin{tabular}[b]{cc}
    \fbox{\includegraphics[width=0.13\textwidth]{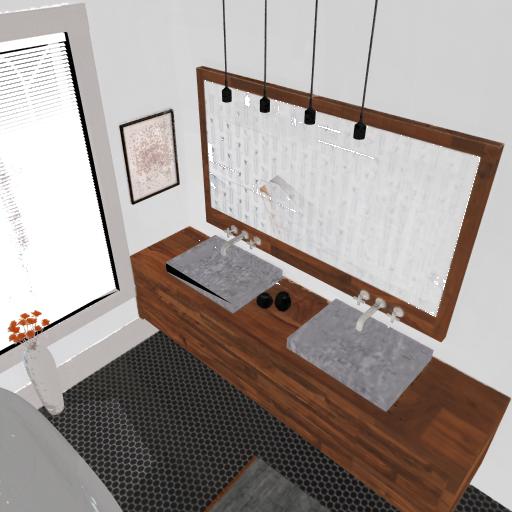}} &
    \begin{tabular}[b]{c}
    \fbox{\includegraphics[width=0.06\textwidth]{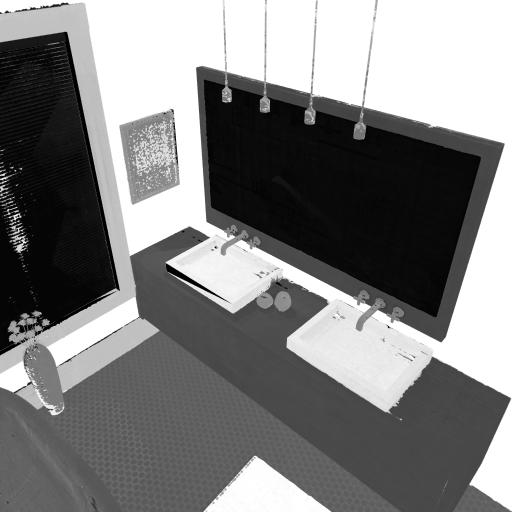}} \\
    \fbox{\includegraphics[width=0.06\textwidth]{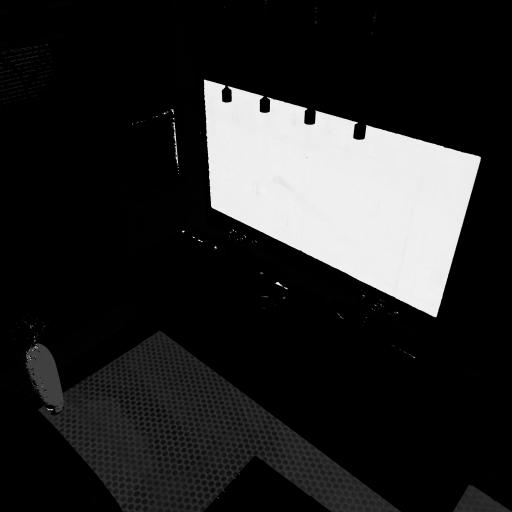}}
    \end{tabular}
    \end{tabular} &
\begin{tabular}[b]{c}
\fbox{\includegraphics[width=0.13\textwidth]{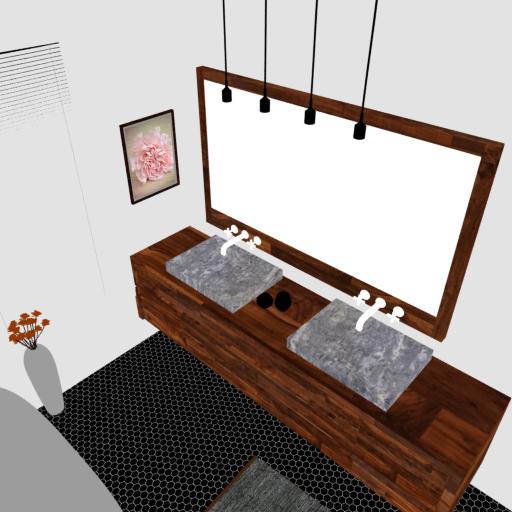}}
\end{tabular} \\
\begin{tabular}[b]{c}
\fbox{\includegraphics[width=0.13\textwidth]{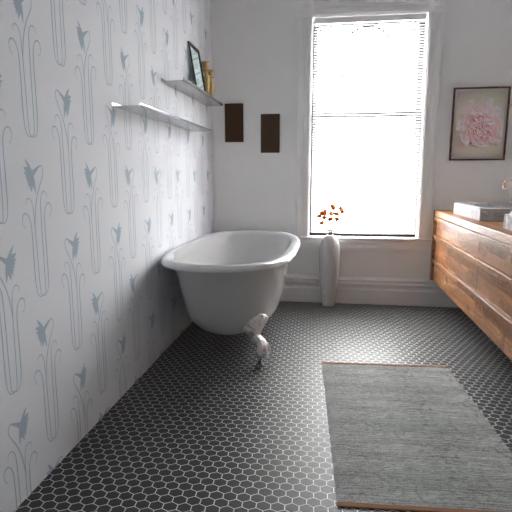}}
\end{tabular} &
\begin{tabular}[b]{cc}
    \fbox{\includegraphics[width=0.13\textwidth]{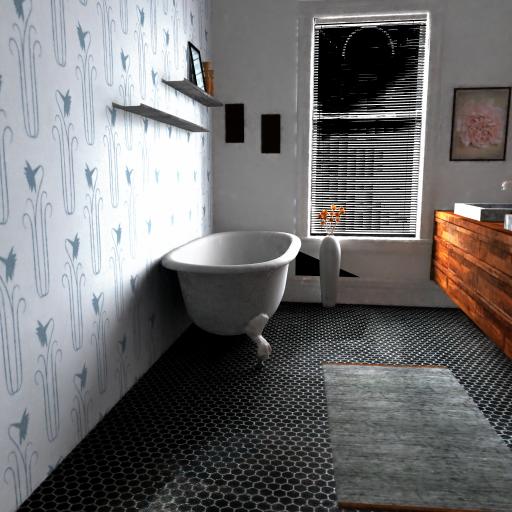}} &
    \begin{tabular}[b]{c}
    \fbox{\includegraphics[width=0.06\textwidth]{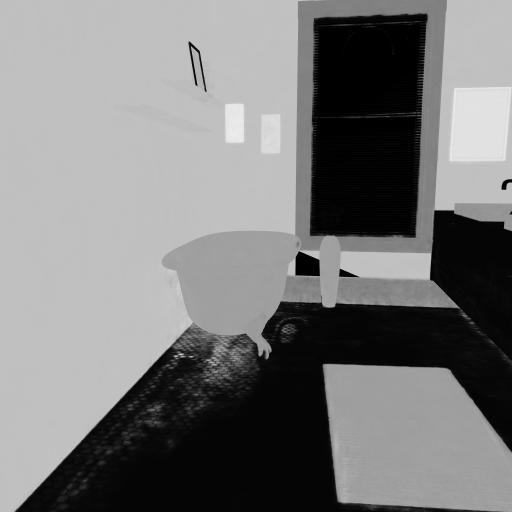}} \\
    \fbox{\includegraphics[width=0.06\textwidth]{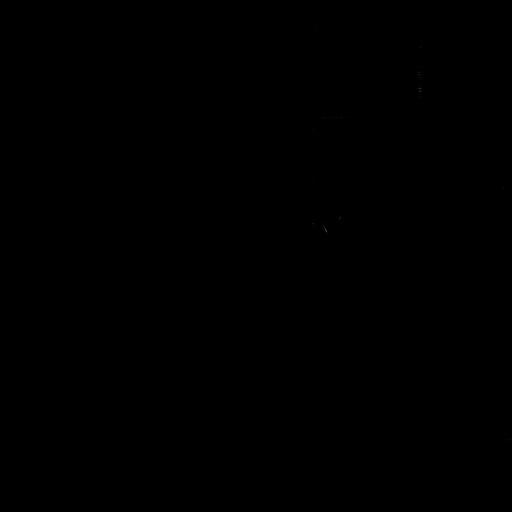}}
    \end{tabular}
    \end{tabular} &
\begin{tabular}[b]{cc}
    \fbox{\includegraphics[width=0.13\textwidth]{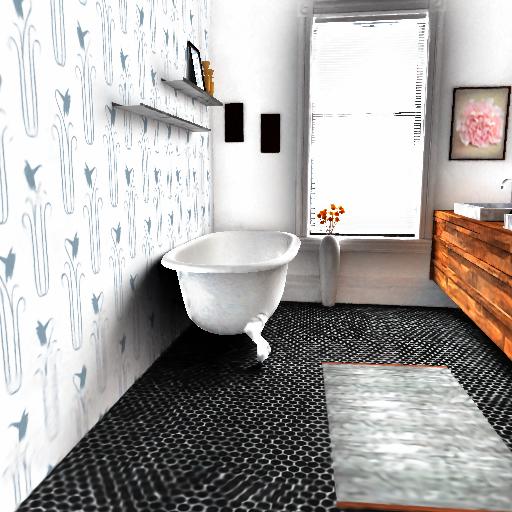}} &
    \begin{tabular}[b]{c}
    \fbox{\includegraphics[width=0.06\textwidth]{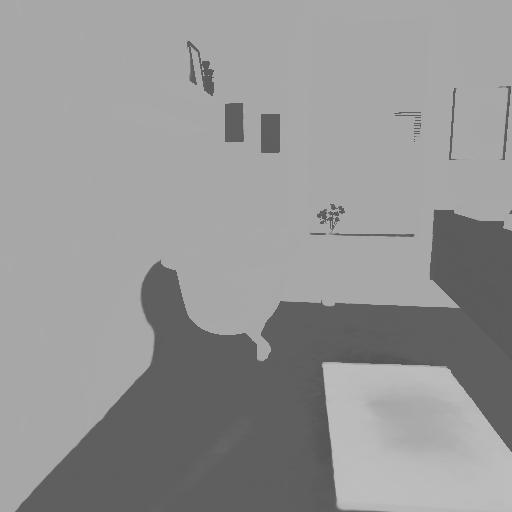}} \\
    \fbox{\includegraphics[width=0.06\textwidth]{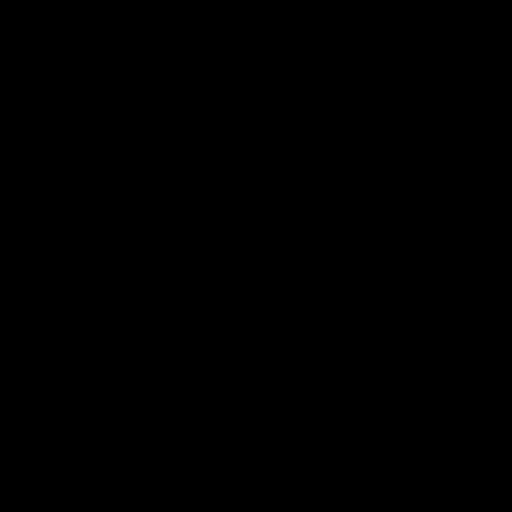}}
    \end{tabular}
    \end{tabular} &
\begin{tabular}[b]{cc}
    \fbox{\includegraphics[width=0.13\textwidth]{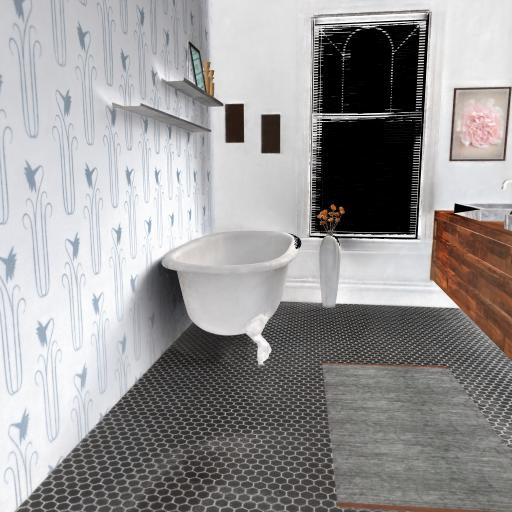}} &
    \begin{tabular}[b]{c}
    \fbox{\includegraphics[width=0.06\textwidth]{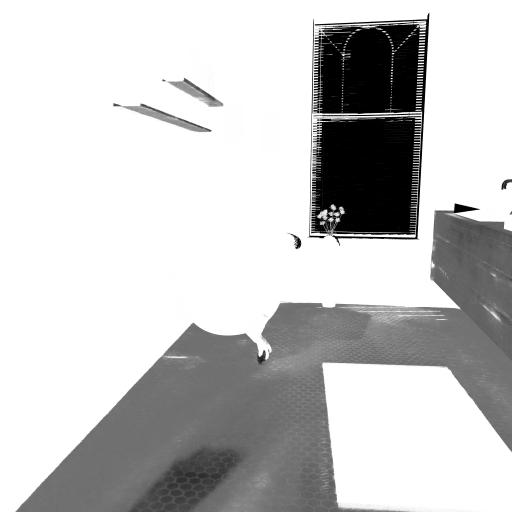}} \\
    \fbox{\includegraphics[width=0.06\textwidth]{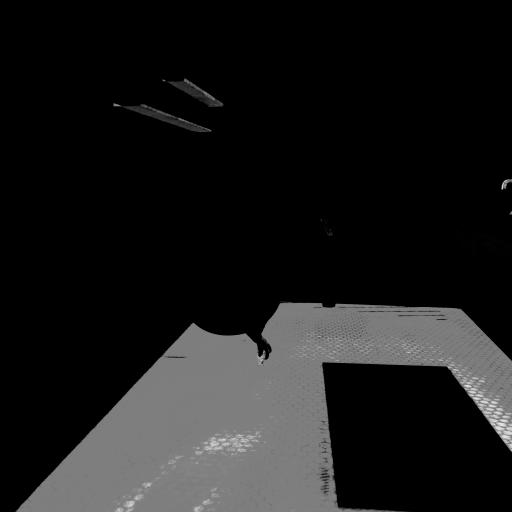}}
    \end{tabular}
    \end{tabular} &
\begin{tabular}[b]{cc}
    \fbox{\includegraphics[width=0.13\textwidth]{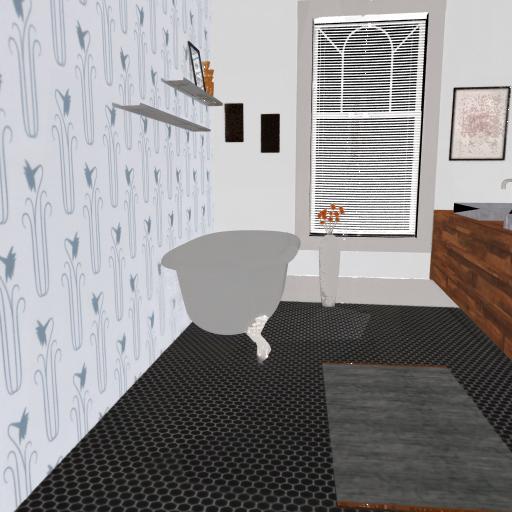}} &
    \begin{tabular}[b]{c}
    \fbox{\includegraphics[width=0.06\textwidth]{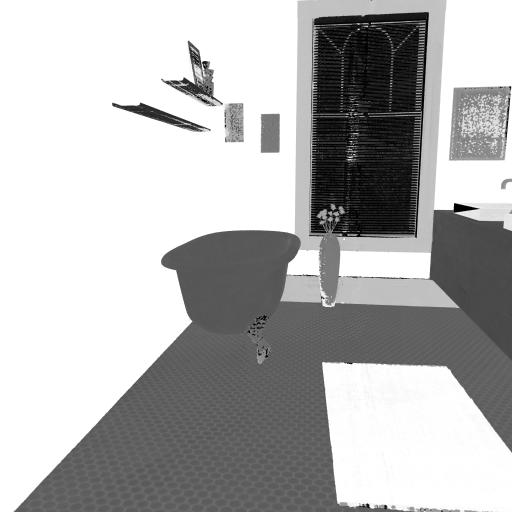}} \\
    \fbox{\includegraphics[width=0.06\textwidth]{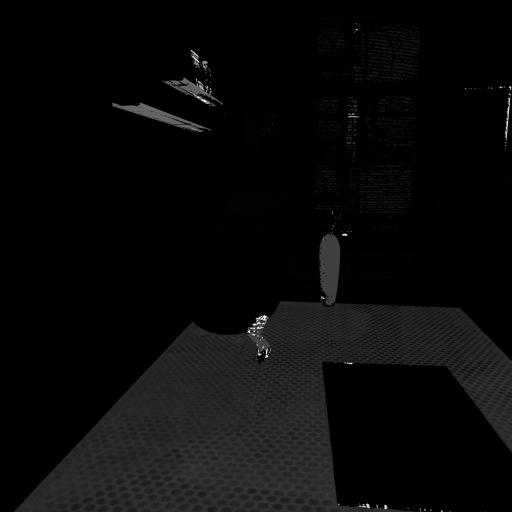}}
    \end{tabular}
    \end{tabular} &
\begin{tabular}[b]{c}
\fbox{\includegraphics[width=0.13\textwidth]{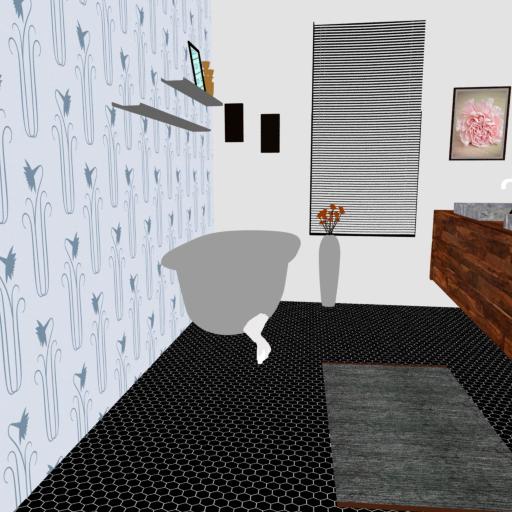}}
\end{tabular} \\
\hline
        % Header row at bottom
        \smash{\scriptsize RGB} &
        \smash{\scriptsize FIPT \cite{FIPT}} &
        \smash{\scriptsize NeILF++ \cite{NeilfPP}} &
        \smash{\scriptsize IRIS \cite{IRIS}} &
        \smash{\scriptsize IIF (Ours)} &
        \smash{\scriptsize GT}
    \end{tabular}}
    \caption{\textbf{Synthetic comparisons.} Additional samples on the synthetic scenes.}
    \label{fig:supp:synthetic_comparisons}
\end{figure*}

\begin{figure*}[t]
    \centering
    \setlength\tabcolsep{1.25pt}
    \resizebox{\textwidth}{!}{
    \fboxsep=0pt
    \begin{tabular}{c|cccc}
\begin{tabular}[b]{c}
\fbox{\includegraphics[width=0.13\textwidth]{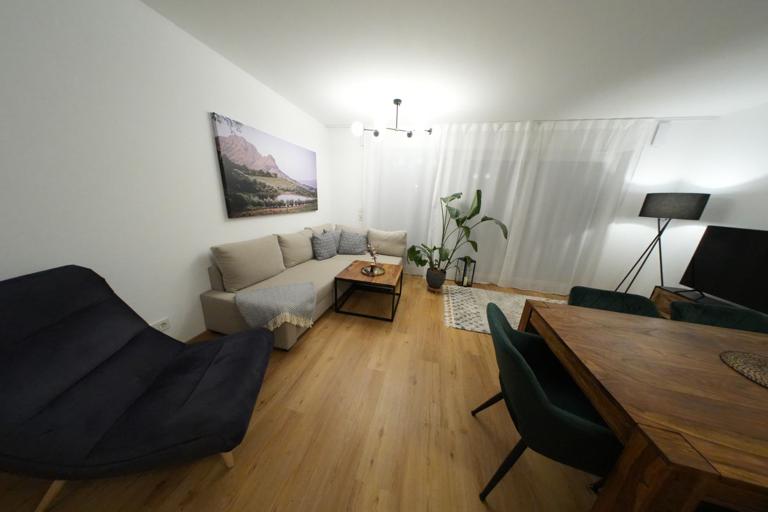}}
\end{tabular} &
\begin{tabular}[b]{cc}
    \fbox{\includegraphics[width=0.13\textwidth]{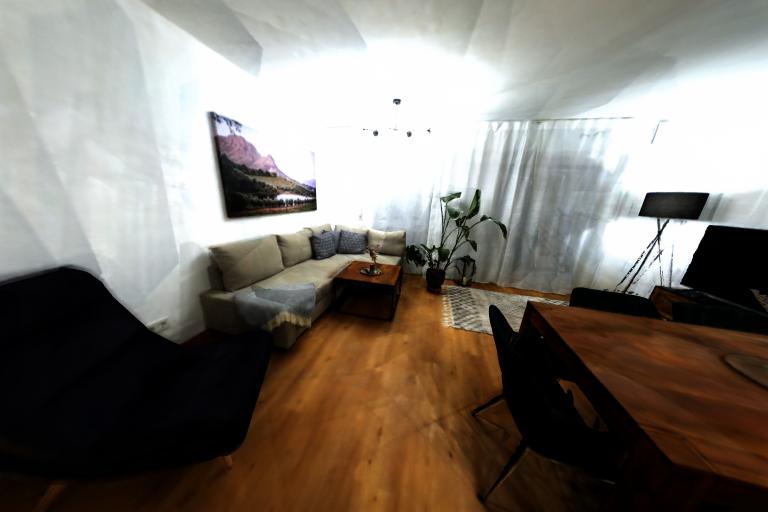}} &
    \begin{tabular}[b]{c}
    \fbox{\includegraphics[width=0.06\textwidth]{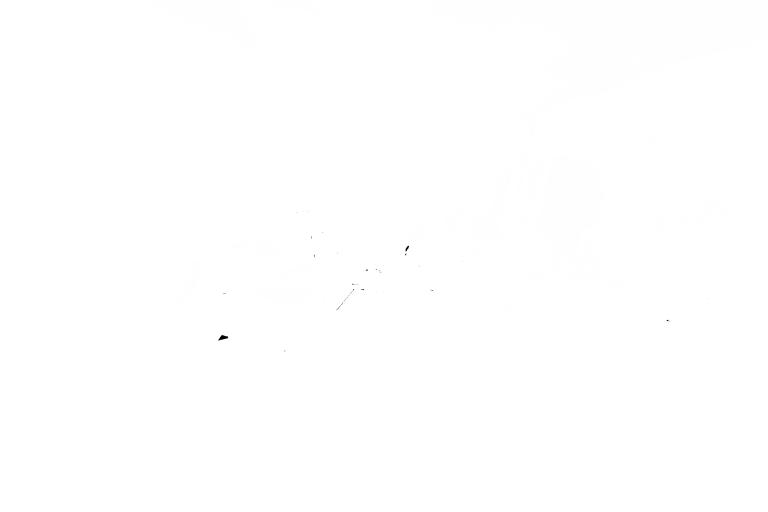}} \\
    \fbox{\includegraphics[width=0.06\textwidth]{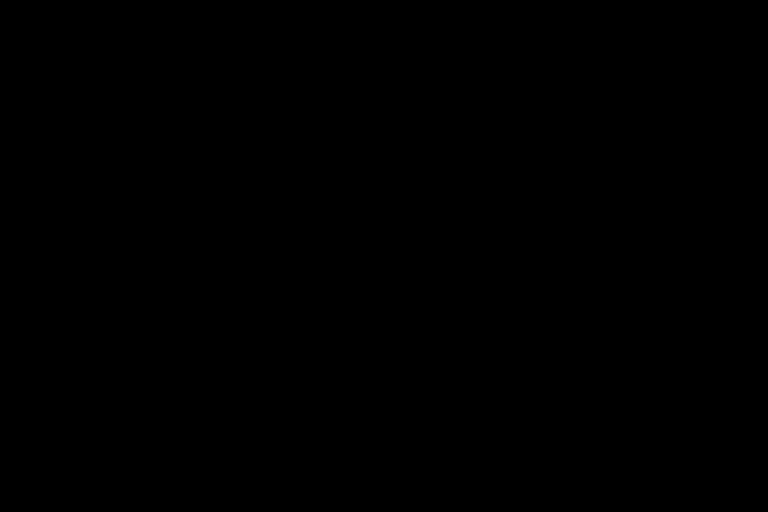}}
    \end{tabular}
    \end{tabular} &
\begin{tabular}[b]{cc}
    \fbox{\includegraphics[width=0.13\textwidth]{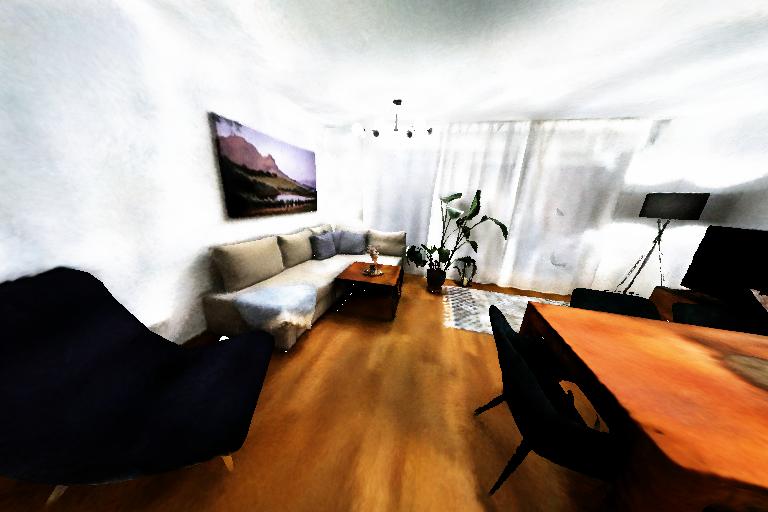}} &
    \begin{tabular}[b]{c}
    \fbox{\includegraphics[width=0.06\textwidth]{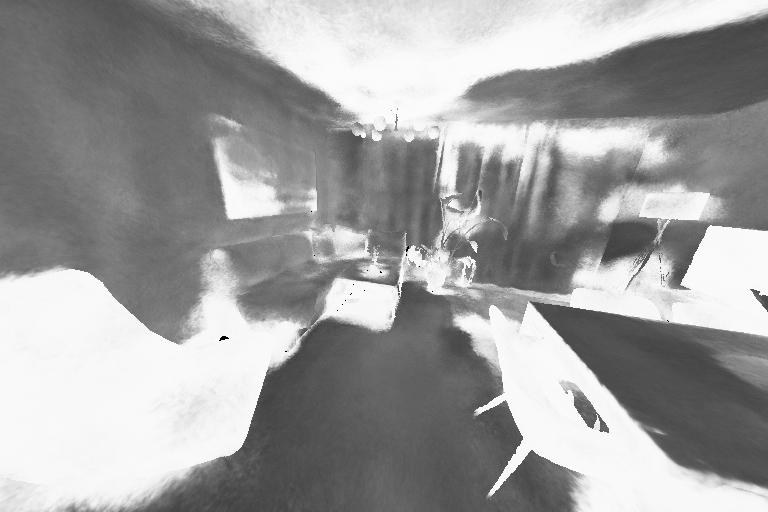}} \\
    \fbox{\includegraphics[width=0.06\textwidth]{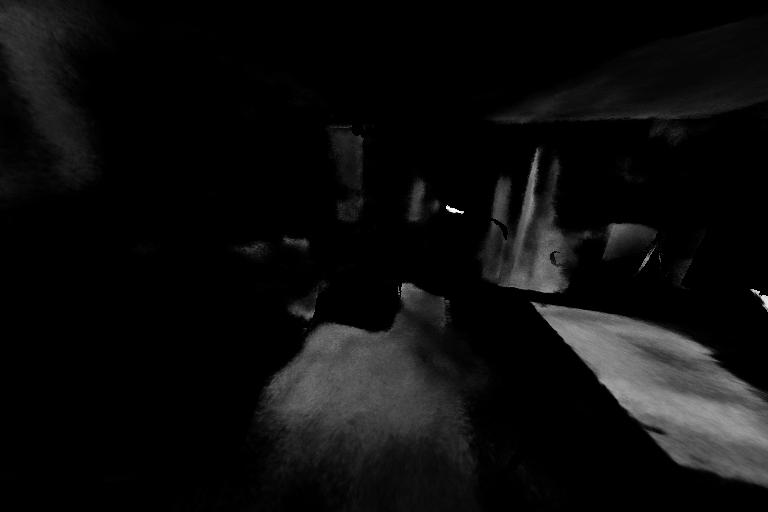}}
    \end{tabular}
    \end{tabular} &
\begin{tabular}[b]{cc}
    \fbox{\includegraphics[width=0.13\textwidth]{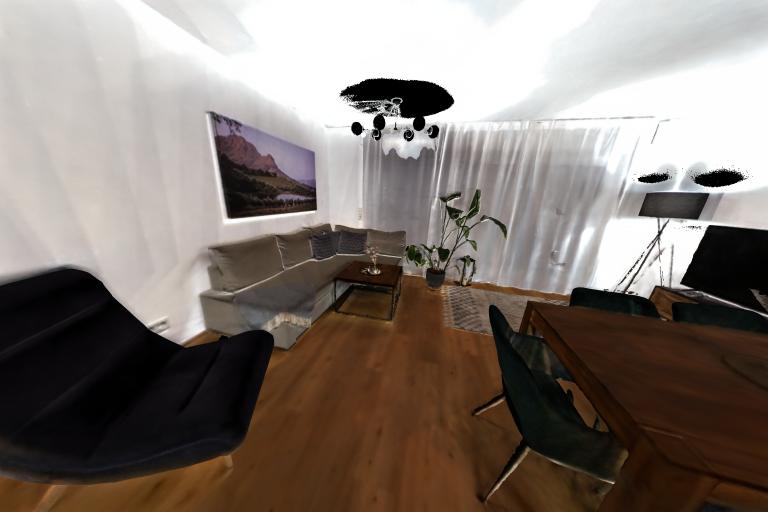}} &
    \begin{tabular}[b]{c}
    \fbox{\includegraphics[width=0.06\textwidth]{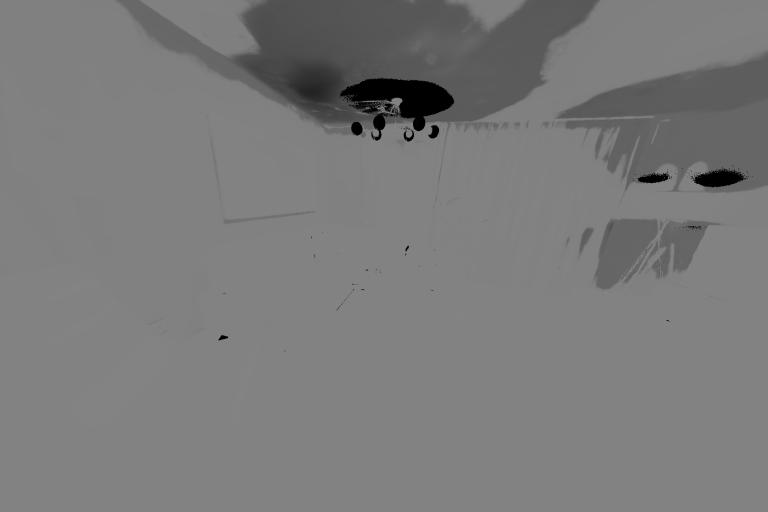}} \\
    \fbox{\includegraphics[width=0.06\textwidth]{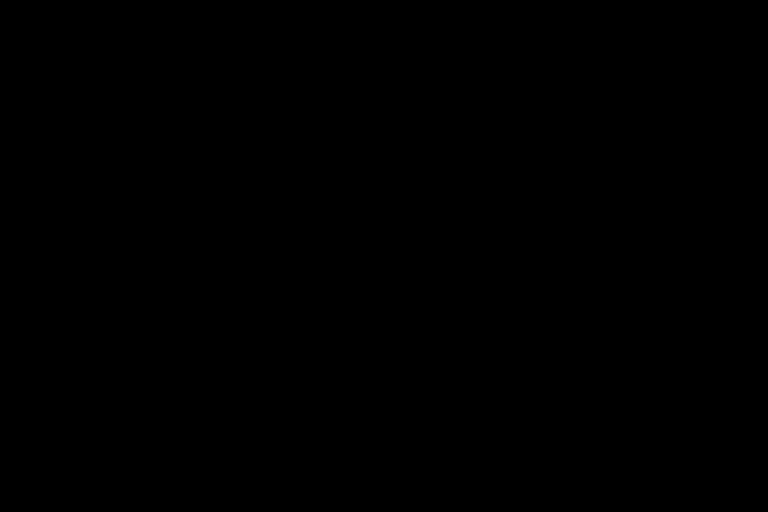}}
    \end{tabular}
    \end{tabular} &
\begin{tabular}[b]{cc}
    \fbox{\includegraphics[width=0.13\textwidth]{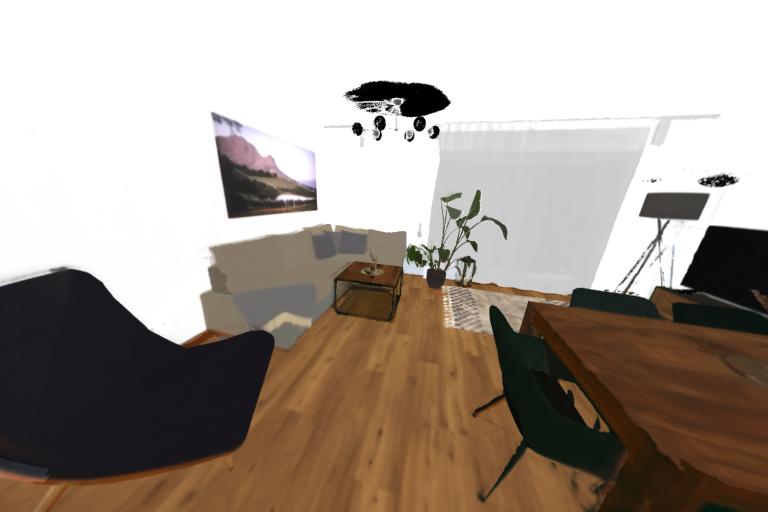}} &
    \begin{tabular}[b]{c}
    \fbox{\includegraphics[width=0.06\textwidth]{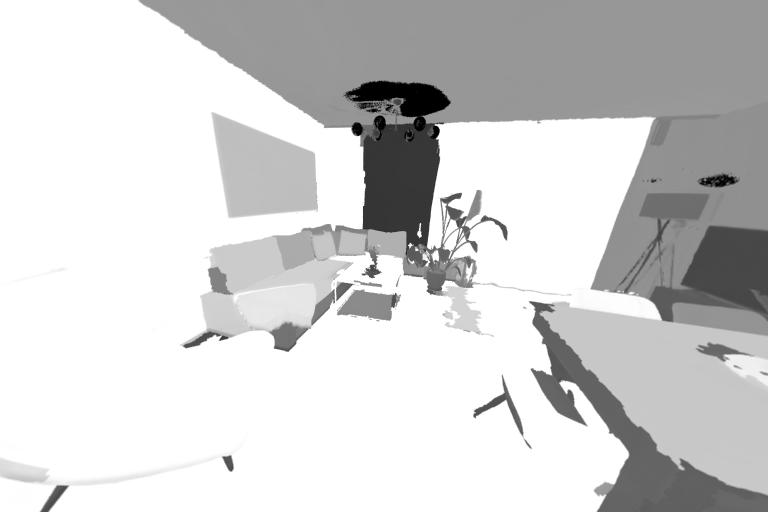}} \\
    \fbox{\includegraphics[width=0.06\textwidth]{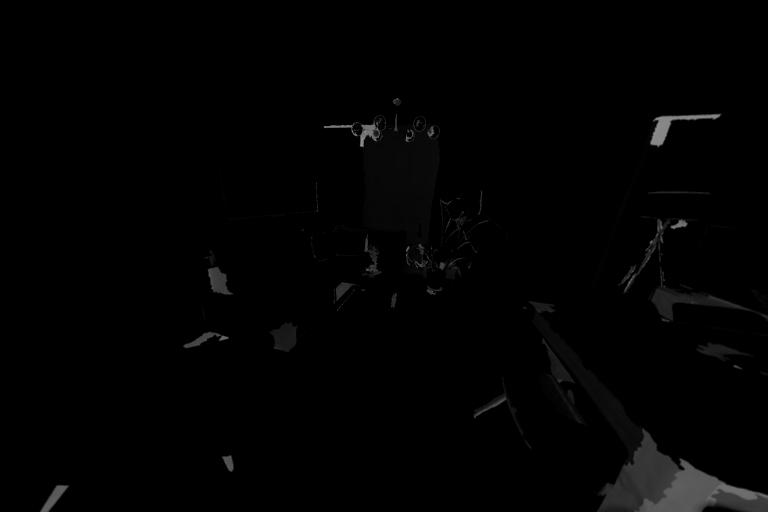}}
    \end{tabular}
    \end{tabular} \\
\begin{tabular}[b]{c}
\fbox{\includegraphics[width=0.13\textwidth]{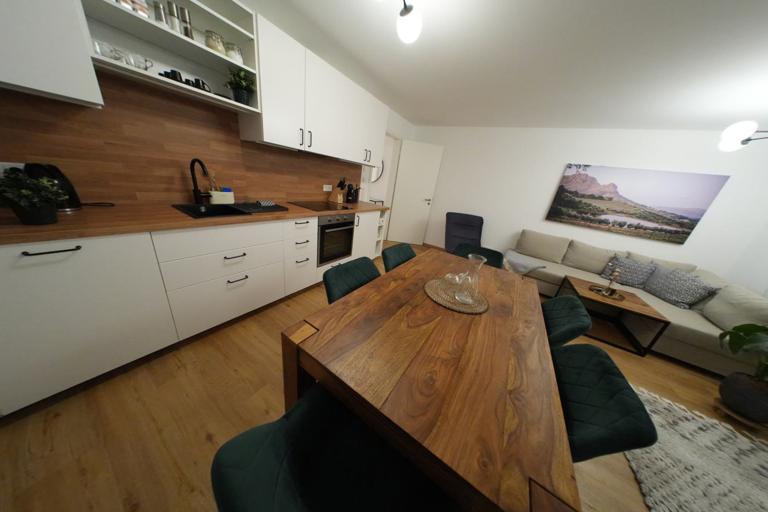}}
\end{tabular} &
\begin{tabular}[b]{cc}
    \fbox{\includegraphics[width=0.13\textwidth]{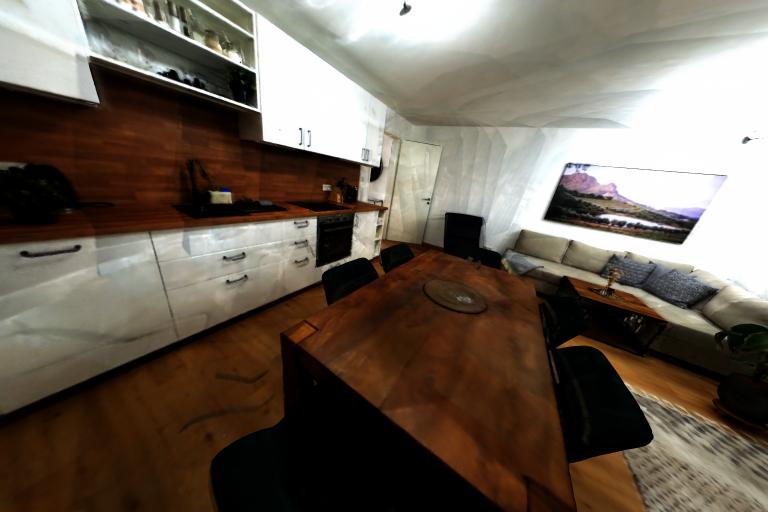}} &
    \begin{tabular}[b]{c}
    \fbox{\includegraphics[width=0.06\textwidth]{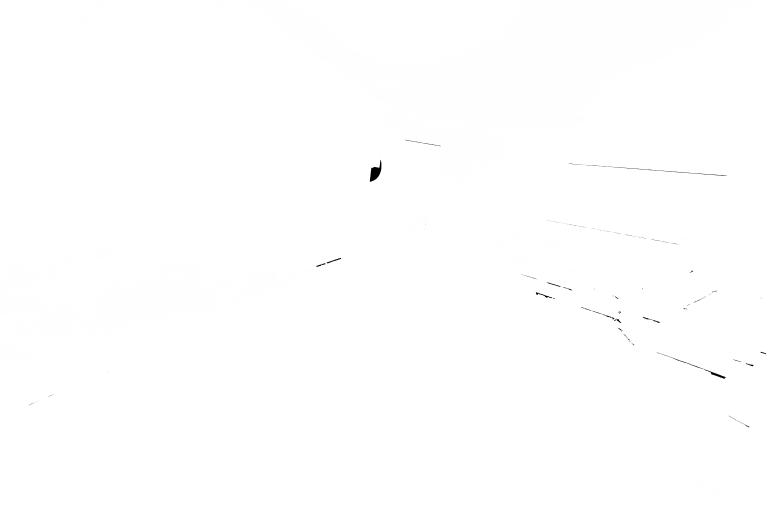}} \\
    \fbox{\includegraphics[width=0.06\textwidth]{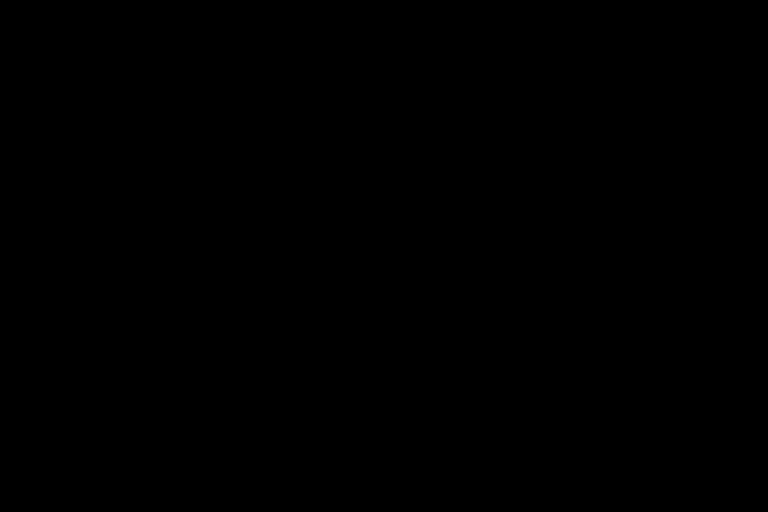}}
    \end{tabular}
    \end{tabular} &
\begin{tabular}[b]{cc}
    \fbox{\includegraphics[width=0.13\textwidth]{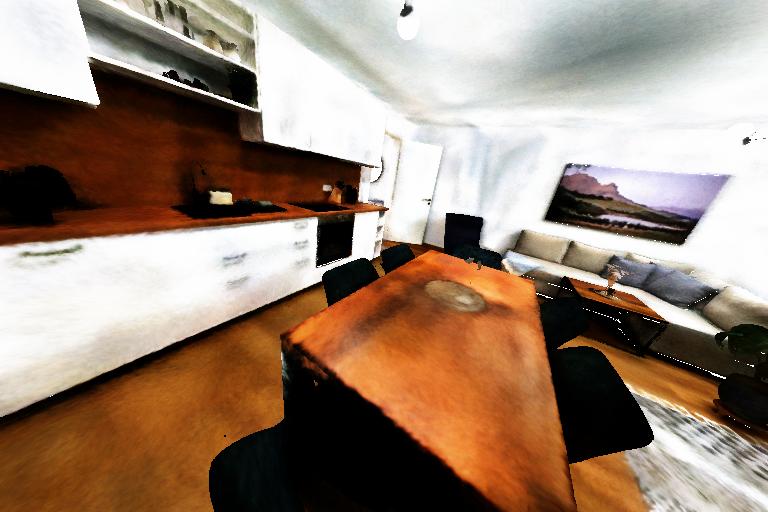}} &
    \begin{tabular}[b]{c}
    \fbox{\includegraphics[width=0.06\textwidth]{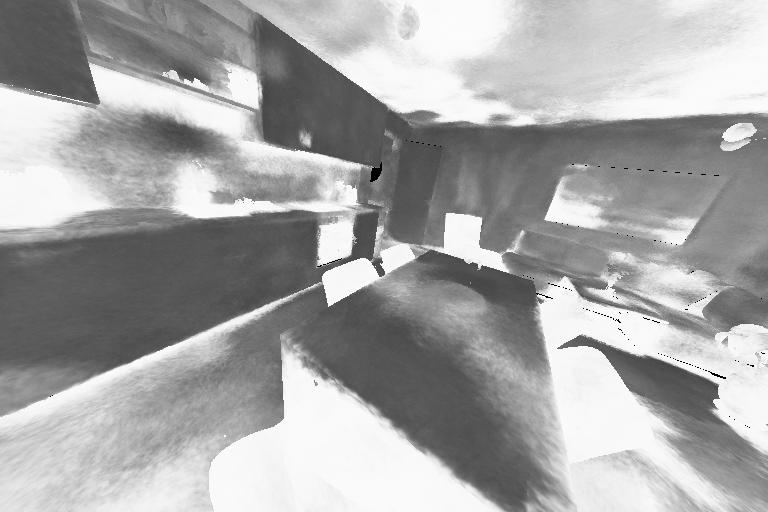}} \\
    \fbox{\includegraphics[width=0.06\textwidth]{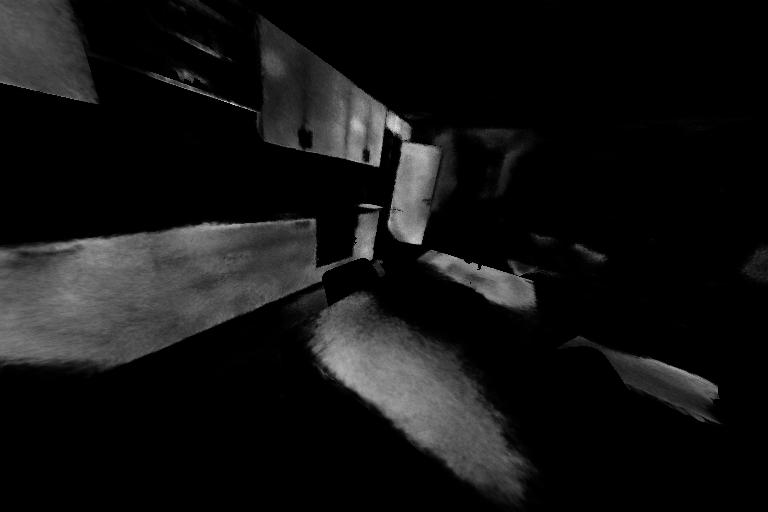}}
    \end{tabular}
    \end{tabular} &
\begin{tabular}[b]{cc}
    \fbox{\includegraphics[width=0.13\textwidth]{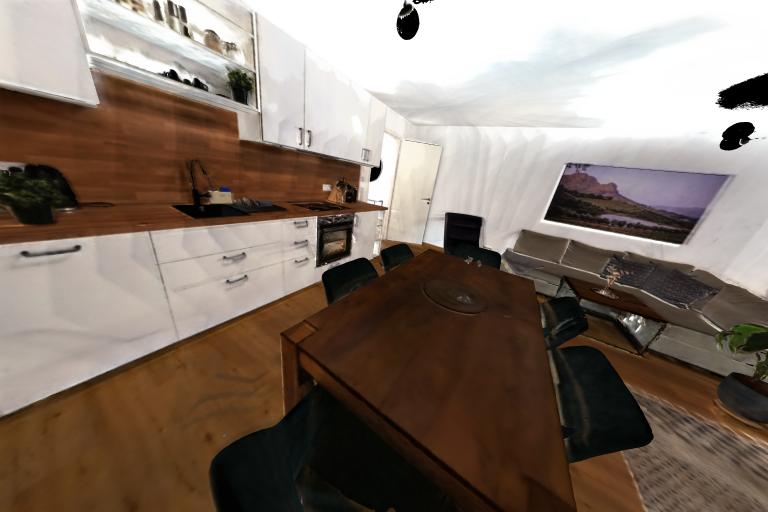}} &
    \begin{tabular}[b]{c}
    \fbox{\includegraphics[width=0.06\textwidth]{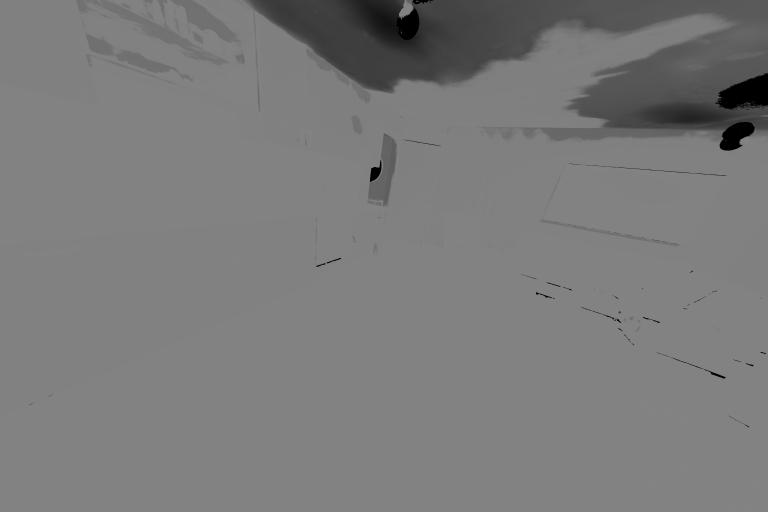}} \\
    \fbox{\includegraphics[width=0.06\textwidth]{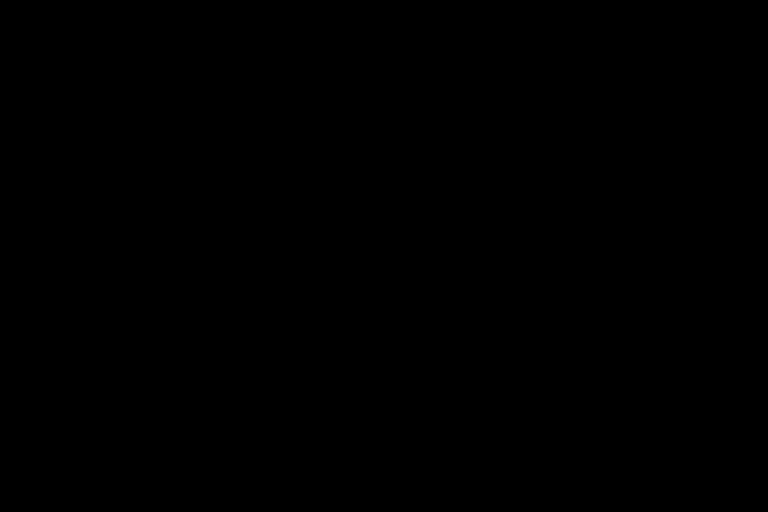}}
    \end{tabular}
    \end{tabular} &
\begin{tabular}[b]{cc}
    \fbox{\includegraphics[width=0.13\textwidth]{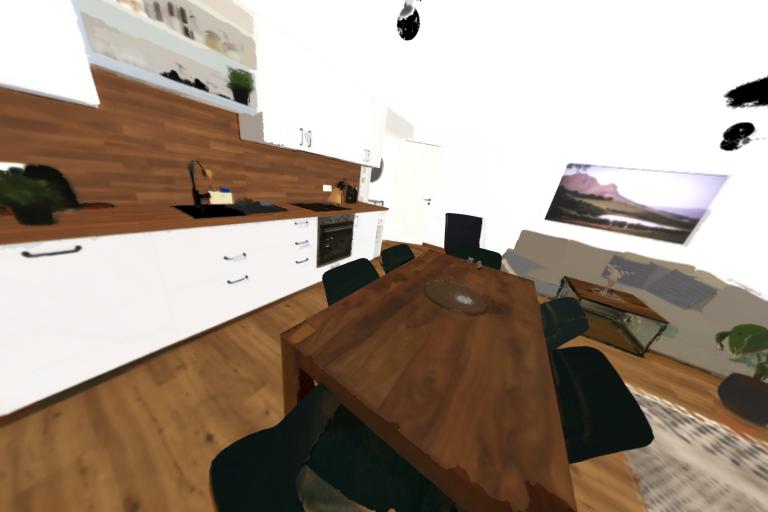}} &
    \begin{tabular}[b]{c}
    \fbox{\includegraphics[width=0.06\textwidth]{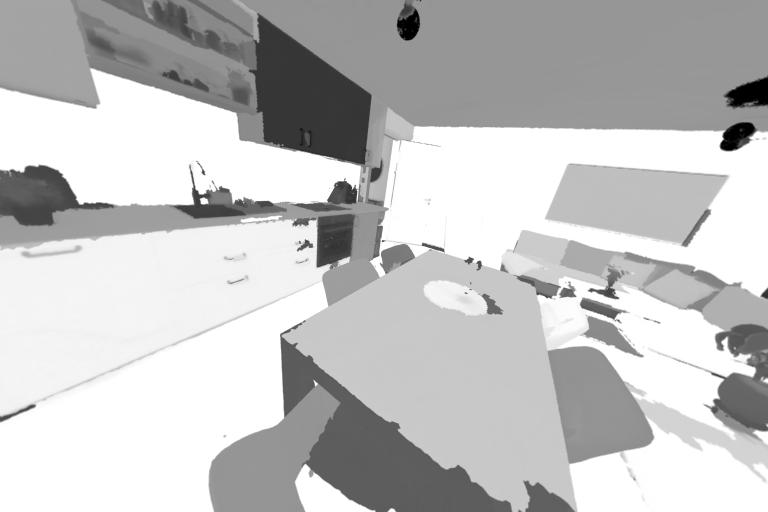}} \\
    \fbox{\includegraphics[width=0.06\textwidth]{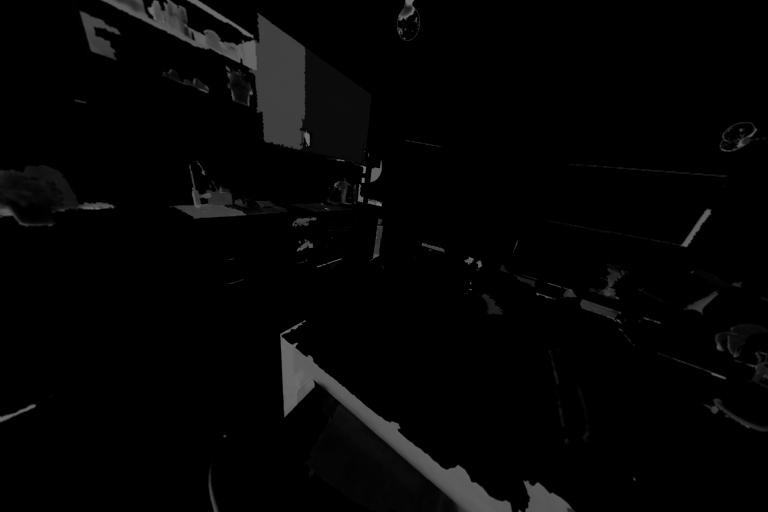}}
    \end{tabular}
    \end{tabular} \\
\hline
\begin{tabular}[b]{c}
\fbox{\includegraphics[width=0.13\textwidth]{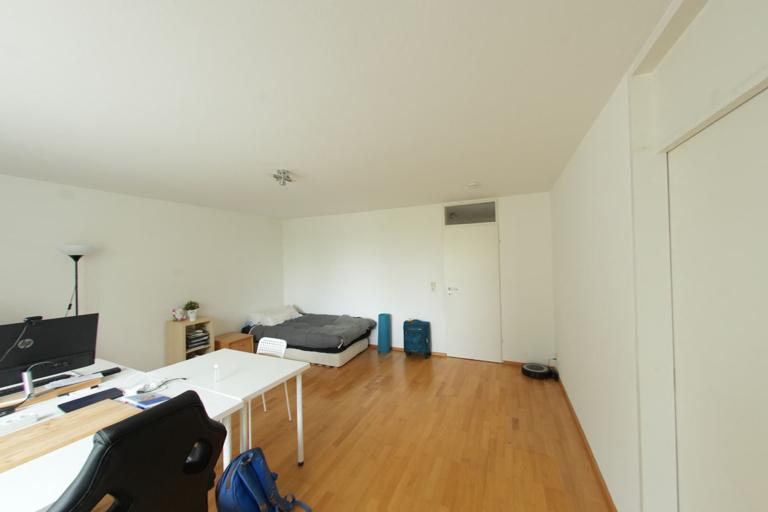}}
\end{tabular} &
\begin{tabular}[b]{cc}
    \fbox{\includegraphics[width=0.13\textwidth]{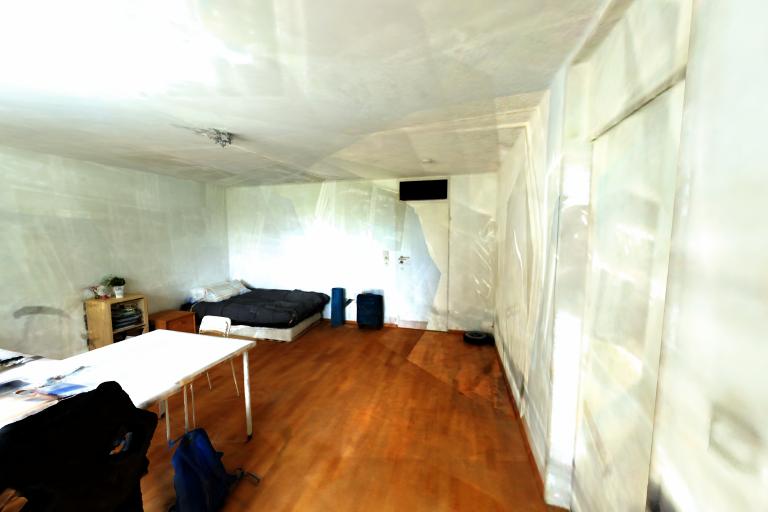}} &
    \begin{tabular}[b]{c}
    \fbox{\includegraphics[width=0.06\textwidth]{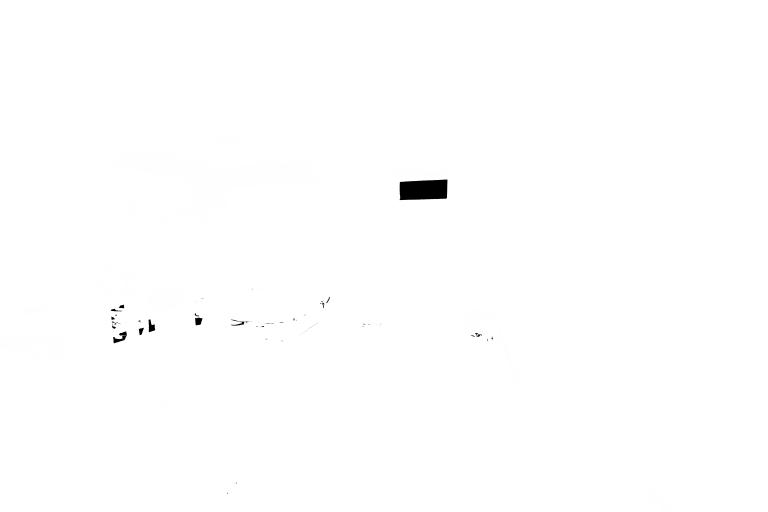}} \\
    \fbox{\includegraphics[width=0.06\textwidth]{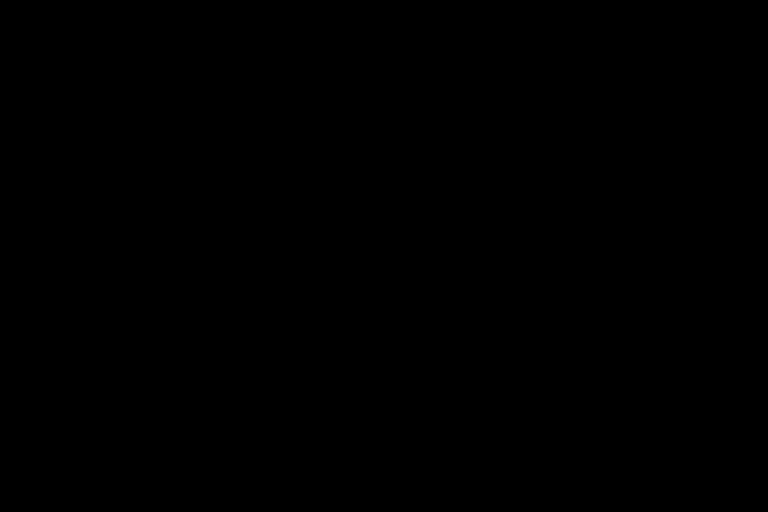}}
    \end{tabular}
    \end{tabular} &
\begin{tabular}[b]{cc}
    \fbox{\includegraphics[width=0.13\textwidth]{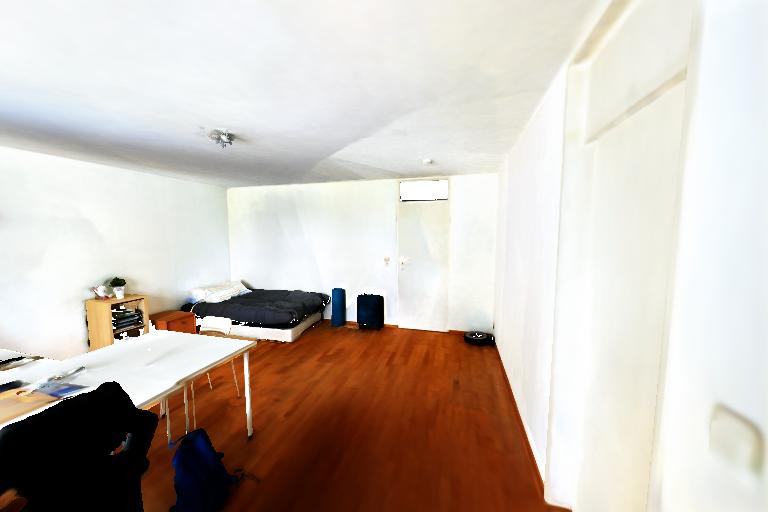}} &
    \begin{tabular}[b]{c}
    \fbox{\includegraphics[width=0.06\textwidth]{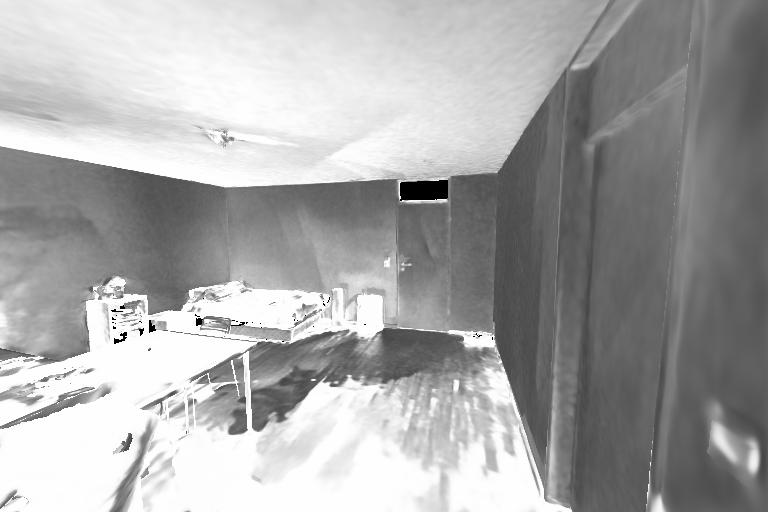}} \\
    \fbox{\includegraphics[width=0.06\textwidth]{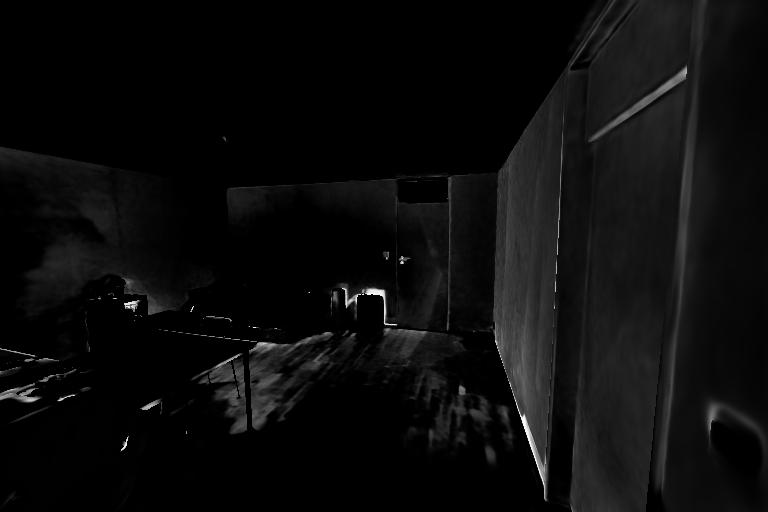}}
    \end{tabular}
    \end{tabular} &
\begin{tabular}[b]{cc}
    \fbox{\includegraphics[width=0.13\textwidth]{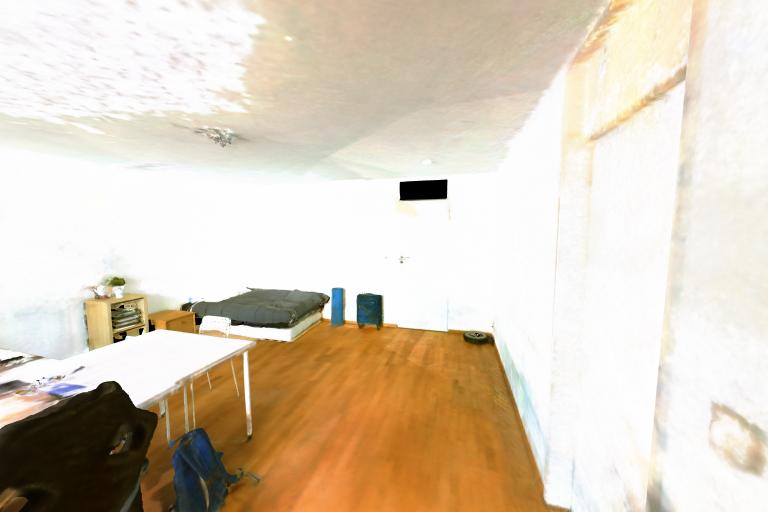}} &
    \begin{tabular}[b]{c}
    \fbox{\includegraphics[width=0.06\textwidth]{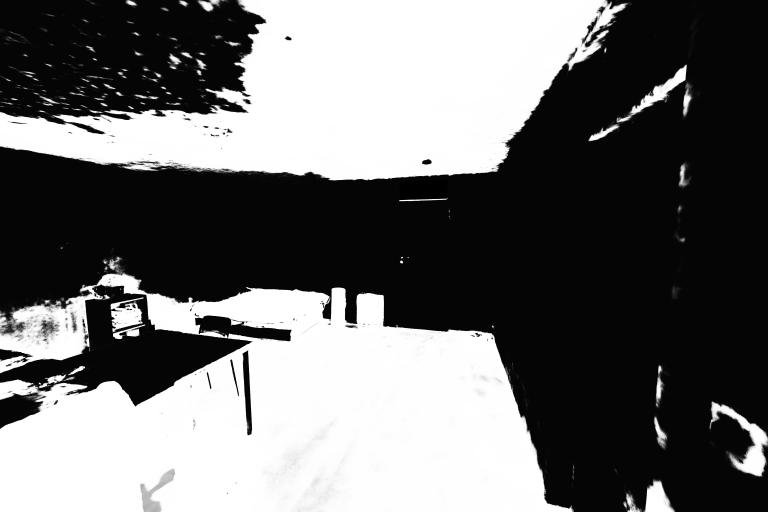}} \\
    \fbox{\includegraphics[width=0.06\textwidth]{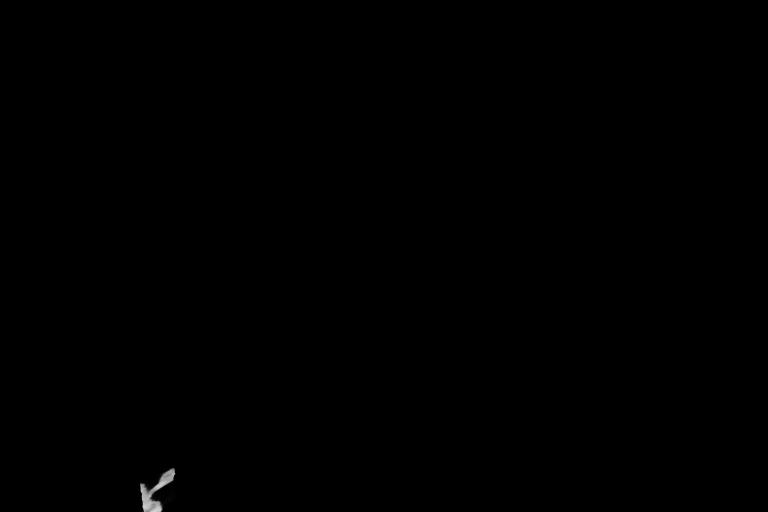}}
    \end{tabular}
    \end{tabular} &
\begin{tabular}[b]{cc}
    \fbox{\includegraphics[width=0.13\textwidth]{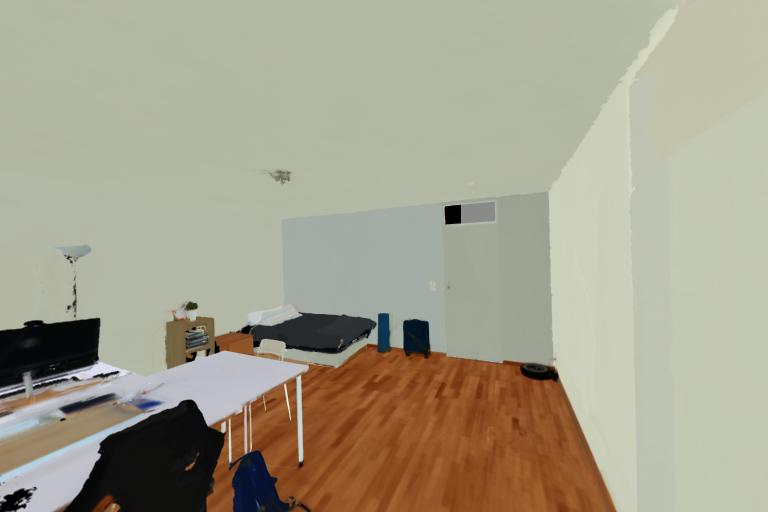}} &
    \begin{tabular}[b]{c}
    \fbox{\includegraphics[width=0.06\textwidth]{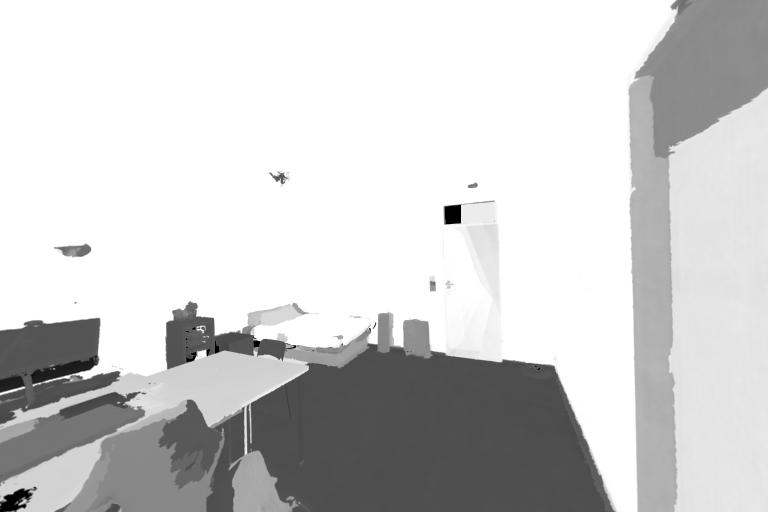}} \\
    \fbox{\includegraphics[width=0.06\textwidth]{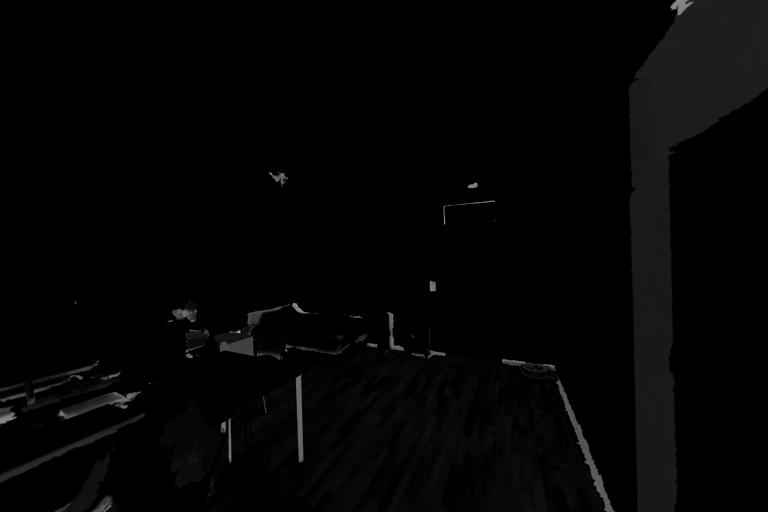}}
    \end{tabular}
    \end{tabular} \\
\begin{tabular}[b]{c}
\fbox{\includegraphics[width=0.13\textwidth]{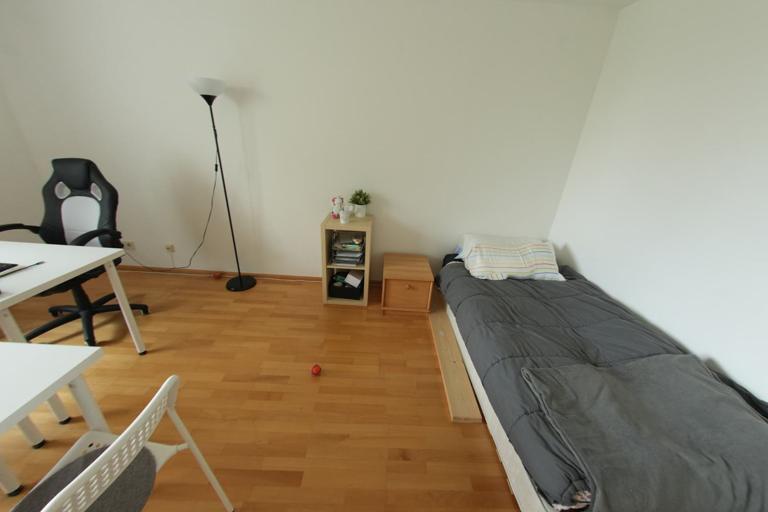}}
\end{tabular} &
\begin{tabular}[b]{cc}
    \fbox{\includegraphics[width=0.13\textwidth]{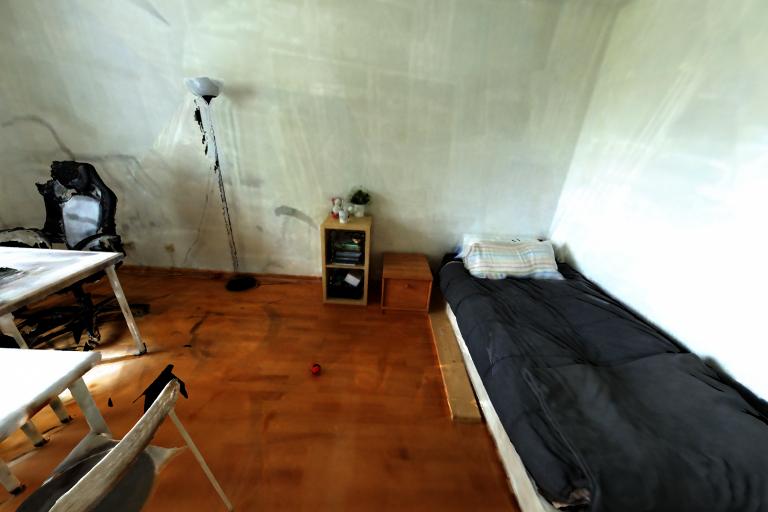}} &
    \begin{tabular}[b]{c}
    \fbox{\includegraphics[width=0.06\textwidth]{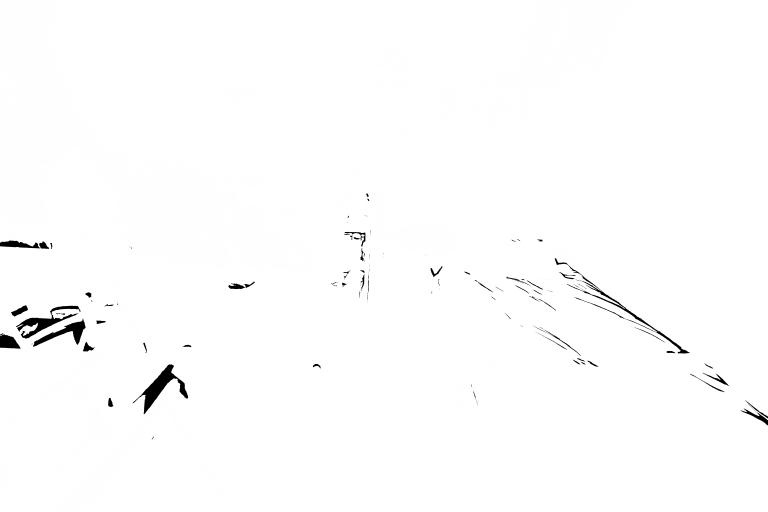}} \\
    \fbox{\includegraphics[width=0.06\textwidth]{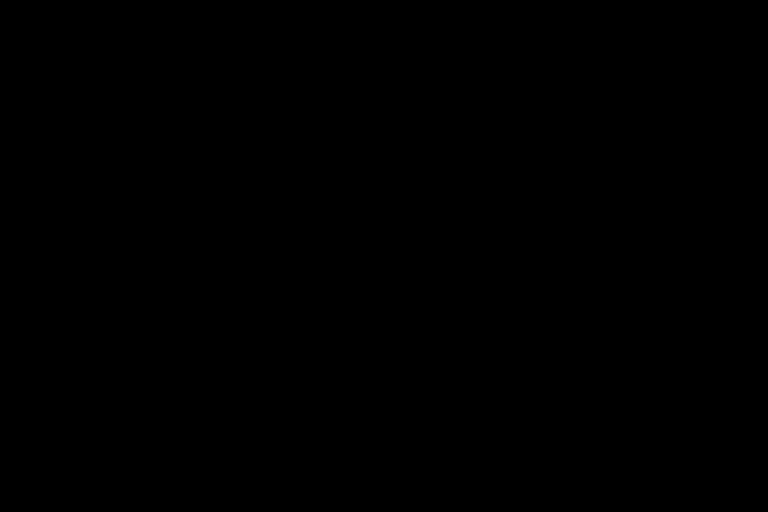}}
    \end{tabular}
    \end{tabular} &
\begin{tabular}[b]{cc}
    \fbox{\includegraphics[width=0.13\textwidth]{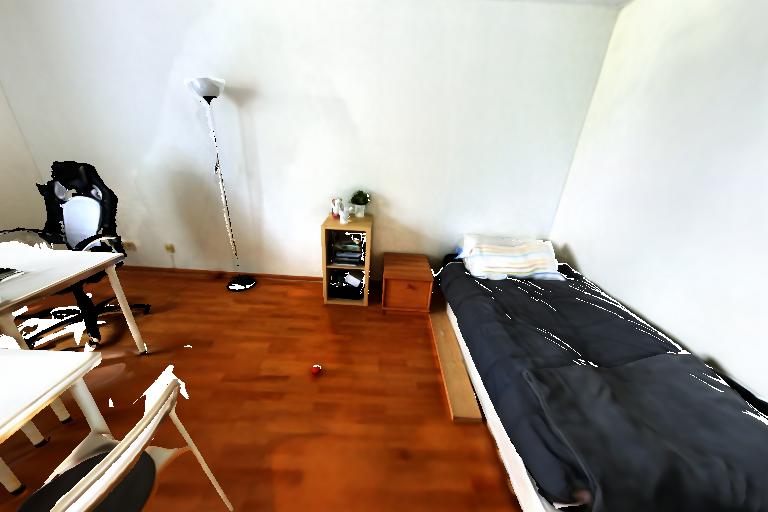}} &
    \begin{tabular}[b]{c}
    \fbox{\includegraphics[width=0.06\textwidth]{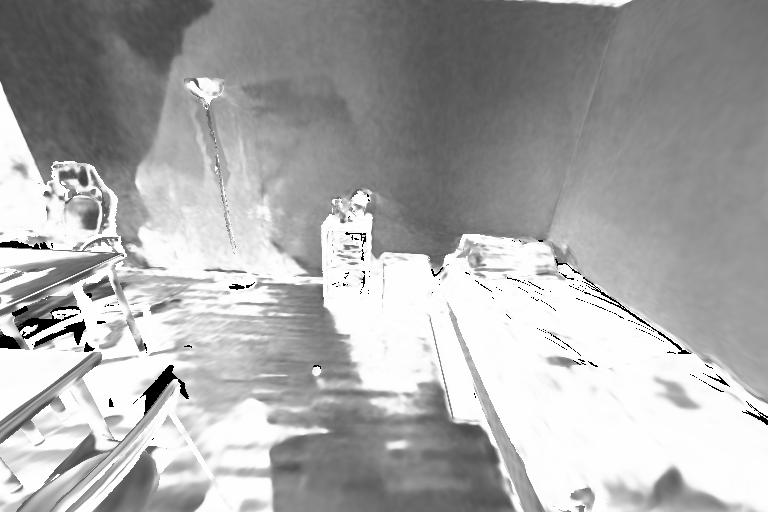}} \\
    \fbox{\includegraphics[width=0.06\textwidth]{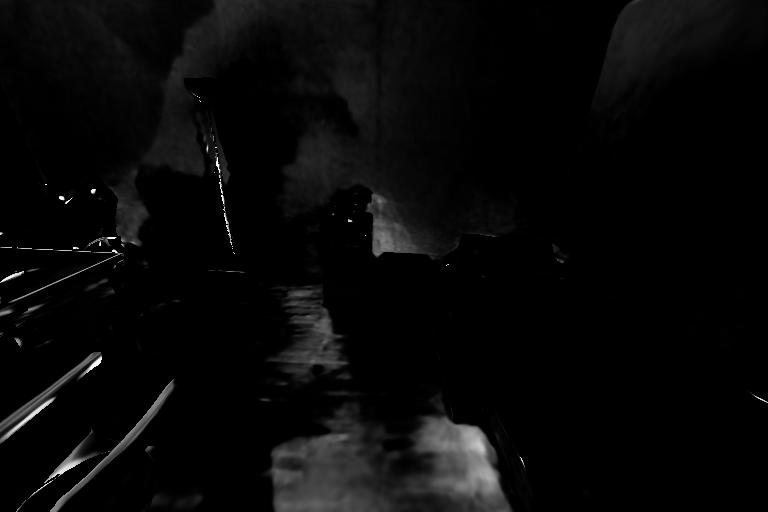}}
    \end{tabular}
    \end{tabular} &
\begin{tabular}[b]{cc}
    \fbox{\includegraphics[width=0.13\textwidth]{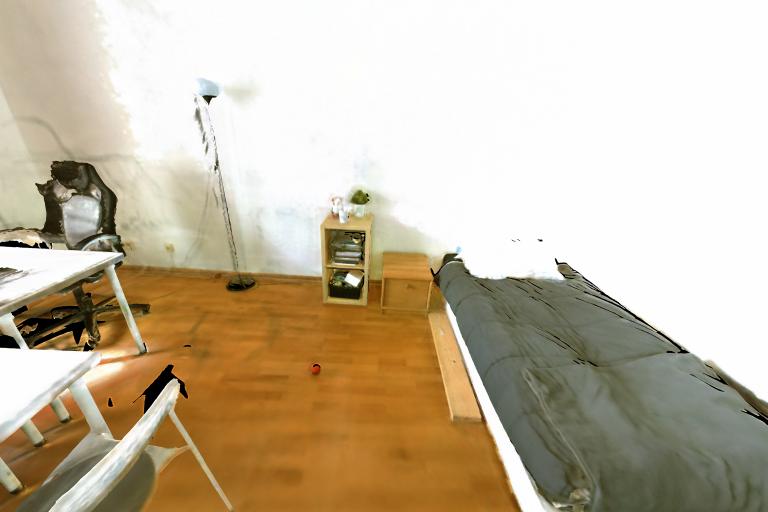}} &
    \begin{tabular}[b]{c}
    \fbox{\includegraphics[width=0.06\textwidth]{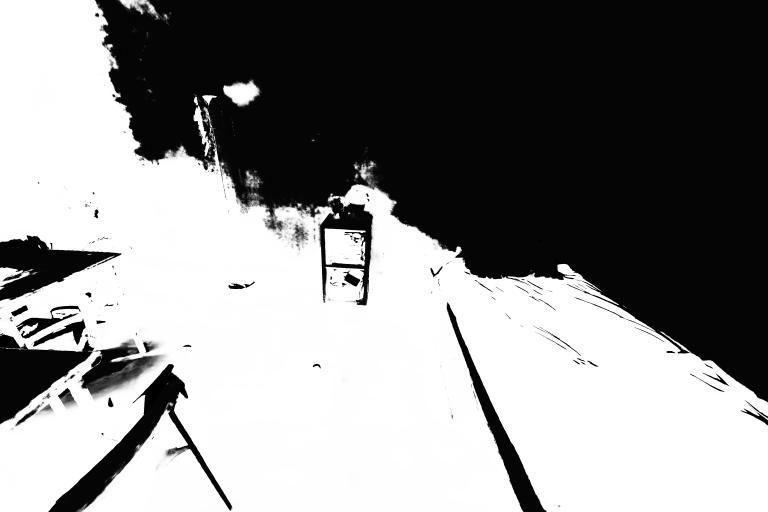}} \\
    \fbox{\includegraphics[width=0.06\textwidth]{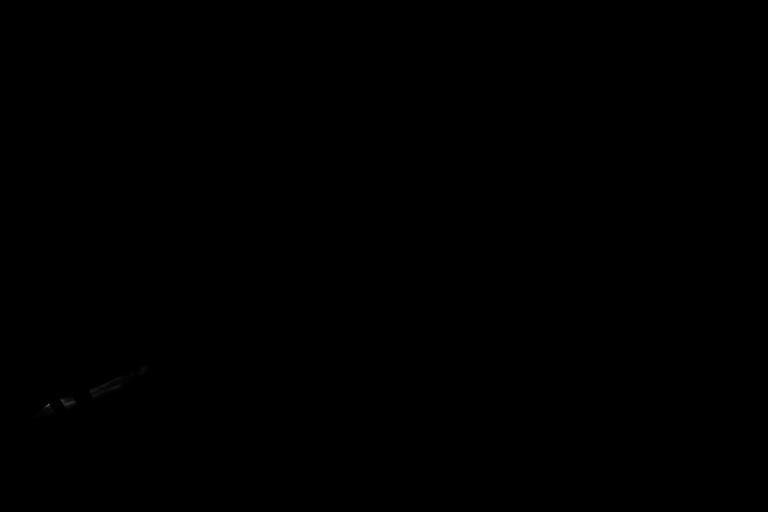}}
    \end{tabular}
    \end{tabular} &
\begin{tabular}[b]{cc}
    \fbox{\includegraphics[width=0.13\textwidth]{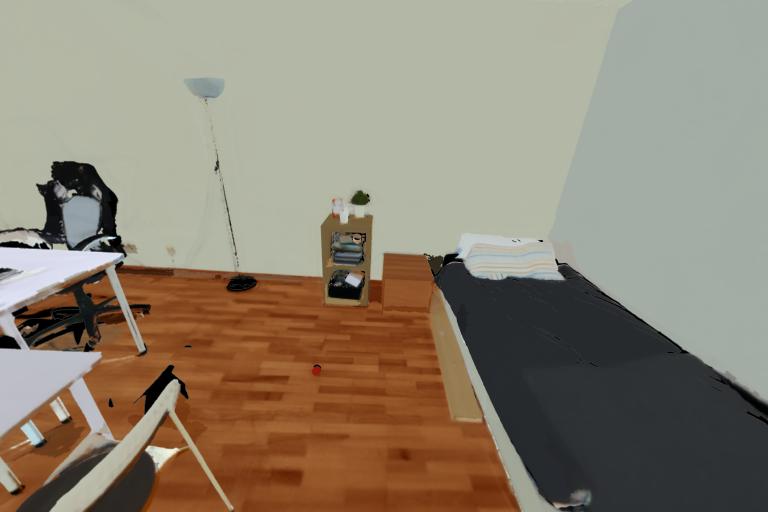}} &
    \begin{tabular}[b]{c}
    \fbox{\includegraphics[width=0.06\textwidth]{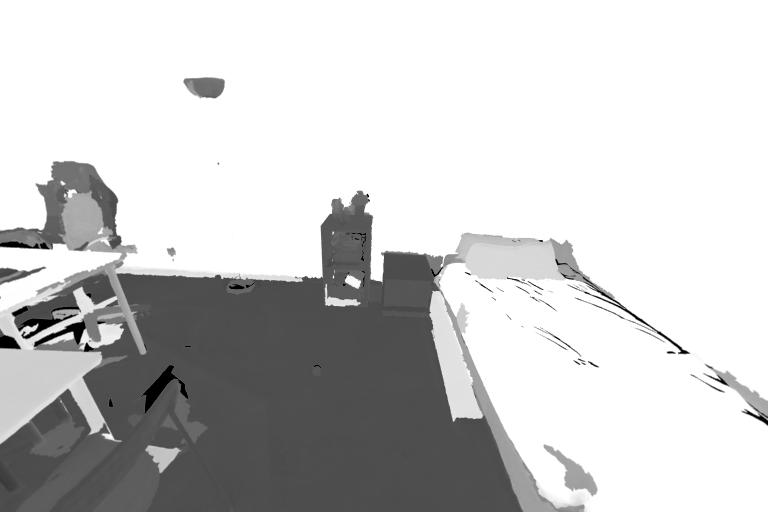}} \\
    \fbox{\includegraphics[width=0.06\textwidth]{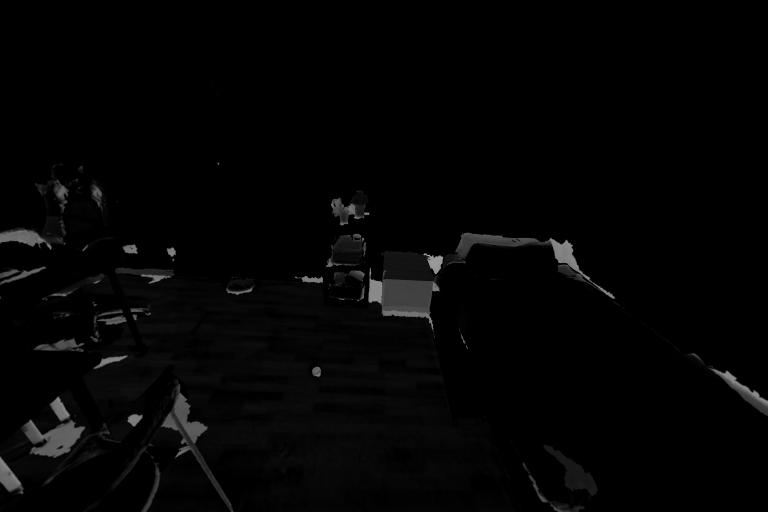}}
    \end{tabular}
    \end{tabular} \\
\hline
\begin{tabular}[b]{c}
\fbox{\includegraphics[width=0.13\textwidth]{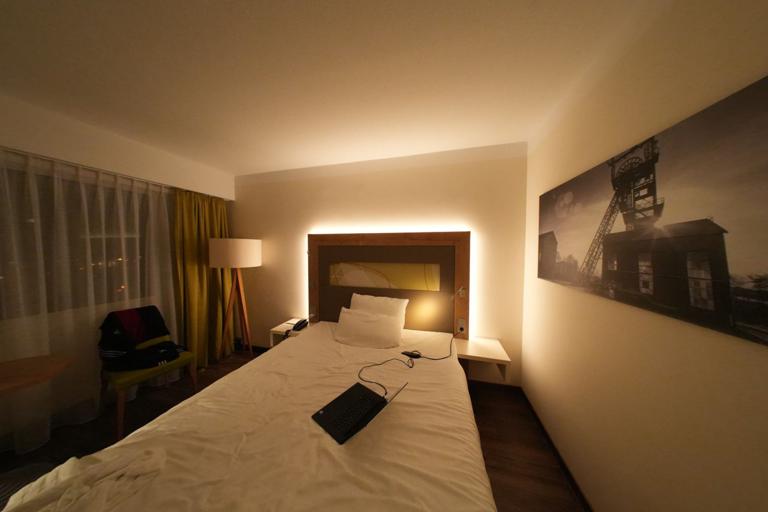}}
\end{tabular} &
\begin{tabular}[b]{cc}
    \fbox{\includegraphics[width=0.13\textwidth]{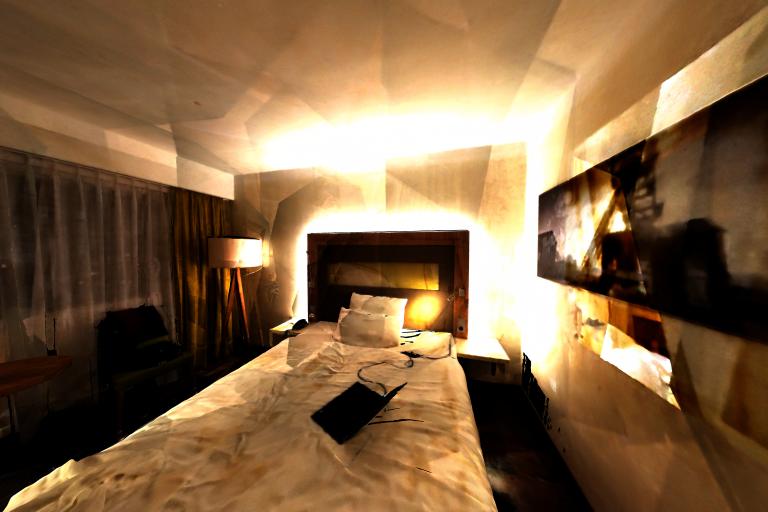}} &
    \begin{tabular}[b]{c}
    \fbox{\includegraphics[width=0.06\textwidth]{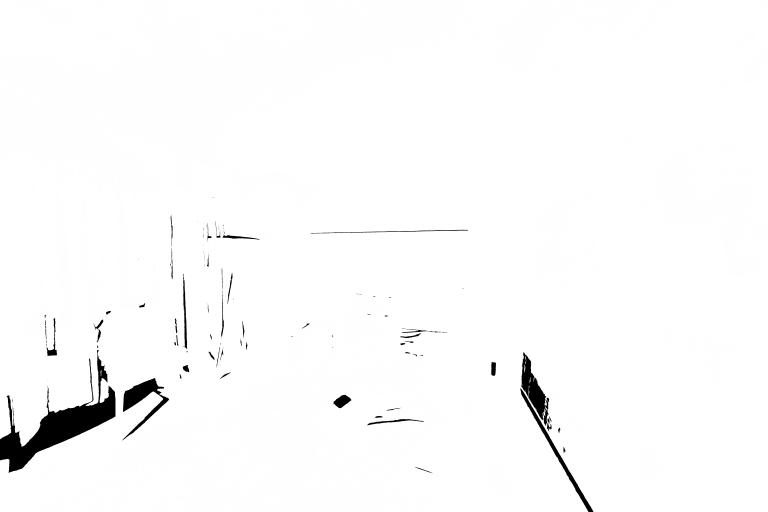}} \\
    \fbox{\includegraphics[width=0.06\textwidth]{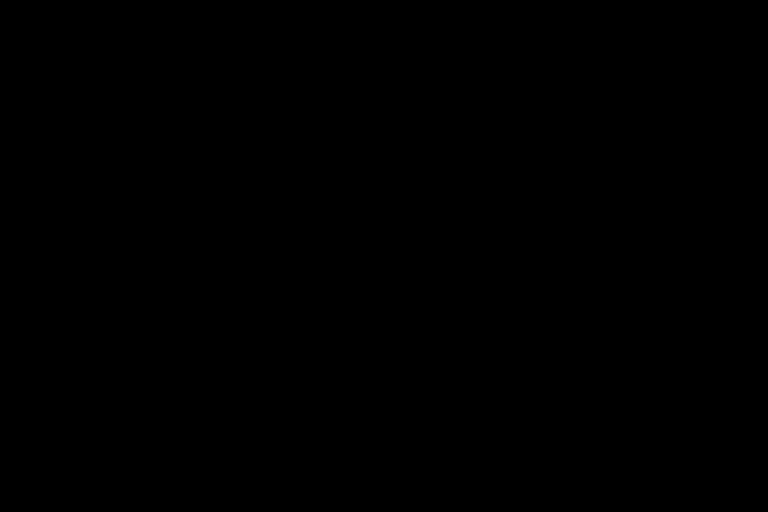}}
    \end{tabular}
    \end{tabular} &
\begin{tabular}[b]{cc}
    \fbox{\includegraphics[width=0.13\textwidth]{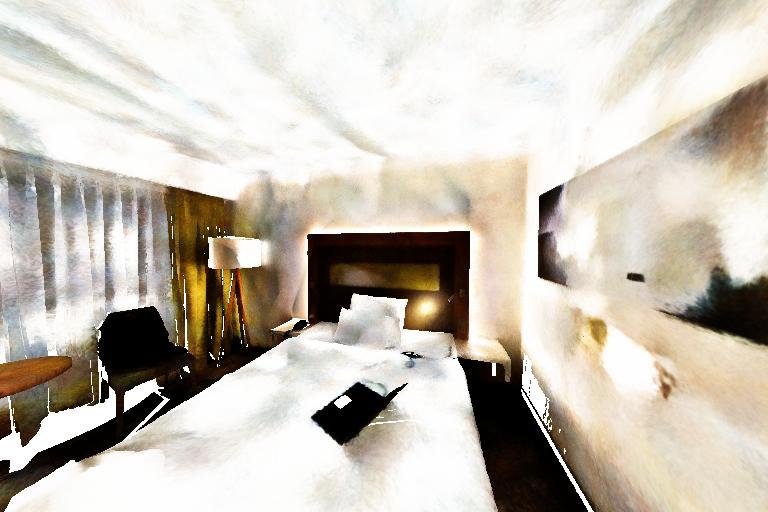}} &
    \begin{tabular}[b]{c}
    \fbox{\includegraphics[width=0.06\textwidth]{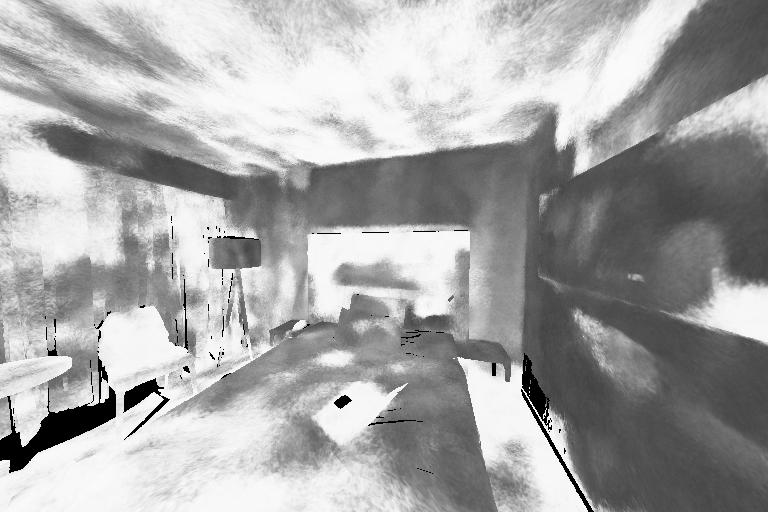}} \\
    \fbox{\includegraphics[width=0.06\textwidth]{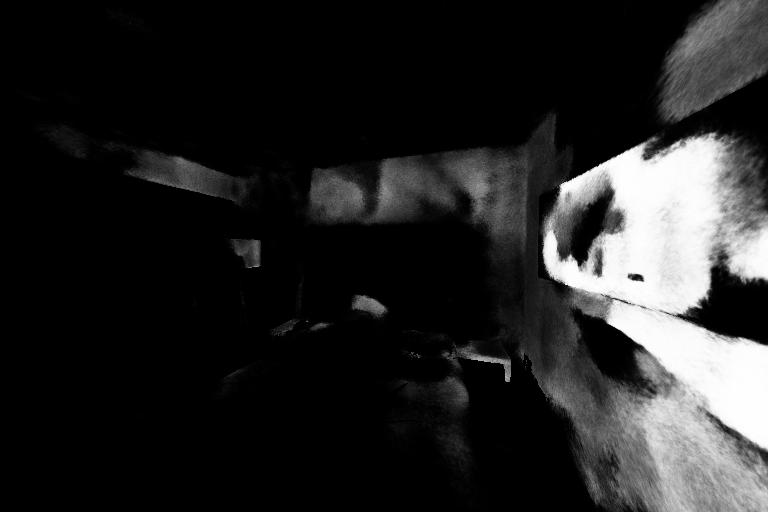}}
    \end{tabular}
    \end{tabular} &
\begin{tabular}[b]{cc}
    \fbox{\includegraphics[width=0.13\textwidth]{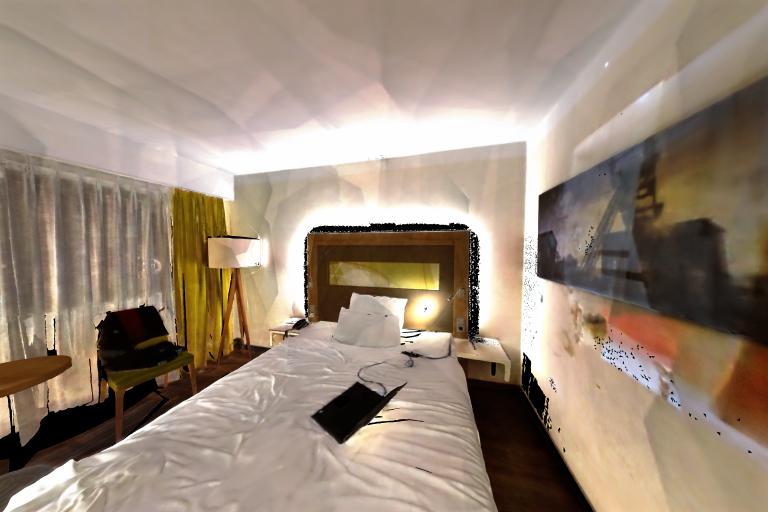}} &
    \begin{tabular}[b]{c}
    \fbox{\includegraphics[width=0.06\textwidth]{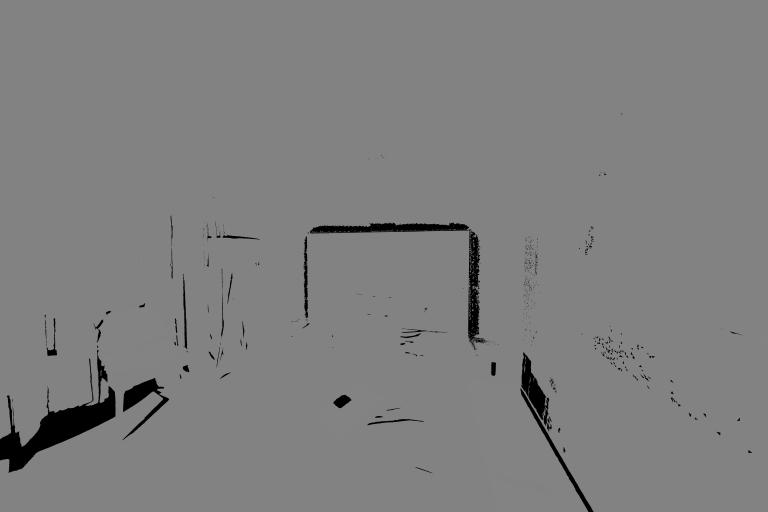}} \\
    \fbox{\includegraphics[width=0.06\textwidth]{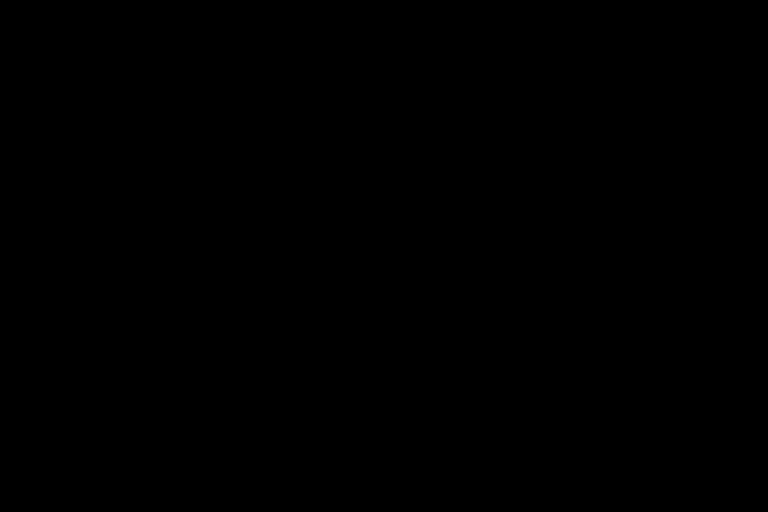}}
    \end{tabular}
    \end{tabular} &
\begin{tabular}[b]{cc}
    \fbox{\includegraphics[width=0.13\textwidth]{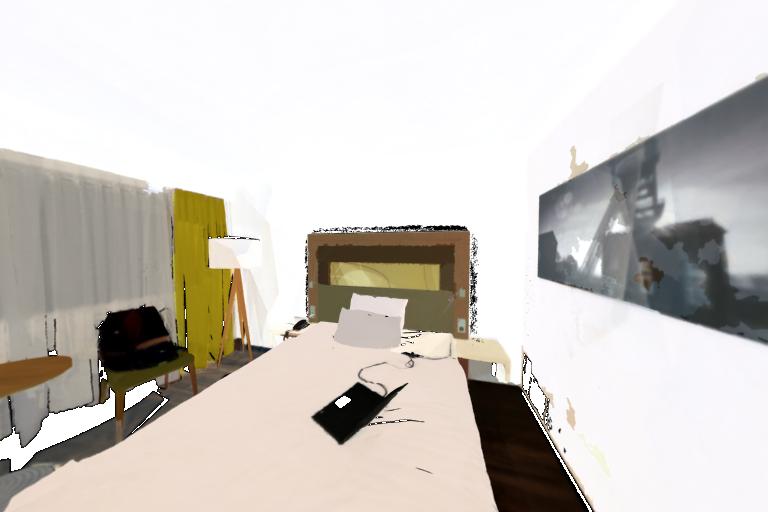}} &
    \begin{tabular}[b]{c}
    \fbox{\includegraphics[width=0.06\textwidth]{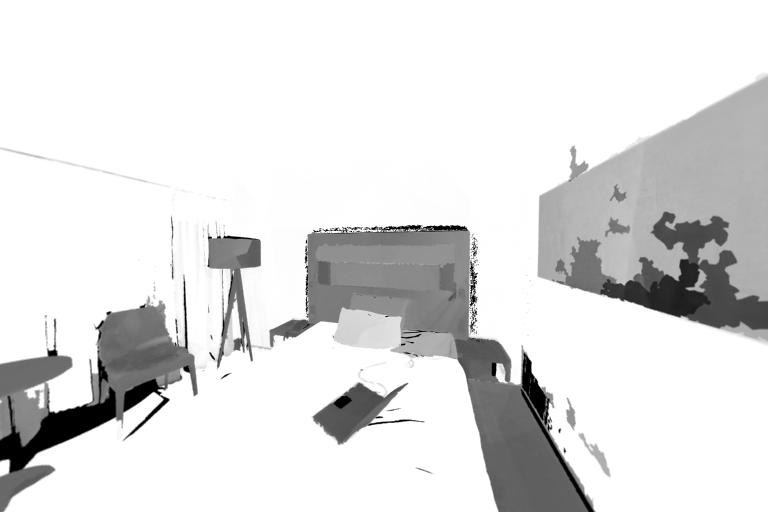}} \\
    \fbox{\includegraphics[width=0.06\textwidth]{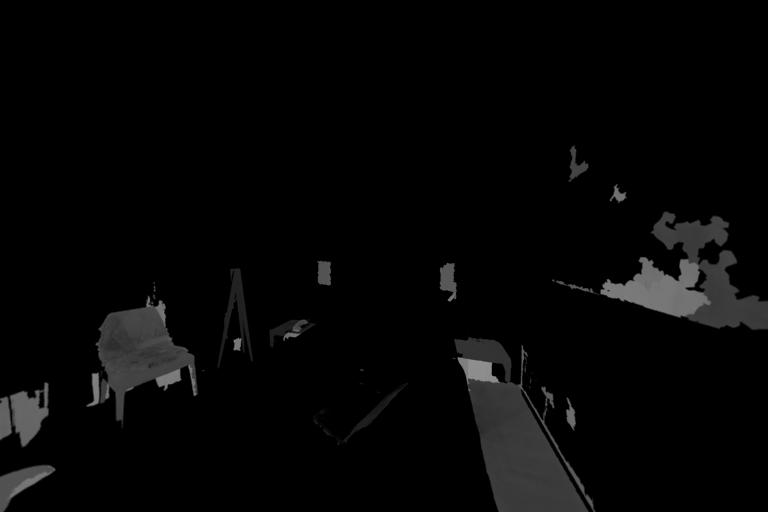}}
    \end{tabular}
    \end{tabular} \\
\begin{tabular}[b]{c}
\fbox{\includegraphics[width=0.13\textwidth]{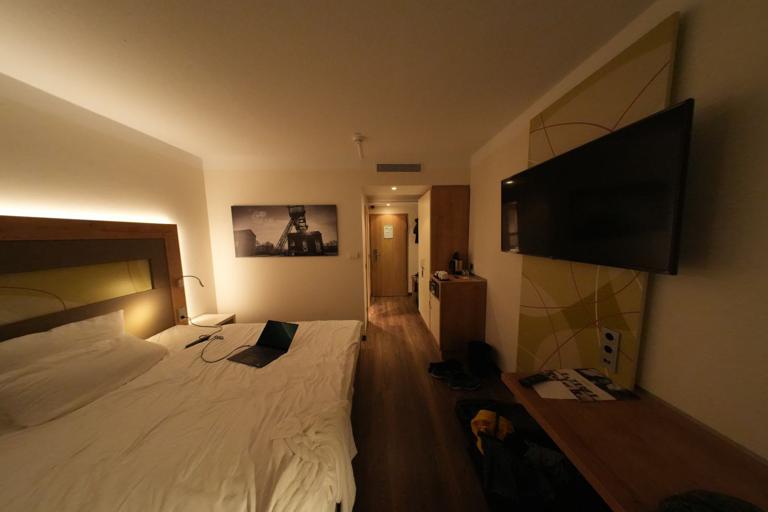}}
\end{tabular} &
\begin{tabular}[b]{cc}
    \fbox{\includegraphics[width=0.13\textwidth]{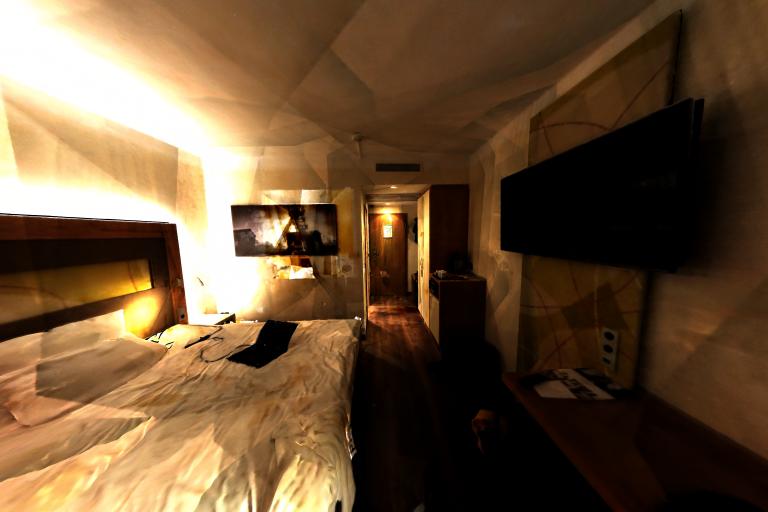}} &
    \begin{tabular}[b]{c}
    \fbox{\includegraphics[width=0.06\textwidth]{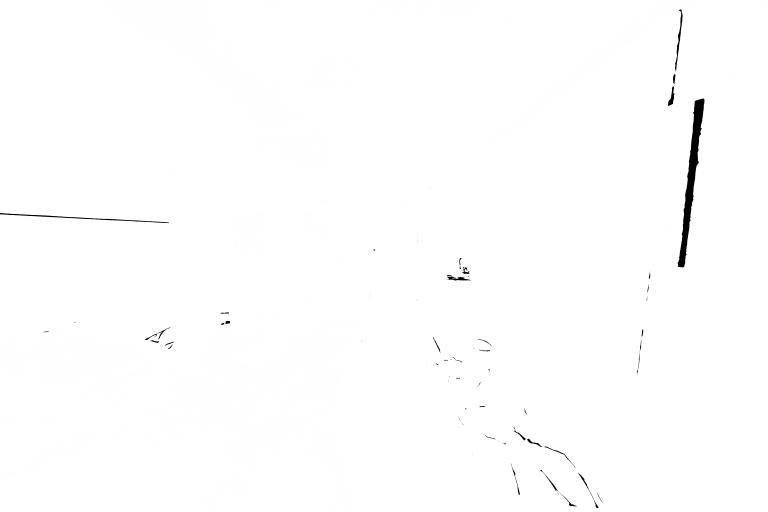}} \\
    \fbox{\includegraphics[width=0.06\textwidth]{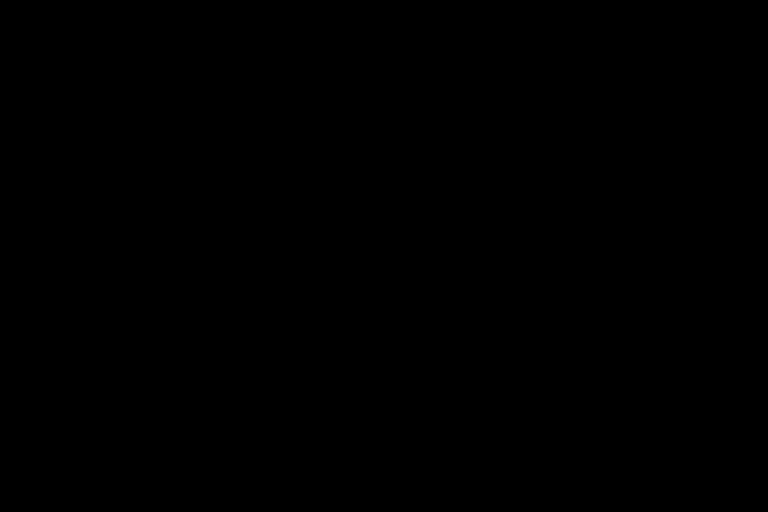}}
    \end{tabular}
    \end{tabular} &
\begin{tabular}[b]{cc}
    \fbox{\includegraphics[width=0.13\textwidth]{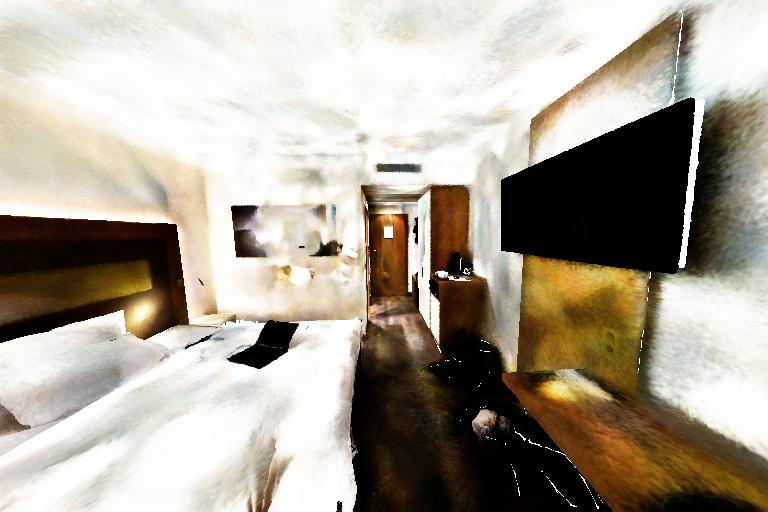}} &
    \begin{tabular}[b]{c}
    \fbox{\includegraphics[width=0.06\textwidth]{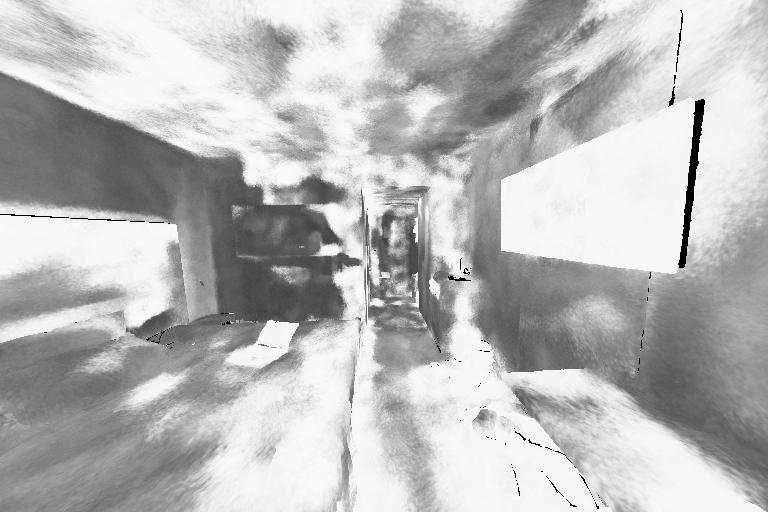}} \\
    \fbox{\includegraphics[width=0.06\textwidth]{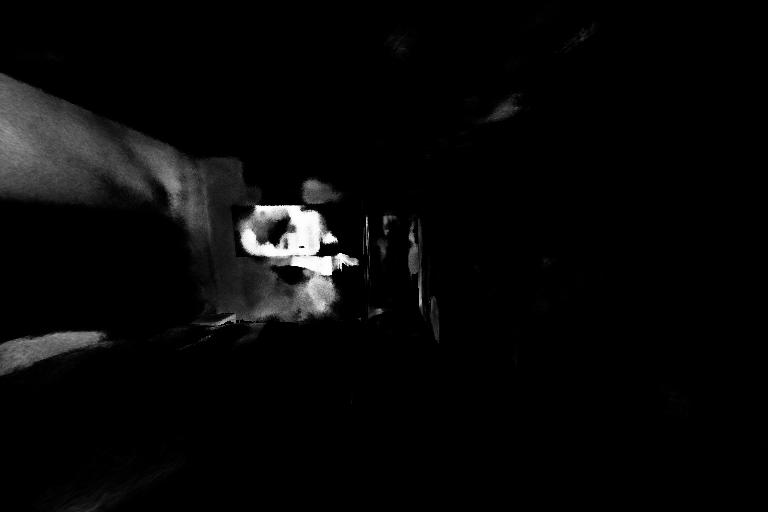}}
    \end{tabular}
    \end{tabular} &
\begin{tabular}[b]{cc}
    \fbox{\includegraphics[width=0.13\textwidth]{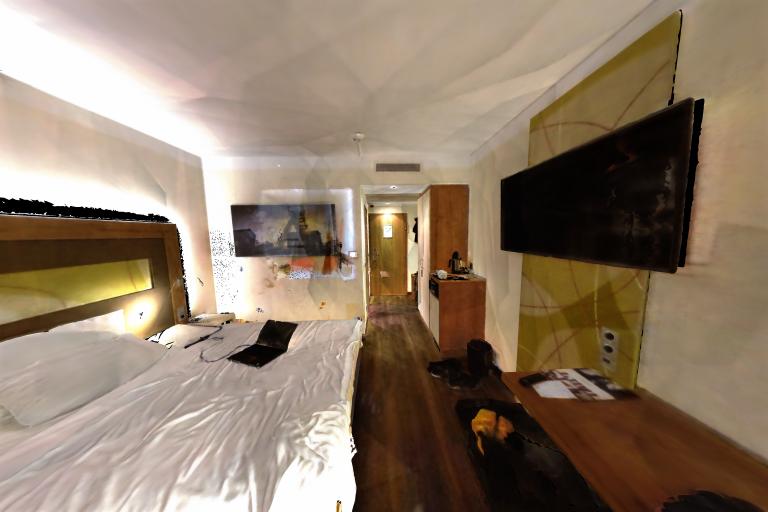}} &
    \begin{tabular}[b]{c}
    \fbox{\includegraphics[width=0.06\textwidth]{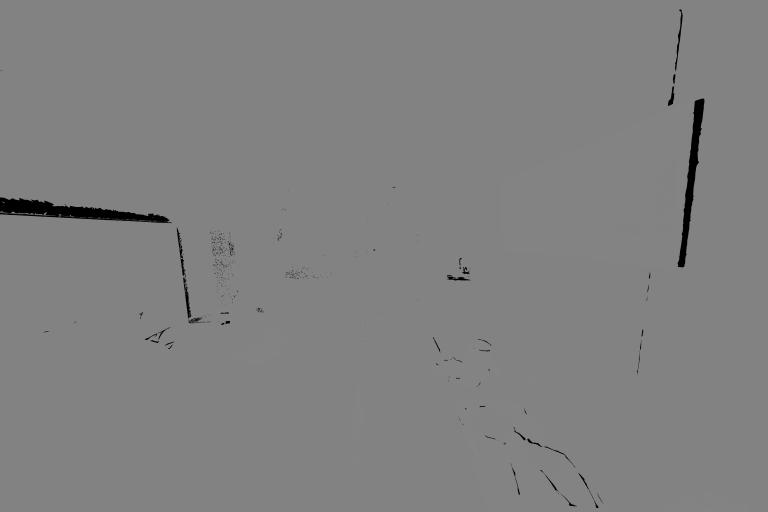}} \\
    \fbox{\includegraphics[width=0.06\textwidth]{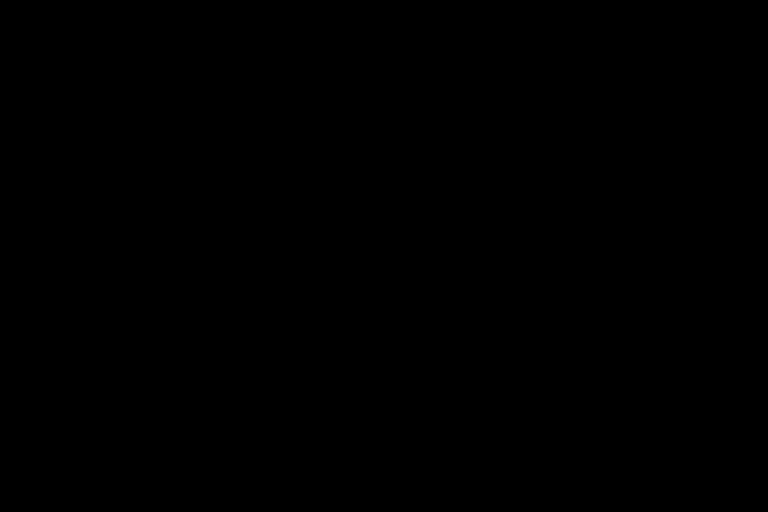}}
    \end{tabular}
    \end{tabular} &
\begin{tabular}[b]{cc}
    \fbox{\includegraphics[width=0.13\textwidth]{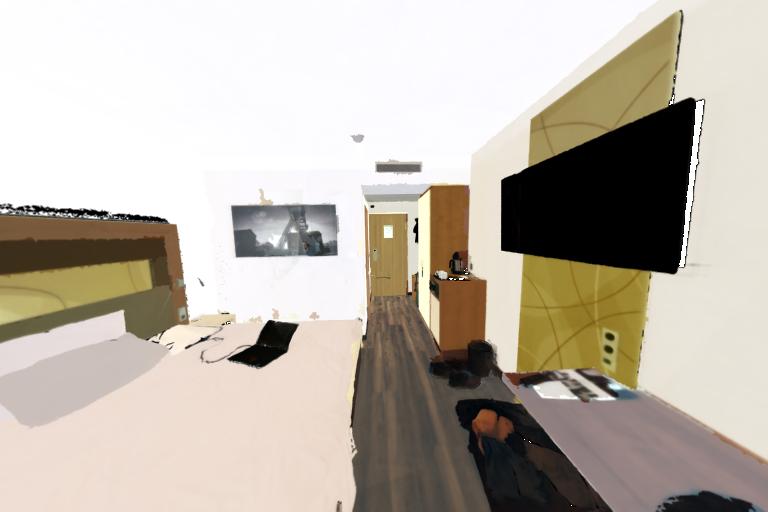}} &
    \begin{tabular}[b]{c}
    \fbox{\includegraphics[width=0.06\textwidth]{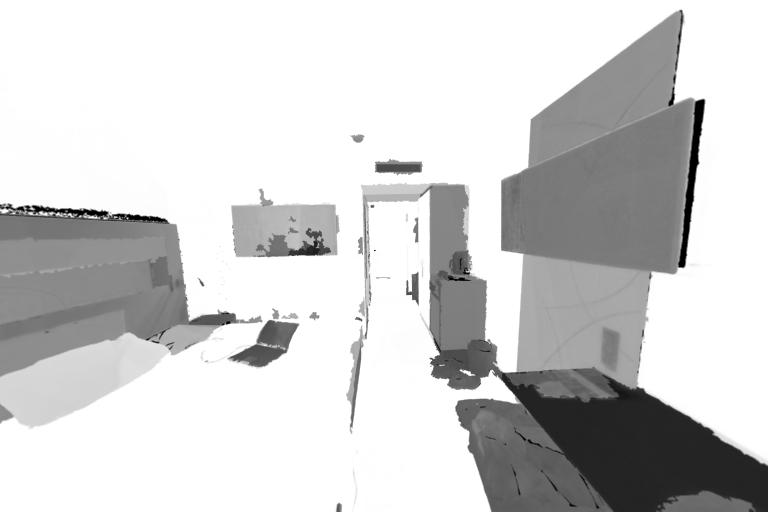}} \\
    \fbox{\includegraphics[width=0.06\textwidth]{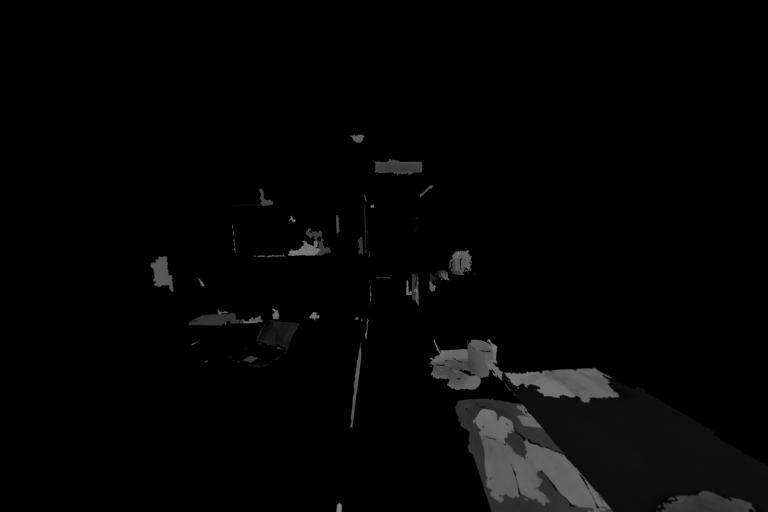}}
    \end{tabular}
    \end{tabular} \\
\hline
        % Header row at bottom
        \smash{\scriptsize RGB} &
        \smash{\scriptsize FIPT \cite{FIPT}} &
        \smash{\scriptsize NeILF++ \cite{NeilfPP}} &
        \smash{\scriptsize IRIS \cite{IRIS}} &
        \smash{\scriptsize IIF (Ours)}
    \end{tabular}}
    \caption{\textbf{Scannet++ \cite{ScanNet++} comparisons.} Additional samples on ScanNet++ \cite{ScanNet++} scenes.}
    \label{fig:supp:scannetpp_comparisons}
\end{figure*}

\begin{figure*}[t]
    \centering
    \setlength\tabcolsep{1.25pt}
    \resizebox{\textwidth}{!}{
    \fboxsep=0pt
    \begin{tabular}{c|cccc}
\begin{tabular}[b]{c}
\fbox{\includegraphics[width=0.13\textwidth]{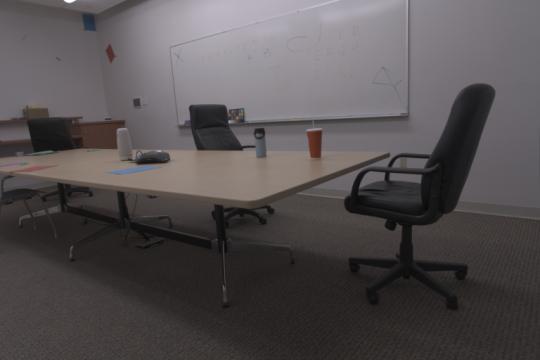}}
\end{tabular} &
\begin{tabular}[b]{cc}
    \fbox{\includegraphics[width=0.13\textwidth]{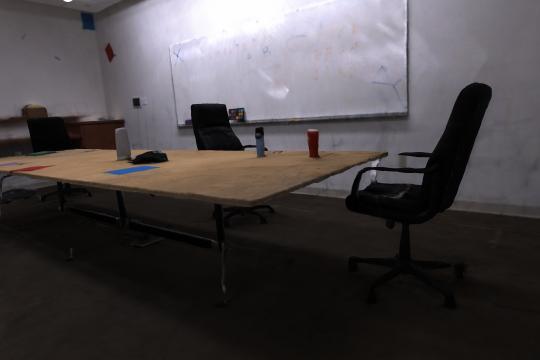}} &
    \begin{tabular}[b]{c}
    \fbox{\includegraphics[width=0.06\textwidth]{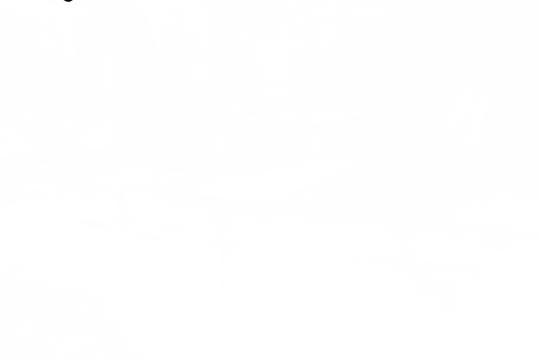}} \\
    \fbox{\includegraphics[width=0.06\textwidth]{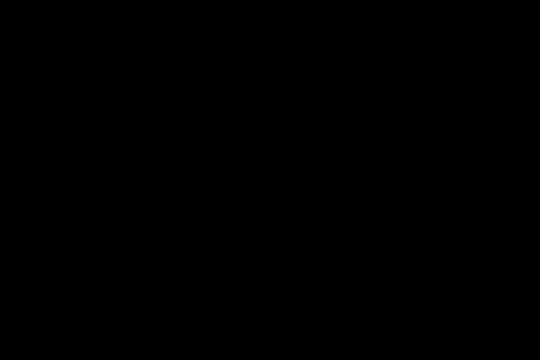}}
    \end{tabular}
    \end{tabular} &
\begin{tabular}[b]{cc}
    \fbox{\includegraphics[width=0.13\textwidth]{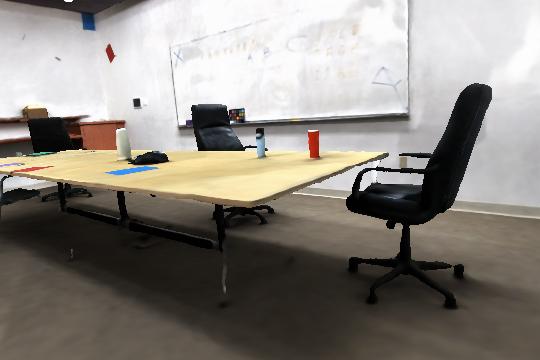}} &
    \begin{tabular}[b]{c}
    \fbox{\includegraphics[width=0.06\textwidth]{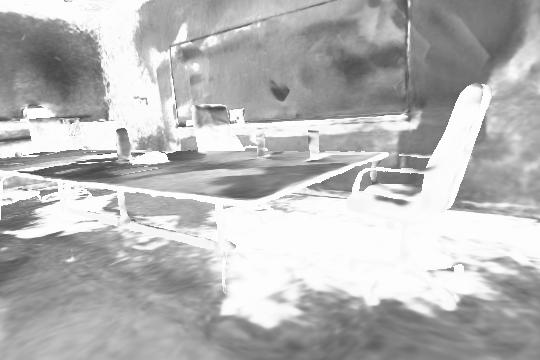}} \\
    \fbox{\includegraphics[width=0.06\textwidth]{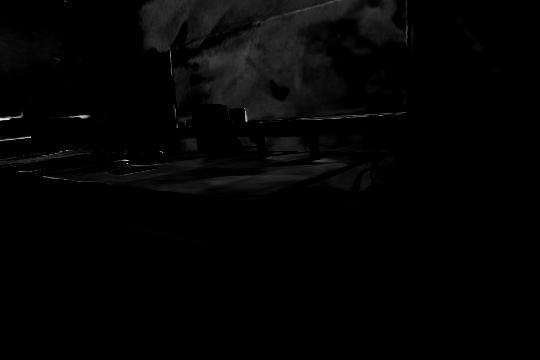}}
    \end{tabular}
    \end{tabular} &
\begin{tabular}[b]{cc}
    \fbox{\includegraphics[width=0.13\textwidth]{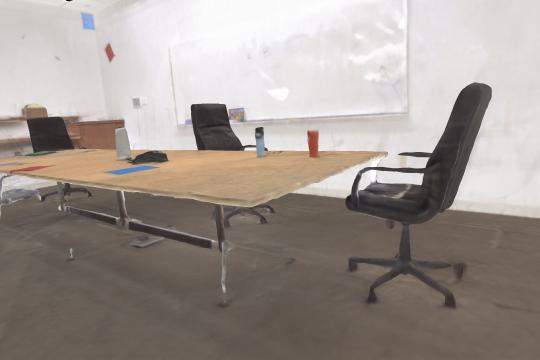}} &
    \begin{tabular}[b]{c}
    \fbox{\includegraphics[width=0.06\textwidth]{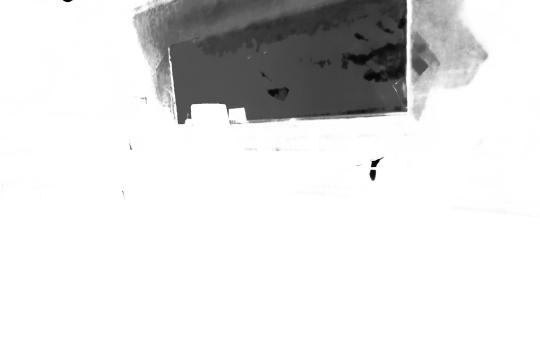}} \\
    \fbox{\includegraphics[width=0.06\textwidth]{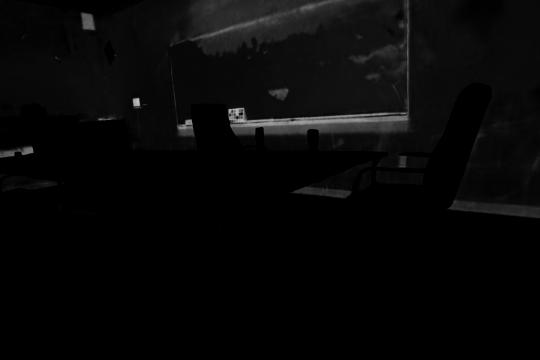}}
    \end{tabular}
    \end{tabular} &
\begin{tabular}[b]{cc}
    \fbox{\includegraphics[width=0.13\textwidth]{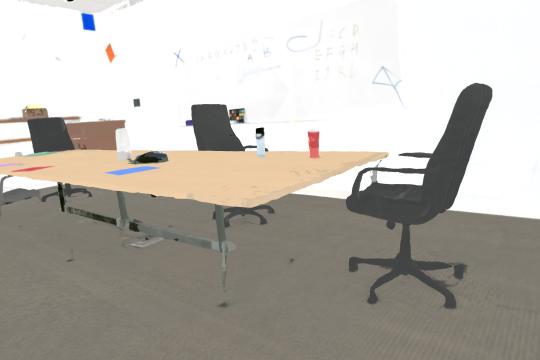}} &
    \begin{tabular}[b]{c}
    \fbox{\includegraphics[width=0.06\textwidth]{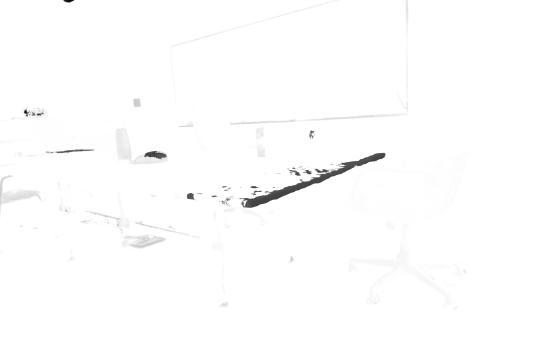}} \\
    \fbox{\includegraphics[width=0.06\textwidth]{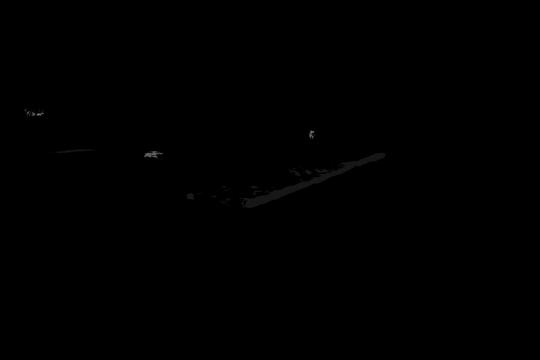}}
    \end{tabular}
    \end{tabular} \\
\begin{tabular}[b]{c}
\fbox{\includegraphics[width=0.13\textwidth]{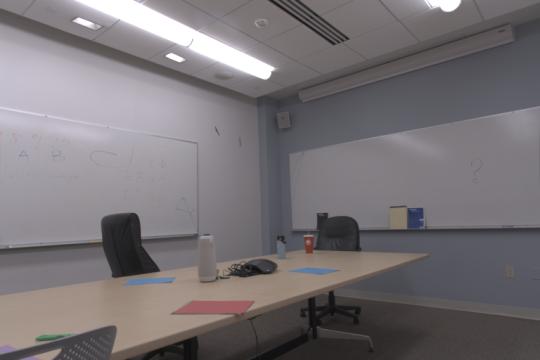}}
\end{tabular} &
\begin{tabular}[b]{cc}
    \fbox{\includegraphics[width=0.13\textwidth]{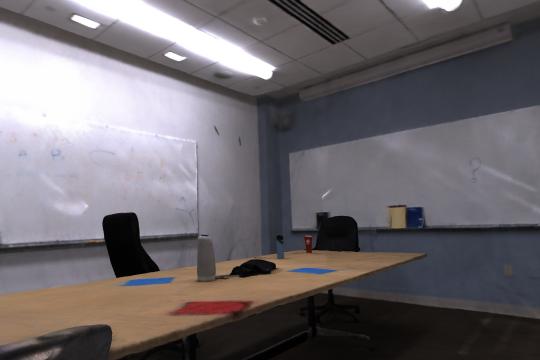}} &
    \begin{tabular}[b]{c}
    \fbox{\includegraphics[width=0.06\textwidth]{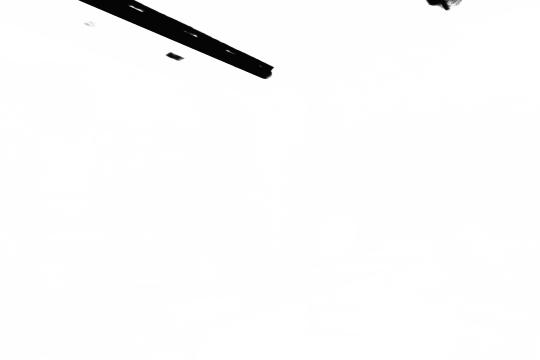}} \\
    \fbox{\includegraphics[width=0.06\textwidth]{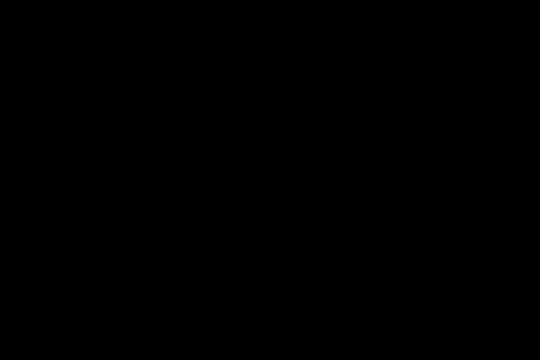}}
    \end{tabular}
    \end{tabular} &
\begin{tabular}[b]{cc}
    \fbox{\includegraphics[width=0.13\textwidth]{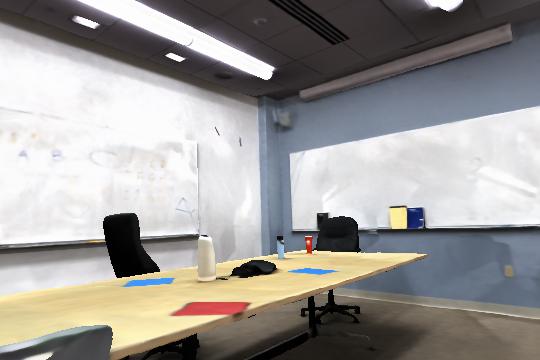}} &
    \begin{tabular}[b]{c}
    \fbox{\includegraphics[width=0.06\textwidth]{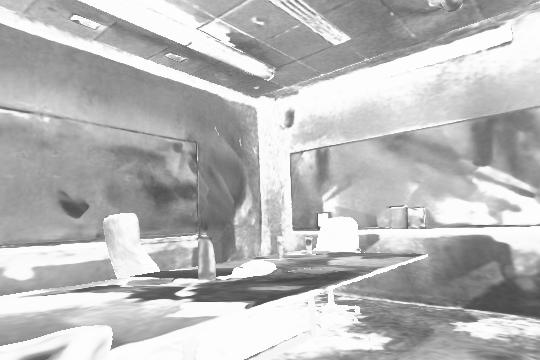}} \\
    \fbox{\includegraphics[width=0.06\textwidth]{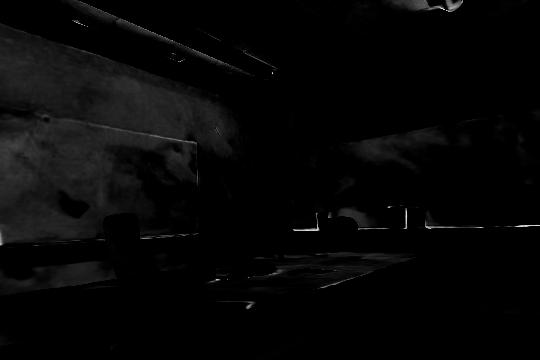}}
    \end{tabular}
    \end{tabular} &
\begin{tabular}[b]{cc}
    \fbox{\includegraphics[width=0.13\textwidth]{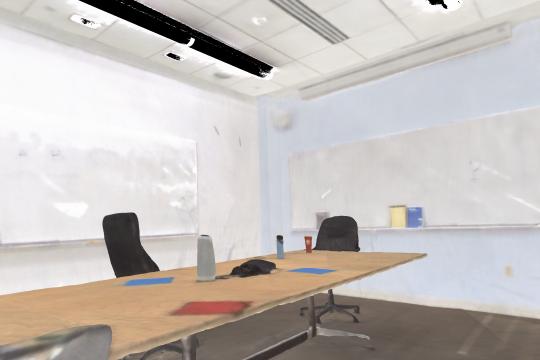}} &
    \begin{tabular}[b]{c}
    \fbox{\includegraphics[width=0.06\textwidth]{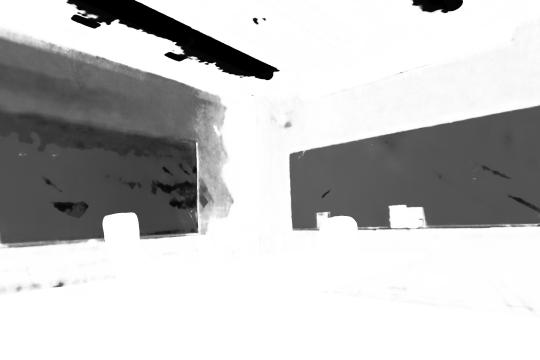}} \\
    \fbox{\includegraphics[width=0.06\textwidth]{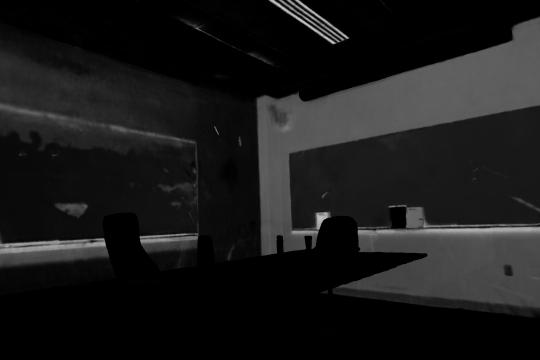}}
    \end{tabular}
    \end{tabular} &
\begin{tabular}[b]{cc}
    \fbox{\includegraphics[width=0.13\textwidth]{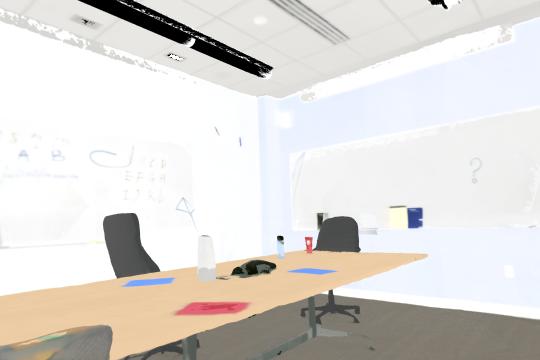}} &
    \begin{tabular}[b]{c}
    \fbox{\includegraphics[width=0.06\textwidth]{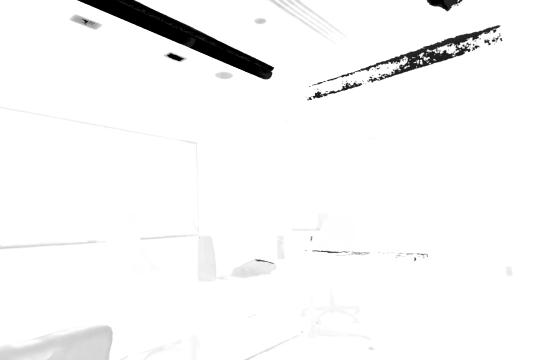}} \\
    \fbox{\includegraphics[width=0.06\textwidth]{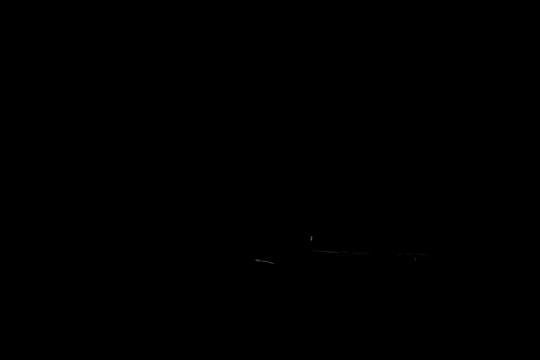}}
    \end{tabular}
    \end{tabular} \\
\hline
\begin{tabular}[b]{c}
\fbox{\includegraphics[width=0.13\textwidth]{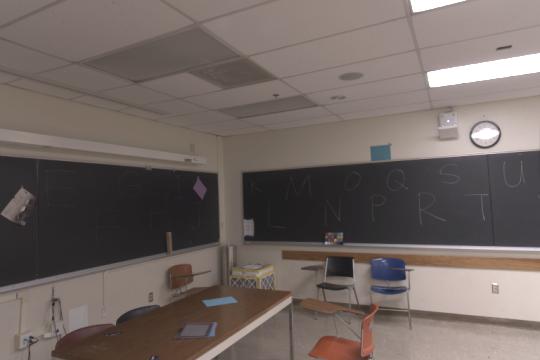}}
\end{tabular} &
\begin{tabular}[b]{cc}
    \fbox{\includegraphics[width=0.13\textwidth]{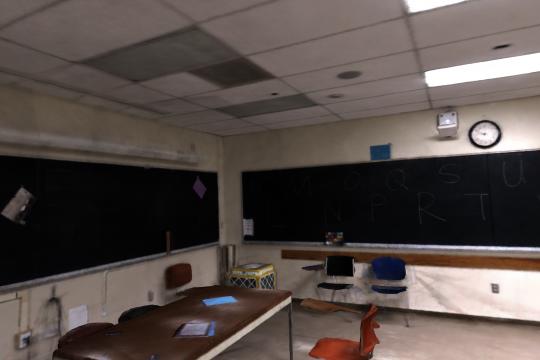}} &
    \begin{tabular}[b]{c}
    \fbox{\includegraphics[width=0.06\textwidth]{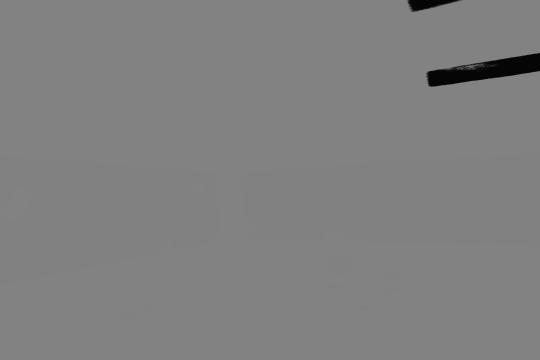}} \\
    \fbox{\includegraphics[width=0.06\textwidth]{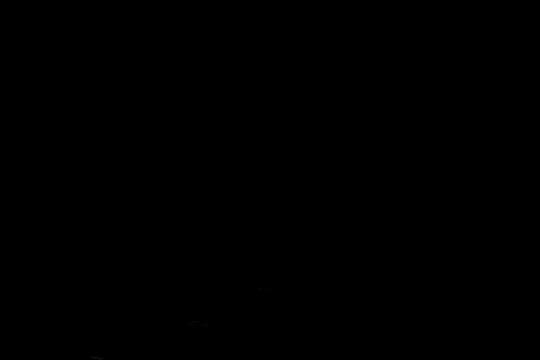}}
    \end{tabular}
    \end{tabular} &
\begin{tabular}[b]{cc}
    \fbox{\includegraphics[width=0.13\textwidth]{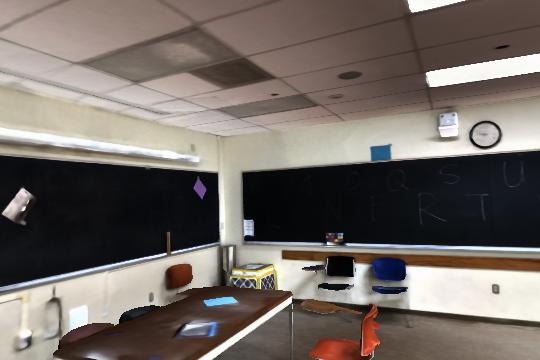}} &
    \begin{tabular}[b]{c}
    \fbox{\includegraphics[width=0.06\textwidth]{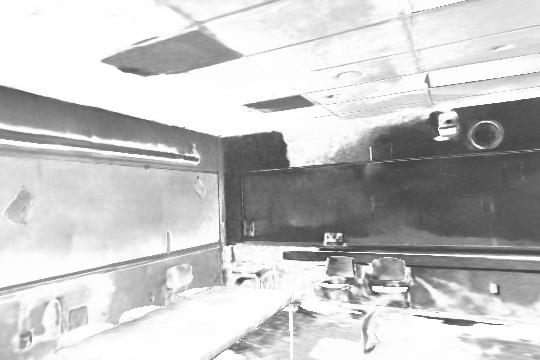}} \\
    \fbox{\includegraphics[width=0.06\textwidth]{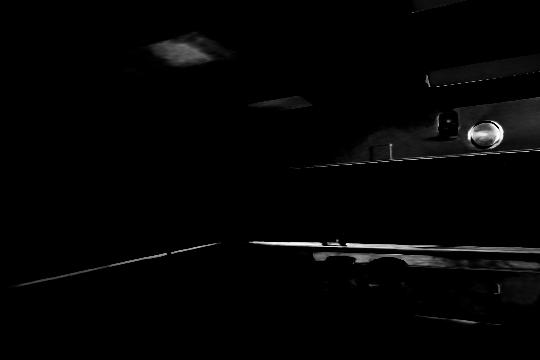}}
    \end{tabular}
    \end{tabular} &
\begin{tabular}[b]{cc}
    \fbox{\includegraphics[width=0.13\textwidth]{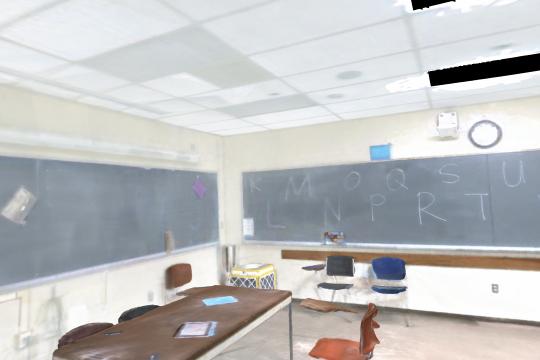}} &
    \begin{tabular}[b]{c}
    \fbox{\includegraphics[width=0.06\textwidth]{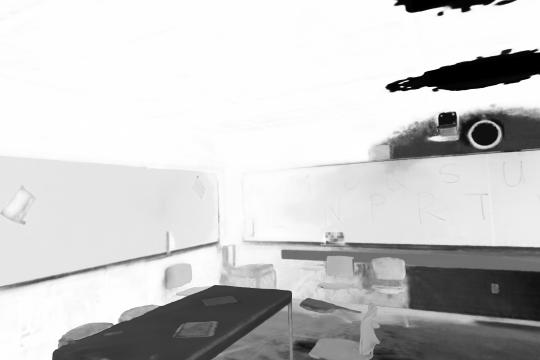}} \\
    \fbox{\includegraphics[width=0.06\textwidth]{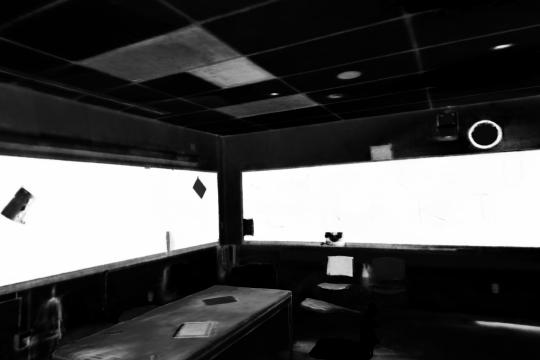}}
    \end{tabular}
    \end{tabular} &
\begin{tabular}[b]{cc}
    \fbox{\includegraphics[width=0.13\textwidth]{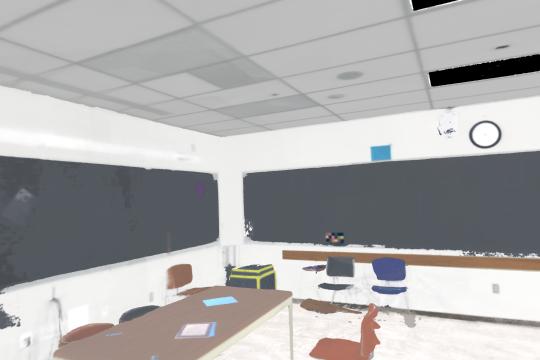}} &
    \begin{tabular}[b]{c}
    \fbox{\includegraphics[width=0.06\textwidth]{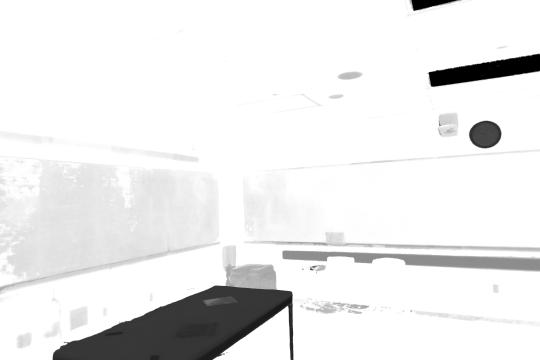}} \\
    \fbox{\includegraphics[width=0.06\textwidth]{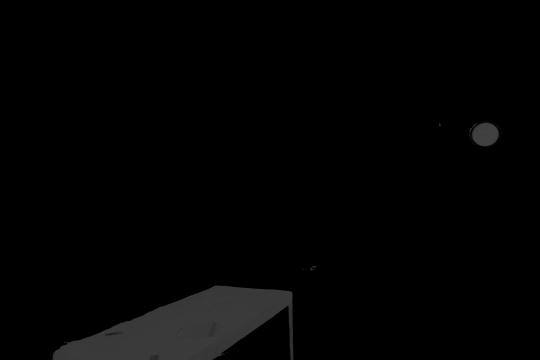}}
    \end{tabular}
    \end{tabular} \\
\begin{tabular}[b]{c}
\fbox{\includegraphics[width=0.13\textwidth]{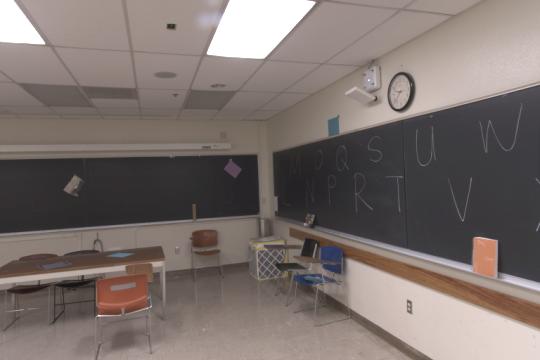}}
\end{tabular} &
\begin{tabular}[b]{cc}
    \fbox{\includegraphics[width=0.13\textwidth]{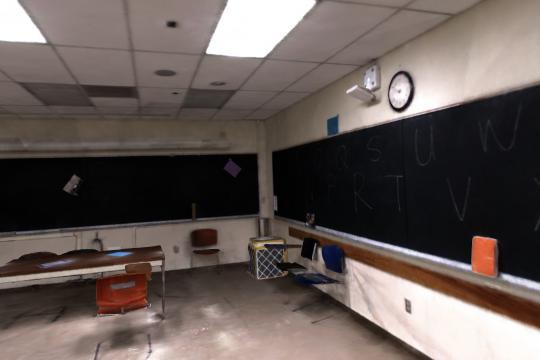}} &
    \begin{tabular}[b]{c}
    \fbox{\includegraphics[width=0.06\textwidth]{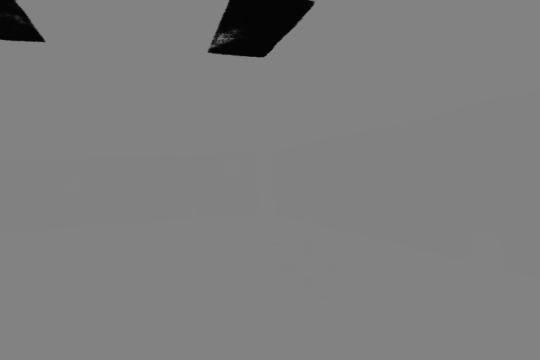}} \\
    \fbox{\includegraphics[width=0.06\textwidth]{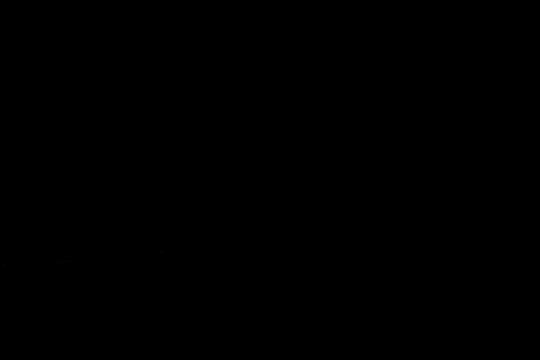}}
    \end{tabular}
    \end{tabular} &
\begin{tabular}[b]{cc}
    \fbox{\includegraphics[width=0.13\textwidth]{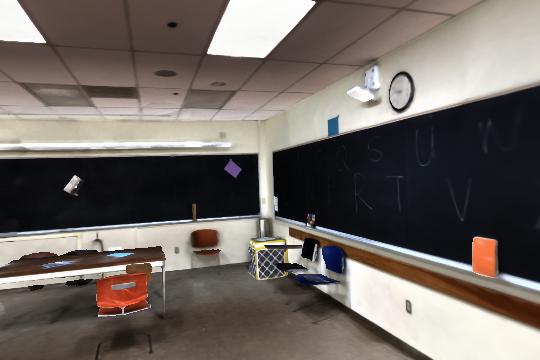}} &
    \begin{tabular}[b]{c}
    \fbox{\includegraphics[width=0.06\textwidth]{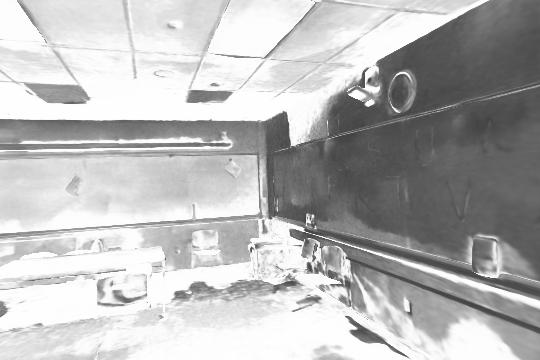}} \\
    \fbox{\includegraphics[width=0.06\textwidth]{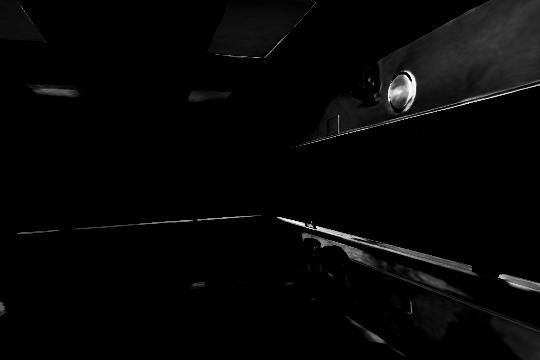}}
    \end{tabular}
    \end{tabular} &
\begin{tabular}[b]{cc}
    \fbox{\includegraphics[width=0.13\textwidth]{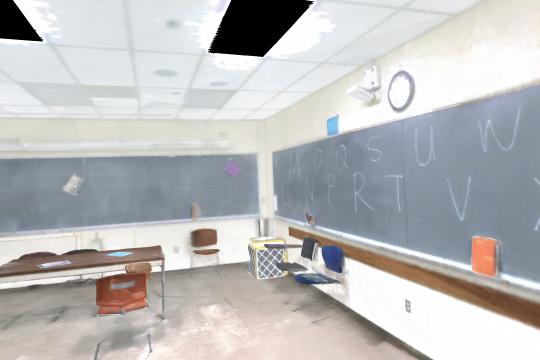}} &
    \begin{tabular}[b]{c}
    \fbox{\includegraphics[width=0.06\textwidth]{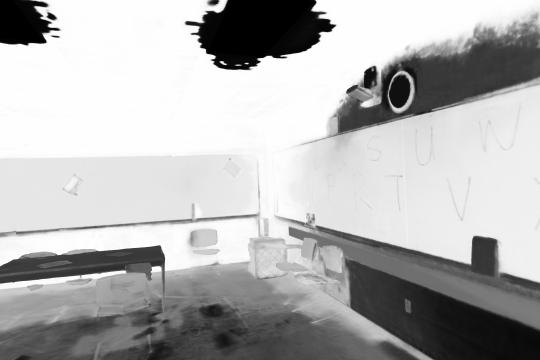}} \\
    \fbox{\includegraphics[width=0.06\textwidth]{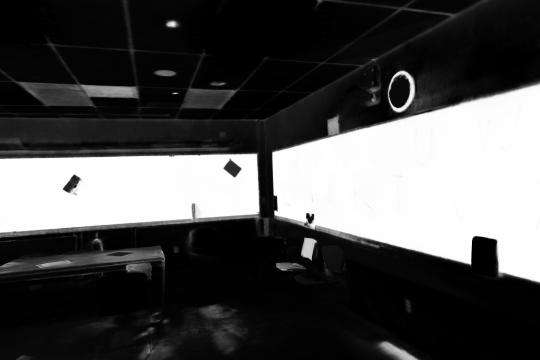}}
    \end{tabular}
    \end{tabular} &
\begin{tabular}[b]{cc}
    \fbox{\includegraphics[width=0.13\textwidth]{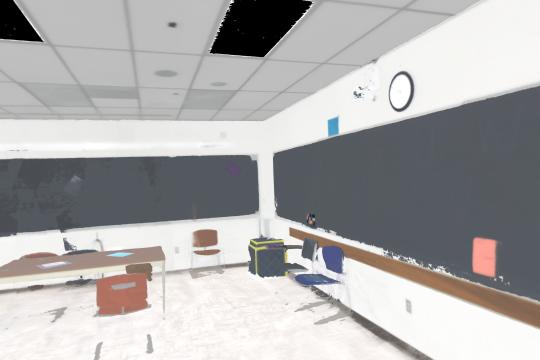}} &
    \begin{tabular}[b]{c}
    \fbox{\includegraphics[width=0.06\textwidth]{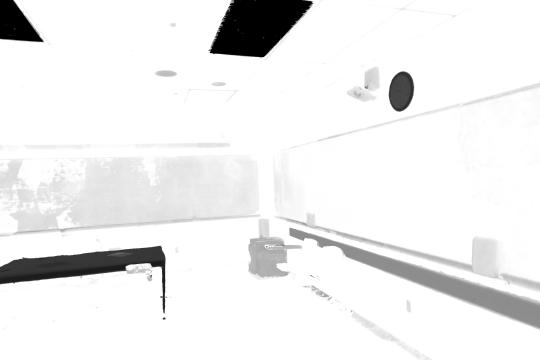}} \\
    \fbox{\includegraphics[width=0.06\textwidth]{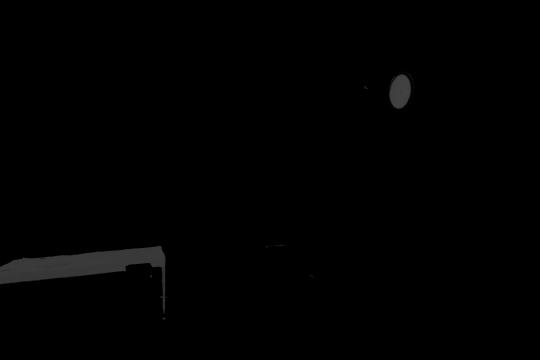}}
    \end{tabular}
    \end{tabular} \\
\hline
        % Header row at bottom
        \smash{\scriptsize RGB} &
        \smash{\scriptsize FIPT \cite{FIPT}} &
        \smash{\scriptsize NeILF++ \cite{NeilfPP}} &
        \smash{\scriptsize IRIS \cite{IRIS}} &
        \smash{\scriptsize IIF (Ours)}
    \end{tabular}}
    \caption{\textbf{Real \cite{FIPT} comparisons.} Additional samples on the real scenes of FIPT \cite{FIPT}.}
    \label{fig:supp:fipt_comparisons}
\end{figure*}

\begin{figure*}
    \centering
    \setlength\tabcolsep{1.25pt}
    \resizebox{\textwidth}{!}{
    \fboxsep=0pt
    
    \begin{tabular}{cccccc}    
        \fbox{\includegraphics[width=0.25\textwidth]{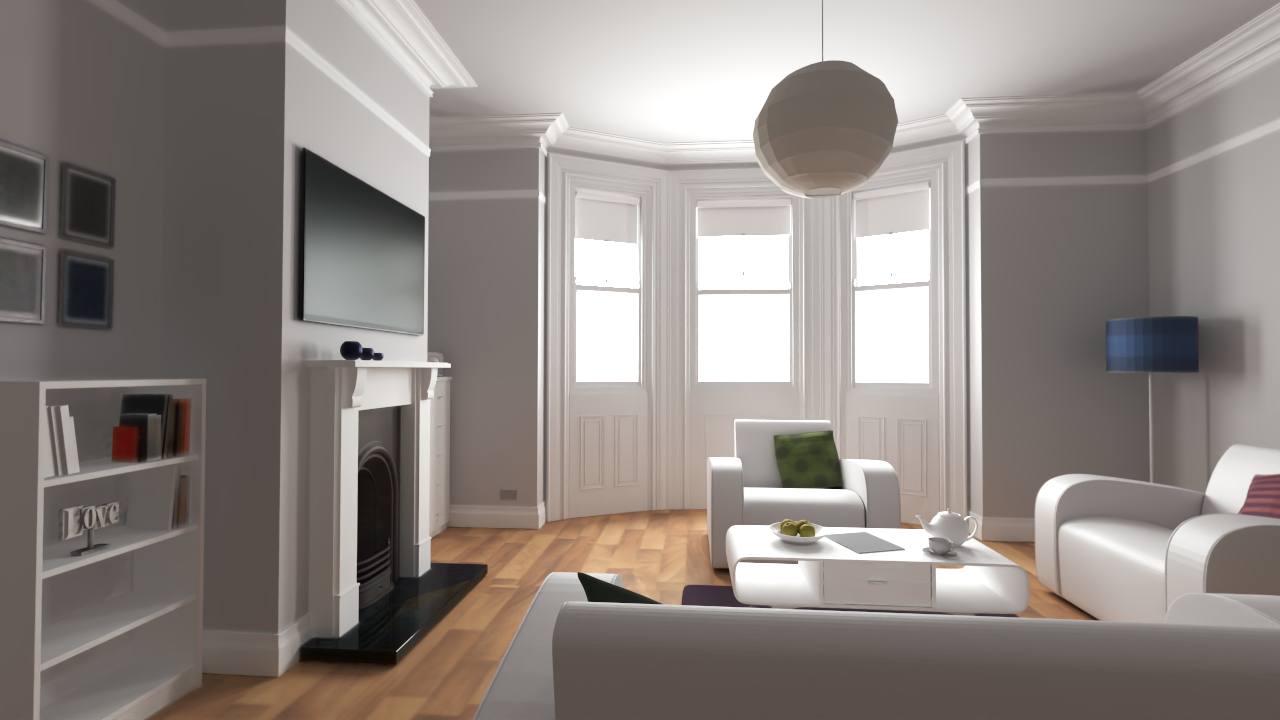}} 
        &
        \fbox{\includegraphics[width=0.25\textwidth]{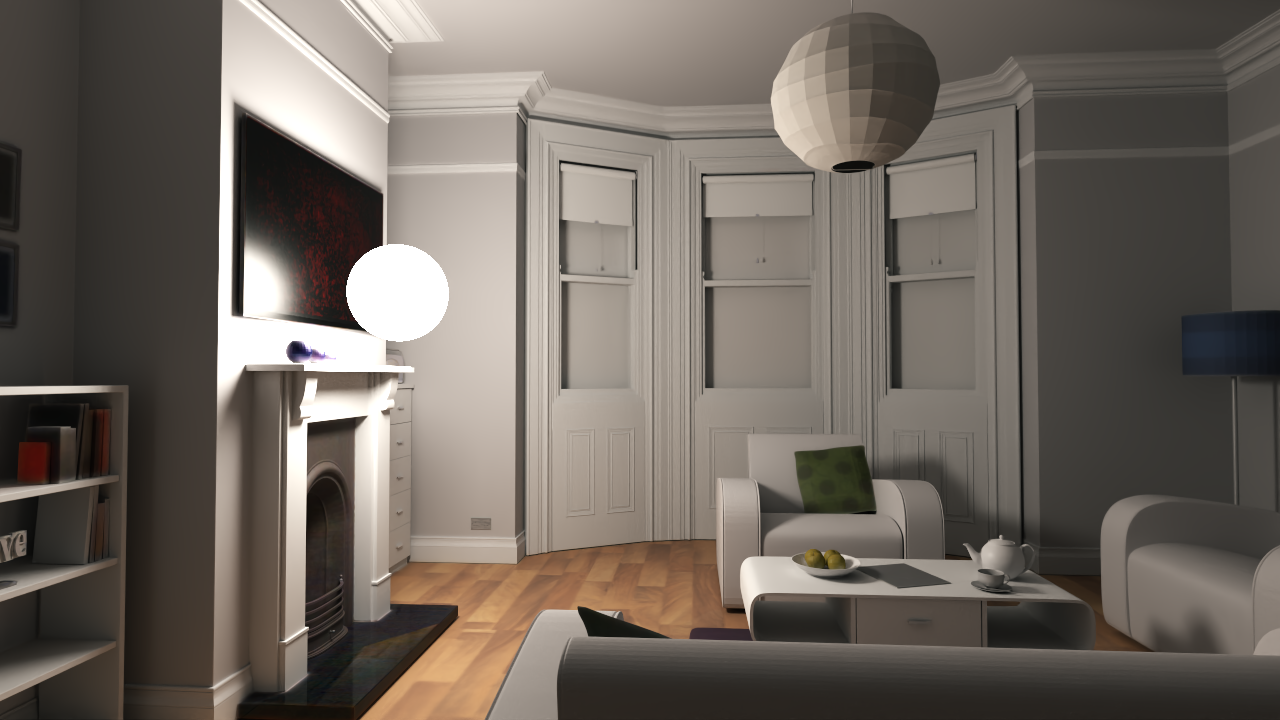}} &
        \fbox{\includegraphics[width=0.25\textwidth]{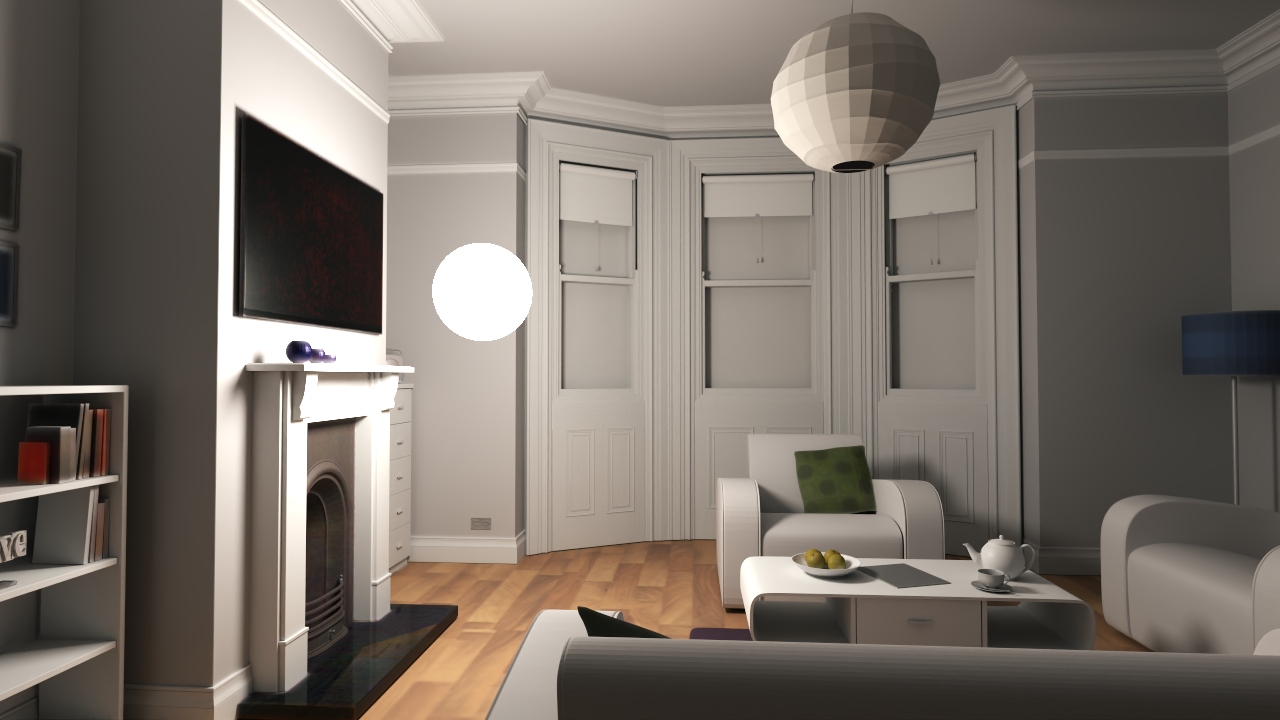}} &
        \fbox{\includegraphics[width=0.25\textwidth]{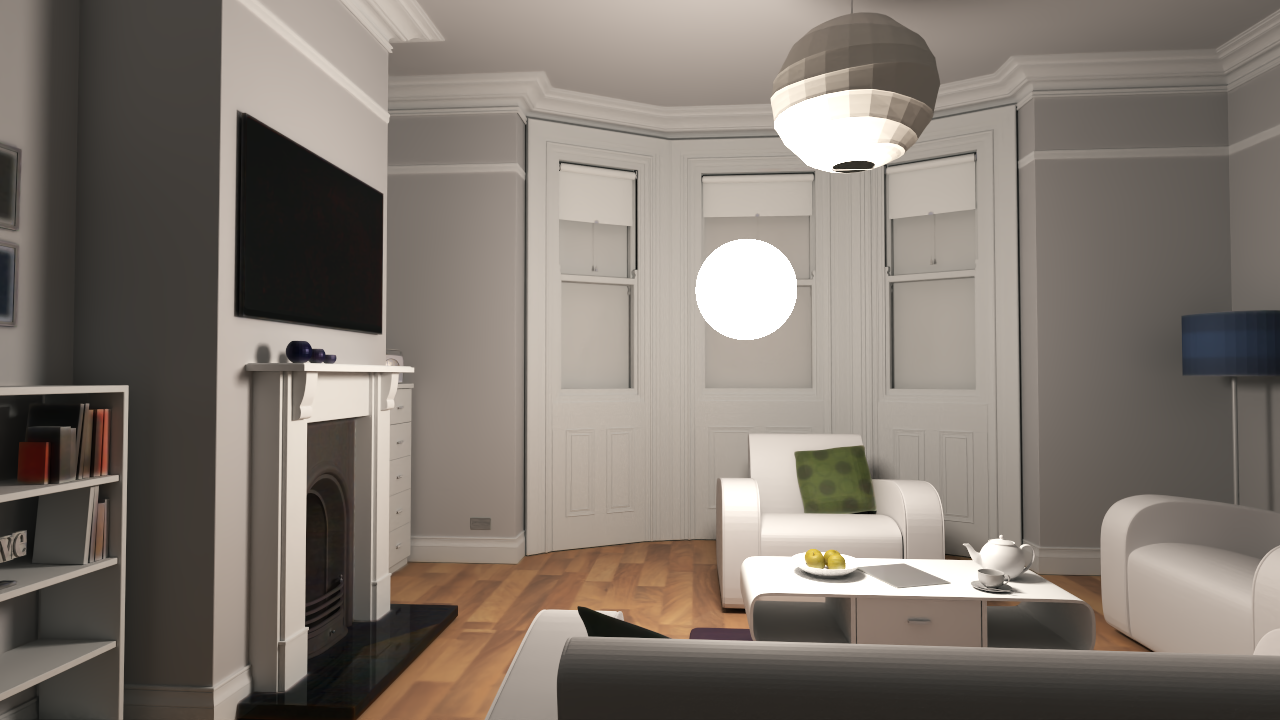}} &
        \fbox{\includegraphics[width=0.25\textwidth]{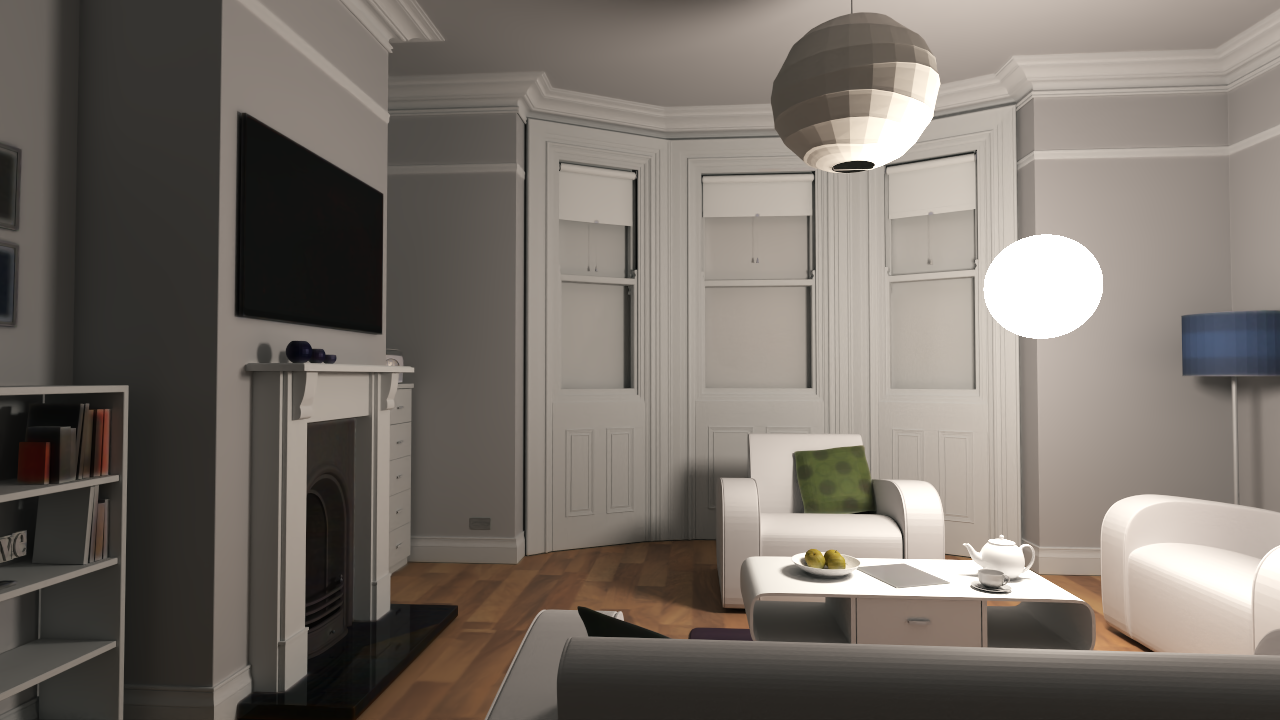}} &
        \fbox{\includegraphics[width=0.25\textwidth]{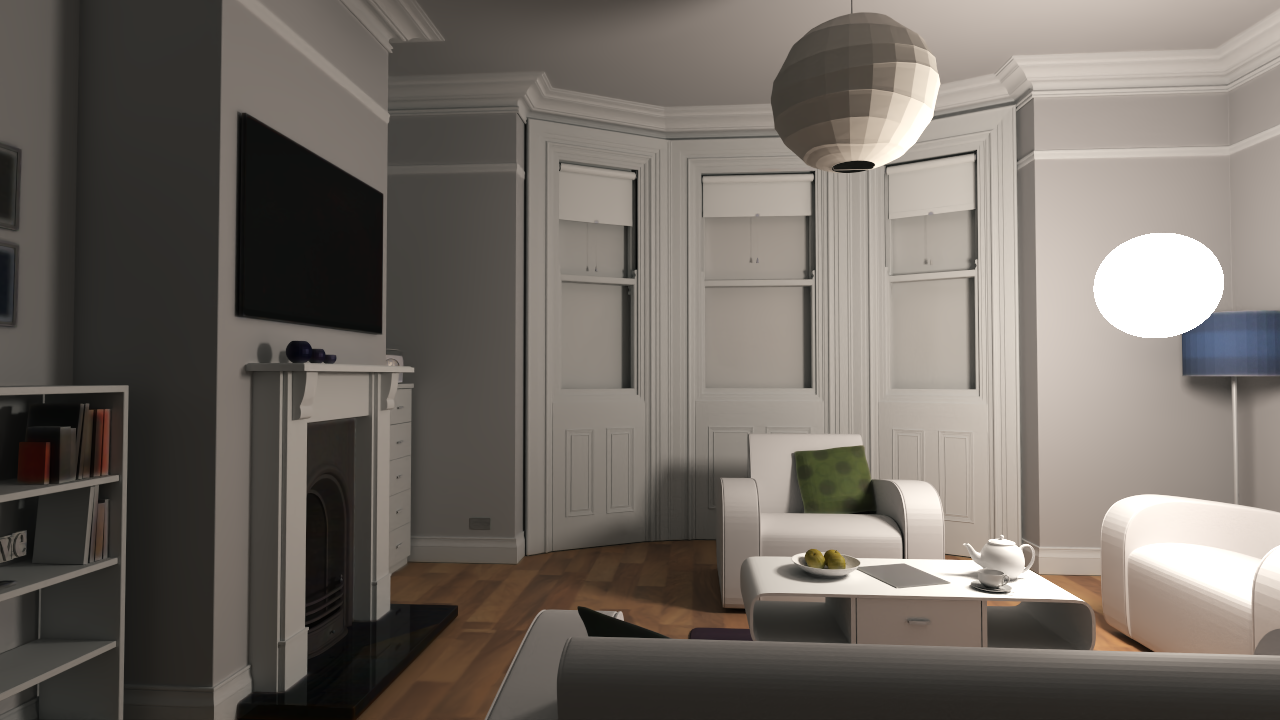}} \\
        \\
        \fbox{\includegraphics[width=0.25\textwidth]{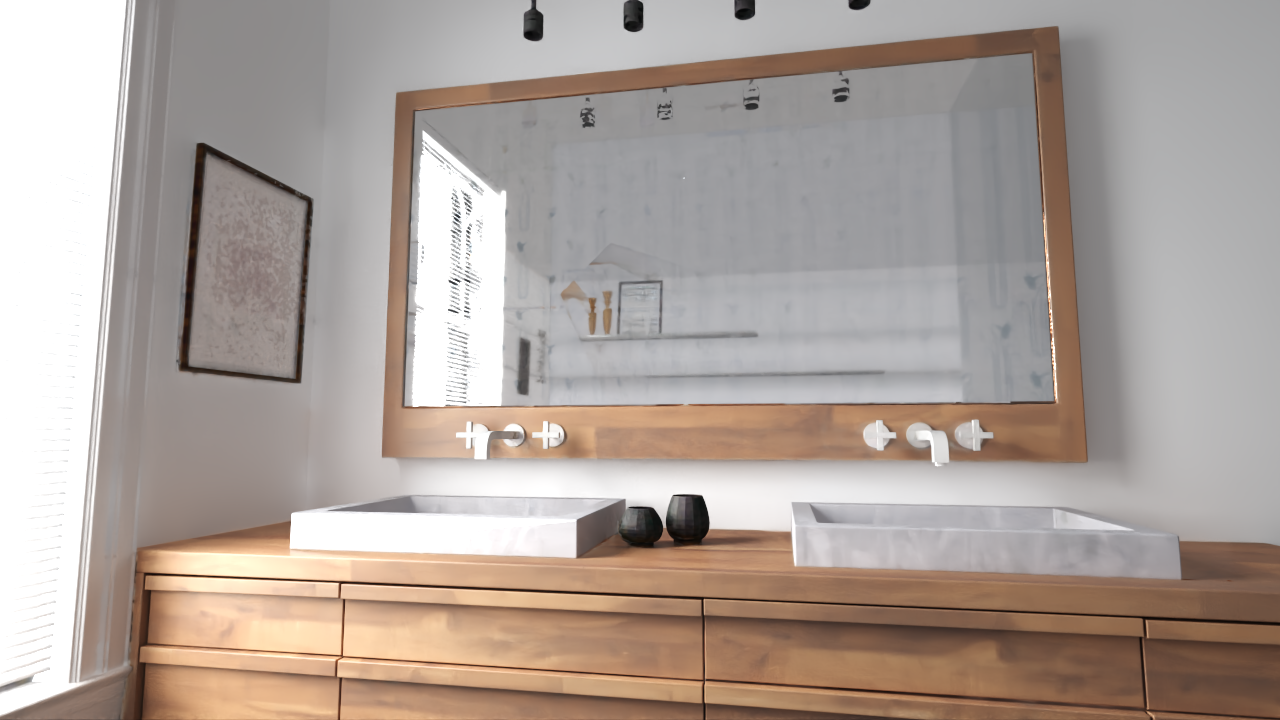}} 
        &
        \fbox{\includegraphics[width=0.25\textwidth]{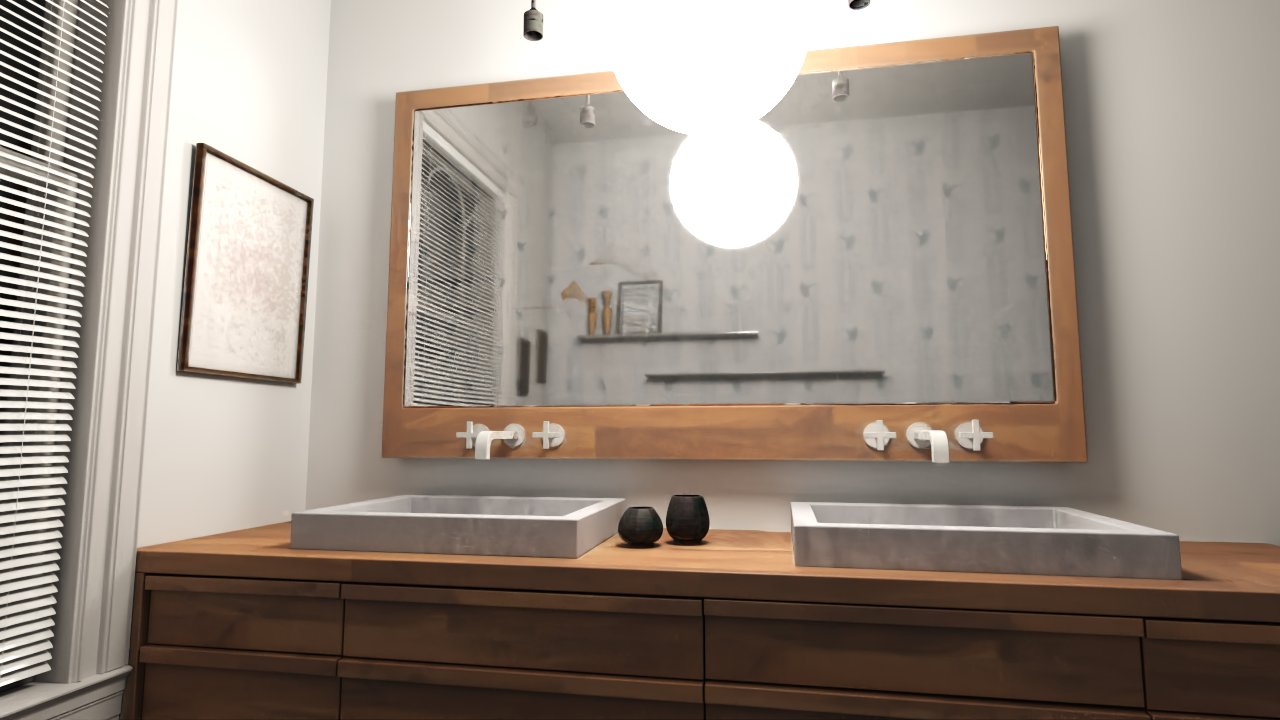}} &
        \fbox{\includegraphics[width=0.25\textwidth]{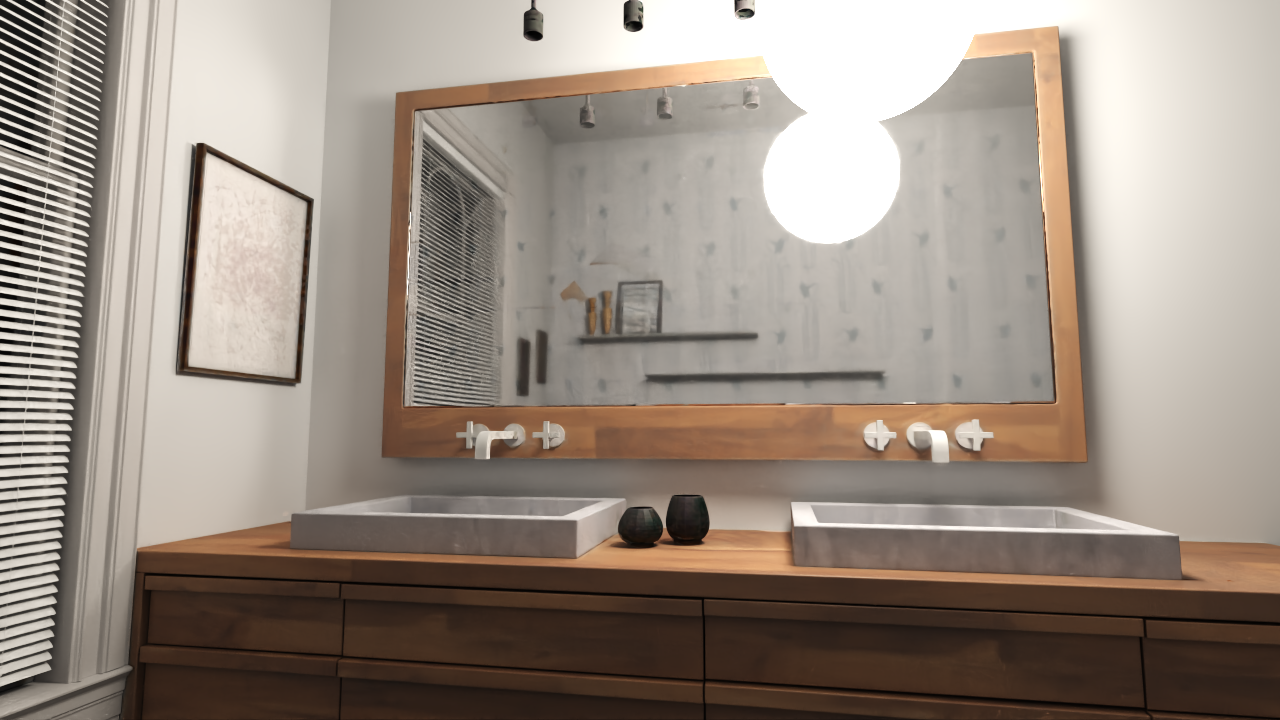}} &
        \fbox{\includegraphics[width=0.25\textwidth]{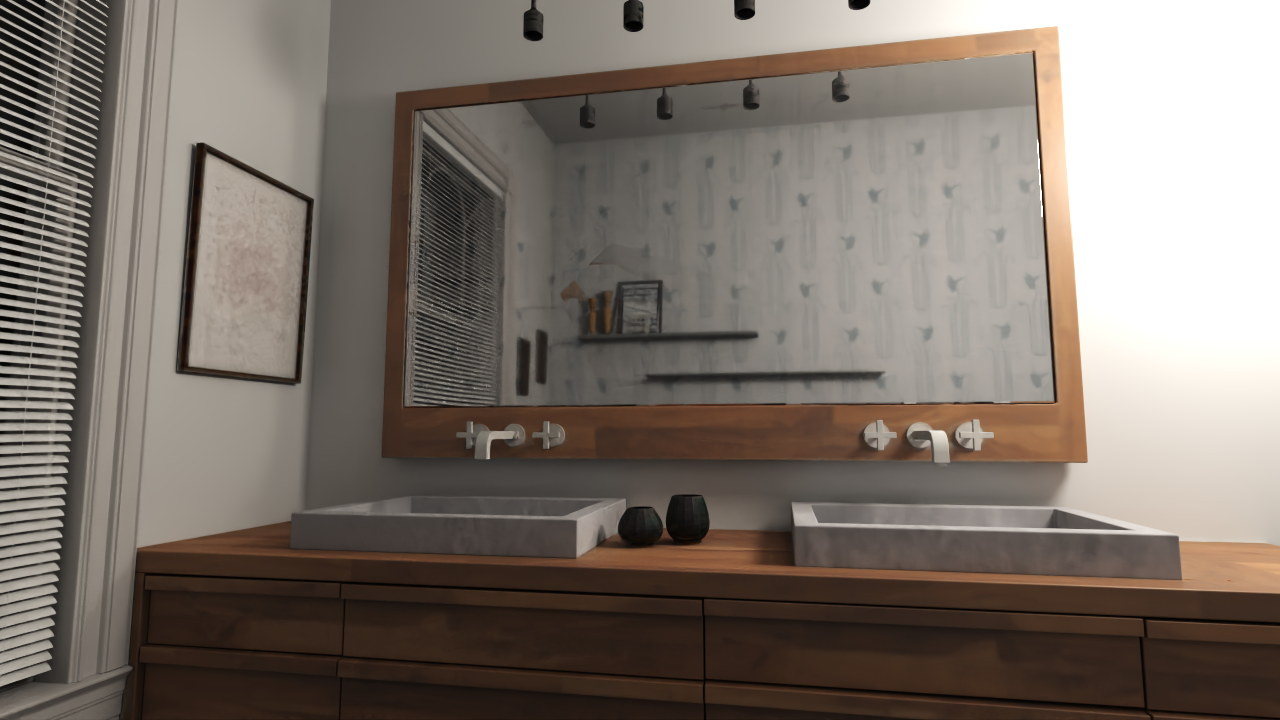}} &
        \fbox{\includegraphics[width=0.25\textwidth]{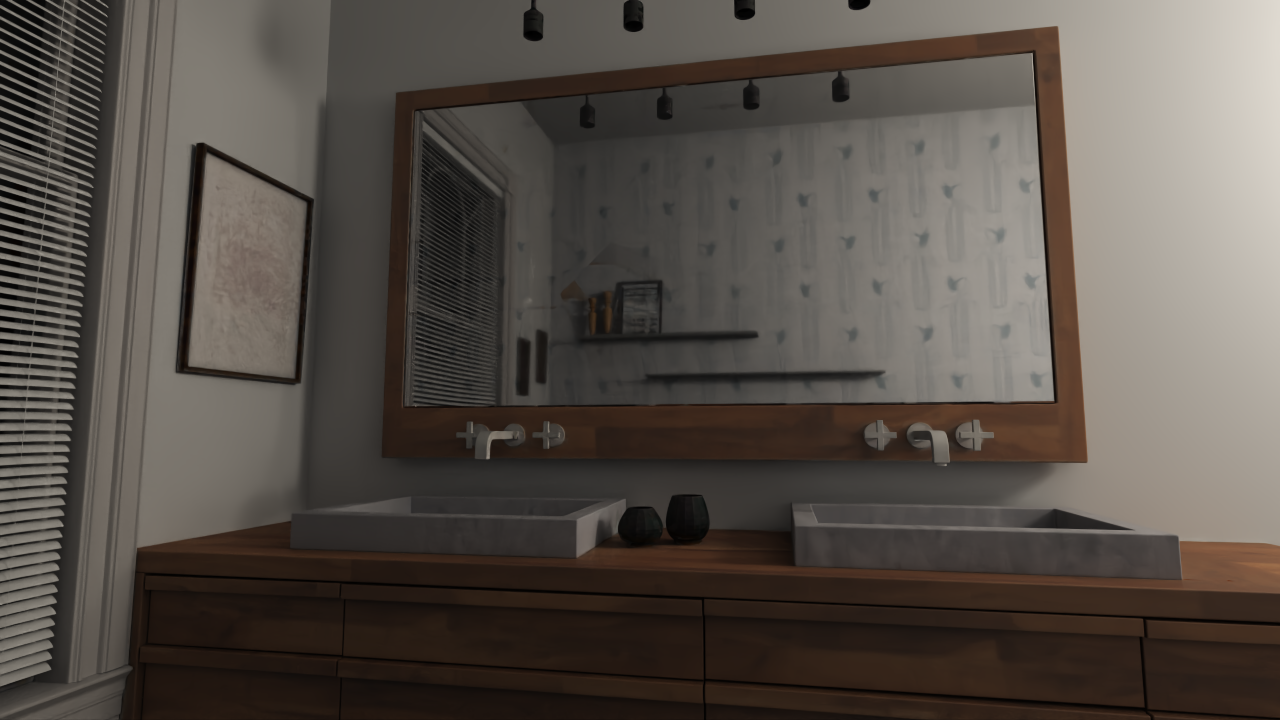}} &
        \fbox{\includegraphics[width=0.25\textwidth]{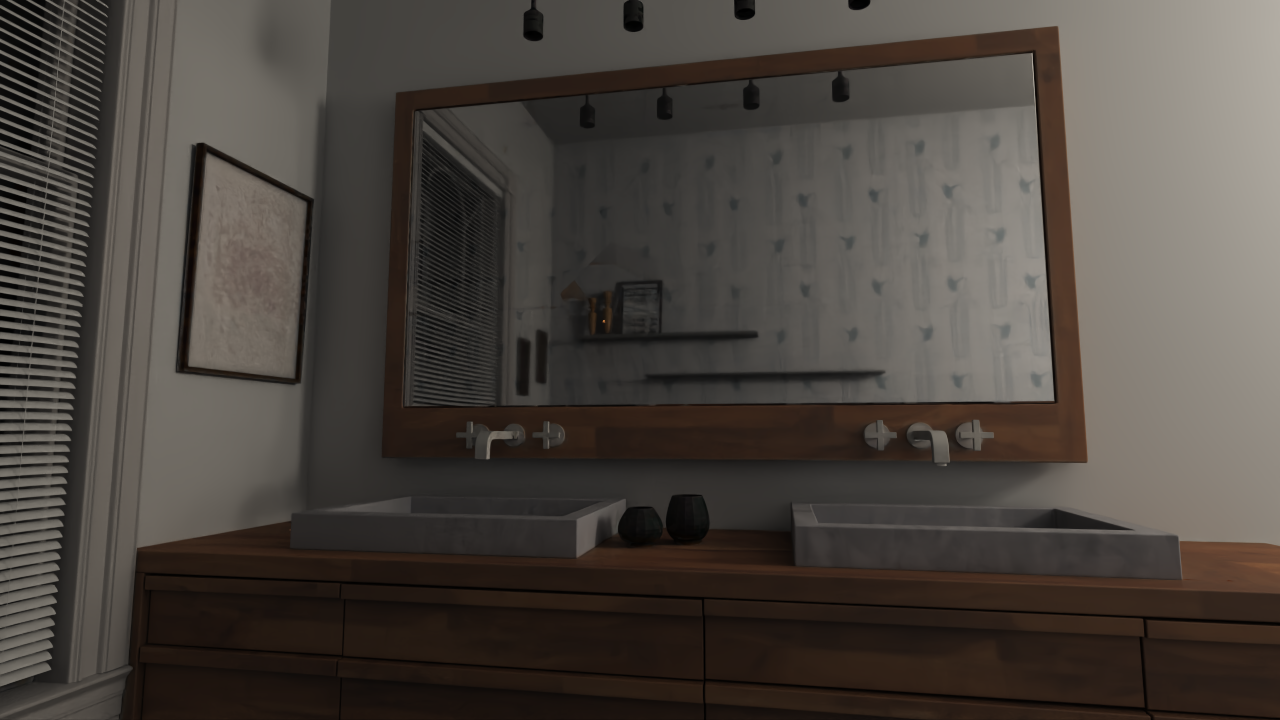}} \\
        \\
        \hline
        \fbox{\includegraphics[width=0.25\textwidth]{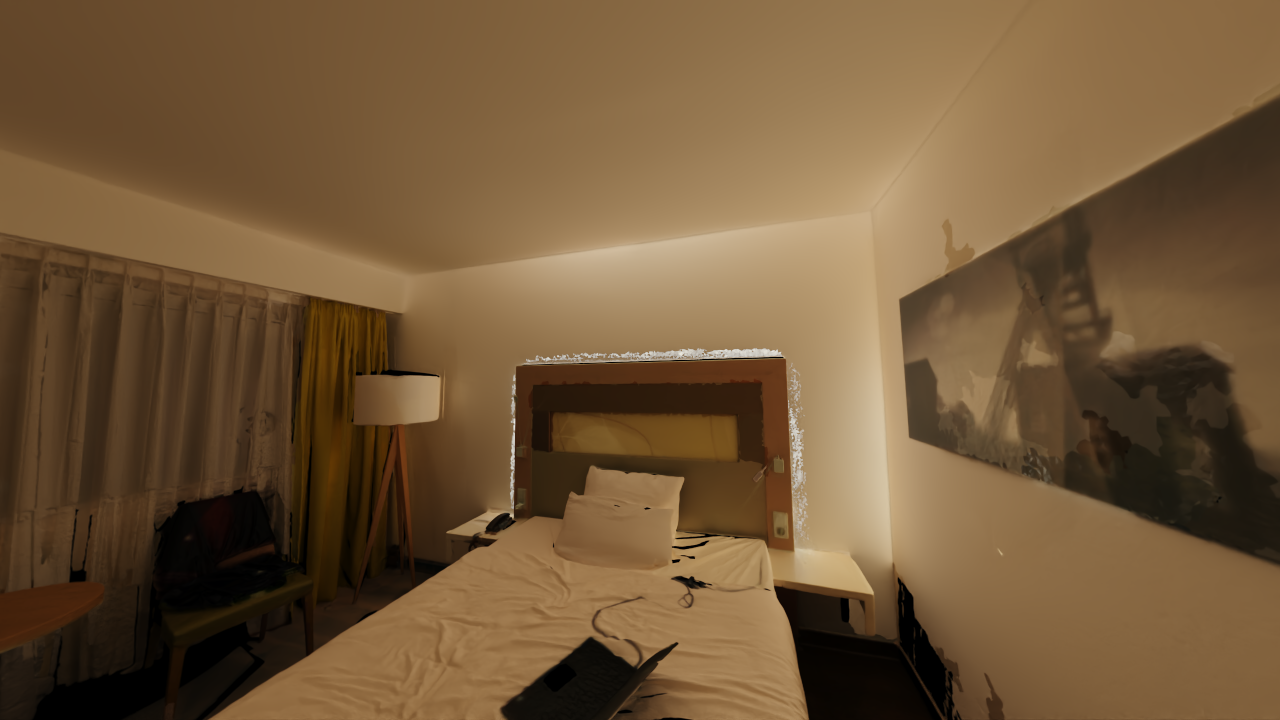}} 
        &
        \fbox{\includegraphics[width=0.25\textwidth]{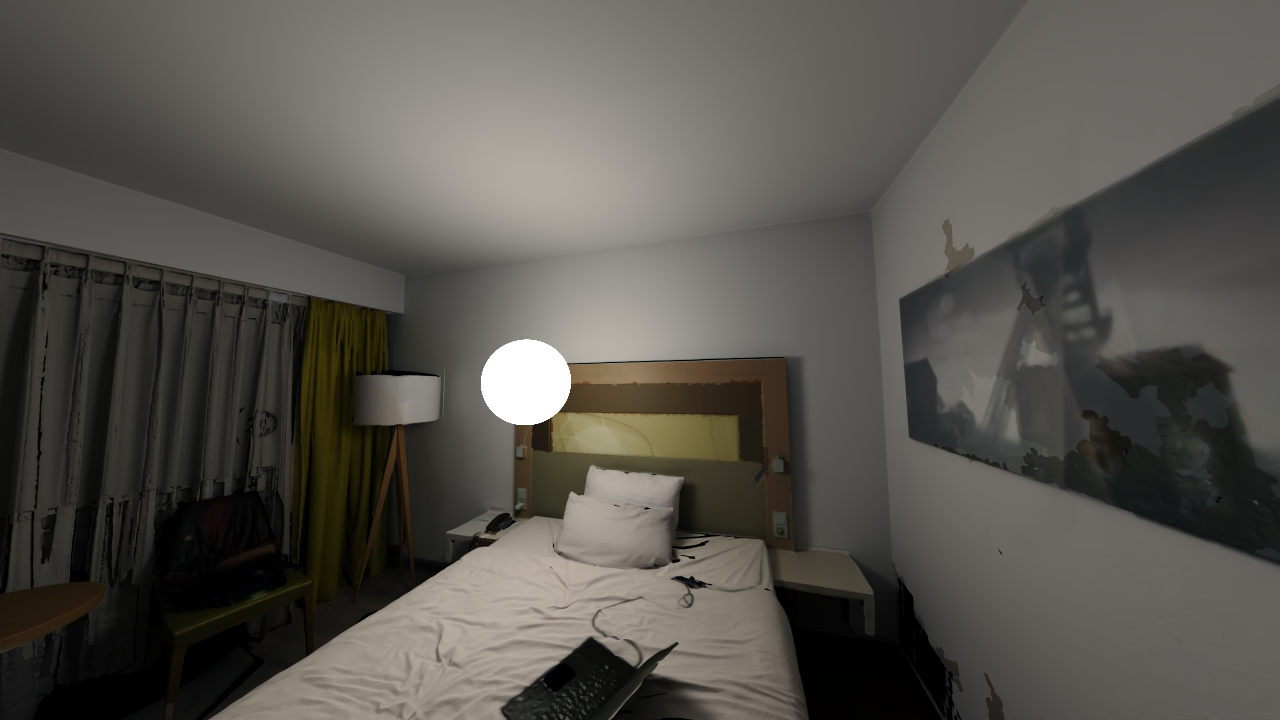}} &
        \fbox{\includegraphics[width=0.25\textwidth]{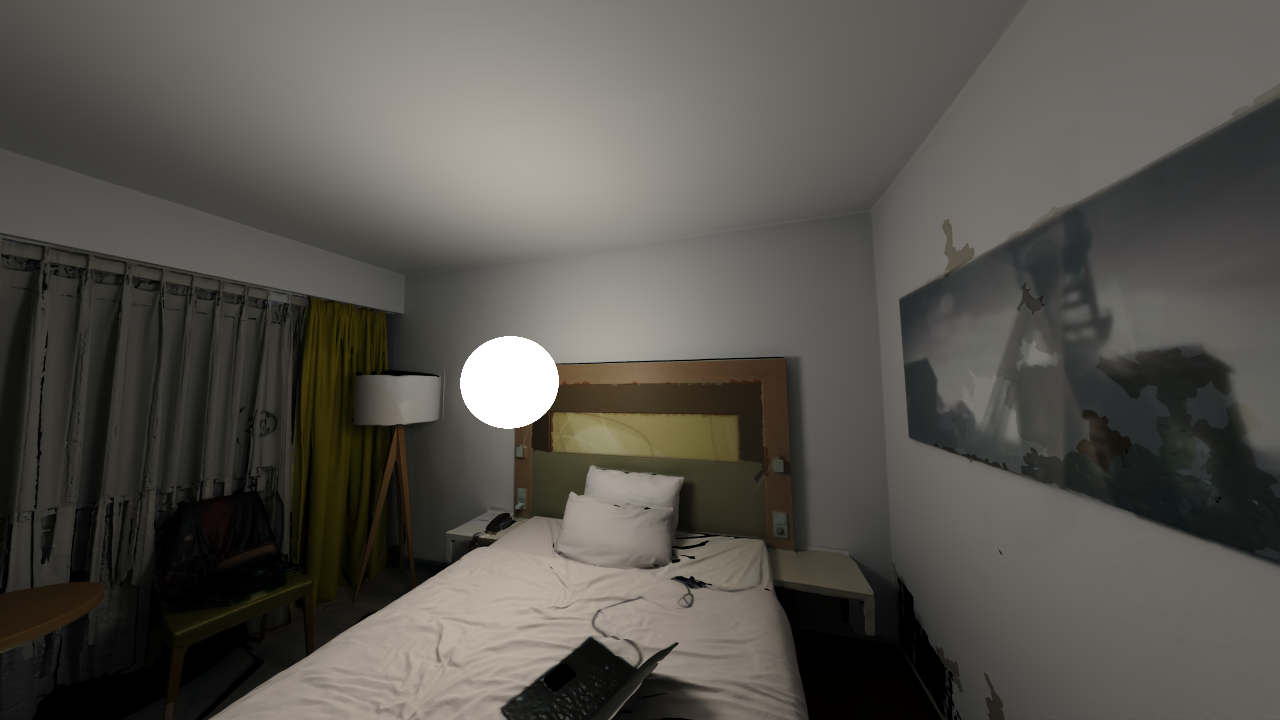}} &
        \fbox{\includegraphics[width=0.25\textwidth]{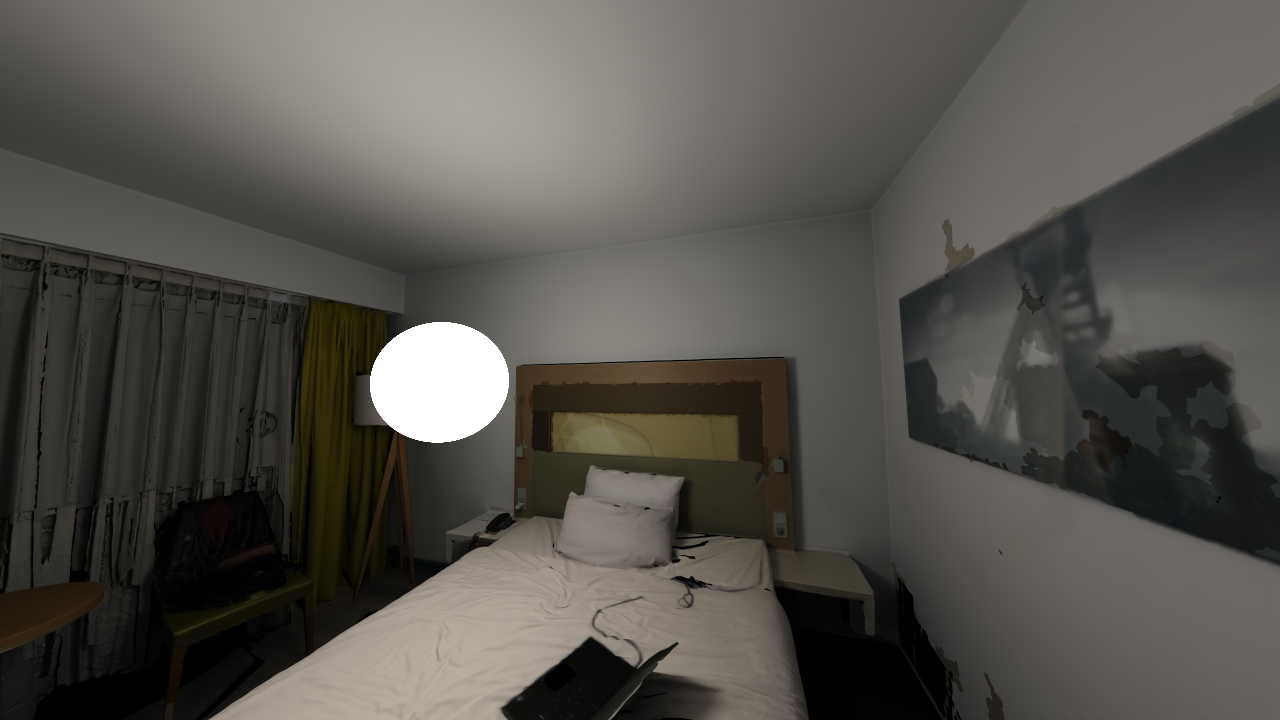}} &
        \fbox{\includegraphics[width=0.25\textwidth]{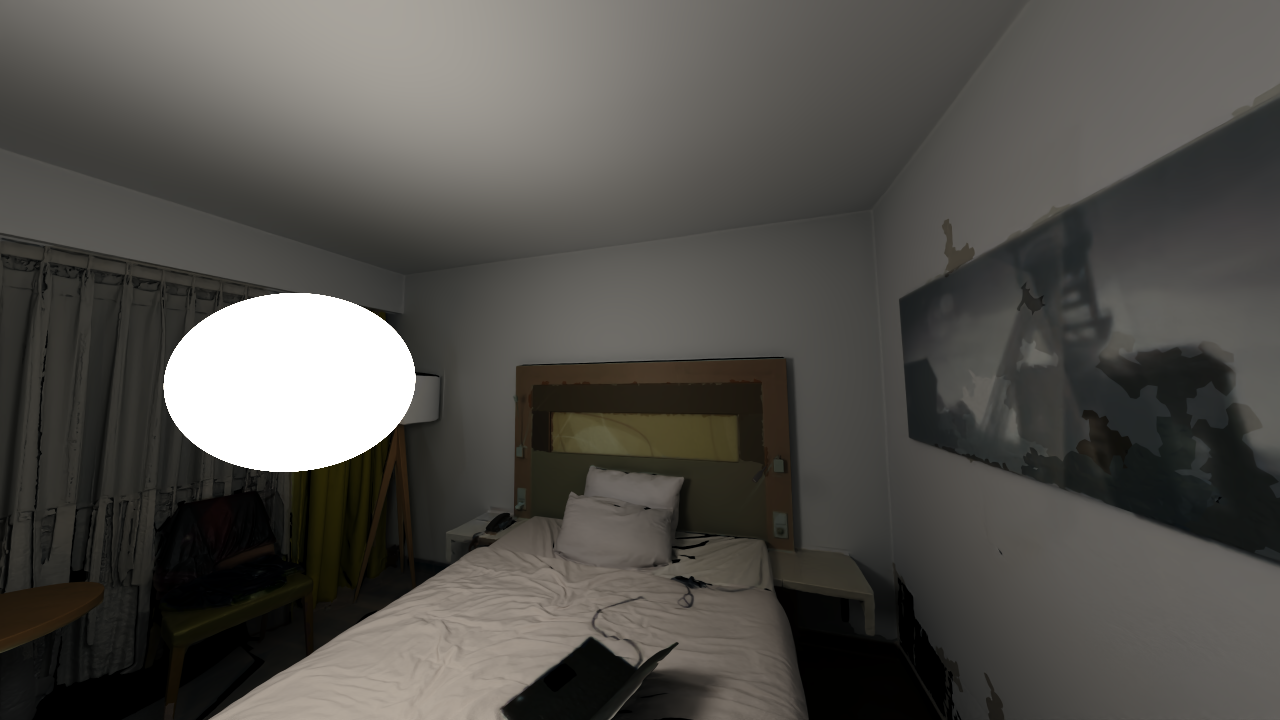}} &
        \fbox{\includegraphics[width=0.25\textwidth]{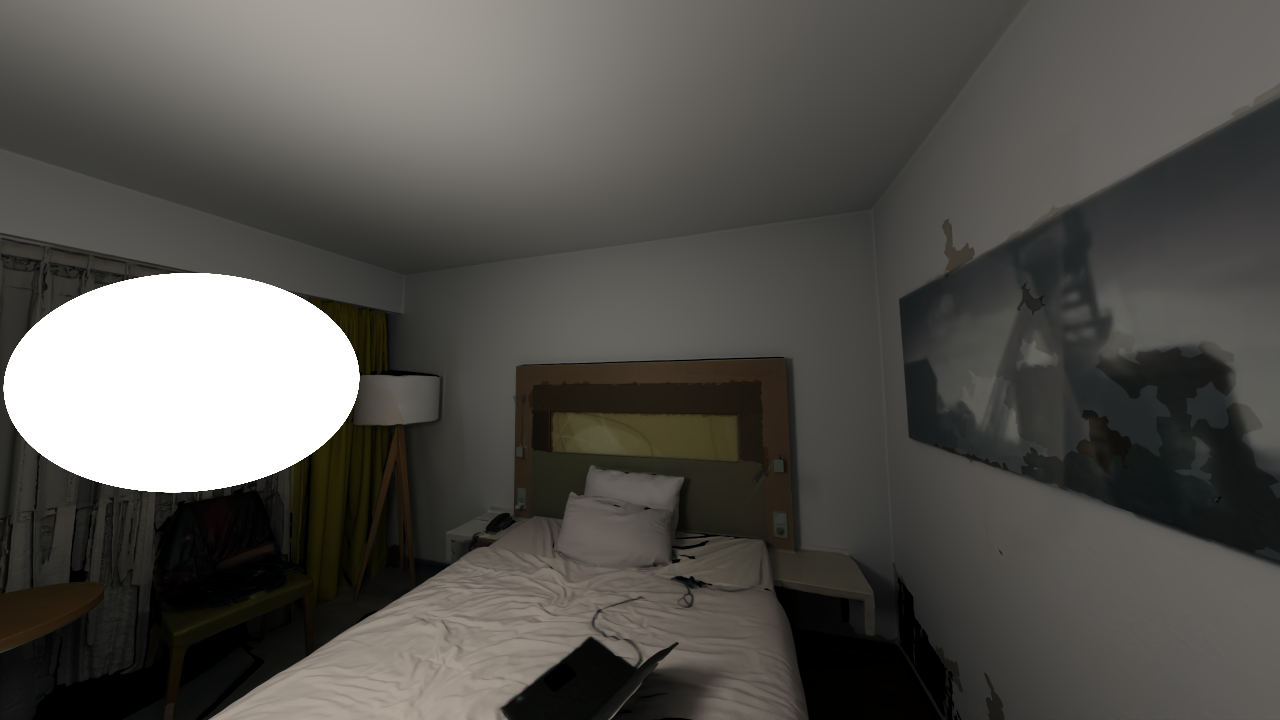}} \\
        \\
        Rendering & 
        Relighting 1 &
        Relighting 2 &
        Relighting 3 &
        Relighting 4 &
        Relighting 5 
    \end{tabular}}
    \vspace{-8pt}
    \caption{\textbf{Additional relightings.} 
    We show additional relighting results on synthetic and ScanNet++ scenes over a smooth trajectory of an emissive sphere. For additional interpolations, please refer to our video. 
    }
    \label{fig:supp:applications}
\end{figure*}

\paragraph{Relighting. }
Our supplementary video provides comparisons and relightings with rendered trajectories. 
We provide additional results for all the synthetic scenes in \Cref{fig:supp:synthetic_comparisons}, for the real ScanNet++ (2a1b555966, 651dc6b4f1, a003a6585e) scenes in \Cref{fig:supp:scannetpp_comparisons}. 
We compare also on the real scenes of FIPT \cite{FIPT} in \Cref{fig:supp:fipt_comparisons}, which uses photo-metric stereo for the mesh reconstruction. 
We provide more results on syntetic and ScanNet++ \cite{ScanNet++} scenes in \Cref{fig:supp:applications}. 

\begin{figure*}
    \centering
    \setlength\tabcolsep{2pt}
    \resizebox{\linewidth}{!}{
    \fboxsep=0pt
    \begin{tabular}{c | cc | ccc}
        % --- Row 1: filename 054_0001 ---
        \fbox{\includegraphics[width=0.16\linewidth]{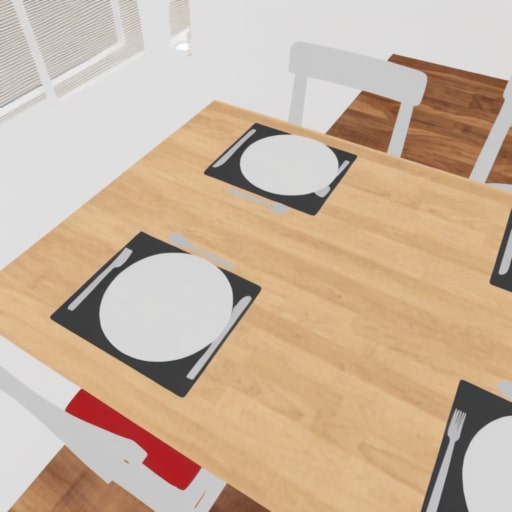}} &
        \fbox{\includegraphics[width=0.16\linewidth]{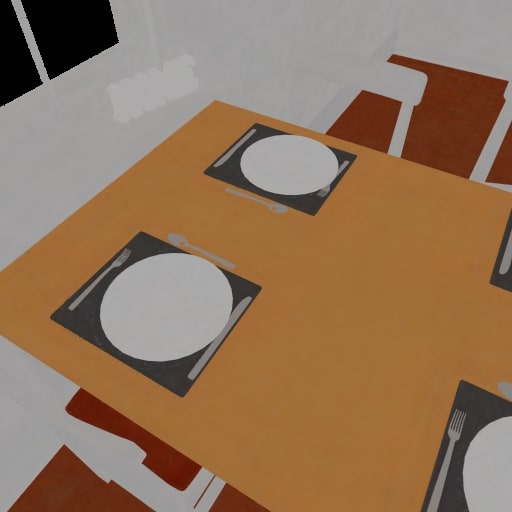}} &
        \fbox{\includegraphics[width=0.16\linewidth]{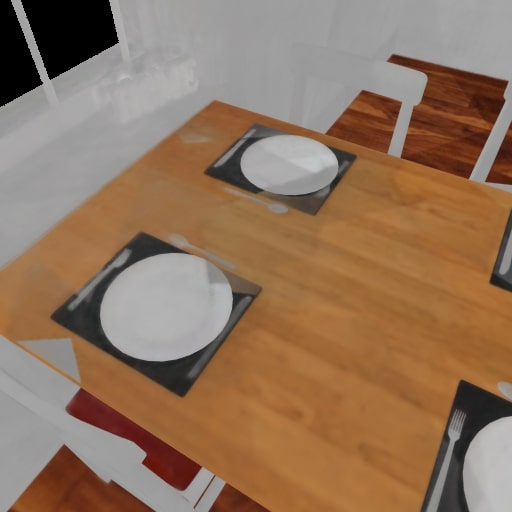}} &
        % \fbox{\includegraphics[width=0.16\linewidth]{res/ablation/synthetic/kitchen/albedo/054_0001_parametric.jpg}} &
        % \fbox{\includegraphics[width=0.16\linewidth]{res/ablation/synthetic/kitchen/albedo/054_0001_ours.jpg}} &

        \begin{tikzpicture}
          \node[inner sep=0pt] (a)
            {\fbox{\includegraphics[width=0.16\linewidth]{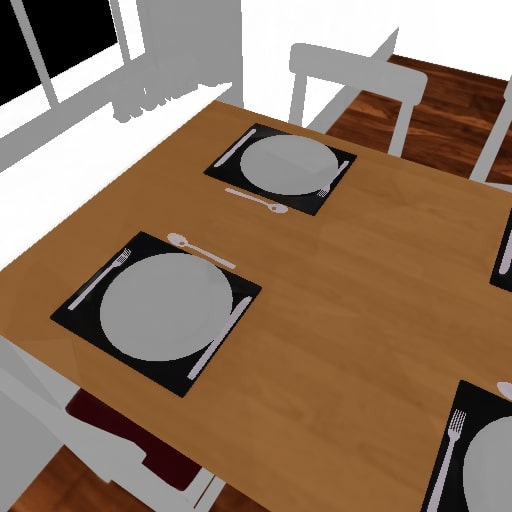}}};
          \begin{scope}
            \clip (a.south west) rectangle (a.north east);
            \node[draw=red, line width=1pt, circle,
                  minimum width=30pt, minimum height=30pt,
                  xshift=+12pt, yshift=-18pt] at (a.center) {};
          \end{scope}
        \end{tikzpicture}
        &
        \begin{tikzpicture}
          \node[inner sep=0pt] (a)
            {\fbox{\includegraphics[width=0.16\linewidth]{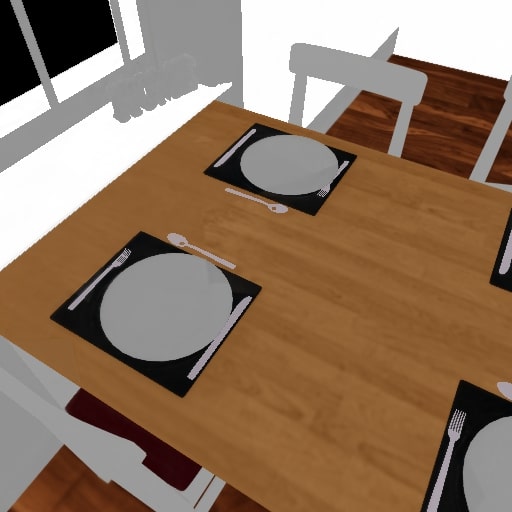}}};
          \begin{scope}
            \clip (a.south west) rectangle (a.north east);
            \node[draw=green, line width=1pt, circle,
                  minimum width=30pt, minimum height=30pt,
                  xshift=+12pt, yshift=-18pt] at (a.center) {};
          \end{scope}
        \end{tikzpicture}
        &
        
        \fbox{\includegraphics[width=0.16\linewidth]{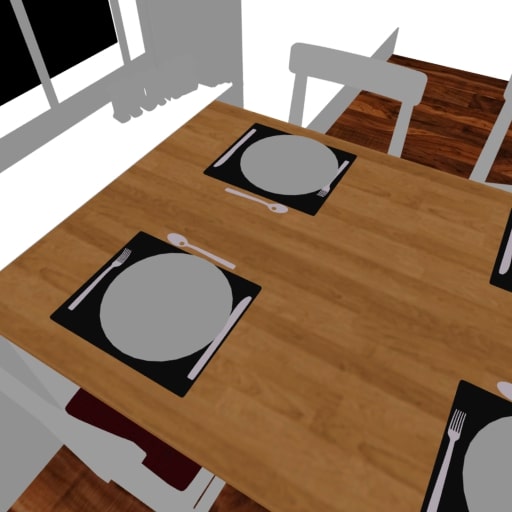}}
        \\[4pt]

        % --- Row 2: filename 003_0001 ---
        \fbox{\includegraphics[width=0.16\linewidth]{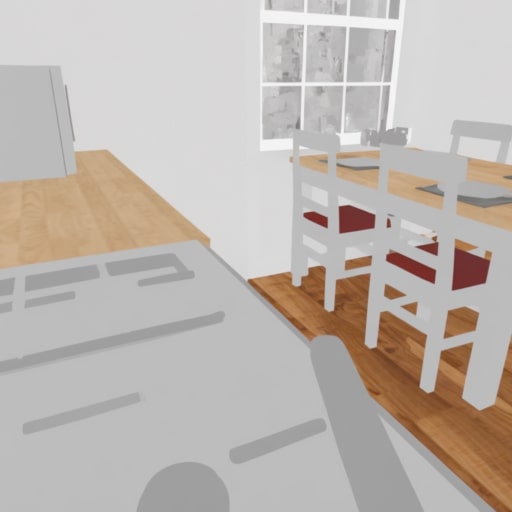}} &
        \fbox{\includegraphics[width=0.16\linewidth]{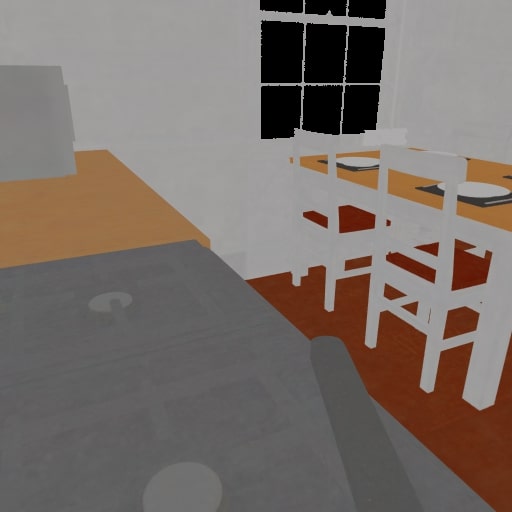}} &
        \fbox{\includegraphics[width=0.16\linewidth]{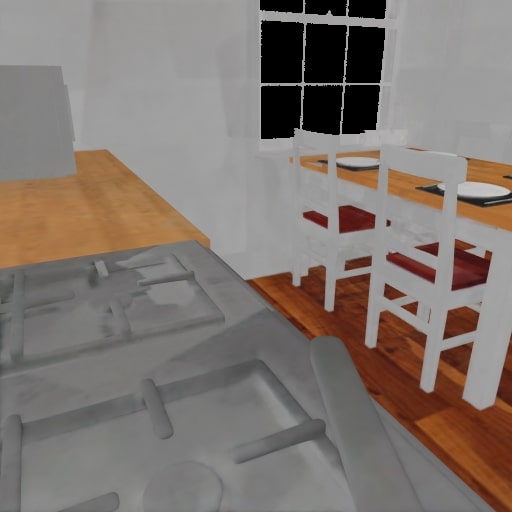}} &
        % \fbox{\includegraphics[width=0.16\linewidth]{res/ablation/synthetic/kitchen/albedo/003_0001_parametric.jpg}} &
        % \fbox{\includegraphics[width=0.16\linewidth]{res/ablation/synthetic/kitchen/albedo/003_0001_ours.jpg}} &

        \begin{tikzpicture}
          \node[inner sep=0pt] (a)
            {\fbox{\includegraphics[width=0.16\linewidth]{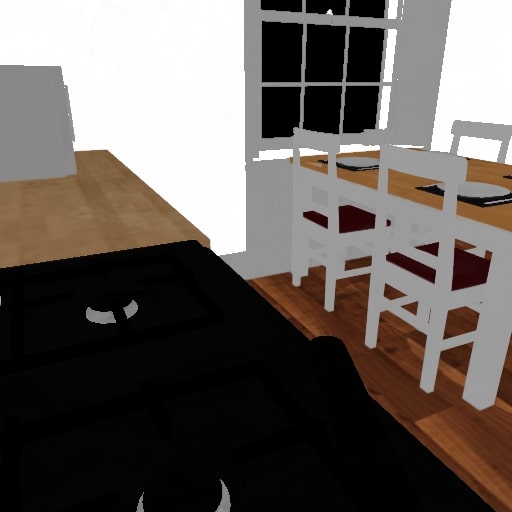}}};
          \begin{scope}
            \clip (a.south west) rectangle (a.north east);
            \node[draw=red, line width=1pt, circle,
                  minimum width=30pt, minimum height=30pt,
                  xshift=-23pt, yshift=+3pt] at (a.center) {};
          \end{scope}
        \end{tikzpicture}
        &
        \begin{tikzpicture}
          \node[inner sep=0pt] (a)
            {\fbox{\includegraphics[width=0.16\linewidth]{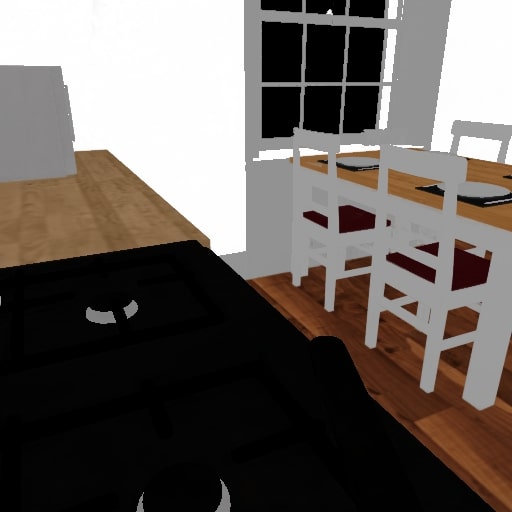}}};
          \begin{scope}
            \clip (a.south west) rectangle (a.north east);
            \node[draw=green, line width=1pt, circle,
                  minimum width=30pt, minimum height=30pt,
                  xshift=-23pt, yshift=+3pt] at (a.center) {};
          \end{scope}
        \end{tikzpicture}
        &
        
        \fbox{\includegraphics[width=0.16\linewidth]{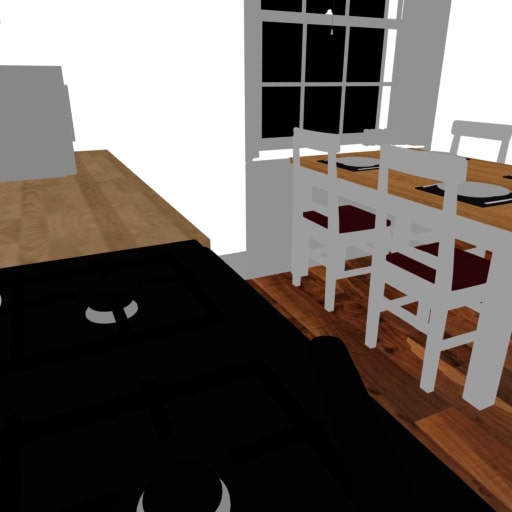}}
        \\[6pt]

        % --- Row 3: category labels at bottom ---
        RGBX \cite{RGBX} &
        Per-Object \cite{IRIS} &
        Per-Texel &
        w/ Parametric &
        Ours &
        GT
    \end{tabular}
    }

    \caption{\textbf{Cross-view aggregation (additional results).}
    Single-view material estimation can yield detailed but inconsistent predictions (\cref{fig:method:motivation}). IRIS \cite{IRIS} uses per-object aggregation, losing patterns. Per-texel aggregation maintains patterns but introduces seams. Our parametric modeling (\cref{sec:method:single_view}) provides a low-dimensional space of consistent 3D aggregations. Distribution matching (\cref{sec:method:cross_view}) selects the best predictions per view to preserve fine details.
    }
    \label{fig:exp:cross_view_supp}
\end{figure*}

\paragraph{Cross-View Aggregation Ablation. }
We provide additional ablation results on our cross-view aggregation strategies in \Cref{fig:exp:cross_view_supp}.

\paragraph{Temperature Ablation. }
We ablate the effect of the temperature annealing factor in \Cref{fig:supp:tau} to evaluate the effectiveness of the  assignments.
The prediction assignments $\alpha_{i,k}$ converge to binary (red=select, blue=drop).
Their entropy (right) drops from 0.5 (naive averaging) to 0 (mode selection); $\tau_\text{anneal}{=}0.85$ balances convergence and exploration.

\paragraph{Geometry Ablation. }
In \Cref{fig:supp:geom} we show qualitative comparison between input geometries obtained with laser scan or with photometric reconstruction on a ScanNet++ scene (7e09430da7), showing that our method gives comparable results even with lower quality geometry.

\paragraph{Failure cases. }
Since our method depends on reconstructed geometry, we inherit their limitations.
Thin structures and semi-transparent object are hard to reconstruct, often leading to missing or incorrect geometry. 
We show such failures in \Cref{fig:supp:fail}
Furthermore, consistently wrong predictions can leak into the reconstruction.
The book in \Cref{fig:method:motivation} shows white and grey caused by incorrect predictions.

% Huge table

% Huge table, left: conditioning (text + component, in and out of domain), right: all the components

% \paragraph{List of things to add}
% \begin{itemize}[leftmargin=*,topsep=1pt, noitemsep]
%     \item data preprocessing, how to obtain segmentations in 2d and 3d. both synthetic and real (only hinted in exp section)
%     \item how is this optimized and used: lighting representation as environment map for real world
%     \item more hparams, runtime breakdown, gpu memory usage
%     \item additional real world and synthetic figures + comparisons
%     \item reference the video in the suppl text
%     \item better ablation images: ours vs w Parametric
% \end{itemize}     % Text for the supplementary

\end{document}